\documentclass[12pt]{report}   
\usepackage{graphicx}  
\usepackage[letterpaper, left=1in, right=1in, top=1in, bottom=1in]{geometry}
\usepackage{setspace}  
\usepackage{times}  
\usepackage[explicit]{titlesec}  
\usepackage[titles, subfigure]{tocloft}  
\usepackage[backend=bibtex, sorting=none, bibstyle=ieee]{biblatex}  
\usepackage{appendix}  
\usepackage{rotating}  
\usepackage{ulem}  
\usepackage{textcomp} 
\usepackage{indentfirst} 
\usepackage{booktabs,array,arydshln} 
\usepackage{amsmath} 
\usepackage[T1]{fontenc} 
\usepackage[utf8]{inputenc} 
\usepackage{newtxmath} 
\usepackage[hidelinks]{hyperref}

\usepackage{bm}
\usepackage{algorithm}
\usepackage{algorithmicx}
\usepackage{algpseudocode}
\usepackage{subfigure}
\usepackage{gensymb}
\usepackage{csquotes}
\usepackage{empheq}
\usepackage{multicol}
\usepackage{multirow}
\usepackage{rotating}

\newcommand{\mystep}[1]{{\vspace{2.5mm}\noindent\textbf{#1}}}
\newcommand{\norm}[1]{\left\lVert#1\right\rVert}
\DeclareMathOperator*{\argmin}{arg\,min}
\algdef{SE}[DOWHILE]{Do}{doWhile}{\algorithmicdo}[1]{\algorithmicwhile\ #1}%

\bibliography{references}

\AtEveryBibitem{\clearfield{issn}}
\AtEveryBibitem{\clearlist{issn}}

\AtEveryBibitem{\clearfield{language}}
\AtEveryBibitem{\clearlist{language}}

\AtEveryBibitem{\clearfield{doi}}
\AtEveryBibitem{\clearlist{doi}}

\AtEveryBibitem{\clearfield{url}}
\AtEveryBibitem{\clearlist{url}}

\AtEveryBibitem{%
  \ifentrytype{online}
    {}
    {\clearfield{urlyear}\clearfield{urlmonth}\clearfield{urlday}}}

\begin{document}
\doublespacing  


\begin{titlepage}
\begin{center}

\begin{singlespacing}
\vspace*{6\baselineskip}
Grasp Stability Analysis with Passive Reactions\\
\vspace{3\baselineskip}
Maximilian Haas-Heger\\
\vspace{18\baselineskip}
Submitted in partial fulfillment of the\\
requirements for the degree of\\
Doctor of Philosophy\\
under the Executive Committee\\
of the Graduate School of Arts and Sciences\\
\vspace{3\baselineskip}
COLUMBIA UNIVERSITY\\
\vspace{3\baselineskip}
\the\year
\vfill

\end{singlespacing}

\end{center}
\end{titlepage}


\begin{titlepage}
\begin{singlespacing}
\begin{center}

\vspace*{35\baselineskip}

\textcopyright  \,  \the\year\\
\vspace{\baselineskip}	
Maximilian Haas-Heger\\
\vspace{\baselineskip}	
All Rights Reserved
\end{center}
\vfill

\end{singlespacing}
\end{titlepage}

\pagenumbering{gobble}

\begin{titlepage}
\begin{center}

\textbf{\large Abstract}

Grasp Stability Analysis with Passive Reactions

Maximilian Haas-Heger
\end{center}
\hspace{10mm}

Despite decades of research robotic manipulation systems outside of highly
structured industrial applications are still far from ubiquitous. Perhaps
particularly curious is the fact that there appears to be a large divide
between the theoretical grasp modeling literature and the practical
manipulation community. Specifically, it appears that the most successful
approaches to tasks such as pick-and-place or grasping in clutter are those
that have opted for simple grippers or even suction systems instead of
dexterous multi-fingered platforms.

We argue that the reason for the success of these simple manipulation systems
is what we call \textit{passive stability}: passive phenomena due to
nonbackdrivable joints or underactuation allow for robust grasping without
complex sensor feedback or controller design. While these effects are being
leveraged to great effect, it appears the practical manipulation community
lacks the tools to analyze them. In fact, we argue that the traditional grasp
modeling theory assumes a complexity that most robotic hands do not possess
and is therefore of limited applicability to the robotic hands commonly used
today.

We discuss these limitations of the existing grasp modeling literature and set
out to develop our own tools for the analysis of passive effects in robotic
grasping. We show that problems of this kind are difficult to solve due to the
non-convexity of the \textit{Maximum Dissipation Principle} (MDP), which is
part of the Coulomb friction law. We show that for planar grasps the MDP can
be decomposed into a number of piecewise convex problems, which can be solved
for efficiently. We show that the number of these piecewise convex problems is
quadratic in the number of contacts and develop a polynomial time algorithm
for their enumeration. Thus, we present the first polynomial runtime algorithm
for the determination of passive stability of planar grasps.

For the spacial case we present the first grasp model that captures passive
effects due to nonbackdrivable actuators and underactuation. Formulating the
grasp model as a Mixed Integer Program we illustrate that a consequence of
omitting the maximum dissipation principle from this formulation is the
introduction of solutions that violate energy conservation laws and are thus
unphysical. We propose a physically motivated iterative scheme to mitigate
this effect and thus provide the first algorithm that allows for the
determination of passive stability for spacial grasps with both fully actuated
and underactuated robotic hands. We verify the accuracy of our predictions
with experimental data and illustrate practical applications of our algorithm.

We build upon this work and describe a convex relaxation of the Coulomb
friction law and a successive hierarchical tightening approach that allows us
to find solutions to the exact problem including the maximum dissipation
principle. It is the first grasp stability method that allows for the
efficient solution of the passive stability problem to arbitrary accuracy. The
generality of our grasp model allows us to solve a wide variety of problems
such as the computation of optimal actuator commands. This makes our framework
a valuable tool for practical manipulation applications. Our work is relevant
beyond robotic manipulation as it applies to the stability of any assembly of
rigid bodies with frictional contacts, unilateral constraints and externally
applied wrenches.

Finally, we argue that with the advent of data-driven methods as well as the
emergence of a new generation of highly sensorized hands there are
opportunities for the application of the traditional grasp modeling theory to
fields such as robotic in-hand manipulation through model-free reinforcement
learning. We present a method that applies traditional grasp models to
maintain quasi-static stability throughout a nominally model-free
reinforcement learning task. We suggest that such methods can potentially
reduce the sample complexity of reinforcement learning for in-hand
manipulation.

\vspace*{\fill}
\end{titlepage}

\pagenumbering{roman}
\setcounter{page}{1} 
\renewcommand{\cftchapdotsep}{\cftdotsep}  
\renewcommand{\cftchapfont}{\normalfont}  
\renewcommand{\cftchappagefont}{}  
\renewcommand{\cftchappresnum}{Chapter }
\renewcommand{\cftchapaftersnum}{:}
\renewcommand{\cftchapnumwidth}{5em}
\renewcommand{\cftchapafterpnum}{\vskip\baselineskip} 
\renewcommand{\cftsecafterpnum}{\vskip\baselineskip}  
\renewcommand{\cftsubsecafterpnum}{\vskip\baselineskip} 
\renewcommand{\cftsubsubsecafterpnum}{\vskip\baselineskip} 

\titleformat{\chapter}[display]
{\normalfont\bfseries\filcenter}{\chaptertitlename\ \thechapter}{0pt}{\large{#1}}

\renewcommand\contentsname{Table of Contents}

\begin{singlespace}
\tableofcontents
\end{singlespace}

\clearpage





\clearpage
\pagenumbering{arabic}
\setcounter{page}{1} 


\titleformat{\chapter}[display]
{\normalfont\bfseries\filcenter}{}{0pt}{\large\chaptertitlename\ \large\thechapter : \large\bfseries\filcenter{#1}}  
\titlespacing*{\chapter}
  {0pt}{0pt}{30pt}	
  
\titleformat{\section}{\normalfont\bfseries}{\thesection}{1em}{#1}

\titleformat{\subsection}{\normalfont}{\thesubsection}{0em}{\hspace{1em}#1}



\chapter{Introduction}\label{sec:introduction}

\section{Background}\label{sec:intro_background}

Since the introduction of the first robotic manipulation systems they have
almost exclusively been confined to specialized spaces. This is true both in
the physical space (as to guarantee the safety of humans in the vicinity) and
the task space. The environments they work in are structured and engineers
design the tools that robots use specifically for a single task. This allows
for the application of robots in highly repetitive tasks where each task
instance is exactly the same as every other instance.  

This paradigm could not be farther removed from the environment us humans
experience. Our everyday environments are complex, unstructured and cluttered.
While some tasks may seem repetitive to a human, instances are never exactly
alike. Attempting to bridge this divide and enabling robotic manipulators to
become helpful in human spaces is perhaps the most important issue facing the
field of robotic manipulation today.

While to date no manipulation system has found success in regular human
environments, efforts at breaching the limitations and introducing functional
robots to everyday environments and tasks are ongoing. The advent of soft
robotics and passively compliant hands has enabled the grasping of isolated
objects. Advances made in computer vision over the last two decades allowed
for significant progress to be made towards solving pick-and-place tasks in
clutter. However, there remains a distinct lack of dexterity. In fact it
appears that the most successful methods for picking objects from clutter
leverage suction grippers instead of multi-fingered
hands~\cite{ijcai2017-676}\cite{7583659}\cite{morrison2018cartman}\cite{ackerman_2020}.
These methods practically reduce the grasping problem to one of computer
vision as they are concerned with identifying surfaces on which to place the
suction gripper and do not require grasp planning or contact modeling.

The majority of works focusing on practical applications of multi-fingered
hands view grasping as the task of placing contacts in order to fulfill a
given task. Despite the advent of practical tactile sensing technology little
attention is being payed to the explicit control of contact forces, which is a
defining characteristic of the human manipulation experience. There is a
wealth of work into the theoretical foundations and the mechanics of grasping
that aim to provide the understanding of the intricate relationship between
contact forces and actuation that is required for truly dexterous
manipulation. However, none of the popular practical manipulation
methods~\cite{mahler2017dex}\cite{1241860}\cite{5152709}\cite{4813847}\cite{SAUT2012347}\cite{8255597}
make use of those theoretical works. In fact, interest in theoretical grasp
modeling seems to have waned since the early 2000s. 

It appears as though the most progress towards dexterous multi-fingered
manipulation has been made in fields, which have only become established
relatively recently compared to the bulk of the manipulation literature. The
fields in question are those of Deep Learning and Reinforcement Learning.
There is promising work indicating that neural network policies can be trained
to process the vast amounts of sensor data and manipulate objects with more
and more
dexterity~\cite{doi:10.1177/0278364919887447}\cite{openai2019solving}.

It strikes the author as fascinating, that the foundational works of grasp
modeling appear to have been largely forgotten while large impact results are
obtained using entirely model-free techniques. We believe there are two
potential reasons for this:

\begin{enumerate}

\item The theoretical works on multi-fingered grasping and unilateral contact
modeling did not capture some important aspect of grasping, made the wrong
assumptions or are too computationally complex to be of practical use;

\item The theoretical works are unknown to the practitioners in the learning
community or their usefulness has not yet been discovered.

\end{enumerate}


We do believe that the first point above as to why the grasping theory has not
been widely adopted by the manipulation community has at least some truth to
it. In Chapter~\ref{sec:related} we will discuss in detail the different
theoretical approaches and their limitations in modeling modern robotic hands.
In short the majority of the grasp modeling literature introduces the
assumption that the contact wrenches\footnote{A wrench is a contraction of a
force and a torque.} of a grasp can be somehow actively controlled in real
time and in reaction to disturbances applied to the grasped object. 

In practice this assumptions means that the robotic hand must be fully
actuated, highly articulated, and equipped with torque or tactile sensors.
Robotic hands of such complexity are few in number and rarely used in
practice. For reasons we will discuss in Chapter~\ref{sec:bridge} the majority
of robotic hands in use today either lack the kinematics, the actuation scheme
or sensing ability to be able to actively modulate contact wrenches. This
means that in practice we rarely explicitly reason about the contact wrenches
when grasping objects with robotic hands. Instead, we rely on either of two
entirely passive effects:

Firstly, there is passive compliance as showcased by underactuated hands or
soft hands. While these hands are complex in their own right they trade off
real-time control of contact wrenches in favor of a priori design complexity.
The second effect is more subtle and applies to all hands with highly-geared
actuators and thus nonbackdrivable joints. Large gearing ratios between the
actuators and the joints coupled with the friction present in the gearbox mean
the fingers cannot be moved through forces applied externally to the finger.
This means that fingers can provide potentially rigid constraints for the
grasped object --- namely when the object is being pushed against that finger
by the applied wrench or contact wrenches from other fingers. While these
passive effects greatly reduce the complexity of control for grasping tasks
they cannot be analyzed using tools derived from the traditional grasping
theory. 

In Chapter~\ref{sec:bridge} we will provide an explanation as to why the
design of robotic hands has diverged so far from the theory developed in order
to model robotic grasping. However, with the increased interest in machine
learning for dexterous manipulation there is also renewed interest into highly
complex hands. Reinforcement learning is a promising candidate for the
synthesis of the highly complex controllers required for truly dexterous
manipulation. Such controllers --- or policies --- require enormous amounts of
data and thus robotic hands providing various modes of sensing technology are
being developed. We believe that with these new hands much of the classical
grasping theory will become relevant again. However, it appears that its
potential has not yet been realized by the learning community. 

\section{Contributions}

As we argue above the lack of ability to model the passive effects provided by
modern robotic hands is one of the most important limitations that have
stymied the success of the grasping theory in practical robotic manipulation.
To the best of the author's knowledge there is no explicit mention of the
implications of nonbackdrivable joints in the literature even though robotic
hands with such properties appear in the majority of practical manipulation
works. We thus develop our own methods of analyzing grasps with passive
effects. Specifically, we are interested in predicting the stability of grasps
as well as the contact forces that arise due to actuation forces or wrenches
otherwise applied to the grasp.

\begin{itemize}

\item We begin by investigating the passive stability of planar grasps and
develop a novel polynomial-time algorithm that makes use of results from
computational geometry. We improve on previous approaches that did not
incorporate accurate frictional constraints including the maximum dissipation
principle. As such, we introduce the first 2D grasp model with Coulomb
friction for grasp stability analysis that is solvable in provably polynomial
time.

\item Moving on to spacial grasps we show that omission of the maximum
dissipation principle introduces unphysical solutions. We thus present a
physically motivated iterative algorithm that mitigates this effect. While the
algorithm is not guaranteed to converge, we verify its accuracy with empirical
data.

\item In order to obtain an algorithm that is guaranteed to converge we
introduce a novel contact model that allows us to efficiently solve for
contact forces obeying Coulomb friction including the maximum dissipation
principle. Specifically, we introduce a convex relaxation of the Coulomb
friction model and an algorithm for the hierarchical successive tightening of
the relaxation. Our method is guaranteed to converge to the exact solution and
is sufficiently efficient for the analysis of practical grasps.

\item We apply our algorithms to answer for the first time grasp stability
queries that take into account the passive behavior of the robotic hand. These
results provide examples of possible applications of our framework to
practical manipulation tasks. We make implementations of the grasp models and
algorithms for spacial grasps publicly available as part of the open source
GraspIt! simulator~\cite{MILLER04}.

\item With the more recent drive towards more dexterous manipulation by the
deep learning community, there is renewed demand for highly sensorized hands
and we believe the rich grasping theory that has mostly been forgotten by the
manipulation community is once again becoming relevant. Our final contribution
is an excursion into a potential application of theoretical grasp models in
ordinarily model-free reinforcement learning of in-hand manipulation.

\end{itemize}

\section{Dissertation outline}

We begin by providing a broad summary of the grasp models and theories
developed by the manipulation community in Chapter~\ref{sec:related}, taking a
closer look at some of these works, which we find to be most pertinent to the
manipulation problems the field faces today. Furthermore, we will introduce
some of the more recent work demonstrating dexterous manipulation -
specifically reinforcement learning for in-hand manipulation.

What follows are the technical chapters of this dissertation. Beginning with
our work on grasp modeling in two dimensions presented in Chapter~\ref{sec:2d}
we then move our attention to problems in three dimensions in
Chapters~\ref{sec:three_dim} \&~\ref{sec:cones}. We will discuss some
preliminary results of our ongoing work investigating the applications of
grasp stability analysis to reinforcement learning of in-hand manipulation
tasks in Chapter~\ref{sec:shield}. 

Finally, we provide concluding remarks in Chapter~\ref{sec:conclusion}.

\chapter{Related Work}\label{sec:related}

Examining the problems the dexterous manipulation community is working on
today --- multi-fingered grasping, pick-and-place, in-hand manipulation,
regrasping, teleoperation, etc. --- they all share one requirement: the
ability to move a grasped object in a desired manner. In manipulation this
most often means maintaining a robust grasp on the object. Thus, the stability
of a grasp is a foundational prerequisite to many forms of dexterous
manipulation.

There is no singular agreed-upon definition of grasp stability in the
literature~\cite{doi:10.1177/027836499601500203}\cite{897777}. Stability might
be simply defined as the existence of equilibrium contact wrenches that obey
some friction law. This is quite different from the notion of stability in
control theory, which distinguishes between stable, unstable and marginally
stable equilibria. In some works the notion of grasp stability also includes
treatment of some region around an original equilibrium state and is thus
similar to that of asymptotic stability commonly encountered in control
theory, which denotes a system that returns to its original state after a
small perturbation is applied in an equilibrium state. Other works use a
broader definition of stability: Contacts may slip leading to the dissipation
of energy such that a system will not, in general, return to its initial state
after a small perturbation. Such a grasp may, however, still be considered
stable if it reaches a new equilibrium point where the grasped object is
relatively at rest with respect to the hand.

Furthermore, some works make use of the quasi-static assumption (all motions
are assumed to be slow enough such that effects due to inertia can be
neglected) while others use the full dynamic equations of motion in their
treatment of grasp stability. The exact definition of grasp stability heavily
depends on the context and the choices made by the researcher in modeling the
grasp. However, they all have in common the study of contact forces and
torques at the interfaces between a robotic hand and a grasped object.

\section{Grasp force analysis}\label{sec:related_closure}

\subsection{Force closure and stability}

The class of \textit{form closure} grasps is trivially stable as it consists
of grasps in which the contacts totally immobilize the grasp object. Analysis
of form closure grasps dates back to Reuleaux (1876)~\cite{REULEAUX1876} who
showed that for planar grasps a minimum of 4 frictionless contacts are
required for form closure. For spacial form closure grasps
Somov~\cite{SOMOV1900} and later Lakshminarayana~\cite{LASHMINARAYANA78}
showed a minimum of 7 frictionless contacts are required. Another class of
grasps is that of \textit{force closure} grasps. These grasps are such that
the hand can apply arbitrary wrenches to the object through the contacts.
There is still disagreement over the exact definitions of the closures
mentioned above. We choose to use the interpretation described by Bicchi and
Kumar in their review on grasp and contact modeling~\cite{BICCHI00}.

Salisbury~\cite{SALISBURY_THESIS} proposed an analytical method to test for
either form or force closure. However, perhaps the most commonly used method
to test for force closure was introduced by Ferrari and
Canny~\cite{FERRARI92}. Their algorithm constructs the total space of possible
resultant object wrenches that can be achieved by contact wrenches that obey
(linearized) friction constraints and whose magnitudes sum to 1. They call
this space the \textit{Grasp Wrench Space} or GWS. The smallest distance from
the origin to the convex hull of the wrenches making up the GWS is a measure
of the magnitude of the contact wrenches required to balance a worst-case
disturbance. Thus, it can also be used as a metric to quantify the
\textit{quality} of a grasp.  

The characteristic of force closure is a necessary requirement for a grasp to
be able to resist and therefore be stable with respect to arbitrary
disturbances. Force closure is, however, by no means a sufficient condition
--- it just indicates the existence of contact wrenches that satisfy the
friction laws that can balance arbitrary disturbances but does not guarantee
they will arise~\cite{10.5555/2887965.2888103}. The choice of specific grasp
forces, or as Bicchi~\cite{BICCHI94} calls it the \textit{force distribution
problem} is crucial for the stability of a grasp. Bicchi~\cite{BICCHI94}
defined two basic questions: 

\begin{displayquote}

"a) when external forces act upon the manipulated object disturbing its
equilibrium, how do they distribute between the contacts, and 

b) how can we modify the contact forces so as to achieve desirable values in
spite of external disturbances."

\end{displayquote}

\subsection{Methods assuming full contact wrench control}\label{sec:related_control}

A complicating factor in this analysis is the unilaterality of contacts in the
context of robotic grasping --- objects in contact may only 'push', not 'pull'
on each other. Furthermore, contact wrenches are subject to friction laws such
as the Coulomb friction model~\cite{COULOMB1781}; providing an upper bound to
the ratio of friction force magnitude to normal force magnitude. If we can
actively control the hand and its kinematics allow us to apply arbitrary
wrenches at the contacts we can ensure that contact wrenches satisfy both
unilaterality and the friction constraints at all times. Thus, this assumption
greatly simplifies analysis and will be made by all of the works discussed in
the rest of Chapter~\ref{sec:related_control}. In this context the question of
grasp stability becomes that of finding contact wrenches that satisfy the
friction law and equations of equilibrium.

A further complication is that in general robotic grasps are statically
indeterminate (or hyperstatic). Even if a grasp has force closure it is
difficult in general to synthesize stabilizing contact forces as there is an
infinite number of internal stresses and hence equilibrium contact wrenches.
Salisbury et al.~\cite{SALISBURY83} investigated the conditions for a grasp to
become overconstrained for a variety of different contact types and was the
first to express contact wrenches as the sum of a particular and a homogeneous
solution. The physical interpretation of these solutions can respectively be
described as the \textit{manipulation forces} balancing the applied wrench and
\textit{internal forces} that do not directly contribute to resisting the
applied wrench but cause internal stresses in the object. This decomposition
allows for the synthesis of stabilizing contact wrenches given a grasp and an
externally applied wrench. First, the particular solution must be computed.
This can be done using a generalized inverse of the grasp map matrix relating
contact wrenches to the overall resultant wrench the hand applies on the
object. The computed manipulation forces will not in general satisfy
unilaterality and friction constraints. Given the grasp has force closure an
appropriate homogeneous solution from the nullspace of the grasp map matrix
can be added to obtain legal stabilizing contact wrenches.

This insight allowed Kerr et al.~\cite{KERR86} to formulate the synthesis of
contact wrenches as an optimization problem providing the foundation for the
large body of work on grasp force optimization, which is concerned with
finding optimal grasp forces in the space of contact forces possible under a
friction law. Linearizing the friction constraints they propose solving a
linear program in order to obtain contact wrenches that are as far as possible
from violating the constraints already mentioned as well as additional
constraints such as actuator torque limits or the dynamic range of tactile
sensors. Nakamura et al.~\cite{NAKAMURA89} used Lagrange multipliers to solve
a nonlinear grasp force optimization problem. In contrast to Kerr et
al.~\cite{KERR86} they argue that contact wrenches should not be maximized for
stability: As Cutkosky et al.~\cite{CUTKOSKY86} pointed out increasing grasp
forces reduces the risk of slipping but may make the grasp less stable to
disturbances. Therefore, they instead find the contact forces that minimize
the internal forces but retain a certain level of contact robustness (i.e. are
far enough from violating contact constraints.)

In order to avoid having to linearize the friction model Buss et
al.~\cite{BUSS96} showed a different optimization formulation applicable when
nonlinear friction constraints can be formulated as positive-definiteness
constraints turning the optimization problem into a linear matrix inequality
(LMI) problem. Their algorithm, however, requires a valid initial guess of
contact wrenches and they do not discuss how to obtain such a guess. Han et
al.~\cite{HAN00} built upon this work and cast the grasp analysis problem as a
set of convex optimization problems involving said LMIs thus eliminating the
need for an initial guess. Boyd et al.~\cite{BOYD07} developed a custom
interior-point algorithm that exploits the structure of force optimization
problems and thus very efficiently solves such problems. In fact, the
complexity of their algorithm is only linear in the number of contacts.

Jen et al.~\cite{doi:10.1177/027836499601500203} take a control theoretic
approach to grasp stability. Starting from a set of initial equilibrium
contact wrenches they derive feedback controllers for the control of contact
wrenches in response to motions of the grasped object such that the grasp
remains asymptotically stable.


\subsection{Compliance methods}\label{sec:related_compliance}

An alternative method to resolve the static indeterminacy is to take into
account the flexibility of the object and the elements of the hand. If we
assume the hand to be a passive fixture we may solve for contact wrenches by
taking into account the flexibility of the object and the fixture through
constitutive equations. This also allows us to analyze grasps without the need
to be able to control contact wrenches explicitly as we have in effect
replaced unilaterality and the Coulomb friction law with a model linearly
relating contact wrenches to displacements.

Salisbury~\cite{SALISBURY_THESIS} used this approach to develop a framework to
test for the stability of a grasp. A grasp is stable if the stiffness matrix,
which characterizes the behavior of the grasp under small perturbations is
positive definite. Nguyen~\cite{1088008} modeled each contact as a spring (a
method first introduced by Hanafusa et al.~\cite{HANAFUSA77}) and showed that
any force closure grasp can be made stable under this definition by choosing
appropriate spring stiffnesses. Cutkosky et al.~\cite{CUTKOSKY_COMPLIANCE}
extended this work to take into account the compliance of structural elements
of the hand and object as well as effects due to changing geometry such as
contact location changes due to rolling contacts. Adding also the compliance
due to servo gains allows for active control of the overall grasp compliance.


Salisbury~\cite{Salisbury1988} pointed out that previous work had focused on
manipulation using only the fingertips of the hand. They suggested using more
of the surface of robotic manipulators, which Salisbury calls 'Whole Arm
Manipulation'. Bicchi~\cite{BICCHI93} noted that in such cases we cannot
assume that the kinematics of the hand allow us to explicitly apply arbitrary
contact wrenches. For enveloping grasps (Bicchi calls them "whole-limb"
grasps) contacts occur on links with limited mobility such as proximal links
or the palm of a robotic hand. In consequence we cannot freely choose
internal forces in order to synthesize stable grasp forces --- only a subset
of the internal forces are actively controllable.

Similarly to the work by Nguyen~\cite{1088008} Bicchi models the contacts as
springs but also uses the compliance matrix introduced by Cutkosky et
al.~\cite{CUTKOSKY_COMPLIANCE} and the principle of virtual work to derive
expressions for the subset of \textit{active internal forces} (internal
forces, which can be actively controlled by commanding the hand actuators) and
the subset of \textit{passive internal forces} (which cannot be actively
controlled and therefore will remain constant at their initial value).
Demonstrating how a particular solution can be found from the compliance
matrix weighted inverse of the grasp map matrix Bicchi proposes a method to
find optimal contact wrenches, which are actually achievable with the
kinematics in question~\cite{BICCHI94}. This was a large step forwards as it
allowed for the analysis of grasps commonly encountered in practice.

With these insights Bicchi proposed a new definition of the term 'force
closure', which takes into account the kinematic capabilities of the grasping
mechanism to actively control the contact wrenches~\cite{BICCHI95}. He
introduced an algorithm to test for force closure under this definition. A
modification to this algorithm allows for a quantitative measure of how far a
grasp with optimal contact wrenches is from violating contact constraints.
Thus, the concept of a metric to quantify the quality of a grasp was extended
from a purely geometric concept~\cite{FERRARI92} to include the kinematic
capabilities of the hand to apply optimal contact wrenches.

\subsection{Quality metrics}\label{sec:related_metrics}

Quality metrics are an essential component of the grasp planning problem,
which can be described as the problem of computing appropriate contacts the
hand should make on the object. Optimal grasp planning can be posed as a
search over the space of possible grasps in order to find an instance that
optimizes a given quality metric. It should come to no surprise that grasp
quality metrics usually relate to the stability of the grasp. We have already
introduced the metric derived from the grasp wrench space~\cite{FERRARI92} and
the metric by Bicchi, which takes into account the hand
kinematics~\cite{BICCHI95}. Note, that the objective value in grasp force
optimization methods such as in~\cite{KERR86} can be understood as a metric of
grasp quality as well. 

In many of the aforementioned works, the objective to be optimized (and hence
the quantitative measure of grasp quality) is related to how far the contact
wrenches are from violating contact constraints. Prattichizzo et
al.~\cite{PRATTICHIZZO97} take a similar approach, however, similarly to
Nakamura et al.~\cite{NAKAMURA89} they relate the distance contact wrenches
are from violating contact constraints to the magnitude of the external
disturbance acting on the object. Given optimal contact wrenches the magnitude
of the smallest external disturbance that will cause a contact constraint
violation is defined as the \textit{Potential Contact Robustness} or PCR. In
contrast to~\cite{NAKAMURA89}, however, contact forces are optimized over the
space of active internal forces as defined by
Bicchi~\cite{BICCHI93}\cite{BICCHI94}\cite{BICCHI95} using the compliance
analysis by Cutkosky et al.~\cite{CUTKOSKY_COMPLIANCE}

Another valuable contribution is founded on the insight that the robustness of
individual contacts is not necessarily required for grasp stability. They
argue that a grasp will remain stable even if some contact constraints are
violated. Contacts may slip or detach entirely --- as long as a sufficient set
of contacts remain that satisfy contact constraints grasp stability may be
maintained. Thus, PCR can be an overly conservative measure of stability. To
achieve a less conservative grasp stability metric --- the \textit{Potential
Grasp Robustness} or PGR --- they make some simplifying assumptions in order
to avoid modeling of the complex frictional behavior at the contacts.
Specifically, they assume a slipping contact may not apply any frictional
forces, which results in a stability measure that is still conservative,
however less so than one based solely on contact robustness. A further
disadvantage of this approach is that every potentially stable combination of
contact states (rolling, slipping, detaching) must be considered. Therefore
the computational complexity of this approach scales exponentially in the
number of contacts, which can be prohibitive for real-time evaluation of
enveloping grasps that make many contacts between the hand and grasped object.

\subsection{Discussion}\label{sec:related_gfa_discussion}

The methods introduced in Chapter~\ref{sec:related_control} assume that we
have some degree of control over the contact wrenches. They deem a grasp to be
\textit{potentially} stable if a solution to the equilibrium equations exists
that also satisfies contact constraints. However, as we have noted before,
force closure is no guarantee for stability. The existence of a solution to
the equilibrium equations is no guarantee that it will arise. Instead, we must
actuate the grasping mechanism accordingly in order to achieve the desired
equilibrium contact wrenches. In Chapter~\ref{sec:bridge} we will show that
this assumption is of only limited applicability for the majority of robotic
hands commonly used as they lack the capabilities to accurately control joint
torques. 

The works discussed in Chapters~\ref{sec:related_compliance}
and~\ref{sec:related_metrics} introduce grasp models that do not necessarily
presuppose the full control of contact wrenches. However, in order to do so
they must introduce treatment of the compliance of the grasp and hence make
another set of assumptions. Specifically, Bicchi~\cite{BICCHI93} made use of
the grasp compliance theory developed by Cutkosky et
al.~\cite{CUTKOSKY_COMPLIANCE}. Assume the grasp is made up of deformable
components such as series-elastic actuators or the contacts themselves are
deformable due to an elastic skin covering the finger links. These deformable
components would provide us with constitutive equations such that we can solve
for displacements and hence contact wrenches given a specific external
disturbance.

One obvious limitation of this approach is that the stiffness parameters can
be difficult to obtain~\cite{TRINKLE95}, particularly for very stiff hands.
Another limitation was pointed out by Prattichizzo et
al.~\cite{PRATTICHIZZO97}: When there is compliance in the grasp the presence
of sliding or breaking contacts does not necessarily mean the grasp is
unstable. Only a subset of contacts may be required for stability. This causes
an issue as the contact wrench at a breaking or sliding contact is clearly not
described by a compliance relation. At a breaking contact a compliance
relation would lead to a negative normal force, which is impossible. At a
sliding contact the contact force would depend linearly on the tangential
displacement, which does not satisfy the Coulomb friction law. While
Prattichizzo et al. find approximations that work well for their purposes,
there is no computationally efficient and accurate way to apply an approach
based on grasp compliance to a problem with potentially breaking or slipping
contacts.

As all of the works discussed thus far do not model the friction forces at
sliding contacts the Coulomb friction model is often reduced to the statement
it makes about static contacts: The magnitude of the friction force may not
exceed the magnitude of the normal force scaled by the coefficient of static
friction. To the reader of the grasp force analysis literature it can often
appear as though this statement was the complete Coulomb friction model.
However, the Coulomb model also includes treatment of sliding contacts.
Specifically, it includes the statement that the friction force, which arises
at a sliding contact is the one, which dissipates the most energy. Thus, for
isotropic friction the friction force must oppose tangential motion at a
sliding contact.

By itself this is known as the \textit{principle of maximum dissipation} or
\textit{Maximum Dissipation Princinple} (MDP)~\cite{Moreau2011}. For a more
recent discussion of the MDP see the review by Stewart~\cite{STEWART00}.
Although many readers will know the MDP to be part of the Coulomb model of
friction, in this dissertation we will explicitly state when we make use of it
due to its absence from much of the grasp force analysis literature.

\section{Rigid body methods}\label{sec:related_dynamics}

A grasp can be viewed as a completely passive fixture. The grasp is made up of
rigid bodies that are either fixed in space (like the 'grounded' base of the
robot), free floating (like the grasped object) or connected by a kinematic
chain (such as the manipulator links.) The bodies comprising the kinematic
chains are connected to each other and --- at least indirectly --- the ground
by bilateral constraints such as hinge joints. Contacts between bodies are
unilateral constraints also satisfying a friction law such as Coulomb
friction. All external forces acting on the bodies such as gravity or
actuation wrenches are known. Such an arrangement is deemed stable if it
remains at rest (all accelerations are zero) or as Palmer~\cite{PALMER_THESIS}
puts it

\begin{displayquote}

"In the context of robotic assembly, we assume that objects are initially fixed
in space, and say that they are stable if the force of gravity cannot cause
the position of any object to change."

\end{displayquote}

\subsection{Rigid body mechanics}

Palmer~\cite{PALMER_THESIS} investigated the stability of arrangements of
rigid polygonal bodies under gravity and showed that the problem of
determining stability is co-NP complete. However, they proposed a class of
stability, which can efficiently be tested for using linear programming
techniques: \textit{Potential stability} is satisfied if contact wrenches
exist within the contact constraints that also satisfy equilibrium. Mattikalli
et al.~\cite{MATTIKALLI96} also noted the difficulty of determining the
stability of assemblies of bodies with frictional contacts and used
\textit{potential stability} as defined by Palmer~\cite{PALMER_THESIS} as an
indicator for stability. They point out that the existence of a solution to
the contact constraints and equilibrium equations is a necessary but not
sufficient condition for stability and introduce conservative approximations
of the friction law to mitigate this. Furthermore, they note that in the
frictionless case a potentially stable arrangement is in fact guaranteed to be
stable.

Erdmann~\cite{doi:10.1177/027836499401300306} investigated the motion of rigid
planar bodies in frictional contact given initial velocities. They note that
in order to predict the motion of a body the relative motion of the contacts
must be known. This apparent circularity stems from the different contact
conditions due to unilaterality and the maximum dissipation principle. They
consider each contact to be in one of several states: A contact can break,
roll or slide. They propose solving the contact problem by first hypothesizing
a combination of contacts states and then verifying that the resulting motion
is consistent with the contact conditions. Thus, they propose enumerating all
possible combinations of contact states, which provides an exponential
complexity method of solving the rigid body contact problem. They note that
the rigid body contact problem does not necessarily have a unique solution or
may even have no solution at all.

Trinkle et al.~\cite{TRINKLE95} also advocate for a rigid body treatment in
order to obtain accurate solutions without having to resort to computationally
taxing finite elements methods to model compliance. They develop theory to
predict the motion of rigid bodies in the plane under a quasistatic assumption
(i.e. inertial effects are considered negligible.) and Coulomb friction. In
their analysis the joints of the manipulator are assumed to be either position
controlled or torque controlled. 

Similarly to the work by Erdmann~\cite{doi:10.1177/027836499401300306} their
initial approach~\cite{TRINKLE95} requires the consideration of every possible
combination of contact states where a contact can either roll, detach or slip
in the tangential plane. While the number of feasible contact state
combinations is quadratic in the number of contacts if the hand is
immobile~\cite{Brost1989a}\cite{MASON_MANIPULATION} they
note that it is exponential in the number of bodies involved and hence in
general the number of combinations is exponential in the number of contacts.

In further work Pang et al.~\cite{PANG1996} avoid this enumeration of contact
state combinations by casting the contact constraints as an uncoupled
complementarity problem (UCP). They show that problems of this type are
NP-complete but present a bilinear programming approach to solving them.
Trinkle et al.~\cite{TRINKLE05} later extended this line of work to grasps in
three dimensions.

\subsection{Rigid body dynamics}

Dropping the quasistatic assumption, Trinkle et al.~\cite{TRINKLE97} worked
with the full dynamic equations instead. They showed that the Coulomb friction
law with maximum dissipation can be cast as a mixed nonlinear complementarity
problem (mixed NCP), which is difficult to solve. Therefore, they propose
linearizing the friction cone by approximating it as a pyramid, which allows
for a formulation of the friction constraint as a linear complementarity
problem (LCP). While problems of this type can be solved with Lemke's
algorithm, a downside of this approximation is that the friction force will in
general lie on the specific edge of the friction cone discretization, which
maximizes the energy dissipation and therefore may not exactly oppose motion.
An immensely influential time stepping scheme for the solution of
multi-rigid-body dynamics with Coulomb friction that makes use of this
framework became known as the Stewart-Trinkle formulation~\cite{STEWART96}.
Another very influential time-stepping scheme --- the Anitescu-Potra
formulation~\cite{Anitescu1997} --- can be obtained by omission of the
constraint stabilization term from the Stewart-Trinkle formulation.

Pang et al.~\cite{ZAMM643} point out that the equations of equilibrium are
insufficient for the determination of stability of rigid body contact problems
due to the possibility of false positives. Therefore they use the methodology
developed in~\cite{TRINKLE97}, which uses the dynamic equations instead with
rigid bodies initially at rest and fixing the position of the elements making
up the hand. Using this framework they build on the stability
characterizations proposed by Palmer~\cite{PALMER_THESIS}, which they describe
as overly conservative. They define the class of \textit{strongly stable
loads} --- disturbances which are guaranteed not to destabilize the grasp in
their framework. They attribute the distinction between strongly and weakly
stable loads (Palmer's \textit{potential stability}) to the nonuniqueness of
the possible contact wrenches; there may be many solutions to the dynamic
equations that satisfy the friction constraints but not all may be stable.
Thus, they define the strongly stable loads as those, where every potential
set of contact wrenches results in \textit{nonpositive virtual work}.

If the solution is unique, then strong and weak stability are equivalent.
Using the uniqueness result from~\cite{TRINKLE97}, they show that this is the
case if the friction coefficient is below a certain bound. Unfortunately, this
bound is not known and the authors expect it to be difficult to compute in
general. Using  the pyramidal linearization of the friction cone they do,
however, present an exponential-time method to check if strong and weak
stability are equivalent for a given grasp geometry. In those special cases
strong stability can efficiently be determined as weak stability can be
determined by solving a linear program.

Song et al.~\cite{SONG04} attribute the nonuniqueness of solutions to both the
statical indeterminacy of general grasping problems and the nonsmooth nature
of the relationship between tangential forces and relative velocities. In
order to overcome these difficulties they propose a compliant model of the
contacts between nominally rigid bodies. Both normal and tangential contact
forces are determined by viscoelastic constitutive relations coupling them to
local deformations.

Anitescu and Tasora~\cite{ANITESCU03}\cite{TASORA10} showed that using the LCP
formulation of the pyramidal friction cone approximation in general leads to
non-convex solution sets. LCP problems with non-convex solution sets contain
reformulated instances of the Knapsack problem and are therefore NP-hard.
Thus, problems of this type are difficult to solve. The authors develop an
iterative algorithm to solve them that converges to the solution of the
original problem. They achieve this through successive convex relaxation
effectively solving subproblems that have the form of strictly convex
quadratic programs. They note that their algorithm is only guaranteed to
converge for 'sufficiently small' friction coefficients but provide a lower
bound for convergence.

In further work Anitescu et al.~\cite{Anitescu2010} argue that the
computational burden of solving LCP based formulations of rigid body dynamics
problems is too high for applications with large numbers of colliding bodies.
They also point out that the linearization of the friction cone as a polygonal
pyramid required for formulation as an LCP violates the assumption of
isotropic friction. They instead propose a cone-complementarity approach and
an iterative method that converges under fairly general conditions but may
allow bodies to behave as if they were in contact although they have drifted
apart, particularly at high tangential velocities~\cite{HORAK19}.

Kaufman et al.~\cite{KAUFMANN08} showed that solving for friction forces when
normal forces are known can be achieved by solving a quadratic program (QP).
Similarly, the normal forces can be solved for when friction forces are know.
However, problems where both are to be solved for simultaneously are
non-convex. The solution of non-convex QPs is generally
NP-hard~\cite{Murty1987}. The authors present an algorithm which iterates
between the two convex QPs until a solution of the required accuracy is found.
Although it works well in practice convergence is not guaranteed with their
approach.

Todorov~\cite{TODOROV10} takes a completely new approach by deriving nonlinear
equations for the dynamic contact problem that implicitly satisfy the
complementarity conditions. In order to solve the resulting nonlinear
equations Todorov proposes a Gauss-Newton approach with specific adaptations
such as a novel linesearch procedure. Todorov~\cite{TODOROV14} and Drumwright
et al.~\cite{Drumwright2011} independently developed formulations that are
relaxing the complementarity conditions such that a convex optimization
problem is recovered. 

More recently, Pang et al.~\cite{PANG18} showed that the LCP formulation for
discretized friction cones first introduced by Trinkle~\cite{TRINKLE97} can be
cast as a Mixed Integer Program instead. While this method still suffers from
the same inaccuracies they demonstrated its applicability for the control of a
robotic gripper in simulation.

An excellent starting point to some of the most influential works towards
rigid body dynamics with frictional contacts is the comparative review by
Horak and Trinkle~\cite{HORAK19}, which elucidates some of the similarities
and differences between many of the approaches listed above.

\subsection{Discussion}\label{sec:related_rigid_discussion}

We introduce the works discussed above because they can be applied to robotic
grasps while making no assumption of active control of contact wrenches. In
fact, the community focusing on the analysis of rigid bodies has developed
very natural complementarity formulations that capture passively stable
grasps. Furthermore, there are numerous relaxations available in order to
improve efficiency.

In order to determine the stability of a grasp with respect to a given
disturbance we could, for example, attempt to use any of the various dynamics
solvers the field has produced to compute contact forces and calculate a
single time step. If all accelerations remain zero, we deem the grasp stable.
Unfortunately, as we will now discuss, this approach will not give us definite
results.

Any algorithm that makes use of relaxations also introduces artifacts as we
are no longer solving the exact problem as outlined above. These artifacts are
acceptable for the purposes of those algorithms: Dynamics simulation that is
fast enough for use in time sensitive applications such as virtual reality or
deep learning. In our application, however, these artifacts mean that we
cannot rely on the simulation to produce exactly zero accelerations even for
grasps that are truly stable.

Thus, we need to focus on the algorithms that solve the exact problem. Perhaps
the best candidate is the formulation as a mixed nonlinear complementarity
problem (mixed NCP) introduced by Trinkle et al.~\cite{TRINKLE97}. In a
previous paper~\cite{PangTrinkle1996} Pang and Trinkle showed that in the case
of systems initially at rest the existence of a solution is guaranteed.
However, they also note a difficulty in using this formulation for stability
analysis: One must show that \textit{every} solution to the complementarity
problem has a zero acceleration.

Generally, when using a complementarity formulation in a dynamics engine only
one solution is required in order to step the simulation forward in time. This
allows the application of solvers such as Lemke's algorithm, which only finds
a single solution --- if one exists. The only algorithms that find all
solutions to general complementarity problems are enumerative in
nature~\cite{murty1988linear} and therefore of exponential computational
complexity.

To the best knowledge of the author, the work which comes closest to answering
the question of stability under the above assumptions is that by Pang et
al.~\cite{ZAMM643}, which makes use of the complementarity formulation due to
Trinkle et al.~\cite{TRINKLE97}. In special cases, in which the solution is
unique, there is no ambiguity due to multiple solutions and the single
solution to the LCP determines stability. A framework that efficiently
determines stability for general grasps under the rigid body assumption has
not yet been found.

\section{Dexterous in-hand manipulation}\label{sec:rel_ihm}

In-hand manipulation is the task of reorienting a grasped object with respect
to the grasping robotic hand. One approach to problems of this kind is to make
use of kinematic models of the hand and object and apply analytic methods from
the realm of optimization and planning algorithms to compute an appropriate
series of behaviors such as contact sliding, contact rolling and finger
gaiting. These approaches (see for
instance~\cite{Erdmann1998}\cite{677060}\cite{6631137}\cite{525325}\cite{795789}\cite{6630637}\cite{Yunfei2014}\cite{10.5555/2422356.2422377}\cite{OdhnerDollar}\cite{8347081}\cite{10.1007/978-3-030-28619-4_39}\cite{Hou2020})
assume some knowledge of the grasped object geometry as well as the ability to
model the dynamics and interactions at the contacts.

In contrast to these model-based methods the advent of efficient Reinforcement
Learning (RL) algorithms has allowed researchers to investigate entirely
model-free approaches. As such, the grasp is treated as a black box and Deep
Reinforcement Learning (DRL) is used to train end-to-end policies directly mapping
from sensor inputs to actuator commands. van Hoof et al.~\cite{7363524}
demonstrated the viability of RL for in-hand manipulation with planar grasps.
They used an underactuated hand for which devising analytical controllers is
difficult and leveraged tactile sensors to provide features for the policy.

More recently results published by groups at OpenAI have garnered large
publicity. They demonstrated reorientation of a block in the palm of a
anthropomorphic robotic hand~\cite{doi:10.1177/0278364919887447} using a
scaled-up implementation of the Proximal Policy Optimization
algorithm~\cite{schulman2017proximal}. They train their policy in simulation
and use domain randomization in order to achieve robust transfer to the real
robotic hand. The reason they train in simulation is the sample complexity of
their method: it takes approximately 100 years of simulated experience to
train a policy that robustly transfers to the real hand. In later work they
extended this work to the solving of Rubik's cubes in a similar experimental
setup~\cite{openai2019solving}. In order to improve sample efficiency Zhu et
al.~\cite{zhu2018dexterous} as well as Rajeswaran et
al.\cite{Rajeswaran-RSS-18} leveraged Imitation Learning in combination with
RL. 

Li et al.~\cite{Li2020LearningHC} point out that none of the model-free
approaches above require the hand to maintain a stable and robust grasp of the
object throughout the grasp. The object is either supported by a
tabletop~\cite{7363524}, the palm of the
hand~\cite{doi:10.1177/0278364919887447}\cite{openai2019solving}\cite{Rajeswaran-RSS-18}
or limited in its degree of freedoms through external
supports~\cite{zhu2018dexterous}. In order to overcome this limitation Li et
al. propose a hierarchical control structure instead of the monolithic
end-to-end paradigm of DRL. They develop analytical torque controllers for
three different manipulation primitives that are intrinsically designed for
grasp stability. A higher level DRL policy is trained to choose the
manipulation primitive as well as its parameters. Using this approach they
achieve stable in-hand manipulation for planar grasps in simulation.

The idea of maintaining a stable grasp throughout an RL task has parallels in
the 'Safe Reinforcement Learning' literature (see Garcia et
al.~\cite{JMLR:v16:garcia15a} for a survey of Safe RL.) Avoiding unsafe states
during training and execution of RL policies is similar to avoiding states
that result in an unstable grasp during an in-hand manipulation task.
Particularly interesting is the work of Dalal et al.~\cite{dalal2018safe} in
which they append a 'safety layer' to their policy network that performs an
action correction. The action correction is formulated as an optimization
problem that uses a linearized model and thus allows for an analytic solution.
The optimization computes the minimum perturbation to the action such that
safety constraints are satisfied. Amos et al.~\cite{10.5555/3305381.3305396}
introduced 'OptNet': A neural network architecture that integrates
optimization problems such that complex constraints can be captured.

Due to the relative novelty of DRL approaches to in-hand manipulation we
believe there are many synergies yet to be discovered in the combination of
the above ideas and approaches from the traditional grasp analysis theory.

\chapter{Grasp stability analysis --- Theory and Praxis}\label{sec:bridge}

In Chapter~\ref{sec:related} we argued that the notion of a grasp's stability
is of foundational importance to the majority of manipulation tasks and hence
many competing definitions of stability have been proposed by the grasp
modeling community. In essence, all these definitions share a common goal: to
maintain a grasp such that the grasped object remains relatively at rest with
respect to the hand throughout the task. Motions in reaction to a disturbance
that lead to another static equilibrium where again the object is at rest with
respect to the hand may be allowed. We will use this informal definition of
stability in the following general discussion of manipulation. In
Chapter~\ref{sec:definition} we will provide a more precise definition of
stability in the context of our own grasp model.

In practice, maintaining stability as defined above throughout a task requires
maintaining stability with respect to every disturbance the grasp will
encounter throughout that task. Determining the ability of a grasp to resist
given disturbances, formulated as external wrenches applied to the grasped
object, is equivalent to computing the stability of a multi-body system with
frictional contacts under applied loads. Problems of this kind are thus
pervasive in grasp analysis and may be encountered in many other scenarios
that rely on the stability of assemblies of general rigid bodies with
frictional contacts.

Inspired by the queries formulated by Bicchi~\cite{BICCHI94} and quoted in
Chapter~\ref{sec:related_closure} we formulate our own pair of queries. We
argue that these queries are foundational to the majority of robotic
manipulation tasks:

\begin{displayquote}

\textit{Given a set of actuator commands and an external disturbance will the
system as described above remain stable?}

\end{displayquote}
We can also formulate the inverse query to obtain useful insight into
how to control the hand: 

\begin{displayquote}

\textit{Given an external disturbance, how must we command the actuators to
guarantee the grasp remains stable?}

\end{displayquote}

In the context of robotic manipulations the ability to answer these queries is
of great practical use. A grasp model that can answer these queries must
capture the interplay of contact forces, joint torques and externally applied
wrenches. It must be able to accurately predict how joint torques and external
wrenches are transmitted through the object and distributed across the
contacts taking into account the unilaterality of contacts, the specific
actuation scheme of the hand as well as the nonlinear nature of friction laws.
So why is there no 'grand unified theory of grasp stability'?

As discussed in~\ref{sec:related_closure} much of the existing grasp analysis
literature makes the simplifying assumption that all contact wrenches are
being actively controlled at all times. This can be done directly through
commanding appropriate joint torques or more indirectly through setting
controllable compliance parameters such as servo gains. This assumption allows
the analysis of grasp stability through the equilibrium equations  and
friction laws alone and allow us to compute optimal contact forces and the
joint torques necessary to balance them. This is incredibly powerful, as for
any specific wrench applied to the object encountered throughout a task we can
compute the specific optimal joint torques for stability.

In order to use this in practice, however, we have to make a string of
assumptions:

\begin{enumerate}

\item We assume perfect knowledge of the disturbance to the object
that must be balanced at all times;

\item We assume that we can actively control the contact wrenches at every
contact;

\item We assume that we can actively control the joint torques required
for equilibrium;

\item We assume that we can accurately control the torque
output of the hand actuators.

\end{enumerate}

In the majority of robotic manipulation tasks these assumptions do not hold. 
First, the exact wrench acting on an object is difficult to compute --- it
requires knowledge of the mass and inertial properties of the object as well
as its exact trajectory. Any additional disturbance cannot be accounted for
unless the fingers are equipped with tactile sensors. Second, many robotic
hands are kinematically deficient and contain links with limited mobility.
This means that we cannot directly control (through controlling joint torques)
the contact wrench at a contact on the palm of the hand for instance. Wrenches
at such a contact can only arise passively by transmission of the disturbance
on the grasped object or wrenches at other contacts through the object.

Third, the kinematics of the hand may not permit explicit control of the
torques at every individual joint. This is the case for the class of
underactuated hands, where joint torques by definition may not be
independently controlled but are also a function of the kinematic composition
and the pose of the hand. Finally, the actuation method of the hand may
prevent accurate torque control at the joints. Most robotic hands use highly
geared motors for instance, which makes accurate sensing and control of the
torques at the hand joints all but impossible.

Therefore, in practice it is rarely the case that contact wrenches are
actively controlled in response to disturbances to the grasp. Unfortunately,
this means that the existence of contact wrenches that satisfy the equilibrium
equations as well as friction constraints is a necessary but by no means
sufficient condition for the ability of a grasp to resist a given disturbance.
In general such contact wrenches will not arise, unless we are actively
controlling the contact wrenches to that end.

Having discussed the limiting assumptions of the grasp analysis theory it is
instructive to think about why robotic hands have not evolved in such a way as
to fulfill the assumptions listed above. We discussed why these assumptions
are difficult to satisfy in practice and why attempts to remove those
assumptions introduce their own limitations. Still, it seems to not provide
enough justification for the schism that separates the theory and the robotic
hands commonly in use today. After all, why build hands that are so far
removed from the theory that allows their analysis? 

In order to answer this question we must revisit a concept, which we call
\textit{passive stability} or \textit{passive resistance}. It denotes the
ability of a grasp to resist a given applied wrench without active control of
the hand, but through purely passive phenomena. In
Chapter~\ref{sec:intro_background} we mentioned two such phenomena
contributing to passive stability: underactuation and nonbackdrivability. Let
us for now focus on nonbackdrivability, which we believe to have had a large
influence on the manipulation field despite the community being largely
unaware of it.

Nonbackdrivability allows practitioners to greatly simplify robotic grasping.
We can often apply actuator torques that close the fingers around the object
without necessarily worrying if these will balance out once contact is made:
the fingers jam as the hand squeezes the object, and the gearboxes between
joints and actuators provide additional structural torques. If the grasp
geometry is adequately chosen, the equilibrium joint torques arise passively
when the fingers squeeze the object between them and a stable grasp arises.

For example, in the grasp in Fig.~\ref{fig:package} it is sufficient to
actively load the joints of one finger. The nonbackdrivability of the other
finger means the object will be stably grasped and equilibrium joint torques
arise passively in the non-actuated finger. The same phenomenon can allow a
grasp to withstand a range of disturbances applied to the object without a
change in the actuator commands. If chosen wisely, the initially applied
actuator forces --- called a \textit{preload} --- are sufficient to balance
the object throughout the task and various corresponding different
disturbances. Again, if the grasp geometry is adequately chosen, equilibrium
joint torques arise passively when disturbance wrenches applied to the object
push it against the fingers. This is the true power of passive resistance.

We believe this characteristic to be the reason why robotic manipulation
without explicit real-time control of the hand actuators has become so
commonplace. Reactions to disturbances arising passively due to
nonbackdrivable actuators remove the need for complex hand control schemes and
the high fidelity sensing they require. If the grasp geometry and actuator
commands are chosen wisely the grasp will be passively stable to a range of
disturbances greatly simplifying the control of the hand. We simply need to
pick a constant command appropriate to the task. So the answer to the question
as to why most robotic hands lack the complexity required for the application
of the grasp theory is that practitioners can rely on passive stability
instead. This allows them to form stable grasps, even though they lack the
tools required to analyze them.

Underactuated hands provide similar advantages: These hands make use of their
passive compliance in order to conform around objects. Soft robotic hands take
this approach to the extreme. Most of the hands commonly found in research
labs today exhibit either underactuation, nonbackdrivability or both. They all
have in common that their passive behavior greatly simplifies control, which
explains their popularity.

\section{A representative example problem}\label{sec:bridge_example}

\begin{figure}[!t]
\centering
\includegraphics[width=0.75\columnwidth]{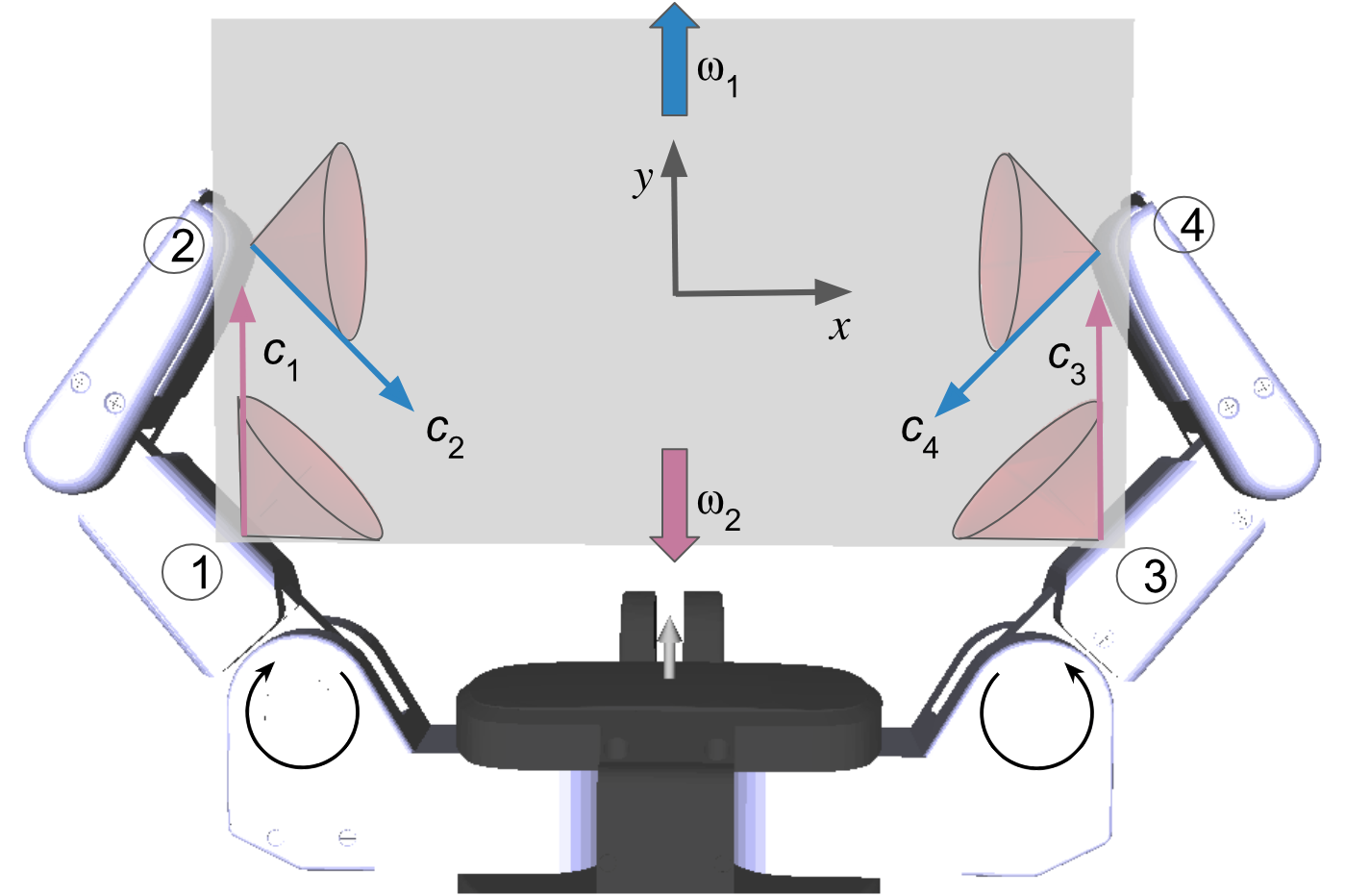}

\caption{A grasping scenario where a hand establishes multiple frictional
contacts (numbered 1-4) with a target object. External disturbance $\bm{w}_1$
can be resisted by contact forces $\bm{c}_2$ and $\bm{c}_4$; $\bm{w}_2$ by
contact forces $\bm{c}_1$ and $\bm{c}_3$.}\label{fig:package}

\end{figure}

In order to illustrate the implications of passive stability on robotic
manipulation let us consider the grasp in Fig.~\ref{fig:package}. Does the
grasp remain stable if we apply either disturbance $\bm{w}_1$ or $\bm{w}_2$ to
the grasped object? In order to resist those disturbances, contact forces must
arise that balance them. Clearly in either case there exist contact forces
that satisfy a simple friction law (illustrated by red friction cones) and
balance the disturbance. In fact, this grasp has \textit{force closure} and
hence contact forces exist that could balance arbitrary disturbances. 

In practice, however, this is not a sufficient criterion for grasp stability.
It is clear that contact forces $\bm{c}_2$ and $\bm{c}_4$ will only arise if
we have previously loaded the grasp such that there is sufficient normal force
at contacts 2 and 4 to sustain the friction forces required: an appropriate
preload is required, or the object will slip out. We could, for example apply
actuator torques at the joints such that the hand 'squeezes' the object and
hence provides such a preload.

We can make a similar argument about contact forces that balance disturbance
$\bm{w}_2$, however if we assume that the joints on the robotic hand are
nonbackdrivable (as is the case with most robotic hands in use today) contact
forces $\bm{c}_1$ and $\bm{c}_3$ will arise entirely passively. The
disturbance will push the object against the fingers, which due to the
nonbackdrivability of the joints provide a rigid support. There is no need for
us to apply any torques at the joints as the grasp will be stable regardless.
The grasp provides \textit{passive resistance} to the applied wrench.

As discussed in Chapter~\ref{sec:related_gfa_discussion} none of the
traditional grasp force analysis works discussed in
Chapter~\ref{sec:related_closure} can accurately capture the passive behavior
of this grasp. A perhaps more natural formulation for problems of this kind
can be found in the rigid body dynamics literature discussed in
Chapter~\ref{sec:related_dynamics}. We mentioned the main complication in
using such a formulation based on the dynamics equations and complementarities
in Chapter~\ref{sec:related_rigid_discussion}. The nonuniqueness of solutions
in the general case means that one has to find all solutions to the problem to
show that all of them result in zero accelerations. 

This ambiguity due to multiple solutions can be well illustrated using again
the grasp in Fig.~\ref{fig:package} assuming nonbackdrivable fingers: Consider
the case of applying $\bm{w}_2$ with a zero preload such that contact forces
are zero prior to the application of $\bm{w}_2$. A formulation based on
complementarities admits at least two solutions:

In solution 1 the box slides upward out of the grasp completely unhindered by
the grasp, as it is not applying any contact forces. This is the intuitively
correct solution to this grasp stability problem. If we do not preload the
grasp it may not withstand $\bm{w}_2$.

In solution 2 contact forces arise at contacts 2 and 4 such that there is
sufficient friction to maintain a stable grasp on the object. This can happen
because, while the contacts are not previously loaded with contact forces,
they are maintained and the complementarity condition for contact forces to
arise is met. Due to the assumption of nonbackdrivability both fingers are
rigidly fixed in space, which allows for arbitrarily large contact forces to
arise without the need to balance them with joint torques.

While this second solution is not useful in practice --- we know it will never
arise --- it is a perfectly valid solution to the formulation used to model
the grasp. Its is simply a feature of the rigid body assumption. In reality,
however, no body is truly rigid, which is why this second solution is never
observed in practice. The existence of such solutions that are mathematically
correct, but physically impossible is problematic. The second solution by
itself would have us conclude the grasp is stable. Thus, in order to be able
to conclude that a grasp is stable from a single solution we must derive a
different formulation that relaxes the rigid body assumption.

The grasp shown in Fig.~\ref{fig:package} is exceedingly simple but still
representative of many robotic manipulation tasks. And yet we lack the means
to analyze the passive effects that make these grasps so powerful. If we want
to move beyond blindly using passive reactions and truly understand how to
leverage them we must be able to model passive effects. We can then answer the
stability queries we defined at the beginning of this chapter with our new
understanding of the mechanics of grasping.


\chapter{Grasp Stability Analysis in Two Dimensions and Polynomial Time}\label{sec:2d}

\section{Introduction}

\begin{figure}[!t]
\centering
\includegraphics[width=0.85\linewidth]{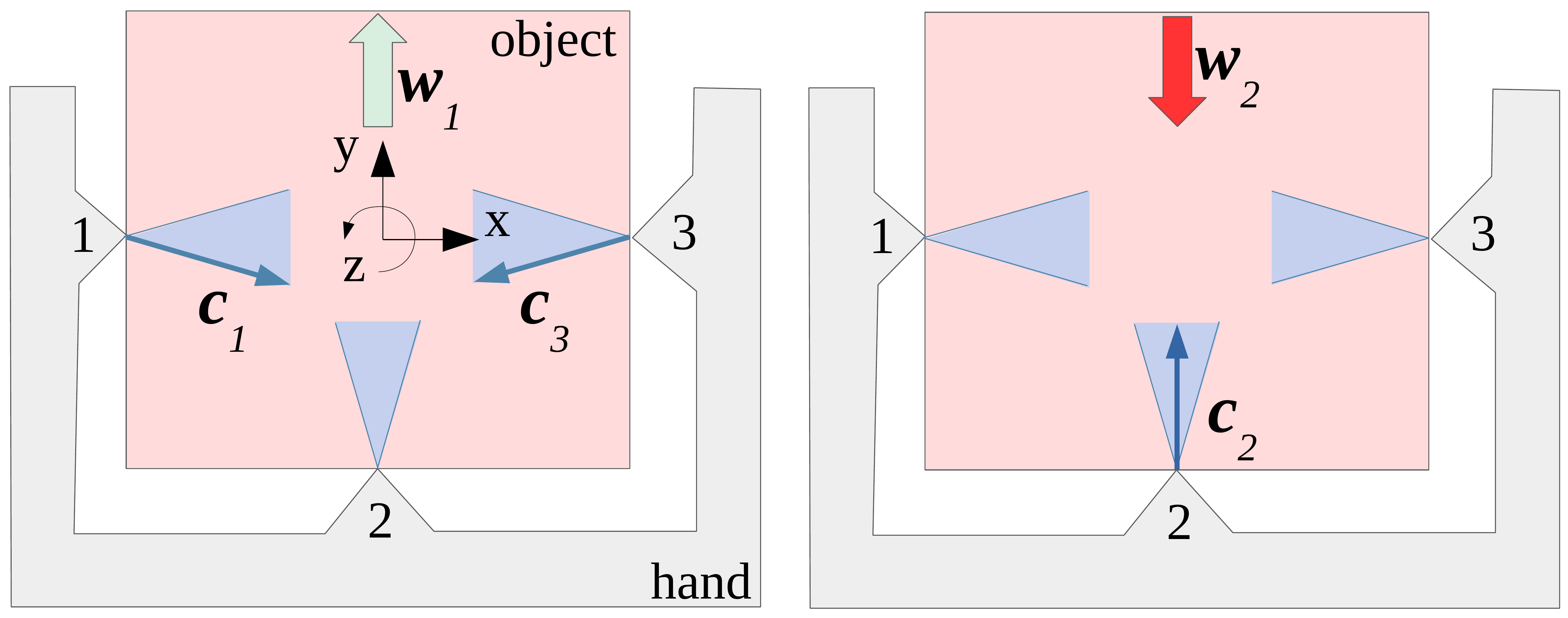}

\caption{A planar grasping scenario where a hand establishes multiple
frictional contacts (numbered 1-3) with a target object. }

\label{fig:grasp}
\end{figure}

In Chapter~\ref{sec:bridge} we argued that for a grasp model to be useful as a
grasp stability analysis tool it must be able to predict the purely passive
behavior of the grasp. In this chapter we will begin to investigate the
modeling of such passive phenomena in the special case of planar grasps. This
simplification is convenient when modeling friction forces under the maximum
dissipation principle. The reason is that in the plane contacts can only slide
in one of two directions. This is illustrated in the grasp in
Fig.~\ref{fig:grasp}: there are no out of plane motions or forces to consider.

This grasp was constructed to exhibit passive characteristics very similar to
those of the grasp in Fig.~\ref{fig:package}: External disturbance $\bm{w}_1$
(left, pushing the object up) can be resisted by contact forces $\bm{c}_1$ and
$\bm{c}_3$, but only if contacts 1 and 3 have been actively pre-loaded with
enough normal force to generate the corresponding friction forces. In
contrast, disturbance $\bm{w}_2$ (right, pushing the object down), regardless
of its magnitude, will always be passively resisted by contact force
$\bm{c}_2$. In this chapter we introduce a computationally efficient method to
predict the stability of passive planar grasps such as the one in
Fig.~\ref{fig:grasp} when applying arbitrary wrenches to the grasped object.

As we are interested in passive stability we cannot make the simplifying
assumption that we have any control over the contact forces. Thus, we take a
similar approach to previous researchers, who leverage the compliance of the
grasp to resolve the static indeterminacy
\cite{SALISBURY_THESIS}\hspace{0pt}\cite{NGUYEN88}\hspace{0pt}\cite{HANAFUSA77}\hspace{0pt}\cite{CUTKOSKY_COMPLIANCE}\hspace{0pt}\cite{BICCHI93}\hspace{0pt}\cite{BICCHI94}\hspace{0pt}\cite{BICCHI95}\hspace{0pt}\cite{PRATTICHIZZO97}.
Modeling the compliance of the grasp provides \textit{constitutive equations}
relating motion of the object to contact forces. This is a common approach in
the solution of statically indeterminate systems, in which contact forces
cannot be uniquely determined from the equations of equilibrium alone and are
therefore difficult to solve. The contact forces are then fully determined by
the relative motion between the hand and the object --- the system becomes
statically determinate. These constitutive relations are desirable, as they
are linear and therefore allow for very efficient solution.

However, as we described in Chapter~\ref{sec:related_compliance} the
unilaterality of contact normal forces as well as the the frictional
components of contact wrenches cannot be accurately described by compliance
relations. Thus, we choose to model only normal forces at persisting contacts
with constitutive relations while modeling friction with the Coulomb friction
model including the maximum dissipation principle (MDP)~\cite{Moreau2011}.

Initially, we assume all bodies making up the grasp to be at rest. Due to the
compliance we introduced to resolve the static indeterminacy, we expect some
motion once an external disturbance is applied to the grasped object. This may
include breaking or sliding contacts. However, we assume all motion to be
small such that we only have to account for first order effects and can ignore
any higher order effects on the grasp geometry due to this motion. Assuming
the grasp geometry to remain constant greatly simplifies the grasp model and
is a reasonable assumption for relatively stiff hands and objects as the
motions observed are indeed very small.

Furthermore, we assume that all dynamic effects are negligible. As such, when
we speak of the motion of a body we really mean the displacement it undergoes
when a wrench is applied to it. As is common when dealing with rigid body
motion we will consider a displacement to include both the translation and
the rotation of the body.



\section{Grasp model}\label{sec:2dmodel}

We first introduce the general framework of our grasp model. Consider a rigid
robotic hand that makes $m$ contacts with a grasped object. The system is
initially at rest. Each contact is defined by a location on the surface of the
grasped object and a normal direction determined by the local geometry of the
bodies in contact. We choose the \textit{Point Contact with Friction} model to
describe the possible contact wrenches that may arise at the interfaces
between the hand and the object. Therefore we only consider contact forces and
do not allow for frictional torques. This is a reasonable assumption for
contacts between smooth and relatively stiff bodies. For any contact specific
vector such as contact force or contact motion we will use subscripts $n$ and
$t$ respectively to denote the components lying in the contact normal and
contact tangent directions. We use the vector $\bm{c} \in \mathbb{R}^{2m}$ to
denote contact forces, where $\bm{c}_i \in \mathbb{R}^2$ is the force at the
i-th contact. Using the notation above, $c_{i,n} \in \mathbb{R}$ is the normal
component of this force, and $c_{i,t} \in \mathbb{R}$ is its tangential (i.e.
frictional) component.

\vspace{2mm} \noindent \textbf{Equilibrium:} Let us define the grasp map
matrix $\bm{G} \in \mathbb{R}^{3 \times 2m}$, which maps contact forces into a
frame fixed to the grasped object. We can now write the equilibrium equations
for the grasped object where we collect all disturbances externally applied to
the object (such as gravitational forces for instance) in $\bm{w} \in
\mathbb{R}^3$.

\begin{equation}
\bm{G}\bm{c} + \bm{w} = 0 \label{eq:equilibrium} 
\end{equation} 

The transpose of matrix $\bm{G}$ also maps object motion
$\bm{r}\in\mathbb{R}^3$ to translational motion of the contacts in the contact
frame $\bm{d}\in\mathbb{R}^{2m}$. Recall that in the context of our grasp
model by motion we mean the displacement of the object and this relation is
only valid for small displacements.

\begin{equation}
\bm{G}^T\bm{r} = \bm{d} \label{eq:contact_motion}
\end{equation}

Note that as we assume a constant grasp geometry matrix $\bm{G}$ is also
constant.

\vspace{2mm} \noindent \textbf{Normal forces:} Due to the unilaterality of
contacts --- we do not concern ourselves with adhesion or similar effects ---
a contact may only push on an object but can never pull. Thus, the normal
force at a contact must be strictly non-negative. Furthermore, if the contact
detaches the contact force must be zero. We model the normal forces by placing
virtual linear springs along the contact normals such that normal forces will
only arise through object motions that compress these springs. As we have
assumed all motions to be small such that grasp geometries are invariant we
are not interested in actually computing the real magnitude of the
displacements of the grasped object. Instead, we can think of the motion as
entirely virtual; it is simply a tool to resolve static indeterminacy and
enforce contact constraints and thus we can forgo having to identify the true
compliance parameters of the hand. To simplify notation we choose the virtual
springs to be of unity stiffness. The result is a compliant version of the
\textit{Signorini Fichera Condition}.

\begin{subequations}
\begin{empheq}[left=\empheqlbrace] {align}
c_{i,n} &= -d_{i,n} &\text{if } d_{i,n} \leq 0 \\
c_{i,n} &= 0 &\text{if } d_{i,n} > 0
\end{empheq}\label{eq:normal_forces}
\end{subequations}

We are here implicitly assuming that all contacts are of equal stiffness. For
grasps with very dissimilar contact compliances it may be necessary to choose
spring stiffnesses with appropriate ratios with respect to each other.
Nonetheless, the absolute magnitude of the spring stiffnesses is irrelevant as
object motion scales inversely proportional to spring stiffness and hence
contact forces are invariant with respect to spring stiffnesses.

Of course only one of the two pairs of constraints in (\ref{eq:normal_forces})
can be active at any contact, the choice of which depends on if the contact
breaks or not. As the object motion is part of the solution we seek it is
unknown which pair of constraints is active. However, these constraints can
also be formulated in terms of two linear inequalities and a non-convex
quadratic constraint that must all be satisfied in either case:

\begin{eqnarray}
c_{i,n} &\geq 0 \\
c_{i,n} + d_{i,n} &\geq 0 \\
c_{i,n} \cdot (c_{i,n} + d_{i,n}) &= 0
\end{eqnarray}

If re-posed as a Linear Complementarity Problem the matrix relating the
vectors of unknowns is non-positive definite and therefore problems including
such constraints are difficult to solve. We can instead construct a system of
equations for each possible combination of contact constraints and search for
solutions in all such problems. This way a problem contains only the active
pair of linear equality and inequality constraints and can thus be readily
solved. Of course we must now solve multiple simpler problems instead of one
complex one, but as we will see we can make use of the structure of the
problem to arrive at an algorithm of polynomial complexity as opposed to a
direct approach to the LCP, which would be worst-case exponential. 

We will therefore work directly with the formulation of the normal force
constraints in (\ref{eq:normal_forces}) after adding one more detail: We may
be interested in grasps where we assume a certain level of contact normal
forces is already present before the application of any external wrench. The
virtual springs in such a grasp are pre-stressed in order to achieve the
desired normal forces --- we call this the preload of a grasp. We denote the
preload at contact $i$ as $p_i$. Adding this term to (\ref{eq:normal_forces})
we obtain the following:

\begin{subequations}
\begin{empheq}[left=\empheqlbrace] {align}
c_{i,n} &= -d_{i,n} + p_i &\text{if } d_{i,n} - p_i \leq 0 \\
c_{i,n} &= 0 &\text{if } d_{i,n} - p_i > 0
\end{empheq}\label{eq:normal_compliance}
\end{subequations}

\vspace{2mm} \noindent \textbf{Friction forces:} We choose the Coulomb model
to describe the friction forces at the contacts. For a stationary contact this
model simply states that the frictional component of the contact force is
bounded by the normal component scaled by the coefficient of friction $\mu_i$.
If a contact slides, however, the \textit{maximum dissipation principle}
(MDP)~\cite{Moreau2011} states that the friction force must oppose the relative
motion at the contact. We can express the Coulomb friction law as follows:

\begin{subequations}
\begin{empheq}[left=\empheqlbrace] {align}
c_{i,t} &\leq \mu_i c_{i,n} &\text{if } d_{i,t} = 0 \\
c_{i,t} &= \mu_i c_{i,n} &\text{if } d_{i,t} < 0 \\
c_{i,t} &= -\mu_i c_{i,n} &\text{if } d_{i,t} > 0
\end{empheq}\label{eq:cone}
\end{subequations}

Similarly to the conditional constraints in (\ref{eq:normal_compliance}) we
will deal with this set of constraints by constructing multiple problems with
different combinations of active constraints such that individual problems
only consist of linear constraints.

\section{Definition of stability}\label{sec:definition}

In Chapter~\ref{sec:2dmodel} we introduced the mathematical constraints
relating object motions and contact forces assuming a constant grasp geometry.
Hence, they define the behavior of a grasp under application of an external
disturbance assuming a constant grasp geometry. In this context we can now
provide a more precise definition of stability.

As we are interested in the stability of a given grasp with respect to a
specific external wrench applied to the grasped object, such a grasp geometry/
wrench pair defines an instance of the stability problem. The question of
stability then becomes one of \textit{solution existence}: A grasp is deemed
stable if a solution exists to the set of constraints described in
Chapter~\ref{sec:2dmodel}. More specifically, a grasp is deemed stable if a
displacement exists such that contact wrenches arise that balance the wrench
applied to the grasped object. As such, the motion can be interpreted as the
(small) movement the bodies undergo under the applied loads before they reach
a new equilibrium.

As the constraints are specific to the initial grasp geometry, non-existence
of a solution only indicates instability of that exact geometry. It does not
indicate if through changes in grasp geometry due to object motion the object
will eventually settle in a new equilibrium. The initial grasp would be deemed
unstable under our definition.

Note also that stability is explicitly defined with respect to a specific
applied wrench as opposed to a more broad definition such as force closure or
asymptotic stability in the control theoretic sense. This is an important
distinction as a grasp can be stable with respect to a specific wrench even
without satisfying force closure. This opens up our definition of stability
and the following analysis to many other problems that may not be immediately
appear related to grasping. A house of cards on a table in a gravitational
field for instance may not intuitively resemble a grasp. The supporting
surface certainly has no force closure over the contacting cards, yet a (well
constructed) house of cards will be stable under the above definition.

We will leave open for now the question of what does and does not constitute a
grasp and focus for the rest of this dissertation on mechanisms we believe
most readers would immediately recognize as grasps with robotic hands.
However, we expect the scope of our definition of stability and the following
work to be much broader. 






\section{Solving for stability given contact states}\label{sec:2d_solving}

The formulation we described in Chapter~\ref{sec:2dmodel} introduces a major
complication as equations (\ref{eq:normal_compliance}) define two possible
constraints and equations (\ref{eq:cone}) define three possible constraints.
Only one constraint of each set can be active a contact but we do not know
which constraints hold a priori. We do not know if a contact detaches and if
it does not if it slips in a negative sense, slips in a positive sense or
rolls.

Note that which constraint is active is entirely determined by the relative
motion at the corresponding contact. A contact may persist or detach, which
determines the contact normal force. The friction force constraint is
determined depending on if the contact slips in a positive sense, a negative
or not at all. This means a contact must be in one of six states.

\begin{table}[h]
\centering
\begin{tabular}{ c | c }
Persists / Detaches & Rolls / Slips \\
\hline
Persists & Rolls \\
Persists & Slips + \\
Persists & Slips - \\
Detaches & Rolls \\
Detaches & Slips + \\
Detaches & Slips - \\
\end{tabular}
\caption{Enumeration of possible states for a single contact}
\label{tab:contact_states}
\end{table}

In order to describe a system with multiple contacts one must know the state
of each contact. This introduces a combinatorial aspect to the problem due to
the different possible combinations of contact states. We will investigate the
number of possible combinations of these states in
Chapter~\ref{sec:state_number} and their enumeration in
Chapter~\ref{sec:enumeration} but for now let us assume that we are given a
valid combination of states such that we can construct the appropriate
constraints and solve for equilibrium.

Note that every constraint possibility in equations
(\ref{eq:normal_compliance})\&(\ref{eq:cone}) consists of a pair of an
inequality and an equality. Thus we can construct a linear system of equations
from the equality components of all the constraints also including the
equations of equilibrium equation (\ref{eq:equilibrium}). Equation
(\ref{eq:contact_motion}) is a simple linear mapping such that all $\bm{d}$
can be expressed in terms of $\bm{r}$. 

\begin{equation}
\bm{A}
\begin{bmatrix}\bm{c} \\ \bm{r}\end{bmatrix} 
= 
\begin{bmatrix}\bm{G} & 0 \\ \bm{A}_1 & \bm{A}_2\end{bmatrix}
\begin{bmatrix}\bm{c} \\ \bm{r}\end{bmatrix} 
= 
\begin{bmatrix}-\bm{w} \\ \bm{p}\end{bmatrix}\label{eq:linear_system}
\end{equation}

Matrix $\bm{A}_1$ and $\bm{A}_2$ collect the equality constraints from
(\ref{eq:normal_compliance})\&(\ref{eq:cone}) that correspond to the set of
contact states we are given. $\bm{p}$ contains the preloads $p_i$ for the
contacts and constraints where they are applicable and zeros elsewhere. For
grasps without a preload it contains all zeros. As we have $2m+3$ equations
for an equal number of constraints matrix $\bm{A}$ is square. Thus, there are
two cases:

\begin{enumerate}

\item if $\bm{A}$ is invertible solve the linear system
(\ref{eq:linear_system}). The solution is unique and if it satisfies the
applicable inequality constraints in equations
(\ref{eq:normal_compliance})\&(\ref{eq:cone}) then it is indeed a valid
solution to the system.
	
\item if $\bm{A}$ is singular solve a linear program with equality constraints
(\ref{eq:linear_system}) subject to the applicable inequality constraints in
equations (\ref{eq:normal_compliance})\&(\ref{eq:cone}). If the linear program
is feasible then the computed solution is valid.

\end{enumerate}

If through either approach no valid solution is found then the set of contact
states given cannot result in equilibrium. This gives us an algorithm that is
exponential in the number of contacts as one could naively enumerate all $6^m$
possible contact states, construct the corresponding system of equations for
each and solve the resulting linear problem as described above. There is,
however, a method to reduce the number of combinations that must be checked
for a solution as only some combinations are valid under rigid body
constraints.

\section{Number of possible contact states}\label{sec:state_number}

In order to analyze the possible contact state combinations it is opportune to
consider the case of contacts detaching and the different slipping conditions
separately. Thus let us define two partitions: First we partition the contacts
into those detaching and those, which persist. We define vector $U_k \in
\{0,1\}^m$. The $i$-th element of $U_k$ is labeled $u^k_i$ and indicates if
contact $i$ detaches.

\begin{subequations}
\begin{empheq}[left=\empheqlbrace] {align}
u^k_i &= 1 &\implies& \text{contact persists, } &d_{i,n} - p_i \leq 0 ~,&~ c_{i,n} = -d_{i,n} + p_i \\
u^k_i &= 0 &\implies& \text{contact detaches, } &d_{i,n} - p_i > 0 ~,&~ c_{i,n} = 0
\end{empheq}
\end{subequations}
We now define the second partition of contacts this time into three sets:
those slipping in a positive sense, those slipping in a negative sense and
those that remain relatively at rest. We define vector $S_k \in \{-1,0,1\}^m$.
The $i$-th element of $S_k$ is labeled $s^k_i$ and indicates the slip state of
contact $i$.

\begin{subequations}
\begin{empheq}[left=\empheqlbrace] {align}
s^k_i &=  0 &\implies& \text{contact rolls, } &d_{i,t} = 0~,&~ c_{i,t} \leq \mu_i c_{i,n} \\
s^k_i &= 1 &\implies& \text{contact slips in $+$ sense, } &d_{i,t} > 0~,&~ c_{i,t} = -\mu_i c_{i,n} \\
s^k_i &= -1 &\implies& \text{contact slips in $-$ sense, } &d_{i,t} < 0~,&~ c_{i,t} = \mu_i c_{i,n}
\end{empheq}
\end{subequations}

Finally $\mathbb{U}$ is the set of all possible system contact detachment
states. Thus $U_k \in \mathbb{U}$ for $k=1..\#(\mathbb{U})$, where
$\#(\mathbb{U})$ is the cardinality of $\mathbb{U}$. Similarly $\mathbb{S}$ is
the set of all possible system contact slip states. At first glance
$\#(\mathbb{U})=2^m$ and $\#(\mathbb{S})=3^m$: since each contact can have two
detachment states and three slip states, the total number of states for the
system is exponential in the number of contacts.

However, it is known that the number of possible slip states for a planar
rigid body in contact with other rigid bodies fixed in space is indeed
quadratic in the number of
contacts~\cite{Brost1989a}\cite{MASON_MANIPULATION}. This perhaps surprising
fact stems from the rigid body constraints that impose relations between the
motions of all contacts on the same body. Trinkle et al.~\cite{TRINKLE95}
point out that this also applies to a grasped object if the hand is immobile.
In the following we shall prove this is indeed the case and show that the
number of possible contact detachment states is also polynomial in the number
of contacts, even for pre-stressed grasps. In Chapter~\ref{sec:enumeration} we
describe an algorithm to enumerate all geometrically possible combinations of
contact states. 

Let us begin by noting that in the three-dimensional space of possible object
motions $\bm{r}=[x,y,r]$, the constraints $d_{i,n} = (\bm{G}^T \bm{r})_{i,n} =
0$ and $d_{i,t} = (\bm{G}^T \bm{r})_{i,t} = 0$ define two-dimensional planes.
We shall in the following only consider the planes generated by the tangential
contact motion constraint as the processing they require is more involved. We
will then describe the simplifications that can be made in processing the
planes due to the normal contact motion constraints.

Consider a tangential motion constraint $(\bm{G}^T \bm{r})_{i,t} = 0$. Any
object motion lying on this plane will result in zero relative tangential
motion at this contact. Motion in the open halfspace where $(\bm{G}^T
\bm{r})_{i,t} > 0$ will result in slip along the tangential axis in the
positive direction, while motion in the complementary open halfspace
$(\bm{G}^T \bm{r})_{i,t} < 0$ will result in slip in the negative direction.
Combining the planes defined by each contact, we can construct an
\textit{arrangement} of planes. These planes segment the space of object
motions into the following partitions:

\begin{itemize}
  \item 3-dimensional \textit{regions} where all contacts are slipping;
  \item 2-dimensional \textit{facets} (region boundaries on a single
  plane) where one contact is rolling;
  \item 1-dimensional \textit{lines} (intersections of multiple
    planes) where two contacts are rolling.
\end{itemize}

\begin{figure}[t]
  \centering
  \includegraphics[width=1.0\linewidth]{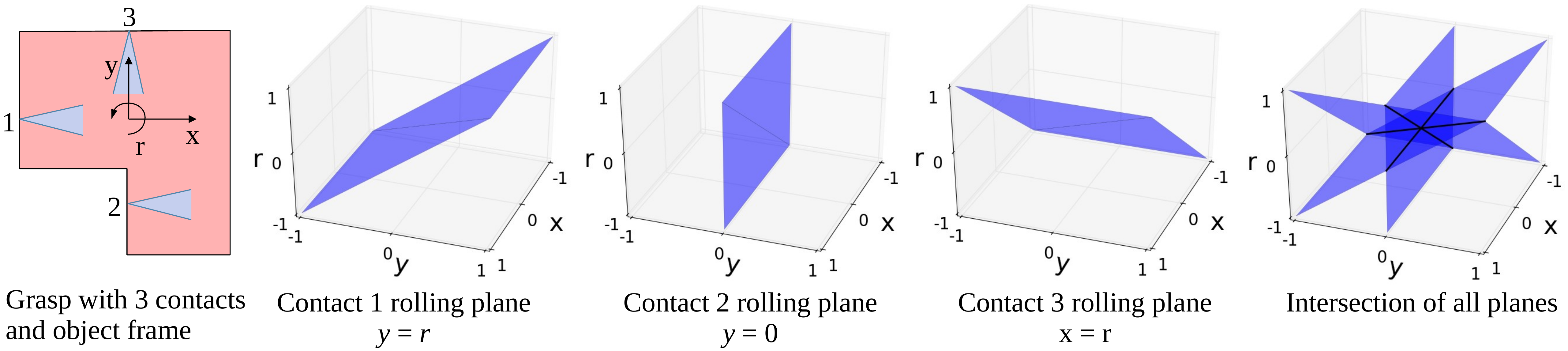}
  \vspace{-5mm}
  \caption{Arrangement of planes equivalent to contact roll-slip
    constraints. For the grasp shown on the left, each contact defines
    a plane in the three-dimensional space of possible object motions
    $\bm{r}=[x,y,r]$. For example, Contact 1 rolls if the translation
    component of $\bm{r}$ along $y$ is counteracted by an equal
    rotational component $r$. Similarly, Contact 2 rolls if the $y$
    component of $\bm{r}$ is 0, and so on.}
  \label{fig:planes}
\end{figure}

By construction, since all of our planes go through the origin, the only
possible zero-dimensional \textit{point} intersection is the origin itself
(see Fig.~\ref{fig:planes}.) Such an arrangement of planes is said to be
\textit{central}~\cite{Ovchinnikov2011}. For the benefit of generality, the
partitions of a space created by an arrangement of hyperplanes are usually
described in terms of the dimensionality of the partition: A $k$-dimensional
partition is known as a "k-face" of the arrangement. Thus, we will refer to
the \textit{regions}, \textit{facets}, \textit{lines} and \textit{points} as
the 3-faces, 2-faces, 1-faces and 0-faces of the arrangement. 

Given an arrangement of planes as described above, it follows that any system
slip state $S_k$ that is consistent with a possible object motion must
correspond to either a 3-face, 2-face, 1-face and 0-face of this plane
arrangement. Finding the maximum number of 3-faces given $m$ planes in a
central arrangement is equivalent to finding the maximum number of
two-dimensional regions on a sphere cut with $m$ great circles, which is known
to be $O(m^2)$~\cite{GREATCIRCLES}. However, the 3-faces do not define all the
combinations of slip states we care about. We must also consider the cases
where at least one contact rolls, namely the 2-faces, 1-faces and 0-faces
defined as above. 

\begin{figure}[t]
  \centering
  \includegraphics[width=0.65\linewidth]{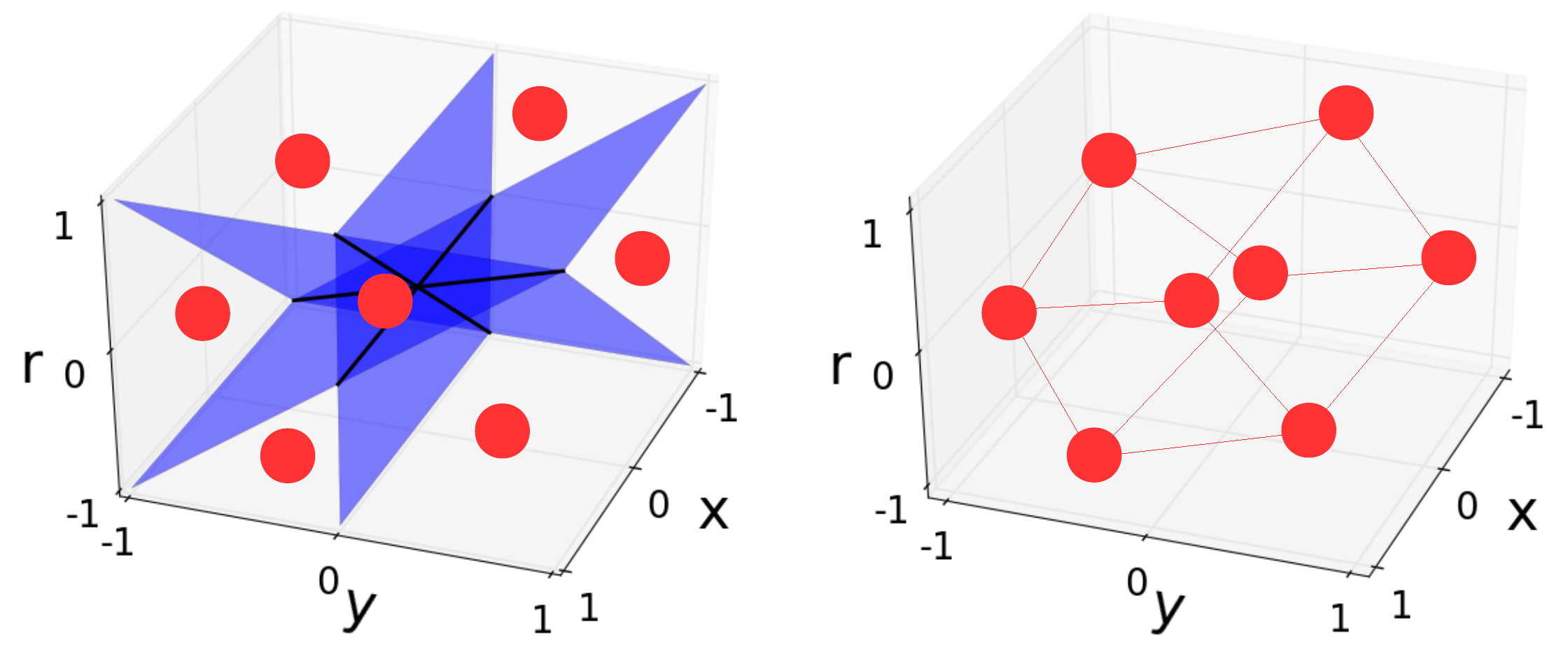} 
  \caption{Dual polyhedron for the arrangement of planes. Each 3-face
    of the arrangement of planes (red dot) corresponds to a
    vertex of the polyhedron (for clarity, only 7 regions are marked
    in the left figure). Vertices are connected if the corresponding
    3-faces share a 2-face.}
  \label{fig:poly}
\end{figure}

Let $f_k^{(d)}(n)$ be the number of k-faces of an arrangement of $n$
hyperplanes in $d$ dimensional space. Zaslavski's formula~\cite{FUKUDA91}
provides the following upper bound:

\begin{equation}
f_k^{(d)}(n) \leq {n \choose d-k} \sum_{i=0}^k {n-d+k \choose i}
\end{equation}

In our case, the total number of slip states we are interested in is equal to
$\sum_{k=0}^3 f_k^{(3)}(m)$ and hence bounded from above by a polynomial in
$m$. This is, however, an upper bound that will not be attained in our case,
as we show next. Let us construct the dual convex polyhedron to our
arrangement of planes, which will be instrumental in enumerating all possible
slip states. Each 3-face of the plane arrangement corresponds to a vertex of
the polyhedron, and each two dimensional boundary (2-face) between 3-faces
corresponds to an edge connecting the vertices corresponding to the
neighboring 3-faces (see Fig.~\ref{fig:poly}.) We can ensure this polyhedron
is convex by selecting, as the representative vertex for each region, a point
where the 3-face intersects the unit sphere. We note that the 1-faces of our
plane arrangement correspond to the faces of the dual polyhedron. The dual
polyhedron thus also fully describes our possible slip states. 

Like any convex polyhedron, the dual we have constructed can be represented by
a 3-connected planar graph. From this result, we can bound the number of slip
states even more closely than with Zaslavski's formula: Any maximal planar
graph with $V$ vertices has at most $3V-6$ edges and $2V-4$ faces and hence
the number of edges and faces of our polyhedron are linearly bounded by the
number of its vertices. Since we already know the number of vertices to be
$O(m^2)$, so are the number of edges and faces. Thus, the number of slip
states we must consider is quadratic in the number of contacts.

For the number of detachment states the argument must be adapted if the grasp
is pre-stressed, as the planes --- due to the normal contact motion
constraints --- do not pass through the origin. The arrangement of planes is
no longer \textit{central}. Thus, we cannot construct a dual polyhedron and
the above result does not hold. However, for the normal motion constraints we
are only concerned with the 3-dimensional regions and none of the lower
dimensional elements of the partition. Zaslavski's formula still applies and
provides an upper bound for the number of contact detachment states that is
cubic in the number of contacts even in the case of preloaded grasps.

\section{Contact state enumeration}\label{sec:enumeration}

We can now present a complete procedure for enumerating all possible
detachment states $U_k$ and slip states $S_k$ of an $m$-contact system that
are consistent with rigid body object motion. Again, we will only describe the
process for the slip states as the procedure for the detachment states is a
simplification (only the first step is required).

\mystep{Step 1.} We begin by enumerating all the slip states $S$ in
$\mathbb{S}$ corresponding to 3-faces in our plane arrangement. We achieve
this using Algorithm~\ref{alg:buildset}. Recall that all such $S \in
\{-1,1\}^m$. 

The algorithm operates by initializing $\mathbb{S}$ with just a single empty
slip state $\{\}$, which corresponds to a 3-face that includes all of
$\mathbb{R}^3$. We then one by one introduce the planes defined by the contact
slip expressions. Whenever we introduce a new plane we iterate through all
slip states in $\mathbb{S}$. At the beginning of the iteration we remove
slip state $S_k$ from $\mathbb{S}$ and check if its corresponding 3-face
intersects with the halfspaces defined by the new plane.

The existence of this intersection can be efficiently tested for by a linear
program: Use the elements in $S_k$ and the corresponding planes to construct
the linear inequalities that define its corresponding 3-face. Then add an
additional inequality constraint for the halfspace either above or below the
new plane. The feasibility of such a linear program tells us if a point exists
that lies in both the original 3-face and the halfspace defined by the new
plane. Thus, a 3-face and a halfspace intersect if the corresponding linear
program is feasible. We then add a new 3-face for each halfspace with which
$S_k$ intersects. The new 3-faces are the same as $S_k$ but we append a 1 or
-1 to indicate if it lies above or below the newly added plane. 

We note that for any state $S$ obtained by this algorithm, all contacts are
slipping, in either the positive or negative direction ($s_i^k=\pm 1$ for all
$i$.) We have not yet considered rolling contacts.

\begin{algorithm}[!t]
\caption{}\label{alg:buildset}
\begin{algorithmic}[0]
\State Initialize $\mathbb{S}$ with empty state $S_0 = \{\}$
\For{$i=1..m$}
  \For{$k=1..\#(S)$}
    \State Remove $S_k$ from $\mathbb{S}$
    \If{3-face $S_k$ intersects halfspace above plane $P_i$}
      \State Create 3-face $S_{k+} = \{S_k, 1\}$
      \State Add $S_{k+}$ to $\mathbb{S}$
    \EndIf
    \If{3-face $S_k$ intersects halfspace below plane $P_i$}
      \State Create 3-face $S_{k-} = \{S_k, -1\}$
      \State Add $S_{k-}$ to $\mathbb{S}$
    \EndIf
  \EndFor
\EndFor
\State\Return $\mathbb{S}$
\end{algorithmic}
\end{algorithm}

\mystep{Step 2.} Now we create the slip states corresponding to 2-faces in the
plane arrangements (one rolling contact). As mentioned before, these
correspond to edges of the dual polyhedron, so we begin this step by
constructing the dual polyhedron. We already have its vertices: each state $S$
created at the previous step defines a 3-face of the plane arrangement, and
thus corresponds to a vertex of the dual polyhedron. Then, for every two
states $S_k, S_l$ in $\mathbb{S}$ that differ by a single $s_i$, we add the
edge between them to the dual polyhedron (note that thus our polyhedron is
also a partial cube where edges connect any two vertices with Hamming distance
equal to 1~\cite{EPPSTEIN08}). Furthermore, we also create an additional state
$S_{kl}$ corresponding to the facet between $S_k$ and $S_l$. This will be
identical to both $S_k$ and $S_l$, with the difference that $s_i=0$ (the entry
corresponding to the plane that this 2-face lies in is set to 0).

\mystep{Step 3.} Next, we must compute the slip states corresponding to
1-faces in our plane arrangement (two rolling contacts), which correspond to
the faces of the dual polyhedron. We obtain the faces of our dual polyhedron
by computing the Minimum Cycle Basis (MCB) of the undirected graph defined by
its edges (computed at the previous step). This gives us $F-1$ of the faces of
our polyhedron; to see why consider that the number of cycles in the minimum
cycle basis is given by $E-V+1$~\cite{MEHLHORN05}. Recall the Euler-Poincar\'e
characteristic $\chi = V - E + F$ relating the number of vertices, edges and
faces of a manifold. For a convex polyhedron $\chi = 2$, and from this we can
derive the number of faces of our dual polyhedron to be equal to $E-V+2$. The
last face is obtained as the symmetric sum of all the cycles in the MCB
(defined as in~\cite{MEHLHORN05}). Once we have the cycles corresponding to
the faces of the dual polyhedron, we convert them into slip states as follows.
For each cycle, starting from the slip state $S_k$ corresponding to any of the
vertices in the cycle, we set $s_i=0$ for any plane $i$ that is traversed by
an edge in the cycle. 

\mystep{Step 4.} Finally, we must add the single 0-face of the plane
arrangement by adding a slip state $\{0\}^m$ to $\mathbb{S}$.

\vspace{2.5mm}

The total number of slip states $S$ we obtain is thus 1 greater than the
number of 3-faces, 2-faces and 1-faces of the plane arrangement, which is the
same as the number of vertices, edges and faces in its dual polyhedron. We
have already shown that this is polynomial (quadratic) in the number of planes
(contacts). We also show that the enumeration algorithm above has polynomial
runtime. We note that Step 1 has two nested loops, with one iterating over
planes and the other one over existing states. The number of states at the end
of this step is bounded by $m^2$, thus the running time of this Step is
$O(m^3)$. Step 2 must check every state against every other one, with $O(m^2)$
states, thus its running time is $O(m^4)$. Finally, the dominant part of Step
3 is the computation of the MCB. We have used an implementation with $O(E^3 +
VE^2\text{log} V)$ running time, where $V$ and $E$ are the number of vertices
and edges of the dual polyhedron. Since both $E$ and $V$ are polynomial in
$m$, the running time of the MCB algorithm is as well. Thus our complete
enumeration method has polynomial runtime in the number of contacts $m$.

Also note that the computation of contact detachment states is much simpler as
we only require the 3-faces of the arrangement of planes defined by the normal
contact motion constraints. Thus, only the first step of the above procedure
is required. In order to obtain the total set of possible combinations of
contact states we must combine the two sets $\mathbb{U}$ and $\mathbb{S}$ such
that for every combination of persisting contacts we superimpose the set of
all combinations of contact slip states. Note, that the slip state information
for a detached contact does not hold any value and hence in practice
Algorithm~\ref{alg:buildset} could be modified to consider both the sets of
planes due to normal and tangential motion constraints simultaneously only
recording the state combinations that are actually useful. Of course even
without this optimization the total number of contact state combinations to be
considered remains polynomial in the number of contacts.

\section{Complete stability determination procedure}

We can now formalize our complete algorithm using the components outlined so
far (Algorithm~\ref{alg:complete}). We first build the total set of possible
contact states $\mathbb{T}=\mathbb{U}\times\mathbb{S}$. Then, for every $T_k
\in \mathbb{T}$, we check for a solution to the system as described in
Chapter~\ref{sec:2d_solving}. If one exists, we deem the grasp stable. If,
after enumerating all possible $T_k$, we do not find one that admits a
solution, we deem the grasp unstable. We make two important observations
regarding Algorithm~\ref{alg:complete}. First, its running time is polynomial
in the number $m$ of contacts. This follows from the results obtained so far:
We know that $\#(\mathbb{T})$ is polynomial in $m$, as is the process for
building it. For each $T_k$, we then solve at most a linear program with
$2m+3$ unknowns, which also has a polynomial runtime, which completes this
result.

Second, Algorithm~\ref{alg:complete} guarantees that, if no solution is found,
none exists that satisfies the constraints of our system. $\mathbb{T}$
provably contains all the contact states consistent with rigid body motion;
for each of these, equilibrium and contact conditions form a linear program
for which we can provably find all solutions (if they exist). So, under the
assumed formulation (virtual springs used to determine contact normal forces,
and frictional constraints including the maximum dissipation principle), if a
solution exists to the equilibrium problem, we must find it.

\begin{algorithm}[!t]
\caption{}\label{alg:complete}
\begin{algorithmic}[0]
\State Build $\mathbb{T}=\mathbb{U}\times\mathbb{S}$, the set of all possible contact states
\For{$k=1..\#(\mathbb{T})$}
\State Given $T_k$, solve system (\ref{eq:linear_system})
\If{solution found}
\State \Return grasp stable
\EndIf
\EndFor
\State \Return grasp unstable
\end{algorithmic}
\end{algorithm}

\section{Applications to planar grasp stability analysis}

\begin{figure}[!t]
\centering
\includegraphics[width=0.85\linewidth]{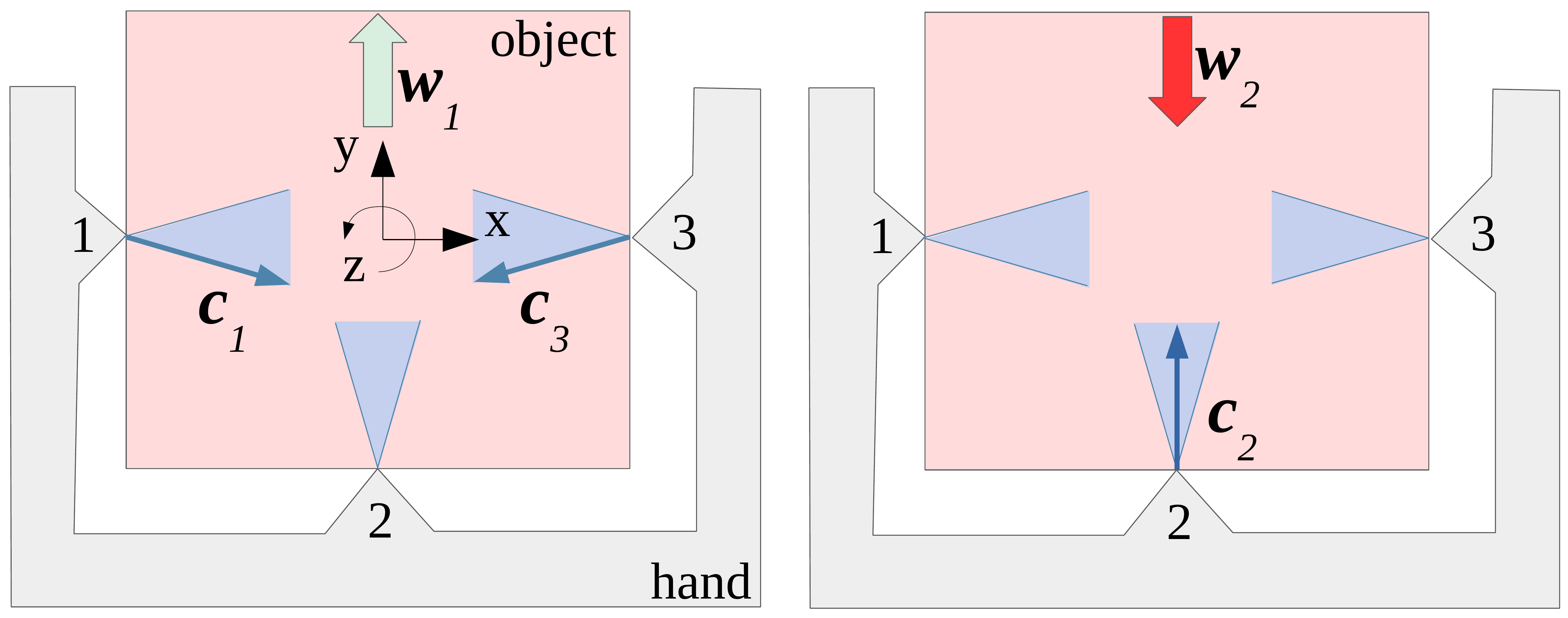} 

\caption{A grasping scenario where a rigid hand establishes multiple
frictional contacts (numbered 1-3) with a target object. External disturbance
$\bm{w}_1$ (left, pushing the object up) can be resisted by contact forces
$\bm{c}_1$ and $\bm{c}_3$, but only if contacts 1 and 3 have been actively
pre-loaded with enough normal force to generate the corresponding friction
forces. In contrast, disturbance $\bm{w}_2$ (right, pushing the object down),
regardless of its magnitude, will always be passively resisted by contact
force $\bm{c}_2$.}
  
\label{fig:grasp2}
\end{figure}

In this section we will demonstrate that our framework predicts the correct
force distributions and makes an accurate prediction on grasp stability. We
will utilize the grasp shown in Fig.~\ref{fig:grasp2} because the correct force
distribution and stability of the grasp is easily understood intuitively.
Specifically, we would like to discriminate which applied wrenches will be
balanced purely passively, and where an active preload of the grasp is
required. We assume the friction coefficient is 0.5 across all contacts such
that $\mu_i=0.5,~i=0..m$.

\begin{figure}[!t]
  \centering
  \includegraphics[width=0.45\linewidth]{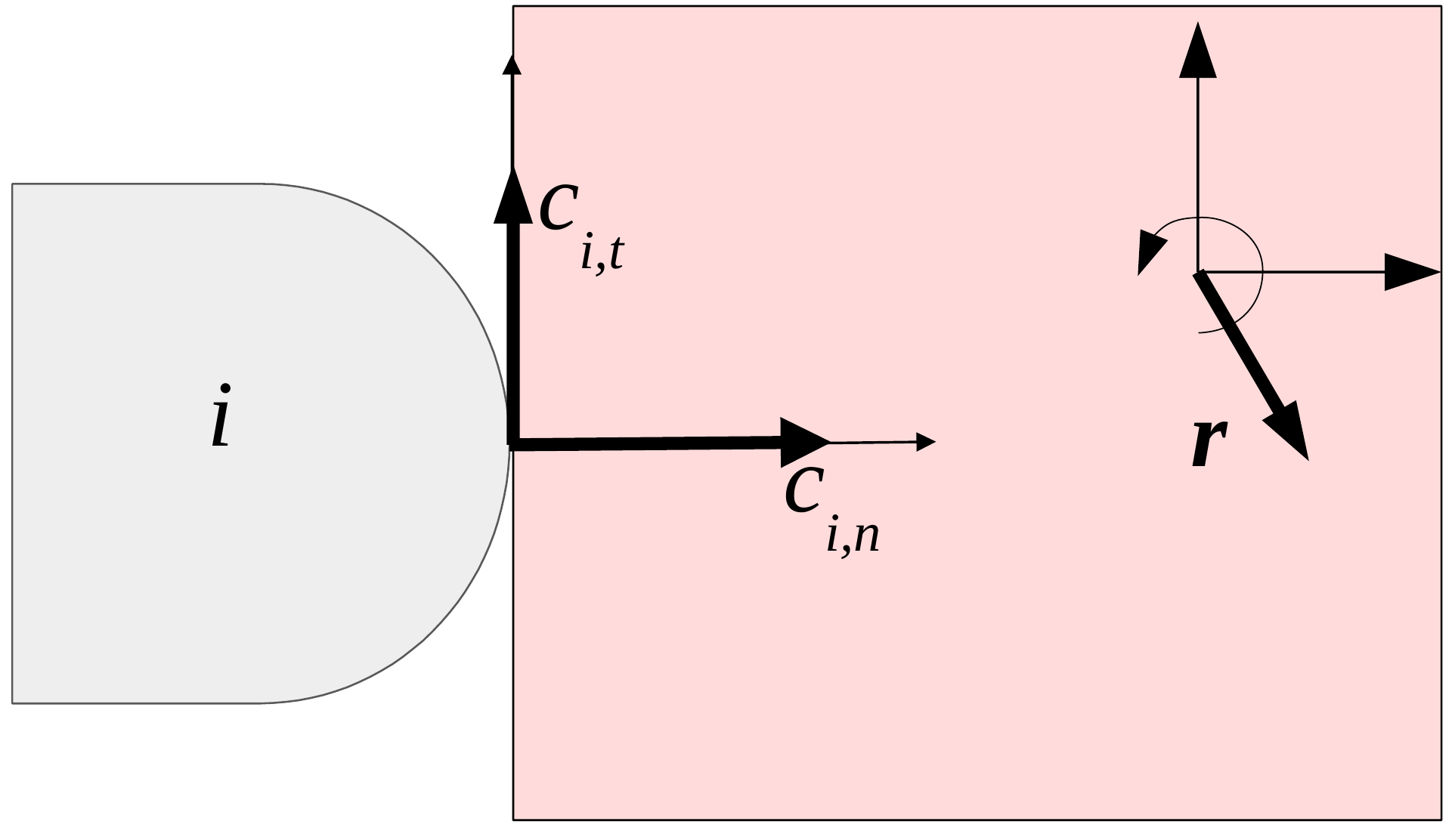} 
  \caption{Illustration of the contact coordinate frame. $\bm{r}$ 
  denotes object motion expressed in the object coordinate frame.}
  \label{fig:detail}
\end{figure}

\begin{figure}[!t]
\centering

\subfigure[Resistible forces with no preload (our algorithm)]
{\label{fig:bin_no_preload}\includegraphics[width=0.45\linewidth]{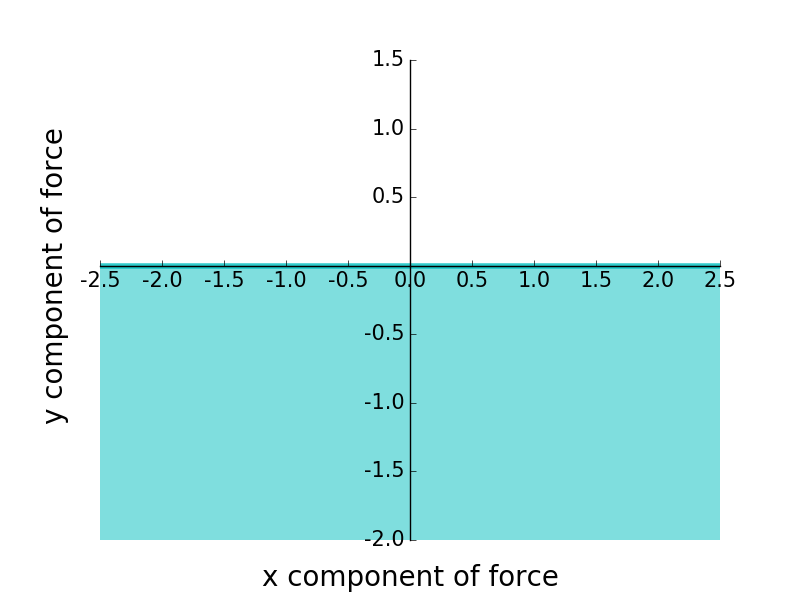}}

\subfigure[Resistible forces with a preload such that the normal force at each contact is 1 (our algorithm)]
{\label{fig:bin_preload}\includegraphics[width=0.45\linewidth]{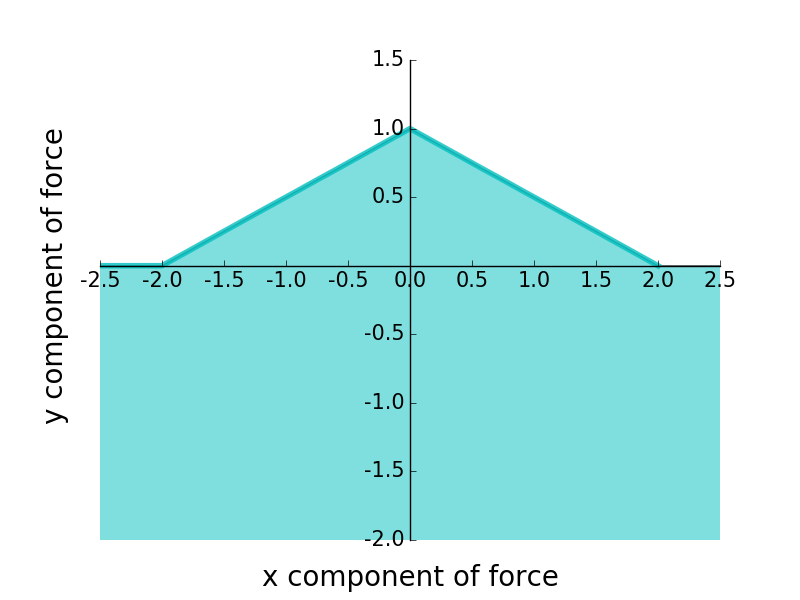}}

\subfigure[Slice through the three dimensional GWS for zero applied torque]
{\label{fig:GWS}\includegraphics[width=0.45\linewidth]{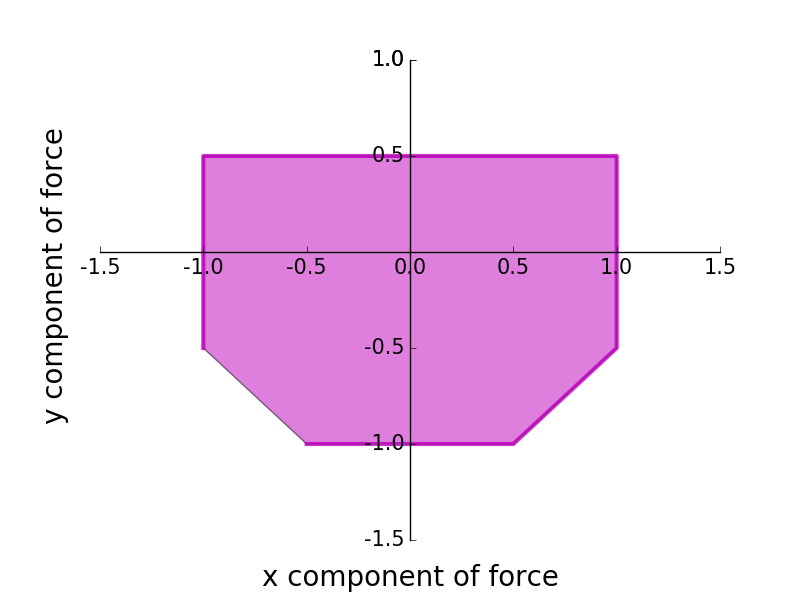}}

\caption{Grasp stability representations for the grasp in Fig.
\ref{fig:grasp2}. Our algorithm (a)(b) captures passive resistance to applied
forces of arbitrary magnitude in directions, that allow for balancing contact
forces to arise passively in response to the disturbance. The GWS
representation (c) shows the space of applied forces the grasp can resist with
contact forces that satisfy only friction constraints. We are using the L1
norm GWS, meaning the normal components of the contact forces sum to 1.}
  
\end{figure}

Recall that there exist contact forces in the interior of the friction cones
that balance both wrenches shown in Fig. \ref{fig:grasp2}. Perhaps the most
commonly used approach to grasp stability analysis is the Grasp Wrench Space
method~\cite{FERRARI92}. Indeed, when we consider the slice through the GWS
visualized in Fig. \ref{fig:GWS} we can see that there exist contact forces
that balance arbitrary forces in the plane. However, we argue that while
$\bm{w}_2$ will always be reacted passively, no matter the preload, in order
to react $\bm{w}_1$ we require the grasp to have been sufficiently preloaded.
The GWS method correctly indicates the existence of equilibrium contact forces
but does not predict if they may arise, and hence does not capture the
necessity of a preload. Thus, in this context force closure is not a reliable
indicator of stability.

Now let us apply our framework to this problem: Using our algorithm, we can
test the resistance of this grasp to forces in the plane. We do this by
discretizing the direction of application of force to the object and finding
the maximum resistible force in each direction using a binary search. Figs.
\ref{fig:bin_no_preload} \& \ref{fig:bin_preload} show the region of
resistible forces for a grasp without and with a preload respectively. As our
algorithm takes into account passive effects it correctly predicts that, in
both cases, forces with non-positive component in the y-axis and arbitrary
magnitude can be withstood. Indeed, without preloading the grasp for any
applied force $\bm{w}=(0,w_y,0), w_y \leq 0$ our framework predicts contact
forces $(0,0)$ at contacts 1 and 3, and contact force $(-w_y,0)$ at contact 2
(see Table \ref{tab:three_contact}). Furthermore, it captures the necessity of
a preload in order to resist forces with positive component in the
y-direction: For $w_y > 0$ our algorithm finds no solution, and hence the
grasp must be unstable to this disturbance, unless an appropriate preload is
applied.

We can compare our results to those obtained with compliance based approaches
\cite{CUTKOSKY_COMPLIANCE}\hspace{0pt}\cite{BICCHI93}\hspace{0pt}\cite{BICCHI94}\hspace{0pt}\cite{BICCHI95}\hspace{0pt}\cite{PRATTICHIZZO97},
which are commonly used to predict contact forces that arise in grasps due to
disturbances. We can now use this grasp (Fig. \ref{fig:grasp2}) and our
algorithm to show that even with the improvements to a linear compliance
suggested by Prattichizzo et al.~\cite{PRATTICHIZZO97} resulting stability
estimates are overly conservative.  Their approach allows each contact to be
in one of three states: rolling, slipping or detached. A slipping contact may
not apply any frictional forces, while a detached contact may not apply any
force at all. If we try every possible combination of states and modify the
compliance of the grasp accordingly, this alleviates some of the problems of
the purely linear compliance approach: If we consider contacts 1 \& 3 to be
slipping, our grasp in Fig. \ref{fig:grasp2} may now withstand arbitrary
forces, where $w_y \leq 0$.

This approach, however, does not allow us to arrive at the correct result in
cases where $w_y \geq 0$. Consider the preloaded grasp before the application
of an external wrench (Table \ref{tab:three_contact}.) The contact forces on
both contacts 1 \& 3 must lie on the friction cone edge in order to balance
the preload applied by contact 2. If we now apply an external wrench
$\bm{w}=(0,1,0)$, there exists no combination of rolling, slipping and
detached contacts (and corresponding modifications of the linear compliance)
that results in legal contact forces. Our algorithm, however, predicts a
stable grasp (Table \ref{tab:three_contact}), showing how important friction
is for grasp stability and why a correct treatment of friction is fundamental
to stability analysis. Furthermore, we have arrived at this result in
polynomial time --- we did not have to consider exponentially many slip
states, as in~\cite{PRATTICHIZZO97}.

\begin{table}[!t]
\centering
\begin{tabular}{l|l|l|lll|l}
$\bm{w}$ & $P$ & Stable & $\bm{c}_1$ & $\bm{c}_2$ & $\bm{c}_3$ & $\bm{r}$ \\
\hline
$(0, 0, 0)$ & 0 & Yes & $(0,0)$ & $(0,0)$ & $(0,0)$ & $(0,0,0)$ \\
$(0, 0, 0)$ & 1 & Yes & $(1,-0.5)$ & $(1,0)$ & $(1,0.5)$ & $(0,0,0)$ \\
$(0, -1, 0)$ & 0 & Yes & $(0,0)$ & $(1,0)$ & $(0,0)$ & $(0,-1,0)$ \\
$(0, -2, 0)$ & 0 & Yes & $(0,0)$ & $(2,0)$ & $(0,0)$ & $(0,-2,0)$ \\
$(0, 1, 0)$ & 0 & No & --- & --- & --- & --- \\
$(0, 1, 0)$ & 1 & Yes & $(1,-0.5)$ & $(0,0)$ & $(1,0.5)$ & $(0,1,0)$ \\
$(0, 1.1, 0)$ & 1 & No & --- & --- & --- & --- \\
\end{tabular}

\caption{Contact forces $\bm{c}_i=(c_{i,n},c_{i,t})$ and (virtual) object
motion $\bm{r}=(x,y,r)$ for the grasp in Fig. \ref{fig:grasp2} and a range of
applied wrenches $\bm{w}=(w_x,w_y,w_z)$. The preload $P$ is such that the
normal force at each contact is equal to either 0 or 1 before any wrench is
applied. The object motion and applied wrenches are expressed in the
coordinate frame shown in Fig. \ref{fig:grasp2} and contact forces are
expressed in frames as shown in Fig. \ref{fig:detail}.}

\label{tab:three_contact}
\end{table}

\begin{table}[!t]
\centering
\begin{tabular}{l|lllllllll}
$m$              & 2 & 3 & 4 & 5 & 6 & 7 & 8 & 9 \\
\hline
$\#(\mathbb{S})$ & 10 & 26 & 50 & 82 & 122 & 170 & 226 & 290 \\
\hline
$time (s)$       & 0.006 & 0.02 & 0.09 & 0.42 & 1.5 & 4.8 & 13.8 & 35.7 \\
\end{tabular}

\caption{Number of slip states $\#(\mathbb{S})$ and computation time for
grasps with $m$ randomly generated contacts.}

\label{tab:performance}
\end{table}

From a computational effort perspective, a summary of the performance of our
algorithm for enumeration of slip states can be found in Table
\ref{tab:performance}. All computation was performed on a commodity computer
with a 2.80GHz Inter Core i7 processor.

\section{Discussion}\label{sec:2d_furtherwork}

In this Chapter we have described an algorithm that can test the stability of
a rigid planar grasp when a given disturbance wrench is applied to the grasped
object. We propose modeling the contact normal forces through constitutive
relations while maintaining accurate friction constraints. In order to apply
this model we must distinguish between rolling, slipping and detaching
contacts. We have shown that the number of combinations of contact states that
are possible under rigid body motions is quadratic in the number of contacts
and provide a polynomial time algorithm to enumerate all such combinations.

As has been noted in Chapter~\ref{sec:state_number} the insight that the
number of contact slip state combinations between two rigid bodies in contact
is quadratic in the number of contacts is not
new~\cite{Brost1989a}\cite{MASON_MANIPULATION}. However, we have shown that
this remains valid when introducing compliance at the contact normals, as is
required to efficiently test for stability. Our novel algorithm for the
polynomial-time enumeration of contact state combinations also enables the
analysis of grasps that are preloaded before the application of an external
wrench.

A limitation of the work presented above is that we defined a grasp preload in
terms of the contact normal forces present before the application of an
external wrench to the object. In practice, preloads are achieved by setting
actuator commands. Thus, defining preloads in terms of joint torques for
instance is of perhaps greater practical use as it would allow us to use our
framework as a tool in computing appropriate actuator commands for a given
task. 

The required treatment of the kinematics of the hand means we must take into
account not only the motion of the grasped object but also of the links making
up the hand. Our algorithm can be extended to such multi-body systems, which
would allow us to study the stability of multiple objects or objects grasped
by hands with multiple links and torques applied at the joints. The
constraints that describe such systems involve the motion of multiple objects
and thus more motion variables than the three dimensional object motion we
consider in the above treatment. Therefore the planes describing the
constraints become higher dimensional hyperplanes. However, much of the 
theory in Chapter~\ref{sec:state_number}, specifically Zaslavski's formula
remains valid for such arrangements of hyperplanes:

For a given number of objects the number of contact state combinations remain
polynomial in the number of contacts. However, the number of contact state
combinations is exponential in the number of bodies. The enumeration of the
faces of the arrangement that correspond to sets of contact conditions is more
involved but arrangement of hyperplanes are well studied in the field of
computational geometry and efficient algorithms are available for the solution
of such problems~\cite{10.5555/2408018}.

The exponential growth of the number of possible contact state combinations
with the number of bodies involved is a well known problem in the rigid body
dynamics community. In fact, many researchers have independently shown that
the underlying complementarity problems are NP-hard (see
Chapter~\ref{sec:related_dynamics}.) A further complication of these
complementarity approaches is that, in order to determine stability, one has
to show that \textit{every} solution to the dynamics equations is stable. Due
to the possibility of multiple solutions to the complementarity problem with
different contact state combinations this means that unfortunately the only
way to ensure stability is to check all possible combinations. 

An approach to multi rigid body problems that instead uses our normal force
compliance model would suffer from the same worst-case exponential complexity
due to an exponential number of contact state combinations. However, in our
model finding a single solution is sufficient to conclude grasp stability and
thus the computational burden is reversed: You only have to check all
combinations to be able to conclude that a grasp is \textit{not} stable by
checking that none of them have a solution.

Finally, we want to discuss the perhaps most significant limitation in that
this work is only applicable to planar grasps. Unfortunately the key insight
that allowed us to formulate an efficient solution to passive stability
problems does not directly translate to spacial grasps: In 3D the directions a
contact can slip in are no longer discrete and finite. A contact can slide in
an infinite number of directions and hence we cannot break the treatment of
sliding contacts into separate piecewise linear problems as we did above.

However, we were pleased to see that Huang et al.~\cite{Huang2020} adopted an
approach based on arrangements of planes and developed an algorithm for the
enumeration of slip states in three dimensions. They approach the problem of
infinite sliding directions by discretizing the tangent plane into sectors of
equal angles. Thus, sliding states are defined in terms of which sector
contains the tangent velocity.

Despite these limitations, our treatment of the special planar case has proven
to be a valuable first step towards modeling the passive characteristics of
general grasps, which we will investigate in the following chapters.





\chapter{Grasp Stability Analysis in Three Dimensions}\label{sec:three_dim}

\section{Introduction}

In Chapter~\ref{sec:bridge} we described the importance of modeling passive
effects for robotic manipulation and began our investigation with the planar
case in Chapter~\ref{sec:2d}. Of course most robotic grasps are not planar:
The fingers and grasped objects do not in general inhabit the same plane.
Furthermore, even grasps that may look to be planar at first (such as the
grasp in Fig.~\ref{fig:package}) require modeling of all three spacial
dimensions in the presence of out-of-plane forces or moments.

Thus, in order to further our understanding of the passive effects at play in
the grasps commonly encountered in practice we must extend our work from
Chapter~\ref{sec:2d} to spacial grasps. Unfortunately, our treatment of planar
grasps does not easily generalize. In order to enforce the maximum dissipation
principle (friction opposes motion) our framework relied on the fact that in
the plane a contact can only slide in one of two directions. In three
dimensions contacts can slide in an infinite number of directions and thus we
cannot enumerate a finite number of contact states as we did in
Chapter~\ref{sec:2d}.

This means that enforcing the principle of maximum dissipation is more
involved in three dimensions. However, as we will see an accurate treatment of
the friction constraints including the maximum dissipation principle is of
paramount importance for the modeling of passive effects. Simply dropping the
maximum dissipation principle allows for unphysical solutions and thus we
cannot determine stability. This may be somewhat surprising as we are only
trying to solve for the contact wrenches of stable grasps where hence all
motions are minuscule. We will present a physically motivated method for
alleviating the consequences of dropping the MDP such that only physically
realistic solutions are obtained.

Even though our formulation for planar grasps does not directly generalize to
three dimensions, the insights gained in the previous chapter will prove to be
useful also in the three dimensional case. Specifically, we already know that
we can use a compliance model to capture passive effects.

In order to develop a grasp model that is truly general and applicable to the
vast majority of robotic hands currently in use we also choose to take into
account effects previously omitted in Chapter~\ref{sec:2d}. We will no longer
assume the hand to be a rigid fixture but instead take into account the
kinematics of the hand as well as the commanded actuator torques. This allows
us to showcase the use of our framework as a practical tool. 

\section{Grasp model}\label{sec:model3d}

The model we choose to describe a grasp in three dimensions is similar to the
planar model introduced in Section~\ref{sec:2dmodel} and we will make the same
assumptions that initially the bodies are at rest and all motions (in this
context displacements) are small such that the grasp geometry is constant.

We consider a hand making $m$ contacts with a grasped object where we again
use the \textit{Point Contact with Friction} model to describe the contact
forces. We will again denote normal and tangential components of contact
specific vectors with subscripts $n$ and $t$ respectively. However, in three
dimensions the tangential component of such a vector is no longer a scalar but
a vector itself: At a contact $i$ the contact force $\bm{c}_i \in
\mathbb{R}^3$ has normal and tangential components $c_{i,n} \in \mathbb{R}$
and $\bm{c}_{i,t} \in \mathbb{R}^2$ respectively. 

\vspace{2mm} \noindent \textbf{Equilibrium:} In three dimensions the grasp map
matrix also has different dimension ($\bm{G} \in \mathbb{R}^{6\times3m}$). One
addition to the grasp model is that we will also consider the kinematics of
the hand and must therefore also take into account not only object but also
hand equilibrium. For a hand with $l$ joints the transpose of the hand
Jacobian $\bm{J} \in \mathbb{R}^{3m \times l}$ maps contact forces to joint
torques $\bm{\tau} \in \mathbb{R}^l$.

\begin{alignat}{2}
\bm{G}\bm{c} &+ \bm{w} &~= 0 \label{eq:object_eq}\\
\bm{J}^T \bm{c} &+ \bm{\tau} &~= 0 \label{eq:hand_eq}
\end{alignat} 

These equilibrium equations can predict the resultant wrench on the object and
the joint torques that arise in response to given contact forces, but, since
neither $\bm{G}$ nor $\bm{J}^T$ are typically invertible in practice, it has
no predictive capabilities in the opposite directions: we cannot use it to
predict the contact response to known external wrenches, or joint torques. It
thus fails to capture effects where contact forces are transmitted (either
amongst the joints or between the joints and the external environment) through
the object itself.

We could use one of the many methods discussed in
Section~\ref{sec:related_closure} to find a set of contact wrenches
$\bm{c}_{eq}$ that balance a given applied wrench and are achievable with the
kinematic and actuation capabilities of the hand. However, in the vast
majority of cases in applied robotics, this simple method has major
shortcomings. It requires that we know $\bm{w}$ exactly at any point in time,
which is unrealistic in all but the most controlled environments. Furthermore,
most robotic hands are position controlled and hence lack the torque
regulating capabilities necessary to accurately command the actuators for real
time grasp control.

The approach actually used in the overwhelming majority of robotic grasping
tasks is to command a given set of joint torques $\bm{\tau}^c$ and let the
fingers jam around the object. We keep torque commands constant throughout the
task (i.e. we take no further action) and rely on appropriate equilibrium
contact forces $\bm{c}$ and equilibrium torques $\bm{\tau}^{eq}$ to arise
through \textit{passive reactions}. This simple strategy is effective for
hands powered by highly geared motors due to the non-backdrivable nature of
the actuators: At any joint $j$ the equilibrium torque $\bm{\tau}_{j}^{eq}$
can exceed the commanded value $\bm{\tau}_j^c$ in a passive response to
torques $\tau_k^c,~k \neq j$ or the applied wrench $\bm{w}$. An intuitive
example of this phenomenon might be an object being pressed against a finger
by an external wrench. If the wrench is large enough to overwhelm the
commanded joint torques then any additional torques at the finger joints are
provided passively by the gearbox such that the finger remains fixed.

Much the same as in Chapter~\ref{sec:2d} we must introduce the notion of
compliance to our grasp model in order to resolve the structural
indeterminacy. Therefore the motion of the bodies making up the grasp become
part of the solution we are seeking. This is much the same as was previously
presented in Section~\ref{sec:2dmodel}. We denote object motion as $\bm{r} \in
\mathbb{R}^6$, however now the joints of the hand and therefore the hand links
may also exhibit motion denoted by a change in joint angles $\bm{q} \in
\mathbb{R}^l$. 

We have already seen in Section~\ref{sec:2dmodel} how contact motion due to
the object moving allowed us to capture the unilateral contact behavior.
However, we must now argue about \textit{relative} motion at the contacts
$\bm{d} \in \mathbb{R}^{3m}$, as both the object and the contacting link may
move. Recall that in the context of our grasp model by motion we mean the
displacements of the bodies involved and this relation is only valid for small
displacements.

\begin{equation}
\bm{G}^T\bm{r} - \bm{J}\bm{q} = \bm{d} \label{eq:rel_contact_motion}
\end{equation}

Again, as we assume a constant grasp geometry matrices $\bm{G}$ and $\bm{J}$
are also constant. We can now reason about the contact forces and joint
torques in terms of the motion of the object and joints.

\vspace{2mm} \noindent \textbf{Normal forces.} In Chapter~\ref{sec:2dmodel} we
discussed a compliance formulation of the unilateral contact constraint. We
will use the same constitutive equations for the contact normal forces
obtained by placing virtual linear springs of unit stiffness along the contact
normals.

\begin{subequations}
\begin{empheq}[left=\empheqlbrace] {align}
c_{i,n} &= -d_{i,n} &\text{if } d_{i,n} \leq 0 \\
c_{i,n} &= 0 &\text{if } d_{i,n} > 0
\end{empheq}\label{eq:unilaterality_again}
\end{subequations}

In Chapter~\ref{sec:2dmodel} we also included a normal force term for
\textit{preloaded} grasps. We can ignore this term here as we are now also
explicitly modeling the hand kinematics. This allows us to define a preload in
terms of joint torques instead of normal forces, which is how preloads are
applied in practice. These preload joint torques then propagate through the
links making up the hand as well as the object according to the equilibrium
equations (\ref{eq:object_eq}) \& (\ref{eq:hand_eq}) and thus create the
preload contact forces present before the application of any external forces.

\vspace{2mm} \noindent \textbf{Friction forces:} Again, we choose the Coulomb
model to describe the friction forces. The first part of the Coulomb model
provides an upper bound to the magnitude of the friction force given the
normal force and the friction coefficient $\mu_i$. As we are investigating the
static equilibrium of a grasp we consider all friction coefficients to be
those of static friction. This defines a cone $\mathcal{F}_i$ at each contact.

\begin{equation}
\mathcal{F}_i( \mu_i, c_{i,n} ) 
= \{ \bm{c}_{i,t} : \left\lVert \bm{c}_{i,t} \right\rVert \leq \mu_i c_{i,n} \}
\label{eq:friction_bound}
\end{equation}

The second part of the Coulomb model concerns the \textit{Maximum Dissipation
Principle}~\cite{Moreau2011}. Now we will show how the MPD results in friction
forces that oppose the relative motion at the contact. The MDP states that
given a relative contact motion the friction force at that contact must
maximize the dissipation where the friction force is bounded by
(\ref{eq:friction_bound}).

\begin{equation}
\bm{c}_{i,t} \in \argmin_{\bm{c}_{i,t} \in \mathcal{F}_i} \bm{c}^T_{i,t} \cdot \bm{d}_{i,t}
\label{eq:mdp}
\end{equation}

\vspace{2mm} \noindent \textbf{Joint torques:} Finally, we model the joints as
nonbackdrivable, in order to capture the behavior of robotic hands driven by
highly geared motors. This means that a joint $j$ may only exhibit motion in
the direction that its commanded torque is driving it in. As we know the
commanded joint torques we can define motion in such a direction as positive.

\begin{equation}
q_j \geq 0
\label{eq:joint_unilaterality}
\end{equation}
A joint with zero commanded torque may not move, as any torque arising from
external factors will be absorbed by the gearing. 

The joint torque may exceed the commanded level, but only if this arises
passively. This means that a joint that is being passively loaded beyond the
commanded torque levels must be locked in place and may not move. A moving
joint must apply the torque it was commanded to. Thus, we must also
distinguish between two types of joints. 

\begin{subequations}
\begin{empheq}[left=\empheqlbrace]{align}
\tau_j &\geq \tau_{j}^c & \text{if}~q_j = 0\\
\tau_j &= \tau_{j}^c & \text{if}~q_j > 0
\end{empheq}\label{eq:joint_model}
\end{subequations}

So far this joint model has assumed a the kinematics of a fully actuated
direct drive robotic hand, where an actuator command equates to an individual
joint torque command. However, a more general formulation allows modeling hand
kinematics, where the joint torques can be expressed as linear combinations of
actuator forces. This includes underactuated designs with fewer actuators than
degrees of freedom such as, for example, a tendon driven hand with fewer
tendons than joints. This implies, that a tendon --- and hence an actuator ---
can directly apply torques to multiple joints by means of a mechanical
transmission. 

We thus define matrix $\bm{R}$, which maps from forces or torques at the
actuators $\bm{f}$ to joint torques $\bm{\tau}$. Note, that its transpose maps
from joint motion to the motion of the mechanical force transmission at the
actuator, which we assume to be nonbackdrivable. Inequality constraints on a
vector are to be understood in a piecewise fashion.

\begin{eqnarray}
\bm{R} \bm{f} &=& \bm{\tau} \label{eq:torqueRatio}\\
\bm{R}^T \bm{q} &\geq& 0 \label{eq:underac_nonbackdrive}
\end{eqnarray}

Hence, at an actuator $l$ we may see a force $f_l$ that exceeds the commanded
value $f_{l}^c$ --- and again, this can only occur passively. This means, that
an actuator force can only exceed the commanded value, if there is an attempt
to backdrive the actuator, and hence the mechanical transmission does not
exhibit any motion. 

\begin{subequations}
\begin{empheq}[left=\empheqlbrace]{align}
f_l &\geq f_{l}^c & \text{if } (\bm{R}^T \bm{q})_l = 0\\
f_l &= f_{l}^c & \text{if } (\bm{R}^T \bm{q})_l \geq 0
\end{empheq}
\label{eq:underactuation}
\end{subequations}

This actuation model is a generalization of the previously introduced joint
model and will reduce as such if the actuators control individual joint
torques directly.

\vspace{2mm} \noindent \textbf{Complete problem:}  The system comprising
(\ref{eq:object_eq}) -- (\ref{eq:underactuation}) defines the static
equilibrium formulation for an arrangement of bodies; some fixed, some free,
some constrained by unilateral or bilateral constraints. Some bodies may be
acted upon by actuators or externally applied wrenches. The formulation
satisfies all the requirements for determining stability including passive
effects due to nonbackdrivability and underactuation but is also applicable to
many other problems with arrangements of bodies in frictional contact. 

\vspace{2mm} \noindent \textbf{Definition of stability:} In the following we
will make use of the same definition previously introduced in
Chapter~\ref{sec:definition}: For a given grasp geometry, actuator torques and
applied wrench a grasp is deemed to be stable if a solution to the above
system of constraints exists. However, note that the above formulation is very
general in nature, and can not only be used in such \textit{existence
problems} (e.g. given $\bm{f}^c$, determine if $\bm{r}$ and $\bm{c}$ exist
that balance a given $\bm{w}$). With the addition of an objective we can also
apply it in \textit{optimization problems} (e.g. determine the optimal
$\bm{f}^c$ that satisfies the existence problem above).

Remaining agnostic to the exact query that is being solved we will refer to
the \textit{exact problem} as the following query: given a subset of $\bm{f}$,
$\bm{r}$, $\bm{c}$ or $\bm{w}$, determine the rest of these variables
such that (\ref{eq:object_eq}) -- (\ref{eq:underactuation}) are exactly
satisfied (i.e. the grasp is stable).

\section{Formulation as a Mixed Integer Program}

\subsection{Unilaterality}\label{sec:unilaterality}

Let us take a closer look at the types of constraints that make up the exact
problem. The equilibrium equations in (\ref{eq:object_eq}) \&
(\ref{eq:hand_eq}) as well as the kinematics (\ref{eq:rel_contact_motion}) \&
(\ref{eq:torqueRatio}) are linear equality constraints and require no further
attention. The unilaterality constraints in (\ref{eq:unilaterality_again}) \&
(\ref{eq:underactuation}) consist of pairs of linear equalities and
inequalities but exhibit a combinatorial nature: Only one of the constraints
is active at a time, the choice of which is unknown a priori.

In Chapter~\ref{sec:2d} we dealt with constraints of this nature by separately
evaluating the problems arising from each choice of constraint combinations.
As we were only interested in the \textit{existence problem}, each individual
problem involved solving at most a linear program. Therefore this approach
could be efficiently leveraged to solve problems involving such combinatorial
constraints. As we want to develop a formulation that can also be applied to
the more complex \textit{optimization problems} we choose a Mixed Integer
Programming (MIP) approach instead. Both the normal force and the joint model
relationships (\ref{eq:unilaterality_again}) \& (\ref{eq:underactuation}) can
be cast as pairs of convex constraints with binary decision variables in an
MIP that can then be solved using algorithms such as branch and bound.

The perhaps simplest implementation of such constraints is the 'Big M method'.
For instance we can reformulate (\ref{eq:unilaterality_again}) as four linear
inequality constraints with the addition of a binary decision variable $y_i
\in \{0,1\}$ and two large constants $M_{i,1}$ and $M_{i,2}$, which give the
method its name. The binary variable can be interpreted as the variable
deciding if a contact rolls ($y_i = 1$) or breaks ($y_i = 0$).

\begin{eqnarray}
c_{i,n} \geq& 0 \\
c_{i,n} + d_{i,n} \geq& 0 \\
c_{i,n} \leq& M_{i,1} \cdot y_i \\
c_{i,n} + d_{i,n} \leq& M_{i,2} \cdot (1 - y_i)
\end{eqnarray}

The two constants must be chosen to be larger than the largest expected value
of the left hand side of their respective inequality constraint. However,
choosing too large a value can lead to numerical issues. An example of such a
numerical issue is 'trickle flow', in which a small violation of the
integrality constraint can lead to a violation of the inequality. Numerical
optimization tools must work within finite tolerances and thus small
violations are unavoidable. Specifically in our example a $y_i$ that is
slightly larger than zero allows for nonzero contact normal force when
multiplied with a large $M_{i,1}$ even at a breaking contact.

An alternative approach is to use indicator constraints, which avoids these
issues. These constraints allow a Mixed Integer solvers to directly branch on
the constraint choices or derive tight 'big M' values during
preprocessing~\cite{gurobi}.

\begin{subequations}
\begin{empheq}[left=\empheqlbrace]{align}
y_i &= 1 &\implies& &c_{i,n} &= -d_{i,n}~,&~d_{i,n} \leq 0 \\
y_i &= 0 &\implies& &c_{i,n} &= 0~,&~d_{i,n} > 0
\end{empheq}
\label{eq:binary_contact}
\end{subequations}

A potential downside is that indicator constraints tend to have weaker
relaxations and thus may lead to larger MIP optimization times. In practice we
have found that this disadvantage is more than compensated for by their ease
of use as well as numerical stability when compared to the big M method. Thus,
going forward we will use indicator constraints in order to deal with
conditional constraints.

Of course we can also use indicator constraints to model the unilaterality of
nonbackdrivable actuator constraints (\ref{eq:underactuation}) by introducing
binary variable $z_l \in \{0,1\}$, which indicates if the actuator is free to
move ($z_l = 0$) or being overpowered by external forces ($z_l = 1$).

\begin{subequations}
\begin{empheq}[left=\empheqlbrace]{align}
z_l &= 1 &\implies& &f_l &\geq f_{l}^c~,&~(\bm{R}^T \bm{q})_l = 0 \\
z_l &= 0 &\implies& &f_l &= f_{l}^c~,&~(\bm{R}^T \bm{q})_l > 0
\end{empheq}
\label{eq:binary_actuator}
\end{subequations}

\subsection{Friction}\label{sec:3dfriction}

The Coulomb friction law in (\ref{eq:friction_bound}) \& (\ref{eq:mdp})
provides more complexity due to its non-convexity. This is readily apparent as
the the bilinear form in (\ref{eq:mdp}) is not positive
definite~\cite{LIBERTI04}. It has been pointed out that solving for
$\bm{c}_{i,t}$ given $\bm{d}_{i,t}$ or vice versa is a convex problem and can
be done efficiently~\cite{KAUFMANN08}. Solving for both simultaneously is much
harder --- in fact the global optimization of non-convex quadratic programs is
in general NP-hard.

For our purposes it will be useful to reformulate the Coulomb friction
constraints. Note, that in the case of isotropic friction dissipation is
maximized if the friction force is anti-parallel to the relative sliding
motion and lies on the boundary of the cone $\mathcal{F}_i$. Thus, we can also
directly express the friction force in terms of the normal force and the
relative sliding motion. We must distinguish between two cases: 

\begin{itemize}

\item At a rolling contact that does not exhibit relative motion in a
tangential direction (sliding) the friction force is constrained such that the
contact force lies within the cone $\mathcal{F}_i$

\item If a contact does exhibit sliding, the friction force must oppose the
relative direction of motion, and the total contact force must lie on the
friction cone edge. 

\end{itemize}

\begin{subequations}
\begin{empheq}[left=\empheqlbrace]{align}
\left\lVert \bm{c}_{i,t} \right\rVert &\leq \mu_i c_{i,n} & \text{if } \left\lVert \bm{d}_{i,t} \right\rVert = 0 \label{eq:frictioncone} \\
\bm{c}_{i,t} &= - \mu_i c_{i,n} \frac{\bm{d}_{i,t}}{\left\lVert\bm{d}_{i,t}\right\rVert} & \text{otherwise} \label{eq:nonconvex}
\end{empheq}\label{eq:fric_forces}
\end{subequations}
Note, that while the formulation in (\ref{eq:fric_forces}) requires the
distinction between sliding contacts and those remaining at rest the original
formulation (\ref{eq:friction_bound}) \& (\ref{eq:mdp}) holds in both cases.

This reformulation introduces also a discontinuity as the relation between
$\bm{c}_{i,t}$ and $\bm{d}_{i,t}$ is discontinuous at
$\left\lVert\bm{d}_{i,t}\right\rVert=0$. We could approach this discontinuity
in a similar fashion as we did above to distinguish between rolling and
breaking contacts by using indicator constraints. However this does not
resolve the non-convexity as (\ref{eq:nonconvex}) is still non-convex.

Unfortunately the methods we developed in Section~\ref{sec:2d} cannot be
directly applied to three dimensions \textit{even with prior knowledge of
which contacts slip and which do not}. While in two dimensions, a contact can
only roll or slide in one of two directions (and there are hence 3 slip states
per contact), in three dimensions there are infinitely many directions a
contact can slip in and we cannot further break down the sliding case. These
characteristics of the Coulomb model presents much difficulty in the modeling
of contacts with Coulomb friction and many approaches have been proposed to
approximate the friction law. We reviewed a selection of such efforts in
Chapter~\ref{sec:related_dynamics}. 

As all motion in our framework is purely 'virtual' and any real motion is
expected to be minuscule one may be tempted to simply consider all contacts to
be stationary and cut away the discontinuity and non-convexity introduced by
the MDP. Thus we would simply ignore the maximum dissipation principle
(\ref{eq:mdp}) and with it (\ref{eq:nonconvex}). The remaining quadratic
inequality constraint (\ref{eq:frictioncone}) is convex. In fact, it has been
shown that (\ref{eq:friction_bound}) can be cast as a \textit{linear matrix
inequality} (LMI)~\cite{HAN00}. 

\begin{figure}[t]
\centering
\includegraphics[width=0.85\linewidth]{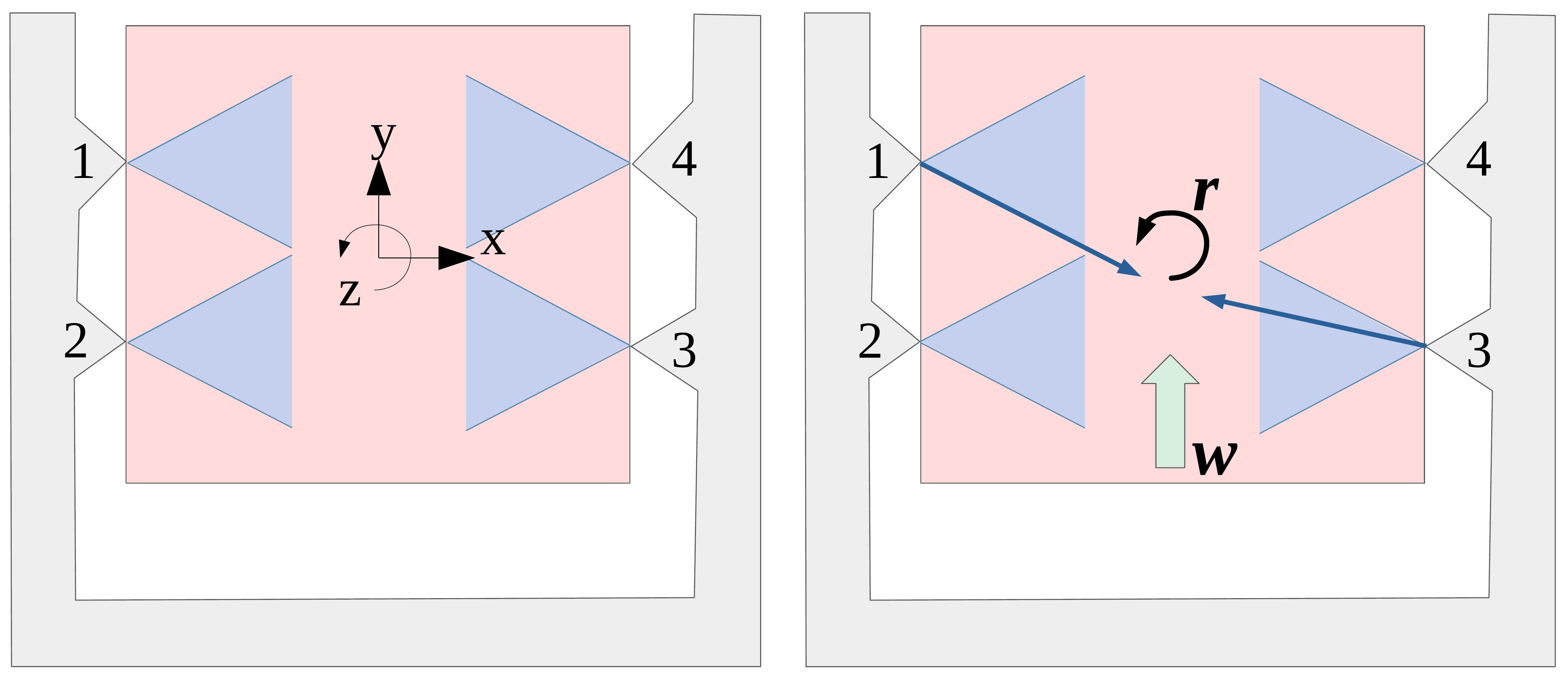}

\caption{Unconstrained friction forces (except for friction cone constraint).
Left: undisturbed system with no contact forces (also showing the object
reference frame). Right: reaction to force $\bm{w}$ applied to the object. A
rotation $\bm{r}$ loads the contacts such that contact forces (blue arrows)
can resist the applied wrench (green arrow), even in the absence of a
preload.}

\label{fig:twist}
\end{figure}

Unfortunately, this approach does not work as we will now illustrate. The
rigid, passively loaded fingers allow an optimization formulation with
unconstrained object movement to 'wedge' the object between contacts creating
large contact forces. This allows the grasp to withstand very large applied
wrenches by performing 'unnatural' virtual displacements that satisfy all our
constraints and lead to equilibrium, but violate the principle of conservation
of energy: the energy stored in the virtual springs and the energy dissipated
due to friction are greater than the work done by the externally applied
wrench and the actuators. In the example shown in Fig.~\ref{fig:twist} the
applied force does no work while energy is being stored in the springs and
being dissipated due to friction at contacts 1 and 3. 

From the principle of virtual work it follows that a solution that satisfies
both equilibrium and the maximum dissipation principle also satisfies
conservation of energy. Hence, in order to enforce conservation of energy and
obtain physically meaningful results we must enforce the MDP or find a
reasonable approximation.

\section{Iterative solution approach}\label{sec:iterative}

Now that we have discussed the consequences of simply dropping the MDP we
describe a physically motivated method to obtain accurate results. The method
described in this section does not offer a guarantee of convergence. In
Chapter~\ref{sec:cones} we will describe a method that is guaranteed to
converge to the exact solution.

Recall that we model some compliance of the grasp in order to be able to
capture passive effects. Thus, we think of a stable grasp as one that moves
from one equilibrium state to another when an external wrench is applied to
the grasped object. The changes in contact forces necessary to balance the
applied wrench arise through the motions of the object and hand links. These
motions are minuscule for the stiff hands we consider here and therefore we
can ignore dynamic effects and argue from a purely quasistatic viewpoint.

Forces applied to an object tend to move that object in the direction of the
applied force. This is why the unphysical motion in Fig.~\ref{fig:twist} seems
intuitively unnatural --- the object moves in a direction orthogonal to the
force applied to it. Of course the complex effects due to grasp geometry as
well as a non-convex friction law mean that in general a grasped object will
not move exactly along the direction of the force applied to it. While these
effects are difficult to model we can devise an iterative scheme to
approximate them.

In this scheme we will take small steps (in terms of object motion) such that
the object may only move in the direction of the net resultant wrench applied
to it. While we cannot enforce the maximum dissipation principle and its
consequence --- friction opposes motion --- we can offer an approximation that
is physically well motivated. We will introduce an objective term to the
optimization that, at every step of the iteration, will minimize the magnitude
of the net resultant wrench on the object. The reasoning is that such forces,
when projected to the objects reference frame, oppose the net resultant wrench
to their best ability. As motion is only allowed in the direction of this net
resultant wrench, the contact forces computed thus oppose this motion to their
best ability. Taking steps that are sufficiently small ensures that the
complex geometric and friction effects of the grasp are accurately captured.

We implement this scheme by removing the object equilibrium constraint
(\ref{eq:object_eq}) and including an objective in the optimization
formulation such as to minimize the net resultant wrench $\bm{w}^k$ (the net
sum of the applied wrench and contact forces) acting on the object at step
$k$.

\begin{eqnarray}
\text{minimize: } \left\lVert \bm{w}^k \right\rVert = \left\lVert \bm{w}+\bm{G}^T\bm{c}^k \right\rVert
\end{eqnarray}

We also constrain the object movement such that motion is only allowed in the
direction of the unbalanced wrench acting on the object remaining from the
previous step. We limit the step size by a parameter $\gamma$ for numerical
stability and accuracy:

\begin{equation}
\bm{r}^k = \bm{r}^{k-1} + s\bm{w}^{k-1}, \quad 0 \leq s  \leq \gamma
\label{eq:movement}
\end{equation}

After each iteration, we check for convergence by comparing the incremental
improvement to a threshold $\epsilon$. If the objective has converged to a
sufficiently small net wrench (we chose $10^{-3}N$), we deem the grasp to be
stable; otherwise, if the objective converges to a larger value, we deem the
grasp unstable. Note that we must use the objective in the optimization
formulation in this approach, which means we are constrained to the solution
of \textit{existence problems}. In this case the problem we are solving is
this: \textit{Given $\bm{f}^c$, determine if $\bm{c}$ will arise that balances
applied wrench $\bm{w}$}. Thus, we formulate a \textit{movement
constrained existence problem} as outlined in Algorithm~\ref{alg:iterative} to
be solved iteratively as outlined in Algorithm~\ref{alg:loop} allowing us to
make a determination as to the stability of the grasp. The computation time of
this process is directly related to the number of iterations required until
convergence.

\begin{algorithm}[!t]

\caption{Inner loop of the iterative solution scheme: Minimize the resultant
wrench, where object motion is only allowed in the direction of the current
net resultant wrench.}\label{alg:iterative}

\begin{algorithmic}[0]
\State \textbf{Input:} $\bm{f}^c$ --- commanded actuator forces, $\bm{w}$ --- applied wrench, $\bm{r}^{k-1}$ --- previous object displacement, $\bm{w}^{k-1}$ --- previous net wrench
\State \textbf{Output:} $\bm{c}^k$ --- contact forces, $\bm{r}^k$ --- next step object displacement, $\bm{w}^k$ --- next step net wrench
\Procedure{Movement Constrained EP}{$\bm{f}^c, \bm{w}, \bm{r}^{k-1}, \bm{w}^{k-1}$}
  \State \textbf{minimize:} $\left\lVert \bm{w}^k \right\rVert$ \Comment{net wrench}
  \State \textbf{subject to:} 
    \State \hspace{\algorithmicindent} $Eqs.\ (\ref{eq:object_eq})\ ,\ (\ref{eq:hand_eq})\ \&\ (\ref{eq:torqueRatio})$ \Comment{object and hand equilibrium}
    \State \hspace{\algorithmicindent} $Eqs.\ (\ref{eq:rel_contact_motion})\ \&\ (\ref{eq:binary_contact})$ \Comment{normal force unilaterality and constitutive relation}
    \State \hspace{\algorithmicindent} $Eq.\ (\ref{eq:binary_actuator})$ \Comment{actuator unilaterality}
    \State \hspace{\algorithmicindent} $Eq.\ (\ref{eq:frictioncone})$ \Comment{friction cone}
    \State \hspace{\algorithmicindent} $Eq.\ (\ref{eq:movement})$ \Comment{object movement}
\State \textbf{return} $\bm{c}^k$, $\bm{r}^k, \bm{w}^k$ 
\EndProcedure
\end{algorithmic}
\end{algorithm}

\begin{algorithm}[!t]

\caption{Outer loop of the iterative solution scheme: Repeatedly solve the
"Movement Constrained EP". Update the current object motion and net resultant
wrench at every step.}\label{alg:loop}

\begin{algorithmic}[0]
\State \textbf{Input:} $\bm{f}^c$ --- commanded actuator forces, $\bm{w}$ --- applied wrench
\State \textbf{Output:} $\bm{c}$ --- contact forces, $\bm{w}_{res}$ --- net resultant
\Procedure{Iterative EP}{$\bm{f}^c, \bm{w}$}
  \State $\bm{r}^0 = 0$
  \State $\bm{w}^0 = \bm{w}$
  \Loop
    \State \textbf{($\bm{c}^k,\bm{r}^k,\bm{w}^k$) = Movement Constrained EP($\bm{f}^c, \bm{w}, \bm{r}^{k-1}, \bm{w}^{k-1}$)} \Comment{Algorithm~\ref{alg:iterative}}
    \If{$ \left\lVert \bm{w}^{k-1}-\bm{w}^k \right\rVert < \epsilon$} \Comment{Check if system has converged}
      \State \textbf{break}
    \EndIf
  \EndLoop
  \State \textbf{return} $\bm{c}^{final}, \bm{w}^{final}$
\EndProcedure
\end{algorithmic}
\end{algorithm}

However, such an iterative approach is not guaranteed to converge, or to
converge to the physically meaningful state of the system. Thus, while this
approach is simple to implement and provides a good estimate of grasp
stability we can make no guarantees of solutions obtained: The solution we
find is not guaranteed to be a good approximation of a solution to the
\textit{exact problem} as defined in Chapter~\ref{sec:model3d}. Furthermore,
if algorithm~\ref{alg:loop} does not return a valid and stable solution we do
not know if such a solution may not exist. Nonetheless, this iterative
approach has proved useful in analyzing the passive behavior of grasps as we
will show in Chapter~\ref{sec:3d_apps_iterative}.

\section{Application to grasp analysis in three dimensions}\label{sec:3d_apps_iterative}

We will illustrate the application of the iterative solution approach
discussed in Chapter~\ref{sec:iterative} on example grasps with the Barrett
hand and an underactuated tendon-driven gripper. We pointed out that a
limitation of the iterative approach is that we may only solve \textit{existence}
type problems as the objective is required in the formulation and is not
available for us to optimize a quantity of our choosing. Recall, however, that
the question of 'will a given grasp with given joint commands resist a
specific applied force' is such an existence problem and hence we can
investigate the range of forces applied to the object our example grasps can
resist.

To facilitate visualization we will confine the forces applied to the object
to the a plane. As our framework can only answer 'spot checks' --- that is
determine stability for a specific applied force --- we discretize this plane
by direction vectors with a spacing of 1\degree~between them. We then
determine the maximum force applied to the object the grasp can withstand for
each of these directions. We do this by performing a binary search converging
on the maximum magnitude of the force before the grasp becomes unstable.

We model the Barrett  hand as having all nonbackdrivable joints. Our
qualitative experience indicates that the finger flexion joints never
backdrive, while the spread angle joint backdrives under high load. For
simplicity we also do not use the breakaway feature of the hand; our real
instance of the hand also does not exhibit this feature. We model the joints
as rigidly coupled for motion, and assume that all the torque supplied by each
finger motor is applied to the proximal joint.

We show force data collected by replicating the grasp on a real hand and 
testing resistance to external disturbances. To measure the maximum force that
a grasp can resist in a certain direction, we manually apply a load to the
grasped object using a Spectra wire in series with a load cell (Futek,
FSH00097). In order to apply a pure force, the wire is connected such that the
load direction passes through the center of mass of the object. We increase
the load until the object starts moving, and take the largest magnitude
recorded by the load cell as the largest magnitude of the disturbance the
grasp can resist in the given direction.

\begin{figure}[!t]
\centering
\includegraphics[width=3.5in]{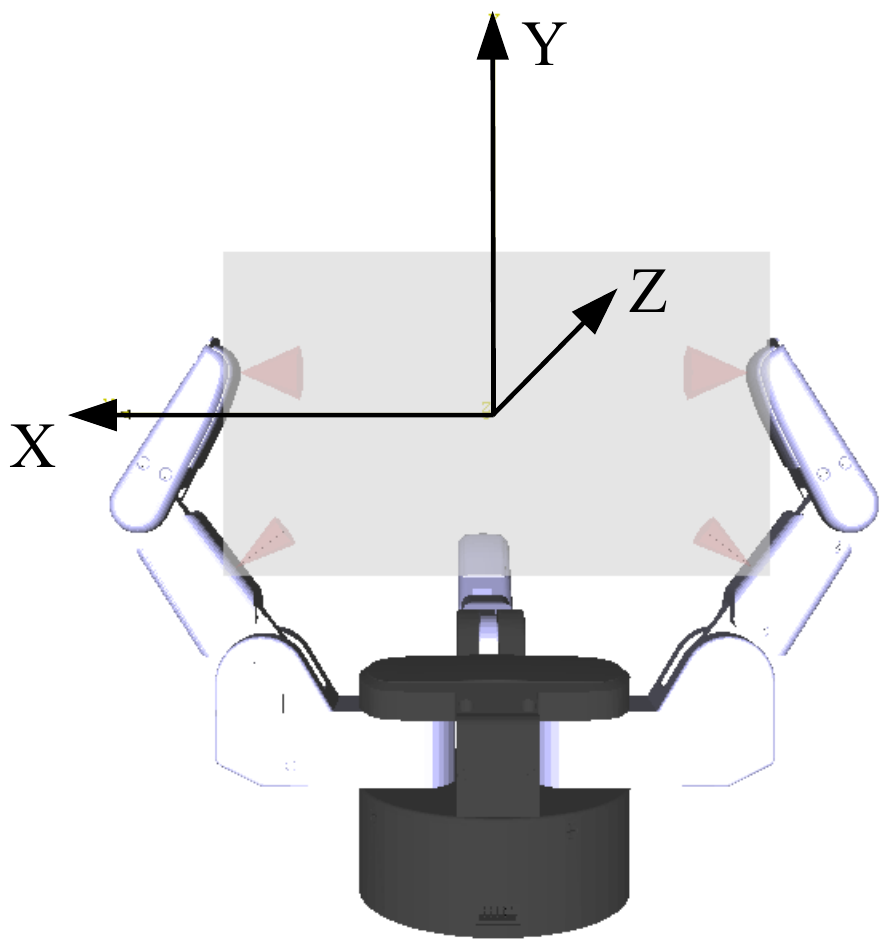}

\caption{Grasp example 1. Note that this is the same grasp we used in 
Chapter~\ref{sec:bridge_example} to illustrate the consequences of passive
reactions.}

\label{fig:2dgrasp}
\end{figure}

\vspace{2mm} \noindent \textbf{Barrett hand:} Let us first consider the grasp
in Fig.~\ref{fig:2dgrasp}, which is similar to the grasp introduced in
Section~\ref{sec:bridge_example} to serve as an example for the importance of
the passive resistance capabilities that nonbackdrivable joints provide. We
now have the necessary tools to analyze such grasps. Considering only forces
in the grasp plane this grasp effectively becomes a 2D problem: simple enough
to be understood intuitively, but still complex enough to give rise to
interesting interplay between the joints and contacts.

Consider first the problem of resisting an external force applied to the
object center of mass and oriented along the $y$-axis. This simple case
illustrates the difference between active and passive resistance. Resistance
against a force oriented along the positive $y$-axis requires active torque
applied at the joints in order to load the contacts and generate friction. The
force can be resisted only up to the limit provided by the preload, along with
the friction coefficient. If the force is applied along the negative $y$-axis,
resistance happens passively, provided through the contacts on the proximal
link. Furthermore, resistance to such a force does not require any kind of
preload, and is infinite in magnitude (up to the breaking limit of the hand
mechanism, which does not fall within our scope here). 

\begin{figure}[!t]
\centering
\includegraphics[width=5in]{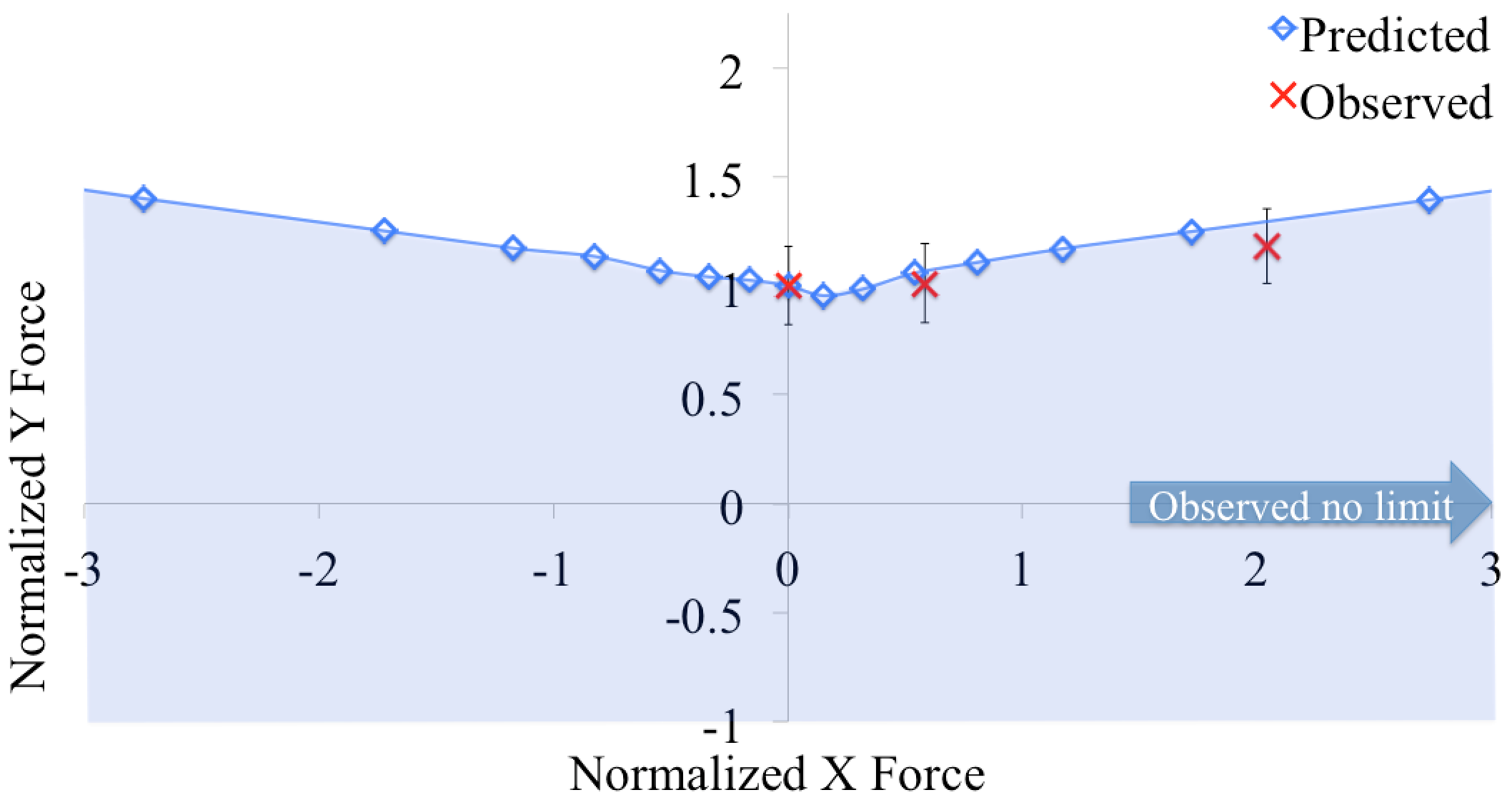}

\caption{Normalized forces in the $xy$-plane that can be resisted by the grasp
in Fig.~\ref{fig:2dgrasp}: observed by experiment (mean $\pm$ one standard
deviation) and predicted by our framework. For both real and predicted data,
we normalize the force values by dividing by the magnitude of the maximum
resistible force along the positive direction of the $y$-axis (note thus that
both predicted and experimental lines cross the $y$-axis at $y=1.0$). In all
directions falling below the blue line, the prediction framework hit the upper
limit of the binary search (arbitrarily set to $10^3$ N). Hence we deem forces
in the shaded area resistible. In the direction denoted by 'Observed no
limit', the grasp was not disturbed even when reaching the physical limit of
the experimental setup.}

\label{fig:2dgrasp_xy}
\end{figure}

For an external force applied along the $x$-axis, the problem is symmetric
between the positive and negative directions. Again, the grasp can provide
passive resistance, through a combination of forces on the proximal and distal
links. For the more general case of forces applied in the $xy$-plane, we again
see a combination of active and passive resistance effects. Intuitively, any
force with a negative $y$ component will be fully resisted passively. However,
forces with a positive $y$ component and non-zero $x$ component can require both
active and passive responses. Fig.~\ref{fig:2dgrasp_xy} shows the forces that
can be resisted in the $xy$-plane, both predicted by our framework and observed
by experiment.

For both real and predicted data, we normalize the force values by dividing by
the magnitude of the maximum resistible force along the positive direction of
the $y$-axis (note thus that both predicted and experimental lines cross the
$y$-axis at $y=1.0$).  The plots should therefore be used to compare trends
rather than absolute values. We use this normalization to account for the fact
that the absolute torque levels that the hand can produce, and which are
needed by our formulation in order to predict absolute force levels, can only
be estimated and no accurate data is available from the manufacturer. The
difficulty in obtaining accurate assessments of generated motor torque
generally limits the assessments we can make based on absolute force values.
However, if one knows the real magnitude of the maximum resistible external
force along any direction, in which this magnitude is finite, one could infer
from these figures the real maximum resistible wrenches in the other
directions.

\begin{figure}[!t]
\centering
\includegraphics[width=5in]{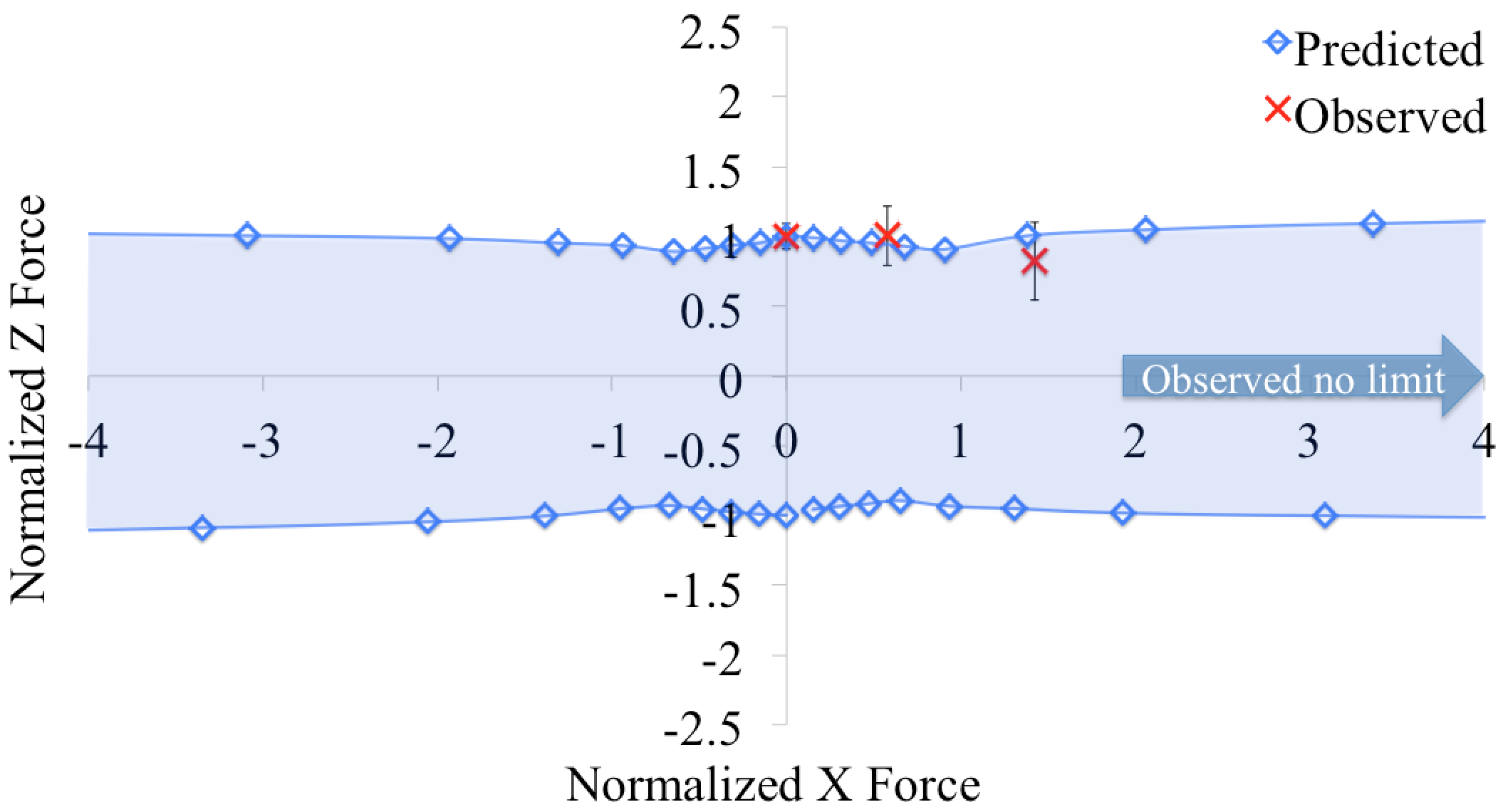}

\caption{Normalized forces in the $xz$-plane that can be resisted by the grasp
in Fig.~\ref{fig:2dgrasp}: predicted by our framework, and observed by
experiment. In all directions falling between the blue lines (shaded), the
prediction framework hit the upper limit of the binary search (arbitrarily set
to $10^3$ N). In the direction denoted by 'Observed no limit', the grasp was
not disturbed even when reaching the physical limit of the experimental setup.}

\label{fig:2dgrasp_xz}
\end{figure}

Moving outside of the grasp plane, Fig.~\ref{fig:2dgrasp_xz} shows predicted
and measured resistance to forces in the $xz$-plane. Again, we notice that
some forces can be resisted up to arbitrary magnitudes thanks to passive
effects, while others are limited by the actively applied preload.

\begin{figure}[!t]
\centering
\includegraphics[width=6in]{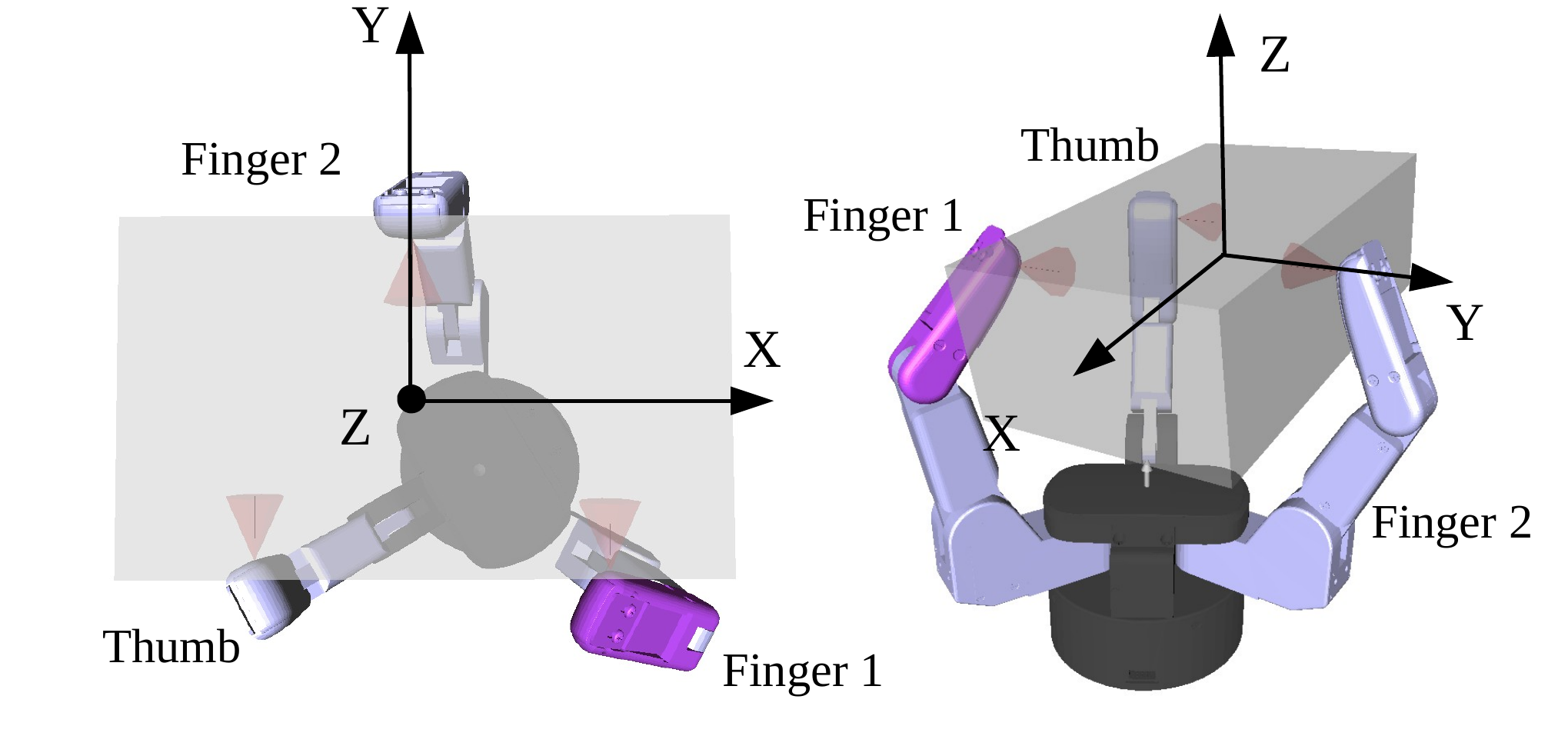}

\caption{Top and side views for grasp example 2 also indicating finger labels.
Note that the spread angle degree of freedom of the Barrett hand changes the
angle between finger 1 and finger 2; the thumb is only actuated in the flexion
direction.}

\label{fig:3dgrasp}
\end{figure}

\begin{figure}[!t]
\centering
\includegraphics[width=3.5in]{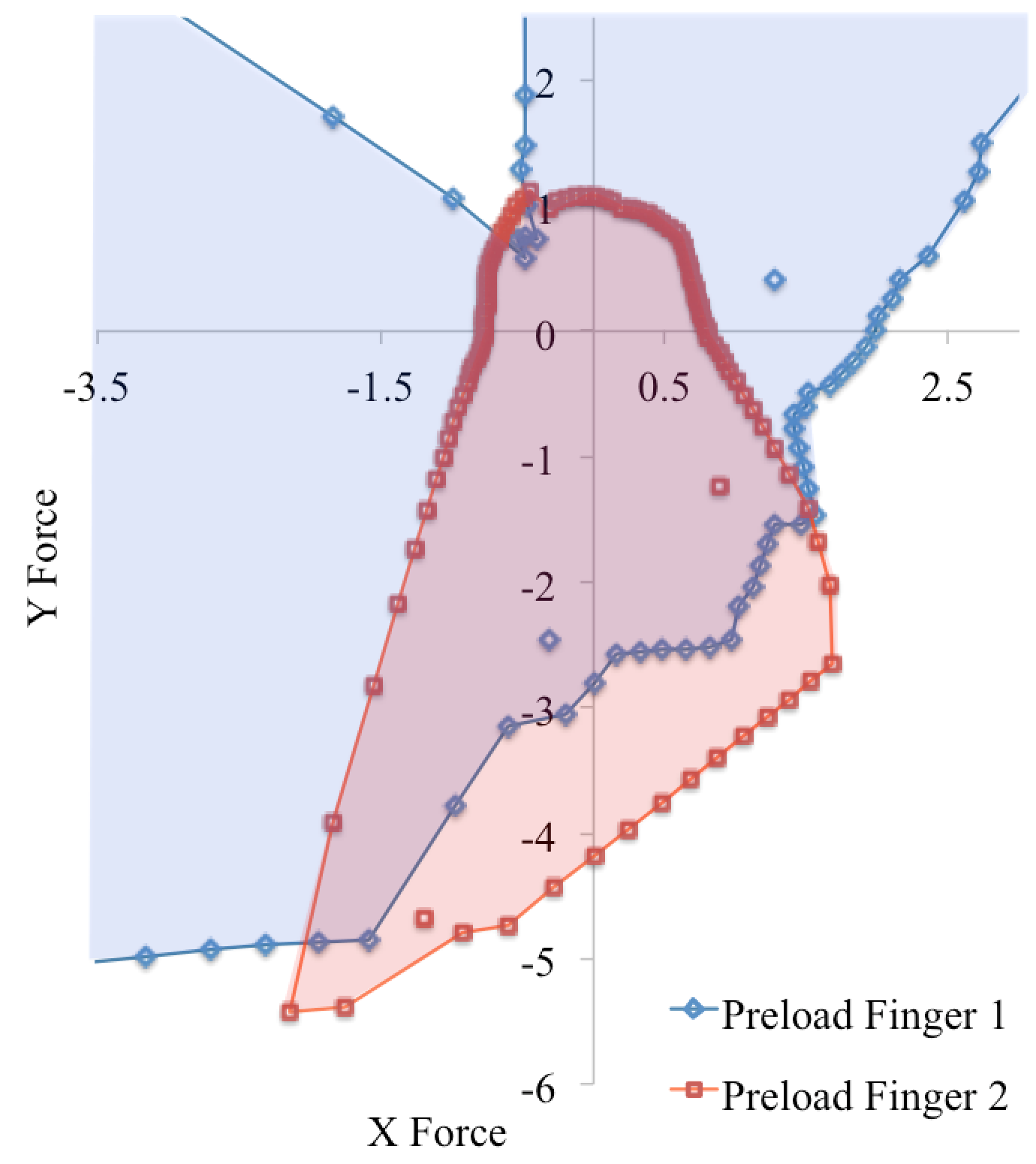}

\caption{Forces in an $xy$-plane that can be resisted by the grasp in
Fig.~\ref{fig:3dgrasp} (shaded) as predicted by our framework, depending on
which finger is preloaded. Note the four outlier results and that they have
not been included in the determination of the regions of resistible forces.
The forces are normalized and hence dimensionless.}

\label{fig:outlier}
\end{figure}

As was mentioned previously, a common method for  grasp creation is to choose
a set of actuator commands to preload the grasp  before applying any external
wrench. These preloads are then maintained  throughout the task trusting in
passive effects to stabilize the grasp.  One advantage of studying the effect
of applied joint torques on grasp stability is that it allows us to observe
differences in stability due to different ways of preloading the same grasp.
For example, in the case of the Barrett hand, choosing at which finger(s) to
apply preload torque can change the outcome of the grasp, even though there is
no change in the distribution of contacts. We illustrate this approach on the
case shown in Fig.~\ref{fig:3dgrasp}. Using our framework we can compute
regions of  resistible wrenches for two different preloads (see
Fig.~\ref{fig:outlier}). 

We compare the ability of the grasp to resist a disturbance applied along the
$x$-axis in the positive direction if either finger 1 or finger 2 apply a
preload torque to the grasp. Our formulation predicts that by preloading
finger 1 the grasp can resist a disturbance that is 2.48 times larger in
magnitude than if preloading finger 2. Experimental data (detailed in
Table~\ref{tab:3dgrasp}) indicates a ratio for the same disturbance direction
of 2.23. The variance in measurements again illustrates the difficulty of
verifying such simulation results with experimental data. Nevertheless,
experiments confirmed that preloading finger 1 is significantly better for
this case.

\begin{table}[!t]
\renewcommand{\arraystretch}{1.3}
\centering
\begin{tabular}{c|cccc|cc}
 & \multicolumn{4}{c}{Measured resistance} & \multicolumn{2}{c}{Predicted} \\
 & Values(N) & Avg.(N) & St. Dev. & Ratio & Value & Ratio\\\hline
&&&&&&\\[-3.5mm]\hline
&&&&&&\\[-2mm]
F1 load & 12.2, 10.8, 7.5, 7.9, 9.3 & 9.6 & 1.9 & 2.23 & 1.98 & 2.48\\
F2 load & 3.7, 4.1, 5.0 & 4.3 & 0.7 &    & 0.80 &
\label{tab:3dgrasp}
\end{tabular}

\caption{Predicted and measured resistance to force applied along the positive
$x$-axis in the grasp problem in Fig.~\ref{fig:3dgrasp}. Each row shows the
results obtained if the preload is applied exclusively by finger 1 or finger 2
respectively. Experimental measurements were repeated 5 times for finger 1 (to
account for the higher variance) and 3 times for finger 2. Predicted values
are non dimensional, and hence the ratio between the two preload cases is
shown.}

\end{table}

This result can be explained by the fact that, somewhat counter-intuitively,
preloading finger 1 leads to larger contact forces than preloading finger 2,
even if the same torque is applied by each motor. Due to the orientation of
finger 1, the contact force on finger 1 has a smaller moment arm around the
finger flexion axes than is the case for finger 2. Thus, if the same flexion
torque is applied in turn at each finger, the contact forces created by finger
1 will be higher. In turn, due to passive reaction, this will lead to higher
contact forces on finger 2, even if finger 1 is the one being actively loaded.
Finally, these results hold if the spread degree of freedom is rigid and does
not backdrive; in fact, preloading finger 1 leads to a much larger passive
(reaction) torque on the spread degree of freedom than when preloading finger
2.

Referring to Fig.~\ref{fig:outlier}, we note that actively preloading finger 
1 results in greater resistance only in some directions. There is much
structure to the prediction made by our framework that could be exploited to
make better decisions when preloading a grasp with some knowledge of the 
expected external wrenches. 

\begin{figure}[!t]
\centering
\includegraphics[width=5.0in]{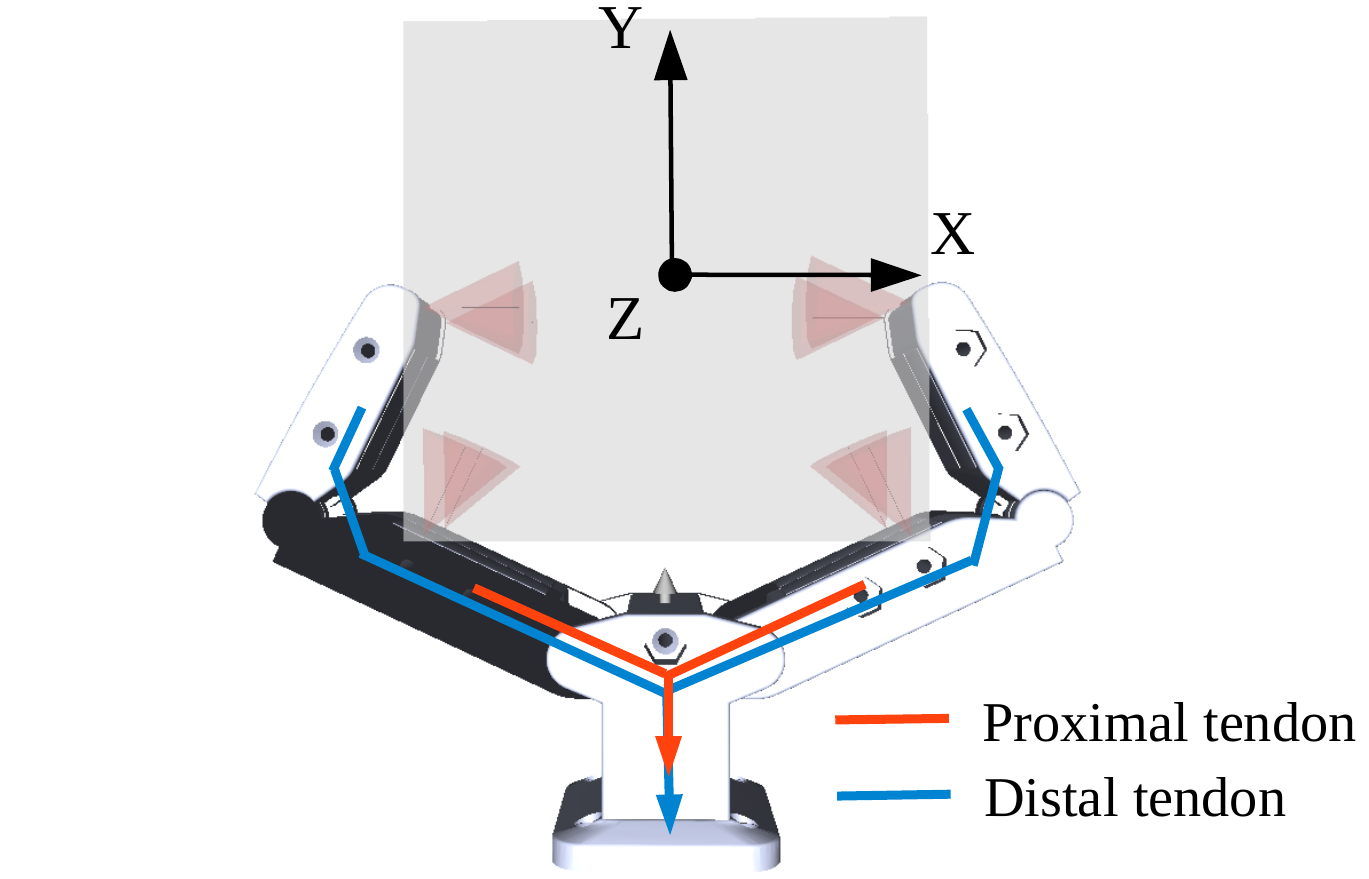}

\caption{Grasp example 3. Note that in terms of contact positions, this grasp
is very similar to that in Fig.~\ref{fig:2dgrasp}. However, the kinematic
differences cause these two grasps to behave very differently.}

\label{fig:underactuatedGrasp}
\end{figure}

\vspace{2mm} \noindent \textbf{Underactuated hand:} We now apply our framework
to a grasp with a two-fingered, tendon-driven underactuated gripper (see
Fig.~\ref{fig:underactuatedGrasp}). The gripper has four degrees of freedom,
but only two actuators driving a proximal and a distal tendon. The proximal
tendon has a moment arm of 5mm around the proximal joints. The distal tendon
has moment arms of 1.6mm and 5mm around the proximal and distal joints
respectively. The actuators are non backdrivable and hence the tendons not
only transmit actuation forces, they also provide kinematic constraints to the
motion of the gripper's links. 

\begin{figure}[!t]
\centering
\includegraphics[width=3.5in]{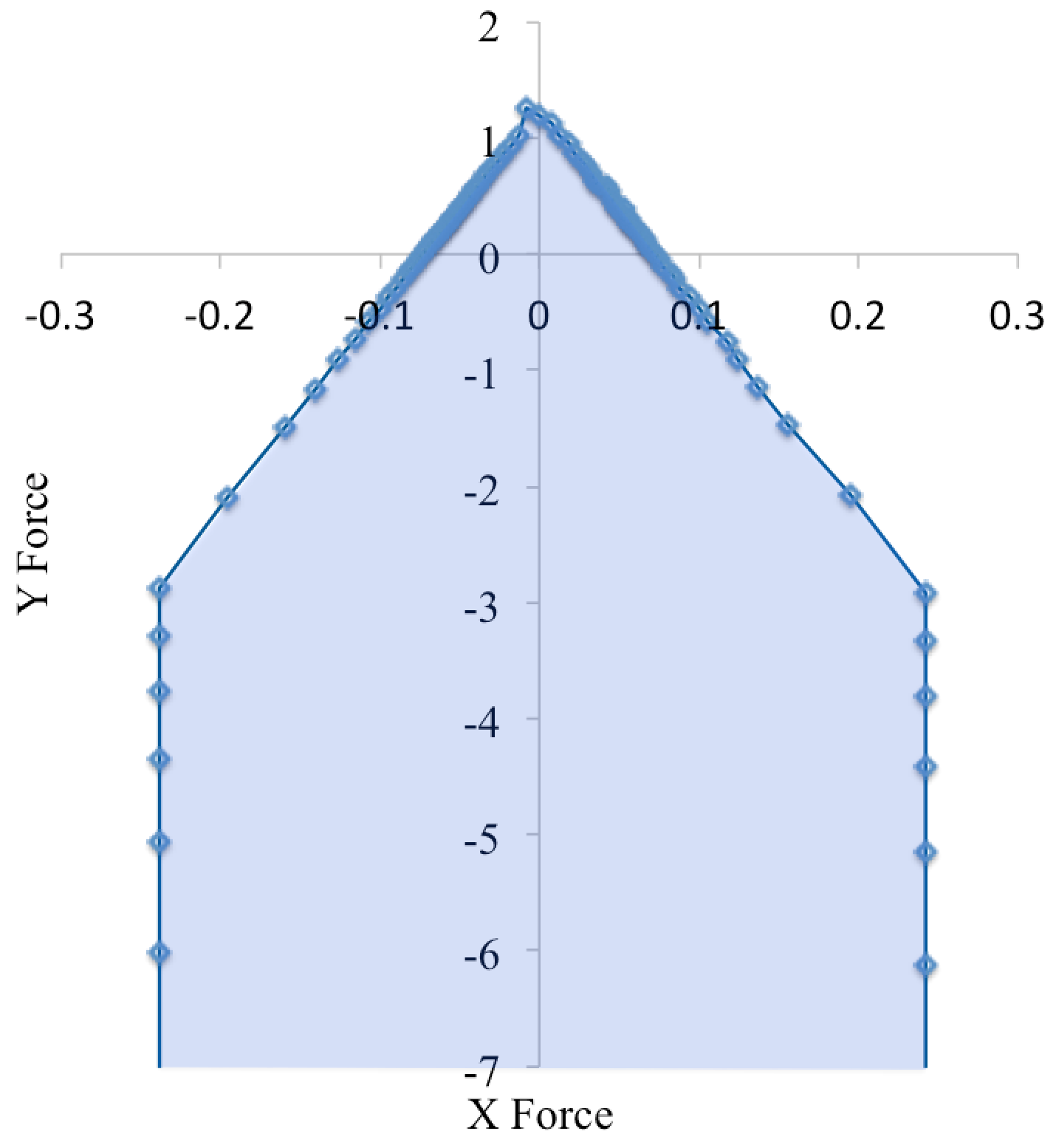}

\caption{Forces in an $xy$-plane that can be resisted by the grasp in
Fig.~\ref{fig:underactuatedGrasp} (shaded) as predicted by our framework. The 
forces are normalized and hence dimensionless. Note the difference in scale on
the $x$-and $y$-axes.}

\label{fig:underactuated}
\end{figure}

As the tendons split and lead into both fingers, we assume that they are  
connected to the actuator by a differential. This introduces compliance to the
grasp: if one proximal joint closes by a certain amount, this will allow the
other proximal joint to open by a corresponding amount.  This compliance means
that the underactuated grasp in  Fig.~\ref{fig:underactuatedGrasp} will behave
fundamentally different than the very similar grasp in Fig.~\ref{fig:2dgrasp}.
To see this,  consider the region of resistible wrenches in the $xy$-plane 
(Figs.~\ref{fig:2dgrasp_xy}\ \&~\ref{fig:underactuated}). In both cases  the
object is gripped by two opposing fingers, however, while in the case  of the
Barrett hand the grasp could withstand arbitrarily large forces pushing the
object  directly against a finger, our framework predicts only limited
resistance to such forces in the case of the underactuated hand. 

We used our framework to apply two different sets of preload actuator commands
$\bm{f}^c$ and analyzed the  resistance of the resulting grasp to an
externally applied wrench. We chose to apply a  torque around the $x$-axis and
considered two preload cases: applying an active load on the proximal tendon
only ($\bm{f}^c = [1,0]^T$), leaving the distal tendon to be loaded  passively
as well as the reverse --- applying an active load on the distal tendon only
($\bm{f}^c = [0,1]^T$), leaving the proximal tendon to be loaded passively.
For equal actuator force, our framework predicts, that a preload created by
only actively loading the distal tendon leads to almost twice as much
resistance to torques applied in the direction of the $x$-axis than only
actively loading the proximal tendon.

Experimental verification of this prediction proved to be difficult, as
results had high variance and application of a pure torque to the object along
an axis that penetrates the distal links of the gripper was  complicated.
However, we mounted the gripper such that the grasp plane was in the
horizontal and placed weights on the top end of the box object. We found the
resistance to these applied wrenches indeed to be much higher when actively
loading the distal tendon as opposed to the proximal.

\begin{figure}[!t]
\centerline{

\subfigure[Actively loading the proximal tendon]{\includegraphics[width=3in]{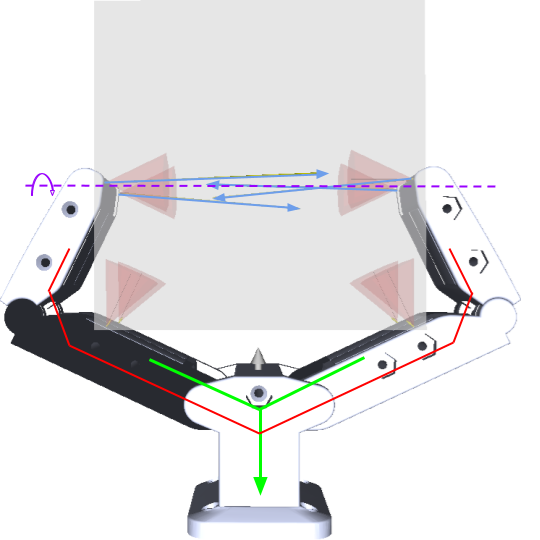}
\label{fig:proximal}}
\subfigure[Actively loading the distal tendon]{\includegraphics[width=3in]{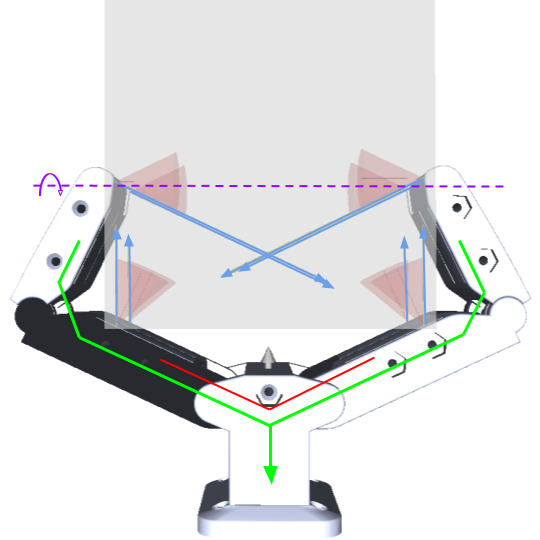}%
\label{fig:distal}}

}

\caption{Different preload contact forces arising from different preload
actuator commands (blue arrows) --- the tendon shown in green is being
actively loaded by the actuator. As a result the red tendon is loaded
passively. The violet line denotes the torque applied to the grasped object.
Note that the forces shown are those that arise purely as a result of the
preload actuator commands and hence \textit{before} the application of any
external wrench to the object.}\label{fig:tendon_loads}

\end{figure}

The reason for this discrepancy in resistance to this wrench becomes apparent
when looking at the contact forces arising from the different preload commands
(see Fig.~\ref{fig:tendon_loads}.) When only actively loading the proximal
tendon (Fig.~\ref{fig:proximal}) the preload contact forces are concentrated
at the distal contacts(!). In fact, the contacts on the proximal links break
entirely. These contacts, however, have the largest moment arm with respect to
the torque applied to the object and are hence crucial for the ability of the
grasp to resist such torques.

When instead only actively loading the distal tendon (Fig.~\ref{fig:distal})
contact forces arise on all four links. The contact forces at the proximal
links particularly allow the grasp to resist torques around the horizontal
object axis. Our framework has captured the passive effects present in this
underactuated hand and helped us to understand how to best make use of the
actuators available.

\section{Discussion}

In this Chapter we have developed a grasp model and the corresponding
algorithms to determine the passive stability of spacial robotic grasps. We
showed the importance of the maximum dissipation principle for the
determination of passive reaction effects. We did so by showing how ignoring
the maximum dissipation principle allows for unphysical solutions such that
stability cannot be determined. We proposed a physically motivated iterative
scheme that allows us to approximate the law of conservation of energy and
hence the maximum dissipation principle.

The resulting algorithm enabled us to determine the passive stability of
spacial grasps for both a fully actuated as well as an underactuated hand. We
showed that our framework correctly captures passive phenomena and
demonstrated a practical application of our algorithm as a tool for choosing
appropriate preload actuator commands. We compared the predictions made by our
framework to data collected from real robotic grasps and found that while
empirical grasp stability analysis is difficult, our framework appears to
agree well with the observed behaviors.

However, our current real underactuated hand only allows experimental
validation of a subset of our analysis results. We would like to design a hand
that we can use to further validate our framework and study the effects of
underactuation on grasp stability. For instance, our framework predicts that
wrench resistance is highly dependent on the torque ratios at the joints due
to the kinematics of the force transmission. We would like to experiment with
a variety of underactuated hands, with varying kinematic and actuation models,
to investigate these effects and demonstrate the effectiveness of our
framework as a design tool. 

An important limitation of our algorithm is that we had to use the objective
in the optimization formulation in order to approximate the maximum
dissipation principle. This means that we could only solve problems of the
\textit{existence} type and perform a spot check if a grasp with given
actuator commands could resist a specific disturbance. As the objective is not
available we cannot use it to do more practical analysis such as finding
optimal actuator commands.

A further limitation is that in a subset of cases, the solver reports maximum
resistible wrenches of very different magniture along some directions relative
to neighboring directions. For example, in the second example grasp from the
previous section (Fig.~\ref{fig:3dgrasp}), when computing resistance to
disturbances sampled from the $xy$-plane (Fig.~\ref{fig:outlier}), we obtain
two outliers for each preload case that do not follow the trend of the
surrounding points. These outliers are quite rare and tend to fall within the
area deemed to contain resistible wrenches (shaded). These effects will
require further investigation.

A promising avenue to pursue in terms of alleviating these outliers might be
to take into account the effect of uncertainties (e.g. in  exact contact
location) on our model. We believe exploring the sensitivity of the model to
such uncertainties may yield many valuable insights and make our framework
even more relevant to practical robotic grasping.

The most significant limitation of our algorithm stems from its iterative
nature. While our iterative approach allows us to approximate the maximum
dissipation principle, such an iterative approach is not guaranteed to
converge, or to converge to the physically meaningful state of the system. As
such, our approach offers an approximation without any formal guarantees. In
practice, however, our algorithm produces physically plausible results that
could be confirmed by empirical validation.


\chapter{Grasp Stability Analysis with the Maximum Dissipation Principle}\label{sec:cones}

\section{Introduction}

In this Chapter we will make use of the same grasp model as described in
Chapter~\ref{sec:model3d}. This model, particularly what we have called the
\textit{complete problem}, is very general and captures all of the
characteristics of robotic grasps we are interested in. It accounts for both
object and hand equilibrium, the unilaterality of both contacts and
nonbackdrivable actuators as well as the Coulomb friction law including the
Maximum Dissipation Principle (MDP).

We showed how the majority of this grasp model can be cast as a Mixed Integer
Program in Chapter~\ref{sec:unilaterality}. The only component that eluded our
formulation was the maximum dissipation aspect of the Coulomb friction law. In
Chapter~\ref{sec:3dfriction} we demonstrated the importance of the MDP for the
determination of passive resistance in robotic grasps. We showed that simply
omitting the MDP allows for unphysical solutions to the grasp model
formulation making a determination of stability impossible. However, we also
discussed the difficulty in accurately modeling the MDP due to its
non-convexity.

In Chapter~\ref{sec:iterative} we proposed a physically motivated iterative
scheme that aims to approximate the MDP. However, this approach does not
explicitly include the MDP and is therefore unlikely to converge to a solution
that actually satisfies the MDP to high accuracy. In fact, our previous
approach provides no formal guarantee of convergence. Furthermore, we had to
use the objective of the MIP optimization formulation in order to approximate
the MDP, which meant that the objective was not available for more involved
analysis. This is unfortunate, as many queries of practical importance can be
posed as optimization problems. In fact, much of the grasp force analysis
literature discussed in Chapter~\ref{sec:related_closure} seeks to find
contact wrenches or actuator torques that are optimal with respect to some
objective.

In this chapter we will develop a Mixed Integer formulation that explicitly
models the MDP. To this end we propose a relaxation of the friction model such
that we can solve the system as an MIP. Through successive and hierarchical
tightening of the relaxation we can efficiently find solutions that satisfy
all constraints --- including the MDP --- to arbitrary accuracy. A very useful
property of tightening approaches is that they allow for strong guarantees:
if, at any stage of refinement our model fails to find a solution, we can
guarantee that no solution exists to the exact problem. This allows for early
exit from computation in cases where equilibrium cannot exist. Furthermore,
tightening approaches can greatly reduce the size of the optimization problem
that has be solved in order to find a solution. We use these properties to
develop a grasp model that, to the best of our knowledge, is the first that
can handle three-dimensional frictional constraints that include the MDP, up
to arbitrary accuracy (and thus approaching the solution to the exact problem)
where solutions can be solved for in a computationally efficient fashion. Our
model allows us to produce solutions on commodity computers fast enough for
practical applicability. 

Our formulation furthermore frees up the objective for any query posed as an
optimization. Depending on the choice of variables and optimization objective,
our model can be used for a wide range of queries. We illustrate its
applicability to grasp stability analysis by answering multiple queries on a
number of example grasps. The queries we show here include those already
introduced in~\ref{sec:3d_apps_iterative}: Given applied joint torques, will
the grasp be stable in the presence of a specified external disturbance,
assuming passive resistance effects? Additionally, the formulation developed
in this chapter allows for queries such as the following: Given applied joint
torques, what is the largest disturbance that can be passively resisted in a
given direction? Given a disturbance, what are the optimal joint torques that
a grasp can apply for stability?

Finally, we develop a method to account for uncertainties in the grasp
modeled. Specifically, we can guarantee robustness to uncertainties in contact
normal direction. We use this algorithm to analyze the passive stability of
spacial grasps similar to those we investigated in
Chapter~\ref{sec:3d_apps_iterative}. We also extend our analysis to the
optimization of actuator torques and demonstrate the practical applicability
of our algorithm. We believe these are all useful tools in the context of
grasp analysis.

\section{A formulation using McCormick envelopes}

As mentioned above we will make use of the same grasp model as described in
Chapter~\ref{sec:model3d} but will deviate in our treatment of the friction
constraints in order to explicitly model the MDP in (\ref{eq:mdp}). Of course
one could cast the MDP in its original minimization formulation which is of
course non-convex as the bilinear form in (\ref{eq:mdp}) is not positive
definite~\cite{LIBERTI04}. However, one could attempt to solve the exact
problem (\ref{eq:object_eq}) -- (\ref{eq:underactuation}) using a global
optimization approach based on a convex relaxation of the bilinear forms using
McCormick envelopes~\cite{McCormick1976}, which relies on upper and
lower bounds on the variables involved in the bilinear forms. In order to
illustrate such an approach let us consider an example of a bilinear form
$w(x,y)$ in two dimensions.

\begin{equation}
w(x,y) = x \cdot y
\end{equation}

Given upper and lower bounds such that $x^L \leq x \leq x^U$ and $y^L \leq y
\leq y^U$ we replace the $xy$ term with variable $w$ and introduce the
following linear inequality constraints:

\begin{align}
w \leq&~x^U y + x y^L - x^U y^L \\
w \leq&~x y^U + x^L y - x^L y^U \\
w \geq&~x^L y + x y^L - x^L y^L \\
w \geq&~x^U y + x y^U - x^U y^U
\label{eq:mccormack}
\end{align}

\begin{figure}[!t]
\centering

\subfigure[Step 1]{\includegraphics[width=0.45\linewidth]{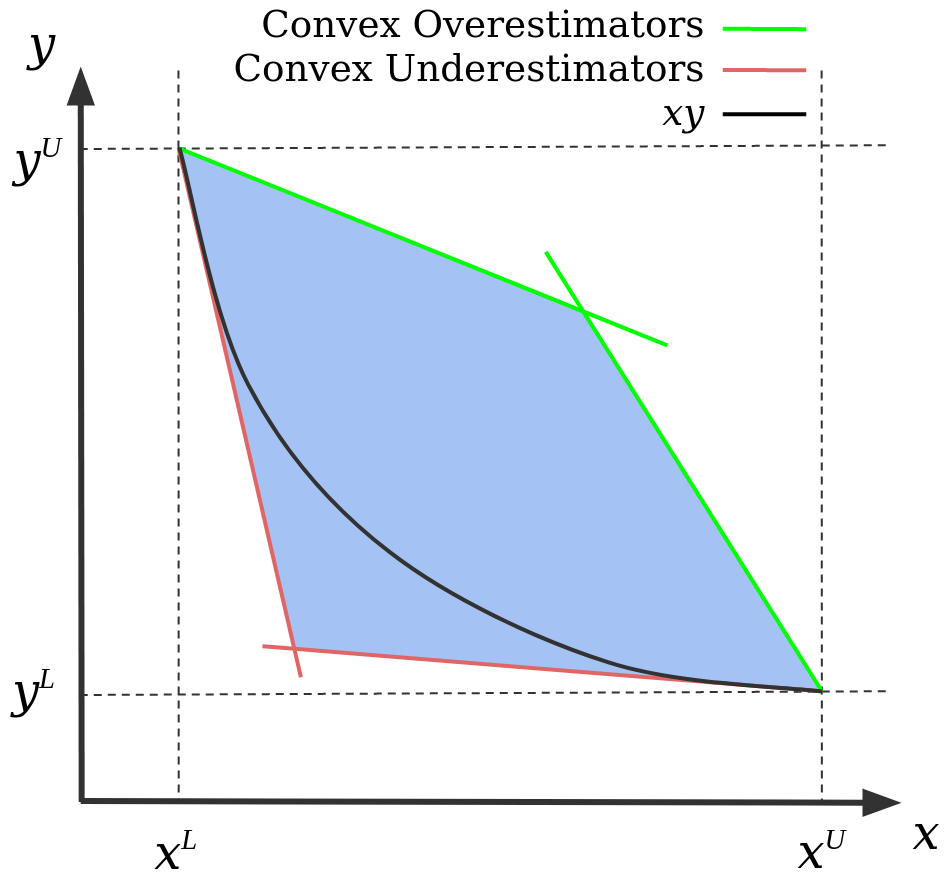}%
\label{fig:step1}}
\subfigure[Step 2]{\includegraphics[width=0.45\linewidth]{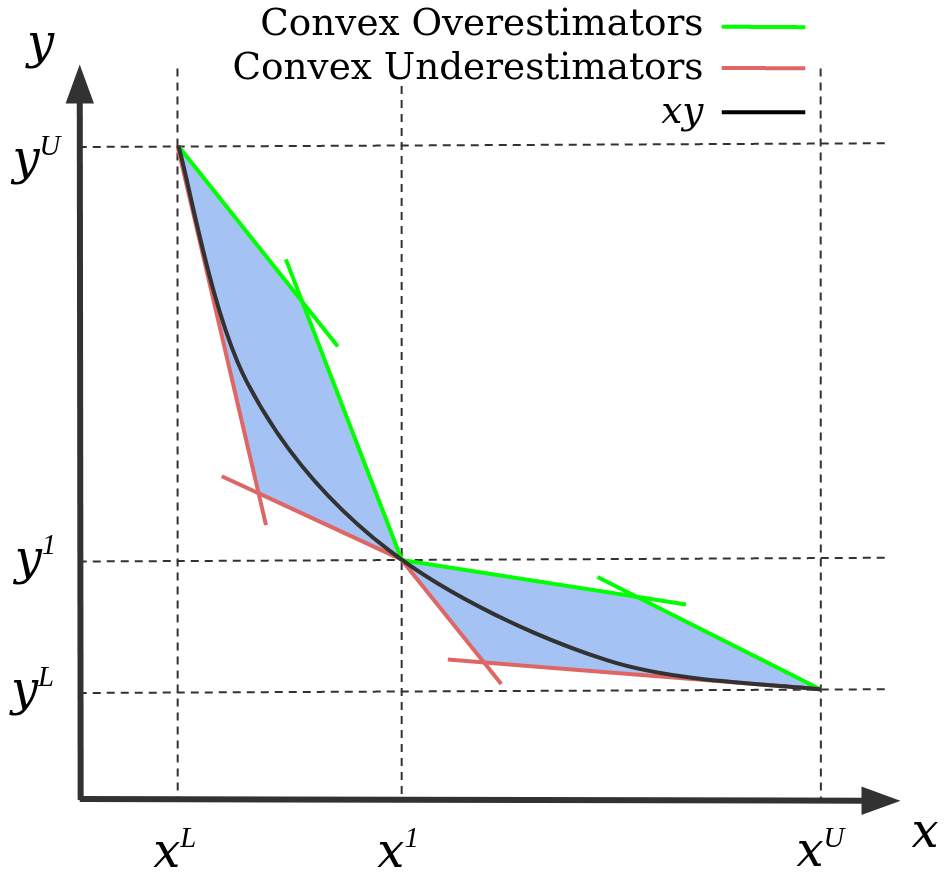}%
\label{fig:step2}}
\subfigure[Step 3]{\includegraphics[width=0.45\linewidth]{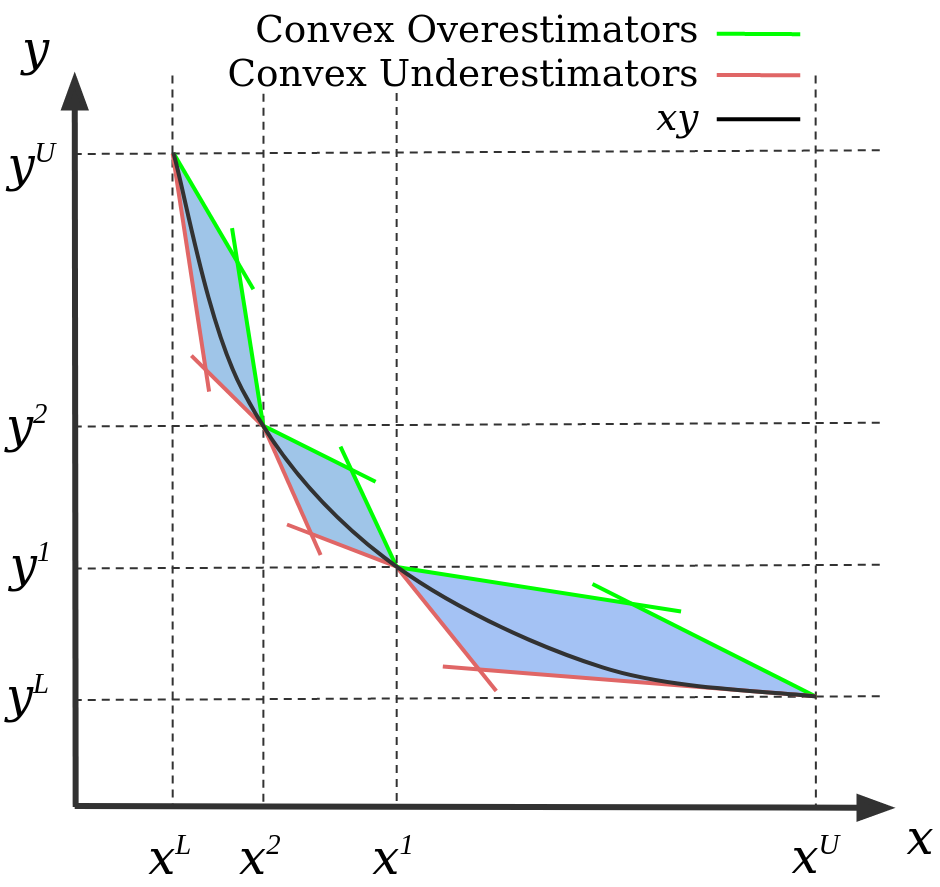}%
\label{fig:step3}}

\caption{McCormick envelopes for the bilinear form $w(x,y) = x \cdot y$.}

\label{fig:mccormick}
\end{figure}

A graphical interpretation of this convex relaxation to the bilinear form is
shown in Fig.~\ref{fig:step1}. The linear inequalities we introduced provide
convex underestimators and overestimators also known as a McCormick envelope.
Of course a solution to this relaxed problem will not in general be a solution
to the exact problem. It will, however, provide a lower bound for the
objective value. Tightening the relaxation will result in a lower bound closer
to the exact solution. Hence, tight relaxations are imperative for obtaining
accurate results.

A common method to obtain such a tight relaxation is its successive
hierarchical tightening. We search for a solution in a coarse relaxation and
the refine just the envelope containing it (see Fig.~\ref{fig:mccormick}).
Algorithms such as Spacial Branch-and-Bound (sBB)~\cite{TUY16} can be used to
efficiently refine and prune parts of the solution space and find global
minima of such non-convex optimization problems. Implementations include the
$\alpha$BB algorithm~\cite{FLOUDAS98} or the BARON~\cite{BARON} software for
global optimization.

Of course this approach generalizes to bilinear forms in higher dimensions
such as that encountered in the maximum dissipation principle (\ref{eq:mdp}).
However, it requires knowledge of upper and lower bounds for the variables
involved in the bilinear form. In the grasping problems we are trying to solve
such bounds are not explicit in the model and are difficult to estimate.

\section{Convex relaxation of the Coulomb friction law}\label{sec:mip}

Taking inspiration from these convex relaxation techniques we make use of the
particular structure of friction cones to derive a relaxation that does not
require variable bounds. This is useful since in the grasping problem there
are no explicit bounds on contact forces and particularly motions and they are
therefore difficult to predetermine. Furthermore, in our approach we formulate
the MDP as a non-convex constraint, which we relax instead of the optimization
objective. This allows us to use the objective to solve for interesting grasp
characteristics such as the optimum actuator commands.

\begin{figure}[t!]
\centerline{
\subfigure[Coulomb friction cone]{\includegraphics[width=0.4\columnwidth]{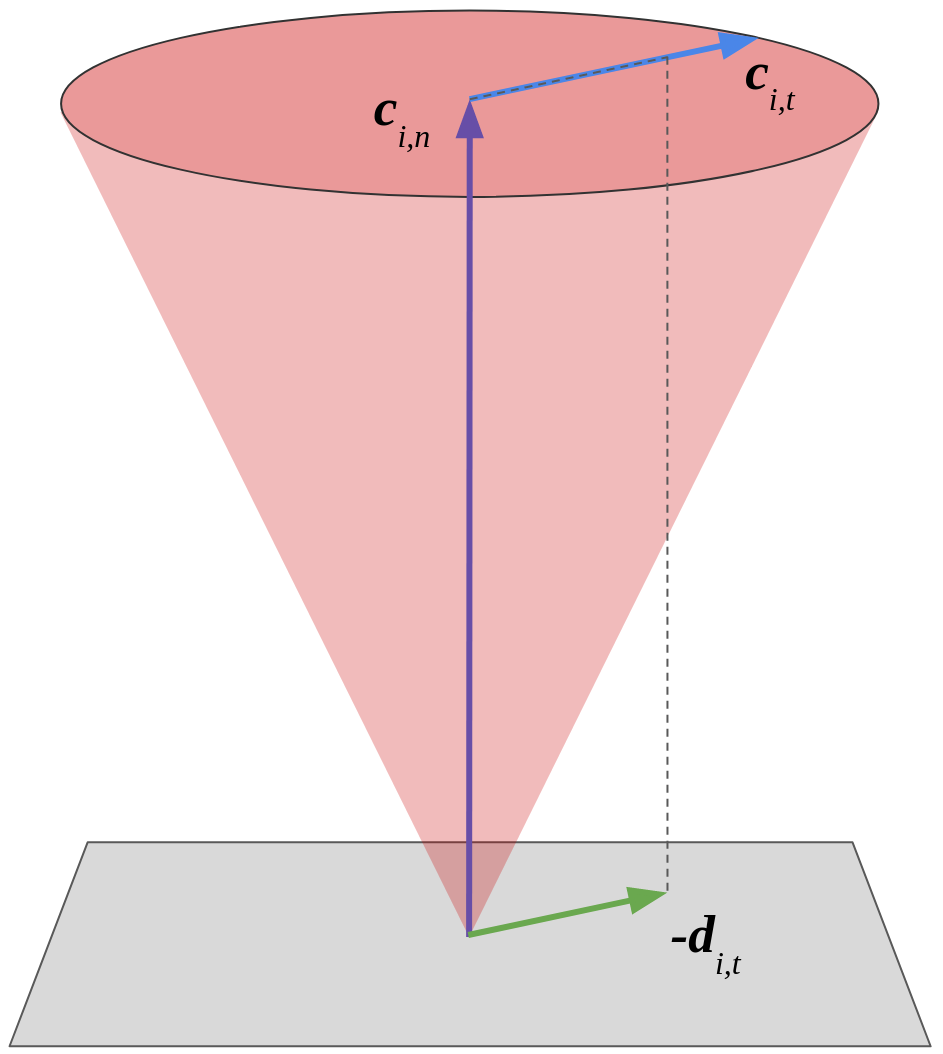}\label{fig:pyramid_a}}\hfill%
\subfigure[Pyramidal approximation]{\includegraphics[width=0.4\columnwidth]{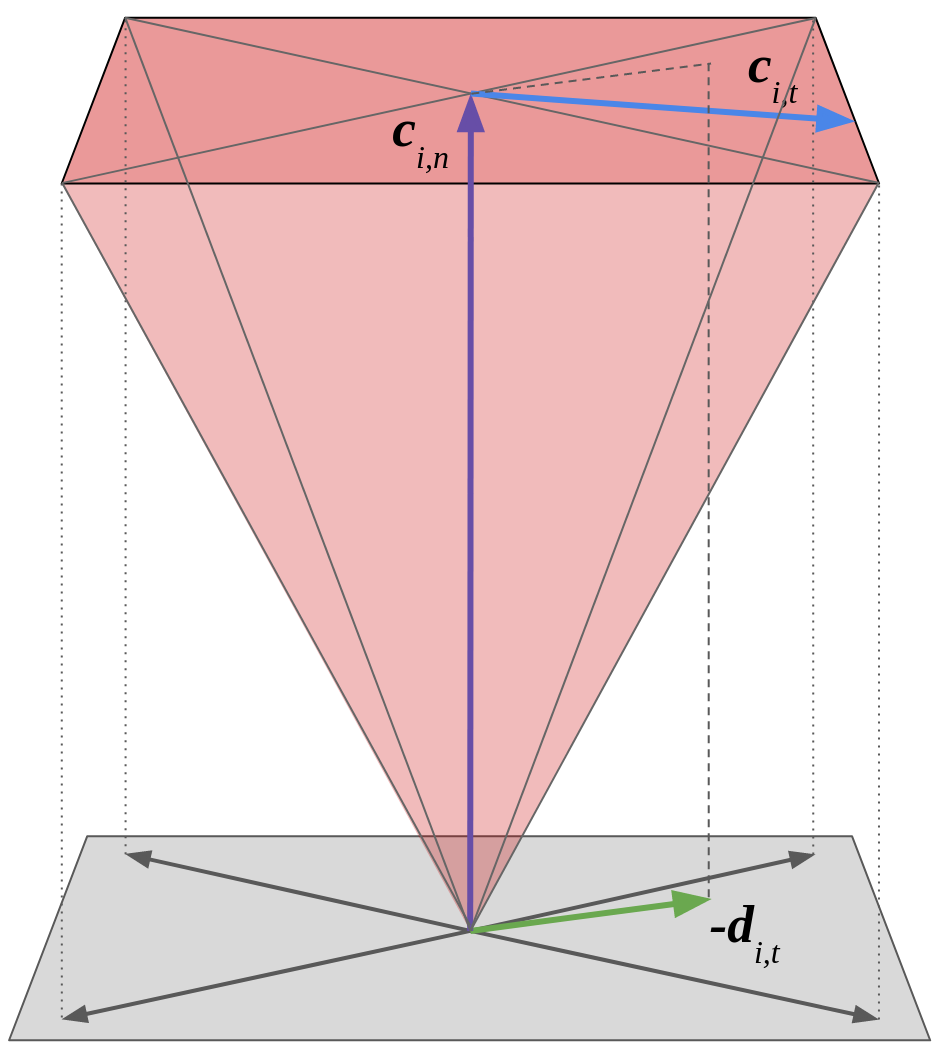}\label{fig:pyramid_b}}
}

\caption{Illustration of the exact Coulomb friction cone and a pyramidal
approximation with $k=4$. The black arrows positively span the contact
tangent plane and make up the matrix of basis vectors
$\bm{D}_i$.}

\label{fig:pyramid}
\end{figure}

We start from the common linearized friction model~\cite{MILLER03B} which
replaces the circular friction cone at contact $i$ with its discretization as
a polygonal cone $\hat{\mathcal{F}}_i$ (see Fig.~\ref{fig:pyramid}.) Matrix
$\bm{D}_i \in \mathbb{R}^{2 \times k}$ contains as its columns a set of $k$
vectors that positively span the contact tangential plane and thus the space
of possible friction forces. Frictional forces can now be expressed as
positive linear combinations of these so called \textit{friction edges} with
weights $\bm{\beta}_i \in \mathbb{R}^k_{\geq 0}$. Vector $\bm{e}=[1,1,...,1]^T
\in \mathbb{R}^k$ sums the weights $\bm{\beta}_i$. Inequality constraints on a
vector are to be understood in a piecewise fashion.

\begin{align}
\bm{c}_{i,t} &= \bm{D}_i \bm{\beta}_i \in \hat{\mathcal{F}}_i\label{eq:betas}\\
\hat{\mathcal{F}}_i( \mu_i, c_{i,n} ) 
&= \{ \bm{D}_i \bm{\beta}_i : \bm{\beta} \geq 0,~\bm{e}^T \bm{\beta} \leq \mu_i c_{i,n} \}
\label{eq:fric_discretization}
\end{align}

\begin{figure}[t!]
\centering

\includegraphics[width=0.6\columnwidth]{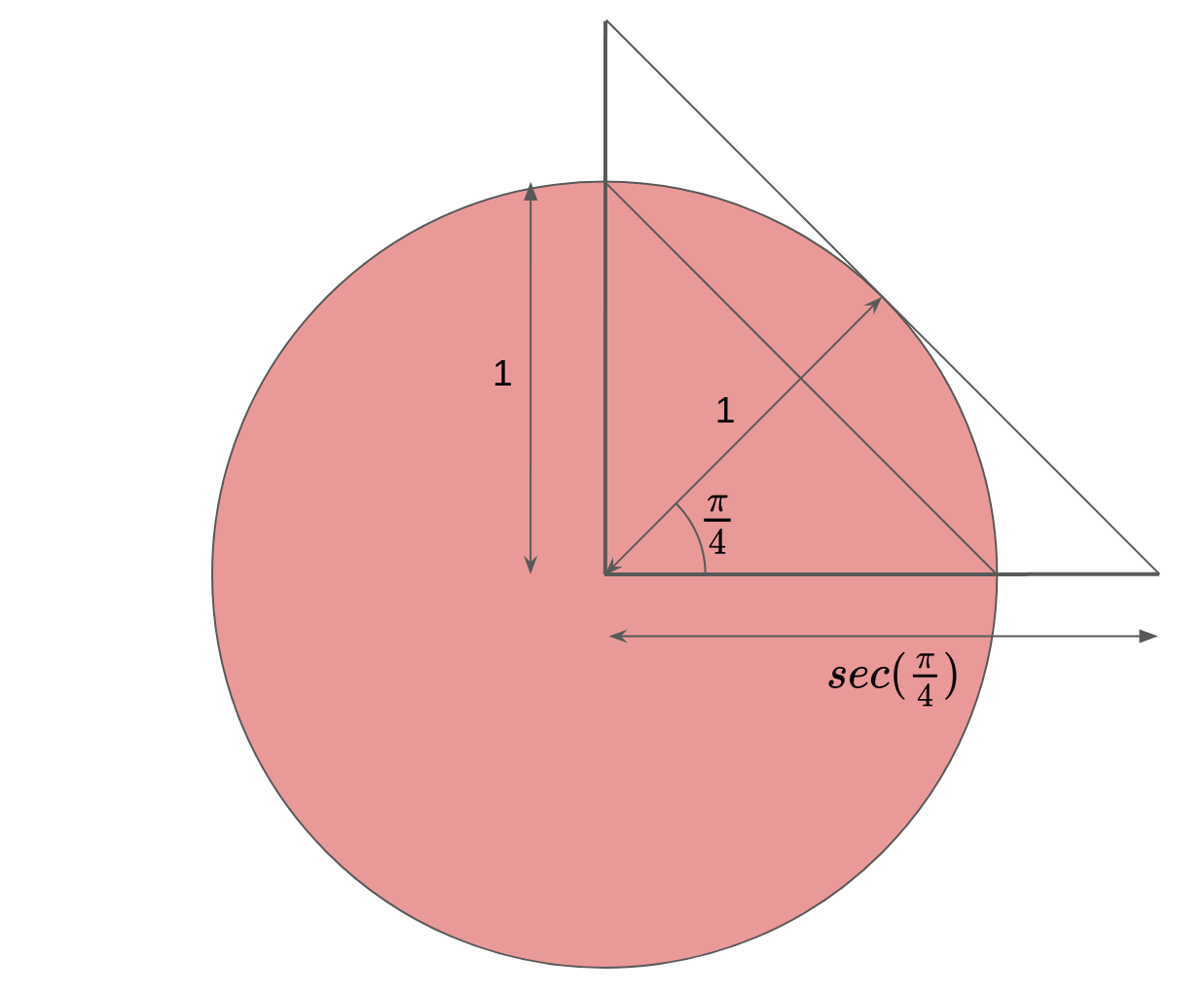}

\caption{Slice through the friction cone showing two possible lengths of
friction edges for a discretization with $k=4$. The shorter friction edge
length leads to a discretized cone fully inscribed by the circular cone,
while the longer friction edge length leads to a discretized cone that itself
fully contains the circular cone.}

\label{fig:relaxation}
\end{figure}

Traditionally, the basis vectors in $\bm{D}_i$ are unit vectors. Thus, the
discretized friction cone is completely inscribed by the exact circular cone
(see Fig.~\ref{fig:relaxation}.) This leads to an approximation to the
possible friction forces that is conservative in the context of grasp
stability. In our case, however, in order to derive a valid convex relaxation
we require the relaxed discretized friction cone to contain the exact cone.
This is important in order to guarantee that we can find a solution to the
relaxed problem if one exists to the exact problem. Thus, we instead choose
basis vectors of length $f$ for Matrix $\bm{D}_i$.

\begin{equation}
f = sec(\frac{\pi}{k})
\label{eq:fric_len}
\end{equation}

Let us now also express relative tangential contact motion as a weighted
combination of the same friction edges, with weights $\bm{\alpha}_i \in
\mathbb{R}^k_{\geq 0}$. For reasons that will soon become apparent, we
actually choose to express the opposite of the tangential motion:

\begin{equation}
\bm{D}_i\bm{\alpha}_i = -\bm{d}_{i,t},~\bm{\alpha}_i \geq 0\label{eq:alphas}
\end{equation}

If the friction edges in $\bm{D}_i$ are arranged in an ordered fashion such
that neighboring friction edges in the tangent plane are also neighbors in
$\bm{D_i}$ we can constrain friction to (approximately) oppose motion by
requiring that the \textit{friction force lie in the same sector of the
linearized friction cone as the negative of the tangential contact motion.}
Without loss of the above properties, we require that all but two components
of $\bm{\beta}_i$ must be zero and that non-zero components are either
consecutive or lie at the first and last positions of vector $\bm{\beta}_i$.
This can be achieved by constraining $\bm{\beta}_i$ with a special ordered set
$\bm{z}_i \in \mathbb{R}_{\geq 0}^{k+1}$ of type 2 (SOS2)~\cite{Beale1976},
which has one more component than $\bm{\beta}_i$ itself.

\begin{equation}
\beta_{i,1} \leq z_{i,1} + z_{i,k+1},~\beta_{i,2} \leq z_{i,2},~...,~\beta_{i,k} \leq z_{i,k},~\bm{z}\in\text{SOS2}\label{eq:sos_beta}
\end{equation}

A special ordered set of type 2 is a set of ordered non-negative numbers of
which at most two can be non-zero. If two numbers in the SOS2 are non-negative
they must be consecutive in their ordering. This type of constraint can be
formulated in the framework of mixed integer problems and therefore is
admissible to problems to be solved by MIP solvers. We now similarly constrain
the weights $\bm{\alpha}$ that determine relative motion with the same SOS2 as
in (\ref{eq:sos_beta}).

\begin{eqnarray}
\alpha_{i,1} \leq z_{i,1} + z_{i,k+1},~\alpha_{i,2} \leq z_{i,2},~...,~\alpha_{i,k} \leq z_{i,k}\label{eq:sos_alpha}
\end{eqnarray}

Note that these constraints hold for both sliding as well as rolling contacts.
If a contact is rolling then all components of $\bm{\alpha}_i$ must be zero.
Thus, any two consecutive components of $\bm{z}_i$ may be non-zero and since
$\bm{\beta}_i$ are the weights of the basis vectors in $\bm{D}_i$ the friction
force may point in any direction in the tangent plane. If the contact slides
then some components of $\bm{\alpha}_i$ must be nonzero. As both
$\bm{\alpha}_i$ and $\bm{\beta}_i$ are constrained by $\bm{z}_i$ and only two
consecutive components of $\bm{z}_i$ may be nonzero, this means the friction
force must lie in the same sector of the friction pyramid as the negative of
the relative tangential contact motion, but may not necessarily be collinear
(see Fig.~\ref{fig:pyramid_b}.)

\begin{figure}[t!]
\centering

\includegraphics[width=0.5\columnwidth]{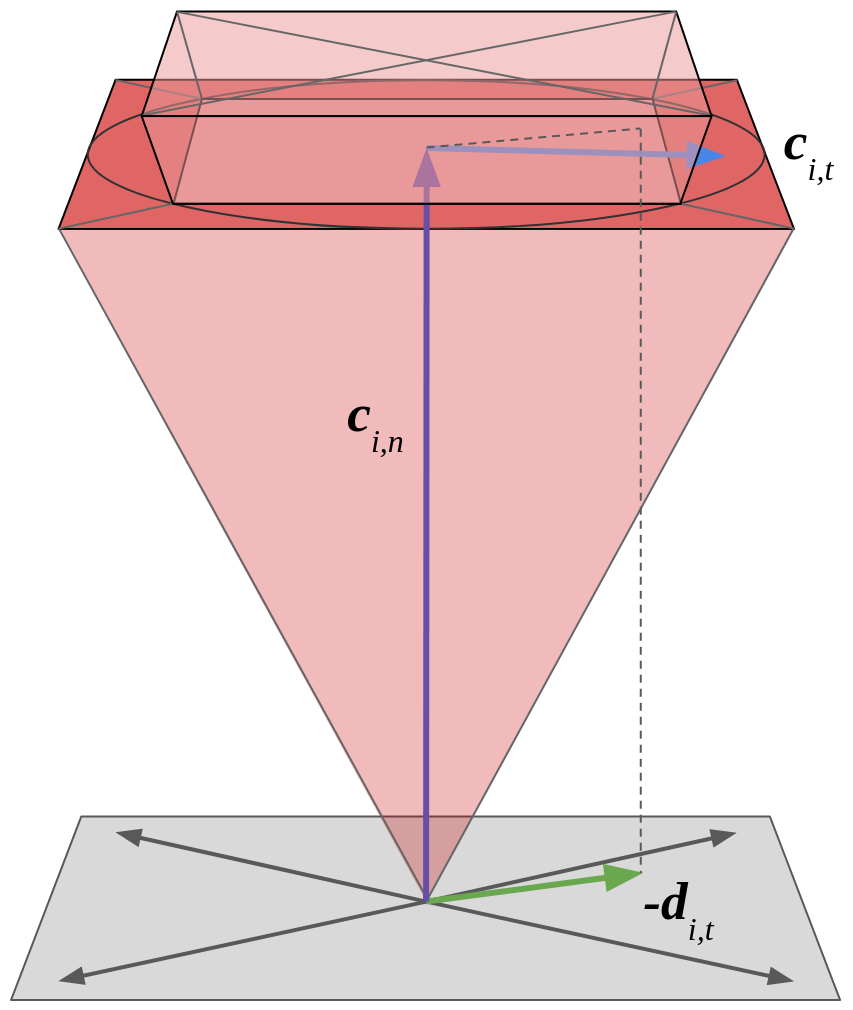}

\caption{Complete convex relaxation of the circular friction cone. At a
rolling contact the contact force may lie anywhere within the larger
discretized cone. At a sliding contact the friction force must lie in the same
sector as $-\bm{d}_{i,t}$. Furthermore the contact force must remain inside
the larger, but outside the smaller discretized cone.}

\label{fig:cone_in_cone}
\end{figure}

Finally, we can constrain the magnitude of the friction force in addition to
its direction. In the exact Coulomb model the friction magnitude must be
maximized at sliding contacts, while for stationary contacts it only has an
upper bound. In our relaxation this means that at rolling contacts the
friction force may lie anywhere within the discretized cone containing the
exact circular cone. At a sliding contact we require that the
\textit{friction force lies within the discretized cone containing the
circular cone, but outside the smaller discretized cone that is itself
contained by the exact cone}. See Fig.~\ref{fig:cone_in_cone} for an
illustration of this relaxation. Note, that in either case the relaxation
contains the solution set of the exact Couliomb friction model.

Defining vector $\bm{f} = [f, f, ..., f]^T \in \mathbb{R}^k$ containing the
lengths of the friction edges (\ref{eq:fric_len}) making up the friction cone
approximation we can express this constraint as follows:

\begin{subequations}
\begin{empheq}[left=\empheqlbrace] {align}
\bm{e}^T \bm{\beta}_i &\leq \mu_i c_{i,n}~, & \text{if}~\bm{e}^T \bm{\alpha}_i = 0\\
\bm{e}^T \bm{\beta}_i &\leq \mu_i c_{i,n}~,~\bm{f}^T \bm{\beta}_i \geq \mu_i c_{i,n}~, & \text{otherwise}
\end{empheq}\label{eq:fric_cases}
\end{subequations}

Constraint (\ref{eq:fric_cases}) can also be included in an MIP as an
indicator constraint using a binary decision variable $w_i \in \{0,1\}$.

\begin{subequations}
\begin{empheq}[left=\empheqlbrace] {align}
w_i = 0 &\implies &\bm{e}^T \bm{\beta}_i \leq \mu_i c_{i,n}~,&\quad \bm{e}^T \bm{\alpha}_i = 0 \\
w_i = 1 &\implies &\bm{e}^T \bm{\beta}_i \leq \mu_i c_{i,n}~,&\quad \bm{f}^T \bm{\beta}_i \geq \mu_i c_{i,n}
\end{empheq}\label{eq:fric_binary}
\end{subequations}

We now have a complete model of friction. For a finite value of $k$, this
model is approximate. However, in the limit as $k \rightarrow \infty$ the
indicator constraints (\ref{eq:fric_binary}) are equivalent to the Coulomb
friction model in (\ref{eq:friction_bound}) \& (\ref{eq:mdp}).

\section{Successive hierarchical refinement}\label{sec:refinement}

We can solve the complete system described by constraints
(\ref{eq:object_eq})-(\ref{eq:unilaterality_again}),
(\ref{eq:joint_unilaterality})-(\ref{eq:underactuation}) and the new piecewise
convex relaxation of Coulomb friction (\ref{eq:fric_binary}) as an MIP with
algorithms such as branch and bound. In order to improve our approximation, we
could choose a high number of edges $k$ for the discretized friction cones. In
practice, however, that approach is not feasible as the time taken to solve an
MIP is sensitive to the number of integer variables in the problem.  As SOS2
constraints are implemented using binary variables a highly refined friction
cone approximation quickly becomes computationally intractable. This is the
trade-off traditionally encountered with discretization methods: coarse
discretizations provide only rough approximations of the exact constraints,
while high resolutions discretizations are computationally intractable.

Our approach is based on the key insight that one can obtain an equally
accurate solution by solving a problem with a coarse friction cone
approximation, and \textit{successively refining the linearized friction
constraints only in the region where friction forces arise.}  Our approach
thus proceeds as follows:

\begin{itemize}

\item We solve our problem using a coarse approximation of the friction cone
(few friction edges).  From the solution, we identify the sector of the
linearized cone (the area between two edges) where  both the friction force
and the negative relative motion lie.

\item To obtain a tighter bound, we add new friction edges that refine
\textit{only the sector identified above}.  We then repeat the procedure with
the new, selectively refined version of the friction cone.

\end{itemize}

\begin{figure}[!t]
\centering

\hspace{0.5in}
\subfigure[Exact]{\includegraphics[width=2.0in]{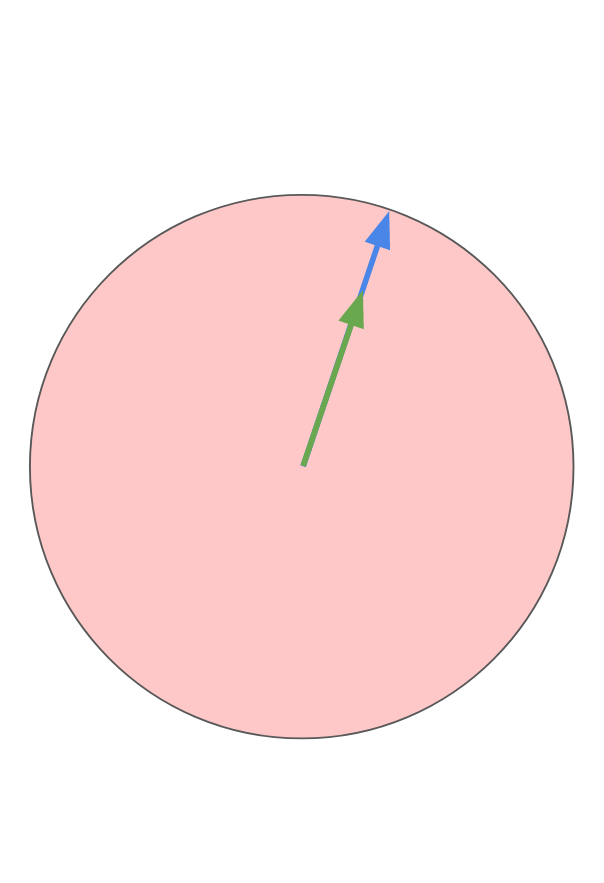}
\label{fig:cone4}}
\hspace{0.5in}
\subfigure[Step 1]{\includegraphics[width=3.1in]{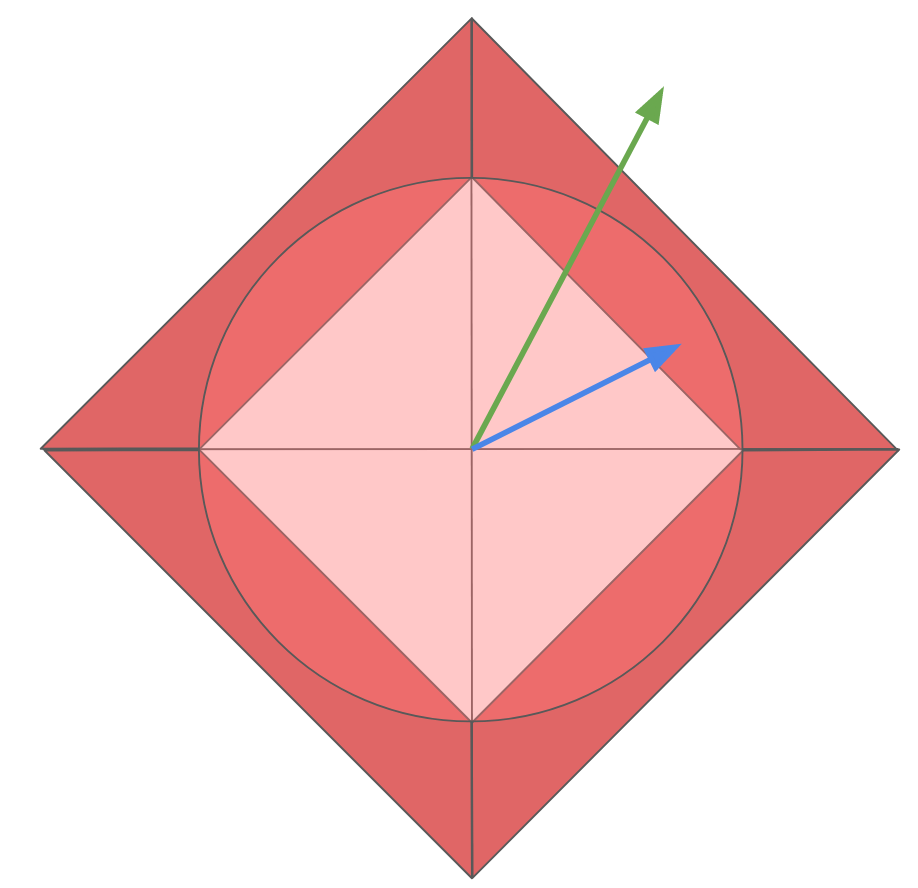}%
\label{fig:cone1}}
\subfigure[Step 2]{\includegraphics[width=3.1in]{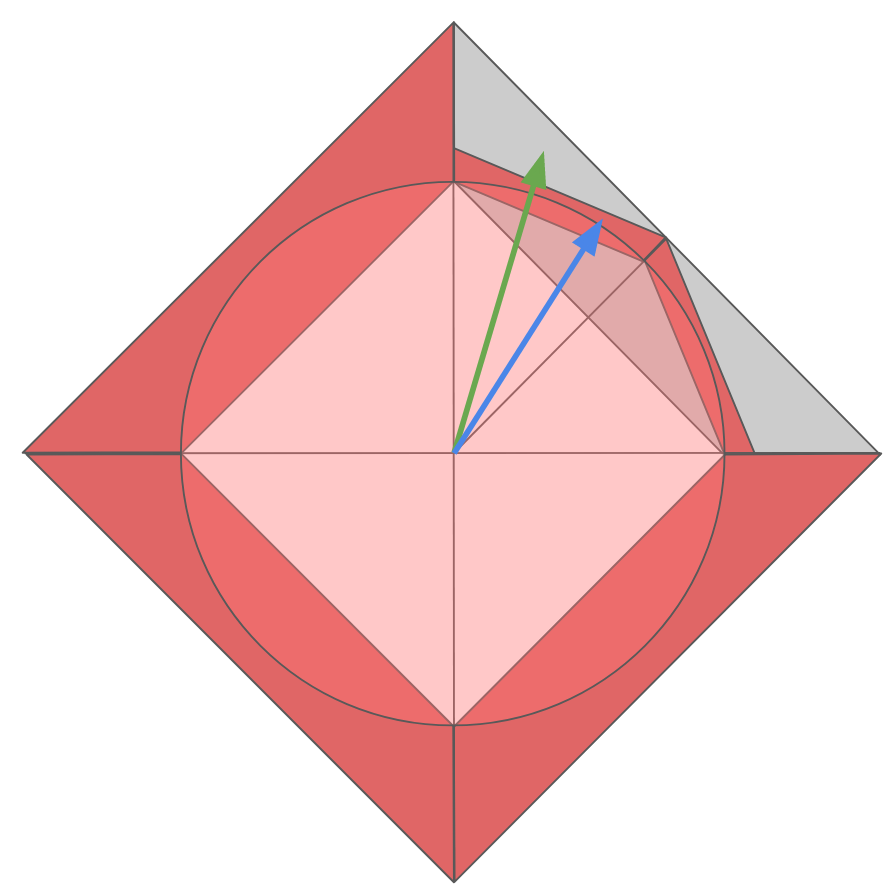}%
\label{fig:cone2}}
\subfigure[Step 3]{\includegraphics[width=3.1in]{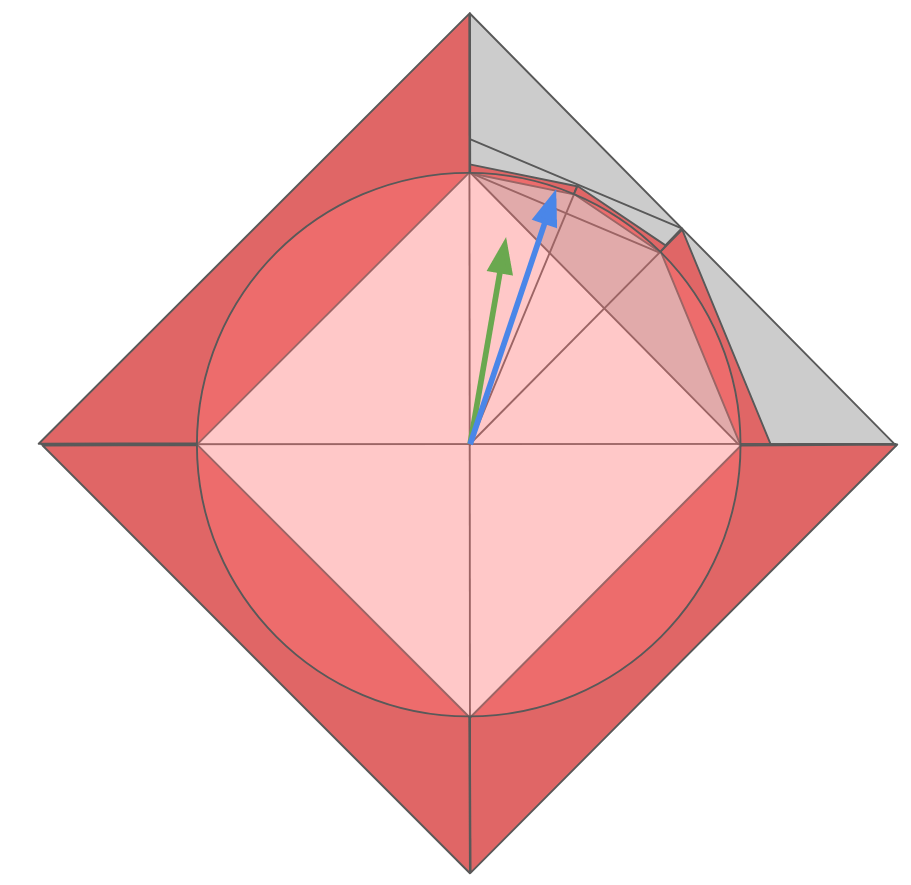}%
\label{fig:cone3}}

\caption{A slice through the exact circular friction cone as well as a
piecewise convex relaxation. Shown are the first three steps of our algorithm
refining the friction law relaxation successively and locally. The dark red
regions are the space of feasible friction forces at a sliding contact. At a
rolling contact the light red regions are added to the feasible friction force
space. The green and blue arrows are $\bm{d}_{i,t} = \bm{D}_i \bm{\alpha}_i$
and $\bm{c}_{i,t} = \bm{D}_i \bm{\beta}_i$ respectively, which drive
refinement of their local friction cone sector. In gray we show the relaxation
from previous steps to show how a coarse relaxation contains the entire
solution space of all successively tighter relaxations.}

\label{fig:cone}
\end{figure}

Consider a rough approximation with four friction edges
(Fig.~\ref{fig:cone1}.) If we allow rolling friction to reside inside the
areas shaded in either shade of red, while sliding friction must lie within
the dark red border, the space of allowable solutions to the exact problem is
contained inside our linear and piecewise convex approximation. Assume that,
at this level of refinement, there is a solution to an equilibrium problem,
with a friction force residing inside the upper right sector. We refine this
sector, taking care that the space of allowable solutions to the exact problem
is contained inside within this new refinement refinement (see
Fig.~\ref{fig:cone2}.) We continue this procedure (Fig.~\ref{fig:cone3} etc.)
until one of two things happen: we either reach a level of refinement where no
solution exists, or we refine down to the point where the active sector is as
small as we want it to be, bringing us arbitrarily close to the solution to
the exact problem.

Note that the friction cone in the coarse approximation of
Fig.~\ref{fig:cone1} differs in a subtle way from that in
Figs.~\ref{fig:relaxation} \& \ref{fig:cone_in_cone}: The edges of the larger
discretized cone containing the exact solution space are no longer tangent to
the exact solution space. Instead, the discretized friction cone is slightly
enlarged. This was done to maintain a crucial characteristic of convex
relaxations in optimization problems: The minimum found in a coarse convex
relaxation should provide a lower bound for the minimum found in a more
refined and thus tighter relaxation. This requires that the solution space at
of a coarse relaxation completely contains the solution space of a tighter
more refined relaxation.

Thus, only at the final level of refinement can the edge of the discretized
friction cone lie tangent to the exact circular cone (Fig.~\ref{fig:cone3}.)
In consequence, the discretized cone at a coarser level (Fig.~\ref{fig:cone2})
must be offset from the exact cone in order for the relaxation to contain the
two finer relaxations. The same argument means that the relaxation in
Fig.~\ref{fig:cone1} must be offset from the circular cone in order for it to
contain the relaxations in Fig.~\ref{fig:cone2} in their entirety.

Thus, we must make a slight modification to the friction edge lengths computed
in (\ref{eq:fric_len}). Let us pick the initial basis vectors in $\bm{D}_i$
such that the angle $\gamma$ between all pairs of successive vectors is equal.
We pick an initial angle $\gamma = \pi / 2$. We refine our polyhedral friction
cone by bisecting sectors defined by the non-zero components of
$\bm{\alpha}_i$ and define the angle at which to stop refinement as $\gamma /
2^q$. We now find the required length $l_1$ of these initial friction edges
such that the initial solution set contains the solution sets at all further
refinement levels.

\begin{equation}
l_1 = \prod_{r=1}^{q+1} \sec(\frac{\gamma}{2^r})
\label{eq:l1}
\end{equation}

Thus, we modify the friction edges in $\bm{D}_i$ accordingly. We define vector
$\bm{f}_i$ to contain the lengths of the friction edges making up the friction
cone approximation in an order corresponding to the order of the weights
$\bm{\beta}_i$. Constraints (\ref{eq:fric_binary}) are modified accordingly.

\begin{subequations}
\begin{empheq}[left=\empheqlbrace] {align}
w_i = 0 &\implies &\bm{e}^T \bm{\beta}_i \leq \mu_i c_{i,n}~,&\quad \bm{e}^T \bm{\alpha}_i = 0 \\
w_i = 1 &\implies &\bm{e}^T \bm{\beta}_i \leq \mu_i c_{i,n}~,&\quad \bm{f}_i^T \bm{\beta}_i \geq \mu_i c_{i,n}
\end{empheq}\label{eq:fric_refine}
\end{subequations}

We are now ready to solve the initial coarse relaxation problem defined by
(\ref{eq:object_eq})-(\ref{eq:unilaterality_again}),
(\ref{eq:joint_unilaterality})-(\ref{eq:underactuation}) and
(\ref{eq:fric_refine}). We find the two active friction edges $d_1$ and $d_2$
and create three new edges that point in the direction of $d_1$, $d_1+d_2$ and
$d_2$ and have magnitude $l_2$. For this and all following refinements we have

\begin{equation}
l_p = \prod_{r=p}^{q+1} \sec(\frac{\gamma}{2^r})\label{eq:length}
\end{equation}

where $p$ is the level of refinement of the sectors to be created. We insert
the new friction edges between $d_1$ and $d_2$ in matrix $\bm{D}_i$ and their
lengths in $\bm{f}_i$. Finally we remove any redundant friction edges (edges
that are identical or edges that lie between any such edges). Solving this new
problem we can continue refining the friction discretization until the angle
of the active sectors at all contacts reach an angle of $\gamma / 2^q$. We do
not further refine any sector that has already reached this threshold. The
overall method is shown in Algorithm~\ref{alg:refinement}.

\begin{algorithm}[t]
\caption{Grasp analysis through successive relaxation}\label{alg:refinement}
\begin{algorithmic}[0]
\Procedure{Relaxation refinement}{}
  \State \textbf{Input:}
  \State ~~~~$O$ --- objective function
  \State ~~~~$C$ --- additional constraints
  \State ~~~~$\gamma$ --- initial refinement level (angle between friction edges)
  \State ~~~~$q$ --- maximum refinement level desired
  \State Initialize $\bm{D}$ with basis vectors of length $l_1$ as defined in (\ref{eq:l1})
  \Do
    \State Minimize $O$ subject to (\ref{eq:object_eq})-(\ref{eq:unilaterality_again}), (\ref{eq:joint_unilaterality})-(\ref{eq:underactuation}), (\ref{eq:fric_refine}) and $C$.
    \If{no solution exists}
      \State \textbf{return} no feasible solution
    \EndIf
    \State $\text{refinement\_needed} \gets \texttt{False}$
    \For{each contact $i$}
      \State Find active edges in $\bm{D}_i$
      \State $\delta_i \gets \text{angle between active edges}$
      \If{$\delta_i > \gamma / 2^q$}
        \State $p \gets \log_2 (\gamma / \delta_i) + 2$
        \State Add new edges to $\bm{D}_i$ of length $l_p$ according to (\ref{eq:length})
        \State Remove redundant edges from $\bm{D}_i$
        \State $\text{refinement\_needed} \gets \texttt{True}$
      \EndIf
    \EndFor
  \doWhile{$\text{refinement\_needed}$}
\State \textbf{return} solution
\EndProcedure
\end{algorithmic}
\end{algorithm}

Note that this convex relaxation does not rely on explicit bounds on the
problem variables as would be the case were we using McCormick envelopes.
Instead, we use the bounds implicit in the friction cone constraint
(\ref{eq:friction_bound}) in our relaxation. This is also a further advantage of
our approach as the relaxation has an easily understood physical
interpretation.

Recall that one of the defining characteristics of our convex relaxation is
that we have chosen our friction edges so that, at any level of refinement,
\textit{the solution set to the approximate problem contains the solution set
of the exact problem}. This means that if, at any point during the refinement,
no solution can be found that satisfies the convex relaxations, we can
guarantee that no solution can exist to the exact version of the problem
either. This guarantee immediately follows from the properties that the
solution set at any refinement level includes the solution set at the next
level, and that, in the limit, our discretization approaches the exact
constraints. (Note that alternative approaches that work with discretized
friction constraints, such as the LCP formulations discussed in
Chapter~\ref{sec:related_dynamics}, do not exhibit this property.) In
practice, this means that, when no solution exists to the exact equilibrium
problem, our algorithm can determine that very quickly, only solving
relatively coarse refinement levels.

The second advantage this scheme provides is that when a solution does exist,
we can typically refine it to high accuracy (a very close approximation to the
solution of the exact problem) using relatively few friction edges. This is
not theoretically guaranteed: in the worst case, our approach could require
all sectors to be fully refined before finding an adequate solution at the
desired resolution, and may hence perform worse than using a fully refined
friction discretization to begin with. However, we have never found that to be
the case. Typically, only a small region of the discretization must be refined
as the contact forces are also constrained by equilibrium relations
(\ref{eq:object_eq}) \& (\ref{eq:hand_eq}) and will generally point in similar
directions at all levels of refinement, leading to a very localized and
targeted tightening of the relaxation. Thus, this algorithm is efficient
enough to analyze complex grasps on a consumer PC to levels of refinement that
are otherwise unachievable.

\section{Robustness to geometrical uncertainties}\label{sec:robust}

In the above we outlined a grasp model that allows us to analyze the grasp
stability given perfect information about the geometry of the grasp. We assume
we know exactly the contact position and orientation. In practice however we
often encounter uncertainties, which can greatly affect the stability of a
grasp. Even when using tactile sensors in order to locate contacts made
between the hand and the object the contact normals (and hence orientation)
are often difficult to obtain accurately. Therefore we would like to make our
framework robust to discrepancies up to a certain magnitude. We introduce the
method  we use for this here, and illustrate its importance in the following
section.

\begin{figure}
\centering

\includegraphics[width=0.5\columnwidth]{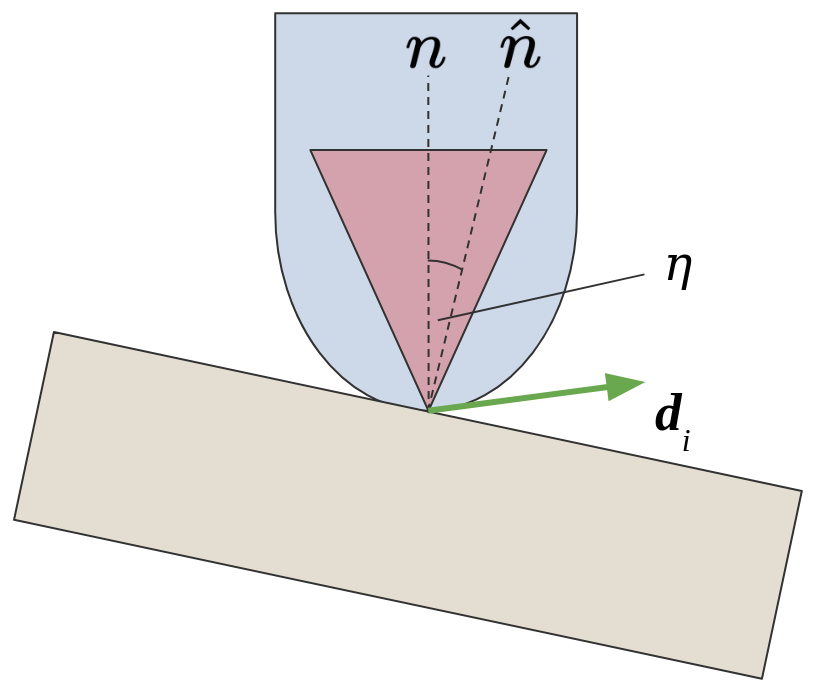}

\caption{Illustration of a worst-case normal within uncertainty $\eta$ given
relative contact motion $\bm{d}_i$: The real normal $\hat{n}$ deviates from
the nominal normal $n$ in such a way as to minimize the contact normal force.}

\label{fig:robustness}
\end{figure}

Let us suppose we have an upper bound on our uncertainty in the contact normal
$\eta$. Thus, the actual contact normal lies in a space of possible contact
normals that deviate by at most angle $\eta$ from the nominal contact normal
$n$. In order for a grasp to be robust to deviations defined by this space we
would like it to be robust in the worst-case. The worst-case contact normal
$\hat{n}$ is one that is at an angle $\eta$ to the nominal contact normal and
such that its projection into the tangent plane points in the same direction
as the relative tangential contact motion. In our space of contact normals
such a normal would be the most effective at unloading the contact and hence
destabilizing the grasp (see Fig.~\ref{fig:robustness}.) The relative contact
motion in the direction of the worst-case normal would then be given by

\begin{equation}
d_{i,\hat{n}} = {d}_{i,n} \cos(\eta) - \left\lVert \bm{d}_{i,t} \right\rVert \sin(\eta)
\label{eq:normal_undertainty}
\end{equation}

However, we need to find a linear approximation for $\left\lVert \bm{d}_{i,t}
\right\rVert$ as including it exactly would introduce a non-convex quadratic
equality constraint. Fortunately we can use the amplitudes of the friction
edges $\alpha$. The summation of the product of all contact motion amplitudes
and the length of the corresponding friction edges gives us an estimate of the
magnitude of the relative tangential contact motion.

\begin{equation}
\left\lVert \bm{d}_{i,t} \right\rVert \approx \sum_{s=1}^{k} f_{i,s} \alpha_{i,s} = \sum_{s=1}^{k} l_p \alpha_{i,s} \label{eq:overestimate}
\end{equation}

Here $p$ is the level of refinement of the active sector (i.e. the sector
corresponding to the nonzero components of $\bm{\alpha}_i$.) The problem with
this formulation is that due to the triangle inequality this overestimates the
relative tangential contact motion. This effect diminishes at finer
resolutions as friction edges become closer to parallel, but the destabilizing
effect is potentially larger at coarser resolution. However, recall that our
refinement method requires that the solution set at coarser levels includes
the solution set at finer levels. We thus require  the destabilizing effect to
be weaker at coarse resolutions and become stronger approaching its exact
value as $p \rightarrow \infty$. 

Note that (\ref{eq:overestimate}) is exact for contact motion parallel to to
any friction edge but for any tangential motion that lies between two edges it
overestimates the magnitude of the relative sliding motion. The overestimation
is most pronounced if the sliding motion exactly bisects the two adjacent
friction edges. Thus, in order to prevent overestimation of sliding motion
while retaining the tightest possible relaxation we must modify
(\ref{eq:overestimate}) such that it instead is exact for motions bisecting
any two adjacent friction edges and underestimates sliding motion everywhere
else.

\begin{figure}
\centering

\includegraphics[width=0.5\columnwidth]{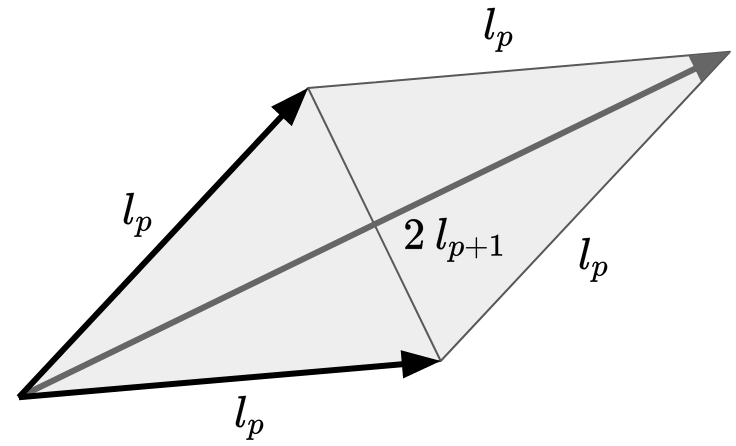}

\caption{The vector addition of two adjacent friction edges of length $l_p$ is
a vector of length $2 l_{p+1}$.}

\label{fig:vectors}
\end{figure}

Let us first find out by how much (\ref{eq:overestimate}) overestimates
tangential motion in the worst case. Recall that $f_{i,s} = l_p$ can be calculated
using the refinement level $p$ of the corresponding friction edges and
(\ref{eq:length}). Consider two generic adjacent friction edges of length
$l_p$. Fig.~\ref{fig:vectors} illustrates that the worst-case overestimation
occurs for vectors bisecting the two friction edges. Furthermore, due to the
geometry of the relaxation described in Fig.~\ref{fig:cone} and
(\ref{eq:length}) we know that the vector addition of two adjacent friction
edges is a vector of length $2 l_{p+1}$. Thus, the factor by which
(\ref{eq:overestimate}) overestimates tangential motion that bisects any two
friction edges of length $l_p$ is

\begin{equation}
\frac{l_p}{l_{p+1}}
\end{equation}

We can now divide (\ref{eq:overestimate}) by this factor such that we obtain
an approximation of the tangential motion magnitude that underestimates
everywhere except at the midpoint between two edges where it is exact.

\begin{equation} 
\left\lVert \bm{d}_{i,t} \right\rVert \approx \sum_{s=1}^{k} l_{p+1} \alpha_{i,s} 
\end{equation}

As $p \rightarrow \infty$ this estimation becomes exact. We can now plug this
back into (\ref{eq:normal_undertainty}) for a linear approximation of the
worst-case relative normal contact motion. 

\begin{equation}
d_{i,\hat{n}} = {d}_{i,n} \cos(\eta) - \sin(\eta) \sum_{s=1}^{k} l_{p+1} \alpha_{i,s}
\label{eq:uncertainty}
\end{equation}

We now replace the normal relative contact motion in
(\ref{eq:unilaterality_again}) with $d_{i,\hat{n}}$ in order to obtain
solutions that are robust to uncertainties in contact normal up to an angular
discrepancy of $\eta$.

\section{Application to grasp analysis in three dimensions}\label{sec:3d_apps}

\subsection{Existence problems}

The range of queries we can answer with the approach described in
Chapter~\ref{sec:refinement} is less limited than that of the iterative
approach discussed in Chapter~\ref{sec:three_dim}. However, existence problems
are still the most simple queries to solve and are a good starting point
before moving on to more complex queries.

\begin{figure}[t!]
\centering

\includegraphics[width=0.75\columnwidth]{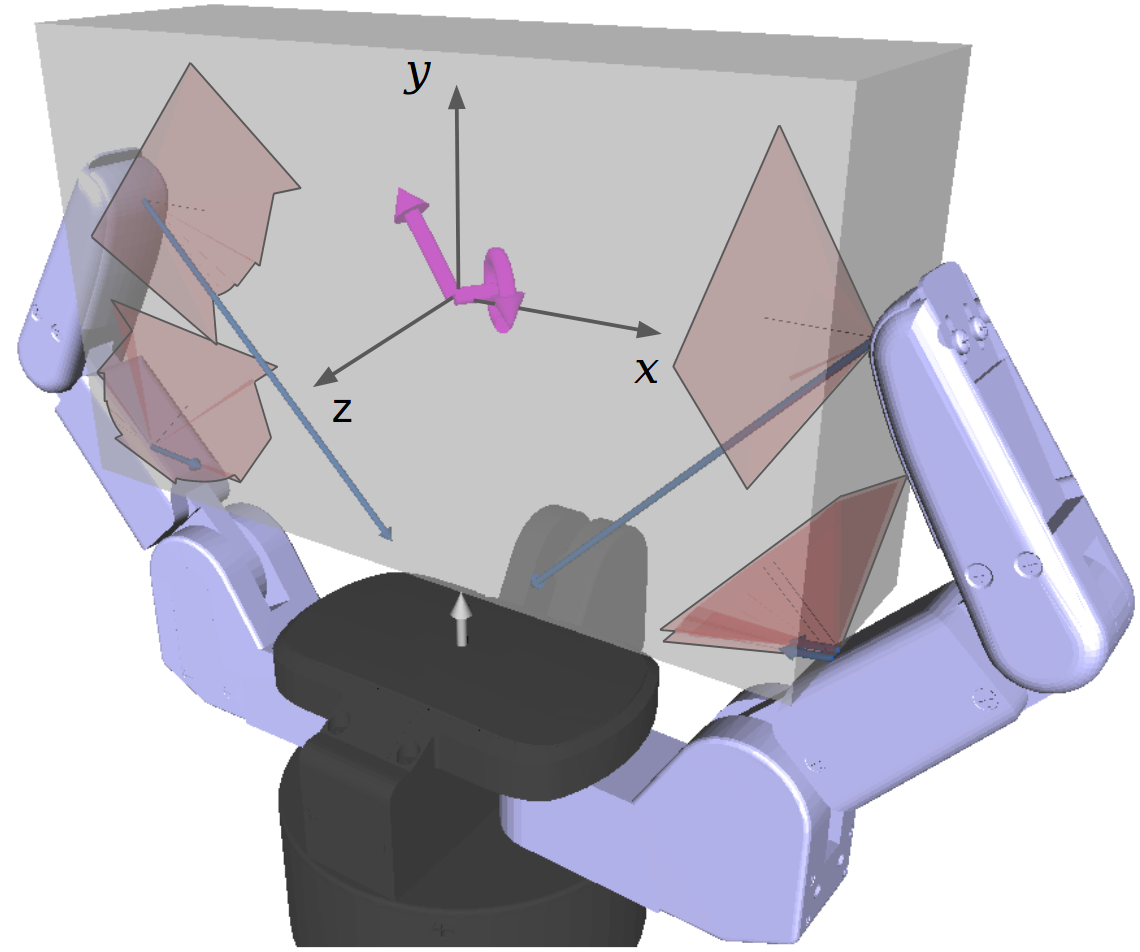}

\caption{Equilibrium contact forces (blue arrows) that are predicted to arise
by our framework when a force of 2.2N is applied to the object in the
$y$-direction. The proximal joints have both been preloaded with 0.1Nm. Note
the refinement of the pyramidal friction cone approximation. Only a minority
of sectors have been refined. Sectors that are not of interest in this
specific grasp problem remain in a less refined state and thus only contribute
little to the complexity of the overall problem. The sectors containing the
equilibrium contact forces are $<1\degree$~small. The maroon arrows denote the direction of
predicted object motion.}

\label{fig:spot_check}
\end{figure}

Let us solve for the stability of the grasp shown in Fig.~\ref{fig:spot_check}
when we apply a force of 1N to the object in the positive $y$-direction
\textit{without preloading the joints at all.} To answer this query, we use
Algorithm~\ref{alg:refinement} as follows:

\begin{eqnarray} 
\text{objective:}~ & \text{none} \\
\text{additional constraints:} & \bm{w} &= [0,1,0,0,0,0]^T \\
& \bm{\tau}^c &= \bm{0}
\end{eqnarray}

Note that, in the absence of an optimization objective, we are simply
asking if a solution exists that satisfies all the constraints of the
problem, equivalent to determining values for all the unknowns
(contact forces, virtual motions, etc.) such that the grasp is
stabilized. Using Gurobi~\cite{gurobi} as a solver for the constituent
MIPs, Algorithm~\ref{alg:refinement} finds no feasible solution to
this problem (i.e. it predicts the grasp is unstable in the presence
of the given disturbance.) This is the expected result: the grasp may
not resist a force of 1N in the $y$-direction without any preloading,
as it is intuitively clear that the object will slide out.

Let us now consider the case where we apply a preload torque of 0.1 Nm at the
proximal joints. Due to the simplicity of the grasp we can analytically
determine the expected maximum force in the $y$-direction the grasp can
withstand: a $\sim$90mm contact moment arm and a friction coefficient of 1.0
results in a maximum total friction force applied to the object of $\sim$2.2
N. Using a similar formulation as before (no objective, $\bm{\tau}^c = 0.1$Nm
at the proximal joints), we indeed find that Algorithm 1 accepts a solution
(i.e. predicts stability, see Fig.~\ref{fig:spot_check}) for a disturbance of 2.2N
in the $y$-direction ($\bm{w} = [0,2.2,0,0,0,0]^T$), but finds no solution for
a disturbance of 2.5N in the same direction ($\bm{w} = [0,2.5,0,0,0,0]^T$).

\subsection{Space of resistible disturbances}

The option to add an objective allows us to formulate more interesting
queries. Algorithm~\ref{alg:refinement} provides us with a simple method of
characterizing the space of possible disturbances on the object a grasp can
withstand through purely passive reaction. We can directly determine the exact
\textit{maximum disturbance} applied to the grasped object in a given
direction that a grasp may resist purely passively. To this end we might
prescribe a preload $\bm{\tau}^m$ for the actuators to be kept constant, and a
direction $\bm{d}$ along which to apply a disturbance to the object. To
compute the largest magnitude disturbance the grasp can withstand in that
direction, we use Algorithm~\ref{alg:refinement} as follows: 

\begin{eqnarray} 
\text{objective:}~ & \text{maximize}~ s \\
\text{additional constraints:} & \bm{w} = s\bm{d} \\
& \bm{\tau}^c = \bm{\tau}^m
\end{eqnarray}

We already analytically estimated the resistance of our example grasp to
disturbances in the positive $y$-direction to be $\sim$2.2N. The maximum
resistible force predicted by our framework is 2.33N and thus very similar.
The 6\% difference is well within the uncertainty introduced by estimating the
exact geometry of the grasp.

\begin{figure}[t!]
\centering
   \begin{subfigure}[No contact normal uncertainty]
   {\includegraphics[width=0.85\columnwidth]{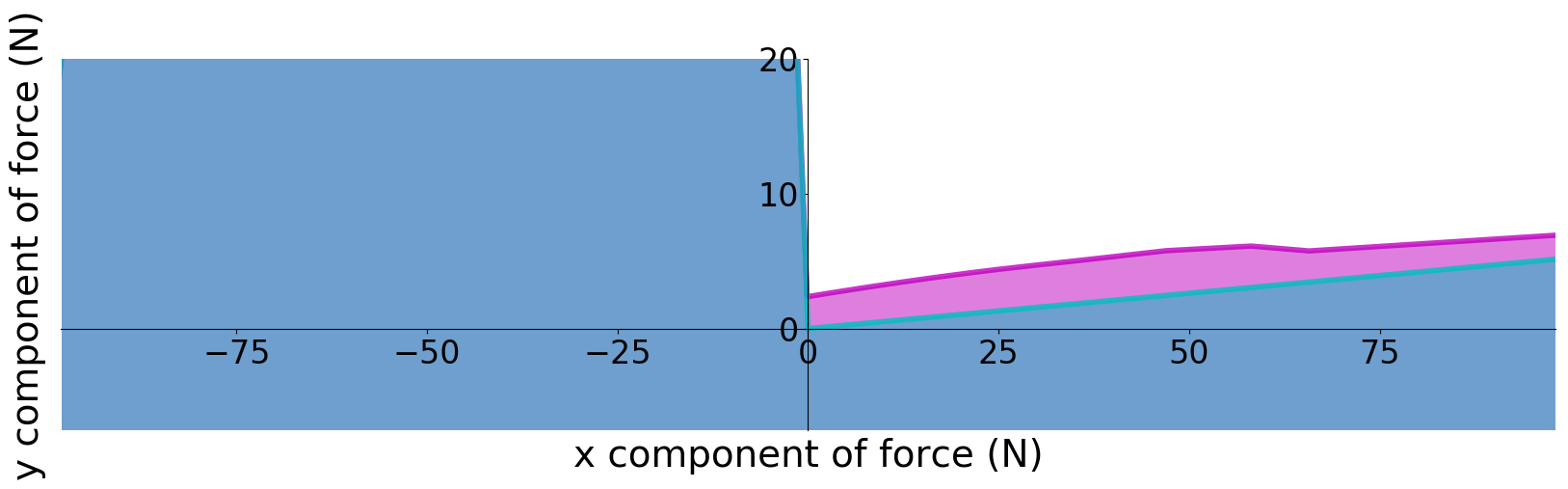}
   \label{fig:package_exact}} 
\end{subfigure}
\begin{subfigure}[Robust to contact normal uncertainties]
   {\includegraphics[width=0.85\columnwidth]{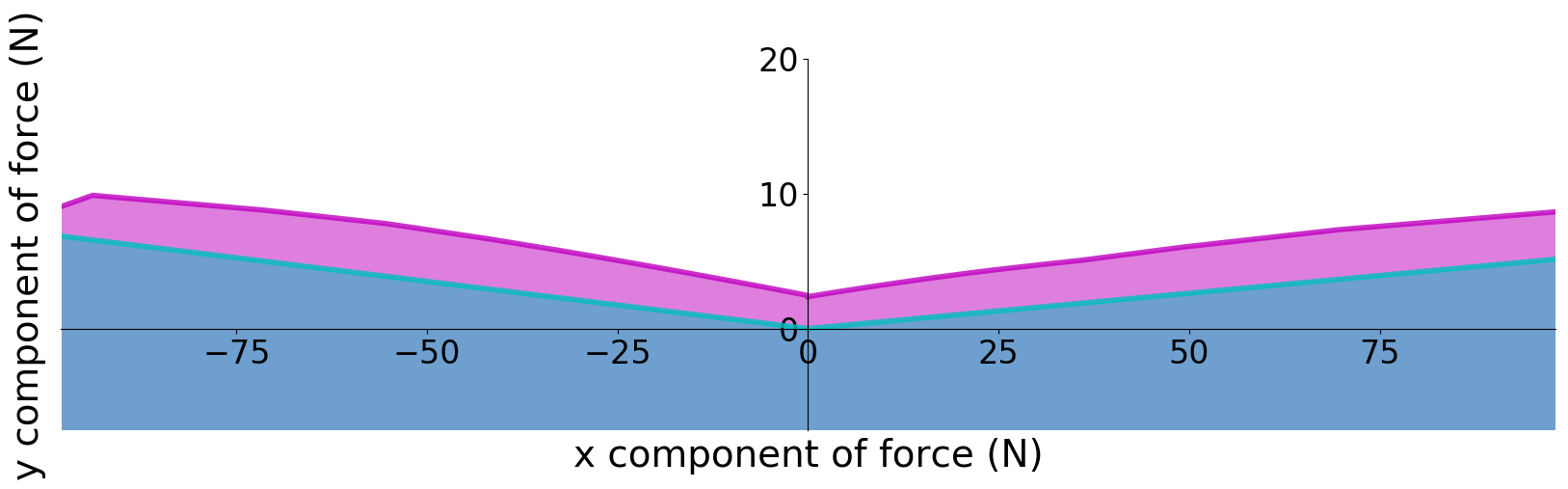}
   \label{fig:package_robust}}
\end{subfigure}

\caption{Resistible forces in the grasp plane of the grasp in
Fig.~\ref{fig:spot_check} as predicted by our model and algorithm. In blue are
forces that can be resisted even without the application of preloading torques
at the joints. When loading the two proximal joints with 0.1Nm the maroon area
is added to the space of resistible forces.}

\label{fig:package_map}
\end{figure}

In order to further investigate the stability of the grasp and also the
capabilities of our framework let us consider the maximum resistible
disturbances in multiple directions. We discretize the grasp plane by
direction vectors with a spacing of 1\degree~between them. We then determine
the maximum force applied to the object the grasp can withstand for each of
these direction. The results are visualized in Fig.~\ref{fig:package_map},
where Fig.~\ref{fig:package_exact} plots results without considering
robustness to contact normal uncertainty, while Fig.~\ref{fig:package_robust}
assumes an uncertainty of 2.5\degree.

The results match our intuition that any downward force can be reacted without
any loading of the fingers. Furthermore the model captures the need for finger
loading in order to resist upward forces. It also shows an effect of passive
finger loading for forces with nonzero $x$ component: pushing sideways
increases the amount of resistance to upwards forces.

\begin{figure}[tbp]
\centering
\includegraphics[width=0.75\columnwidth]{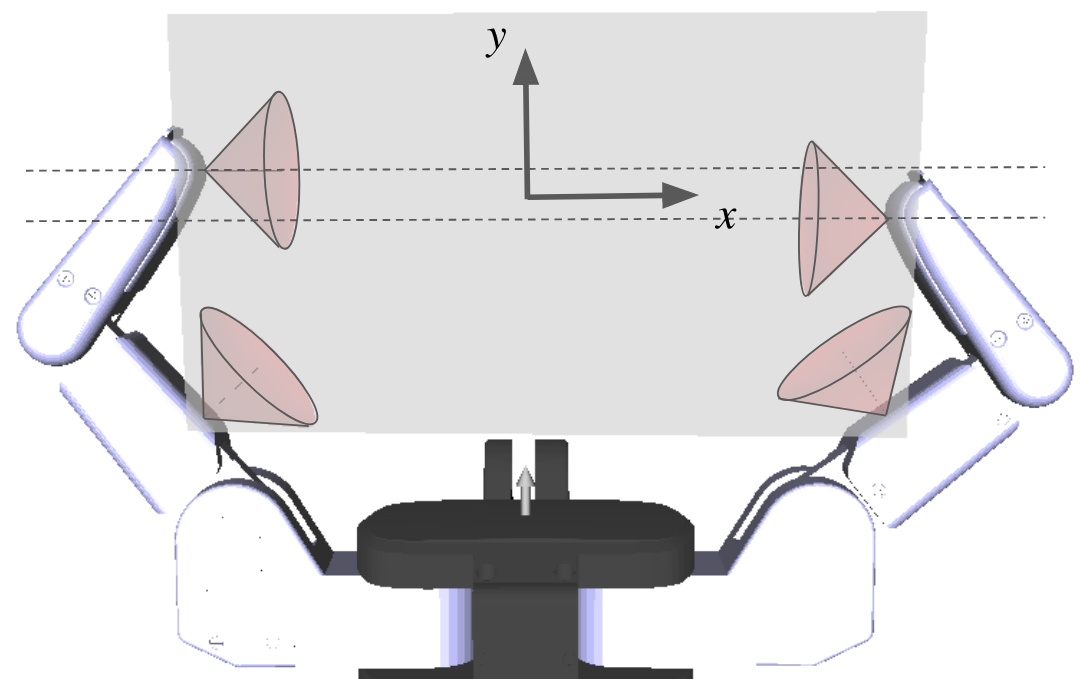}

\caption{A grasp designed to highlight the 'wedging' effect. The two contacts
on the distal link are offset with respect to each other allowing wedging to
occur.}

\label{fig:skewed_grasp}
\end{figure}

The reason for the asymmetry of Fig.~\ref{fig:package_exact} however is not
immediately obvious, as the grasp itself appears symmetric. In fact however,
the two distal contacts are ever so slightly offset, causing the object to pivot about the left distal contact and
wedging itself stuck if enough leftward force is applied. The grasp in
Fig.~\ref{fig:skewed_grasp} makes it clearer why this behavior occurs - here
the contacts are visibly offset. Note, that this wedging behavior is very
different from the wedging behavior discussed in Chapter~\ref{sec:3dfriction},
which occurred when omitting the MDP. There, the wedging occurred no matter
the applied wrench such that arbitrary wrenches could be resisted.
Furthermore, the resulting contact forces did not satisfy energy conservation.

In contrast, in our framework only specific wrenches allow wedging to occur.
These wrenches depend on the geometry of the grasp and are consistent with the
rigid body statics of the grasp problem. The equilibrium contact forces
predicted by our framework satisfy the MDP and hence energy conservation.
However, as our method allows us to solve the rigid body problem very
accurately, only a small offset is required for our model to predict wedging
of the object; an offset that is easily within the accuracy of a typical
triangular mesh. Note, that this behavior is fully consistent with the rigid
body assumption (although we introduced compliance at the contacts we ignore
changes in grasp geometry due to motion) and the predictions made by our
framework are correct, albeit highly sensitive to the grasp geometry.

Of course in practice it is not advisable to rely on such volatile geometric
effects. Therefore taking into account geometric uncertainties is of paramount
importance for practical applications. Fig.~\ref{fig:package_robust} shows
which forces can be robustly resisted when we consider the uncertainty in
normal angle to be no larger than 2.5\degree~(using the approach outlined in
Chapter~\ref{sec:robust}.) The resulting plot of resistible forces is
approximately symmetric corresponding to the near-symmetry of the grasp. The
indicated spaces of resistible forces both with and without a preload are
consistent with our intuition and the empirical data presented in
Chapter~\ref{sec:3d_apps_iterative}.

\begin{figure}[t]
\centering
\includegraphics[width=0.85\columnwidth]{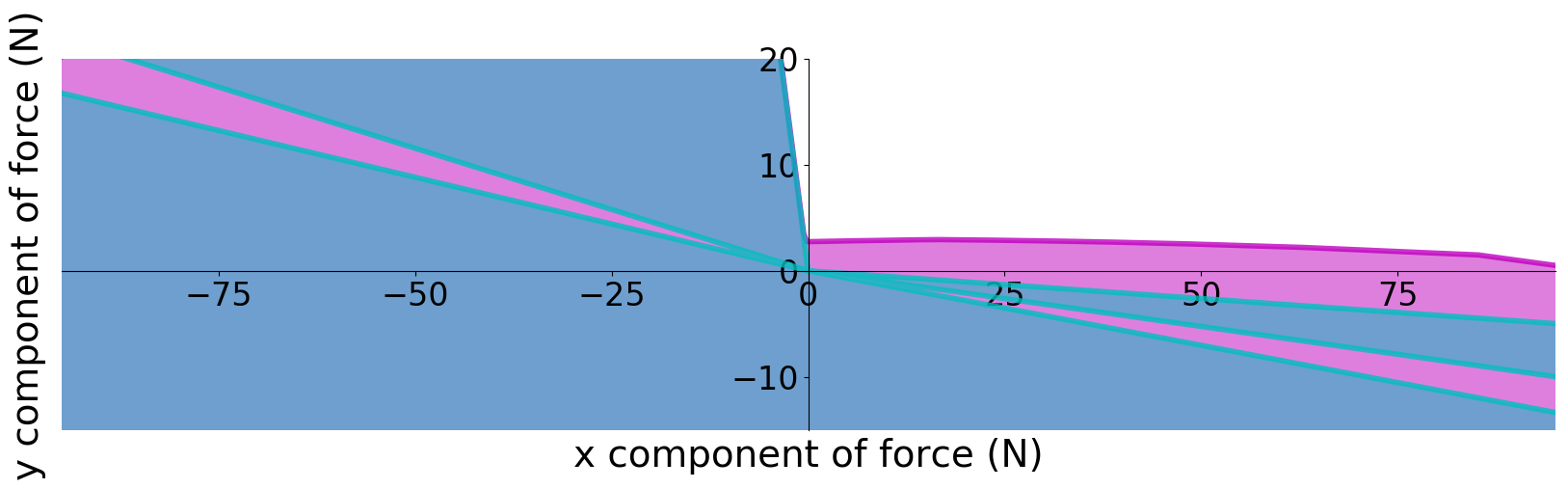}

\caption{Forces in the $xy$-plane predicted resistible for the grasp in
Fig.~\ref{fig:skewed_grasp} using our model and algorithm. In blue are forces
that can be resisted even without the application of preloading torques at the
joints. When loading the two proximal joints with 0.1Nm the maroon area is
added to the resistible forces. We also take into account a contact normal
uncertainty of 2.5 degrees to make sure the grasp is robust to such
discrepancies.}

\label{fig:skewed}
\end{figure}

We showed how we can make the stability predictions less sensitive to the
small scale geometric characteristics of the grasp and thus robust to
uncertainties. At a larger scale, however, wedging effects can be robustly
leveraged. In the grasp in Fig.~\ref{fig:skewed_grasp} the contacts are
significantly offset and the forces this grasp may resist robustly are shown
in Fig.~\ref{fig:skewed}. Thus, if we know the range of disturbances likely to
be encountered during a manipulation task our framework can be a valuable tool
in picking an appropriate grasp.

One feature of Fig.~\ref{fig:skewed} perhaps requires further
elaboration: The gaps in the second and fourth quadrants of
Fig.~\ref{fig:skewed} where forces may only be resisted when a preload is
applied but not otherwise. These gaps stand out because they appear thin
and are surrounded by large areas where applied forces are resistible even
without a preload. We thus investigated what effects cause these wedges in
order to verify if these predictions are physically accurate:

The grasps shown in Figs.~\ref{fig:spot_check} and~\ref{fig:skewed_grasp} were
created such that all contacts are as close as possible to lying in a mutual
plane. This is because three dimensional grasps can be very complex while two
dimensional grasps often allow us to use our intuition to validate the
predictions made by our framework. However, we are using the open-source
GraspIt!~\cite{MILLER04} package in our grasp analysis and a limitation of
this package is that it is difficult to create grasps that are truly two
dimensional in nature. Due to the meshing of the finger and object geometries
as well as the intricacies of collision checking in GraspIt! the contacts
always lie somewhat offset from the central plane.

In the specific case of the grasp in Fig.~\ref{fig:skewed_grasp} the contacts
do not quite lie within the $xz$-plane, in which the forces applied to the
object lie. This means that two contacts are generally not sufficient in order
to balance an applied force. Let us investigate the gap in the second
quadrant of Fig.~\ref{fig:skewed}: When a force is applied along the $(-1,0)$
direction the distal contact of the left finger acts as a fulcrum and the
object rotates clockwise loading the contact on the proximal contact of the
right finger. When a force is applied in the $(-1,1)$ direction the left
finger contact again acts like a fulcrum, however now the object rotates
counter-clockwise loading the distal contact on the right finger.

Somewhere in between those two cases the applied force points almost directly
at the fulcrum contact and instead of rotating the object is mostly pressed
against the left finger breaking both contacts on the right finger. Thus, only
two contacts remain and the grasp becomes unstable. In this particular case
the object would rotate out of the grasp around the y-axis as the two
remaining contacts on the left finger do not lie in the same plane as the
applied force.

\begin{figure}[t]
\centerline
{\includegraphics[width=0.8\columnwidth]{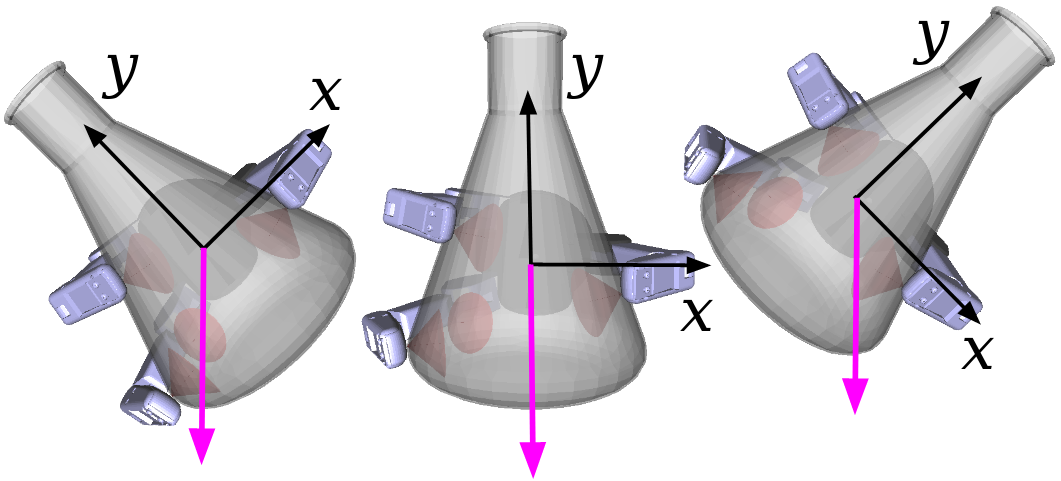}}

\caption{3 dimensional grasp that highlights the necessity of a grasp to be
able to withstand a range of forces applied to the object. During a pouring
task gravity (pink) moves in the $xy$-plane.}

\label{fig:flasks}
\end{figure}

\begin{figure}[t]
\centering
\includegraphics[width=0.75\columnwidth]{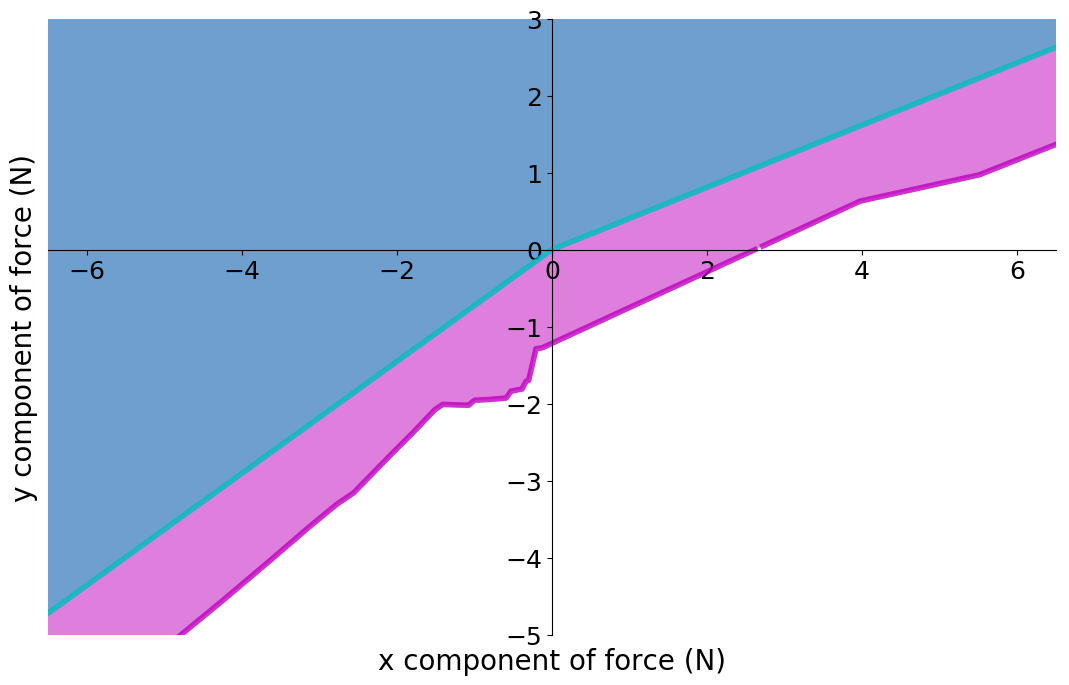}

\caption{Forces in the $xy$-plane predicted resistible for the grasp in
Fig.~\ref{fig:flasks}. In blue are forces that can be resisted even without
preloading joint torques. When loading the proximal joints with 0.1Nm, the
maroon area is added to the resistible forces. We use a contact normal
uncertainty of 2.5 degrees.}

\label{fig:flask_map}
\end{figure}

Let us now consider the grasp in Fig.~\ref{fig:flasks}. Note that this grasp
comprises four contacts (one on each distal link plus one on a proximal link)
which do not lie on the same plane, and thus has to be analyzed in a
three-dimensional framework. We consider here an apparent task the robot
grasping the flask may need to execute. In order to pour a liquid contained in
the flask it is necessary to tip it. If we choose to use the robot wrist for
this tipping motion then the force of gravity acting on the flask and its
contents lies in the $xy$-plane. The grasp must thus be able to resist such
forces in order to complete its task successfully. Furthermore, we have a
choice of direction in which to turn the flask in order to pour its content.
Creating a visualization (shown in Fig.~\ref{fig:flask_map}) as before we can
deduce the need for a preload, and that it is more robust to turn the flask
counter-clockwise. Thus, once a grasp has been established our framework can
help in making decisions as to how a task is to be executed.

\subsection{Actuator command optimization}

\begin{figure}[t!]
\centering
\includegraphics[width=0.6\columnwidth]{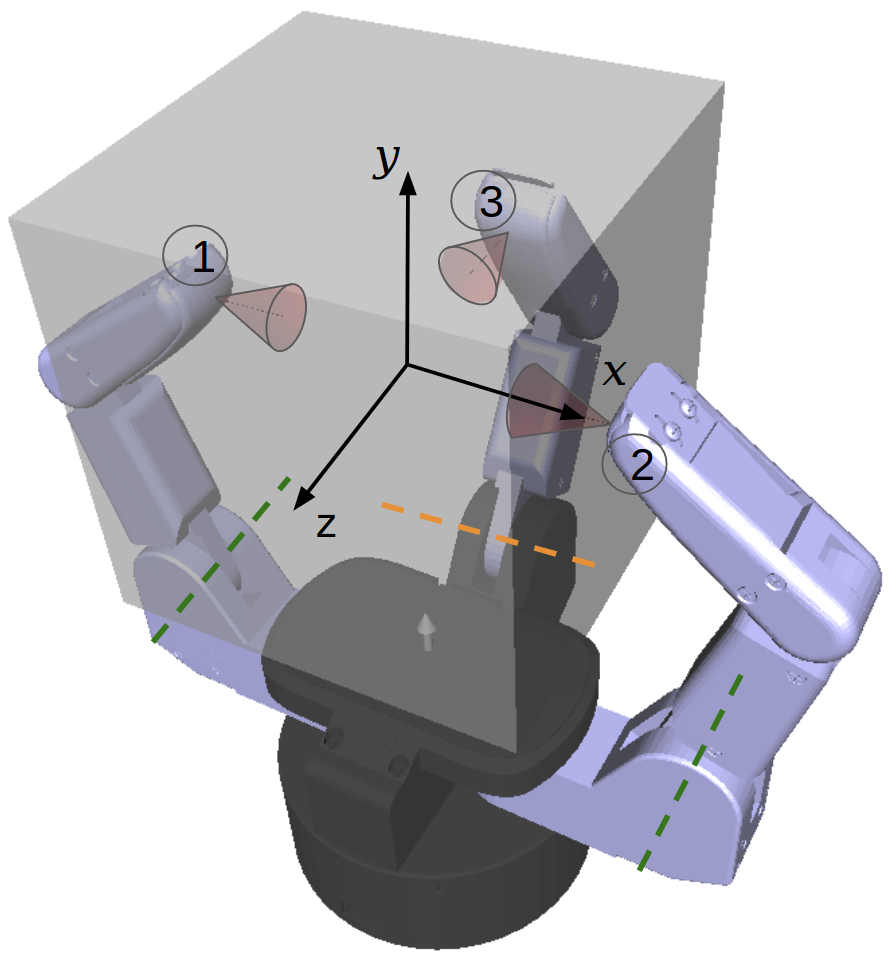}

\caption{Example grasp to illustrate the importance of appropriately choosing
grasp preloads. We assume a preload of 0.1Nm at the proximal joints of fingers
1 and 2 (axes marked as green dashed lines). We apply a range of preload
torques at the proximal joint of finger 3 (axis marked as yellow dashed line)
and evaluate passive stability.}

\label{fig:cube}
\end{figure}

\begin{figure}[t!]
\centering
\includegraphics[width=0.85\columnwidth]{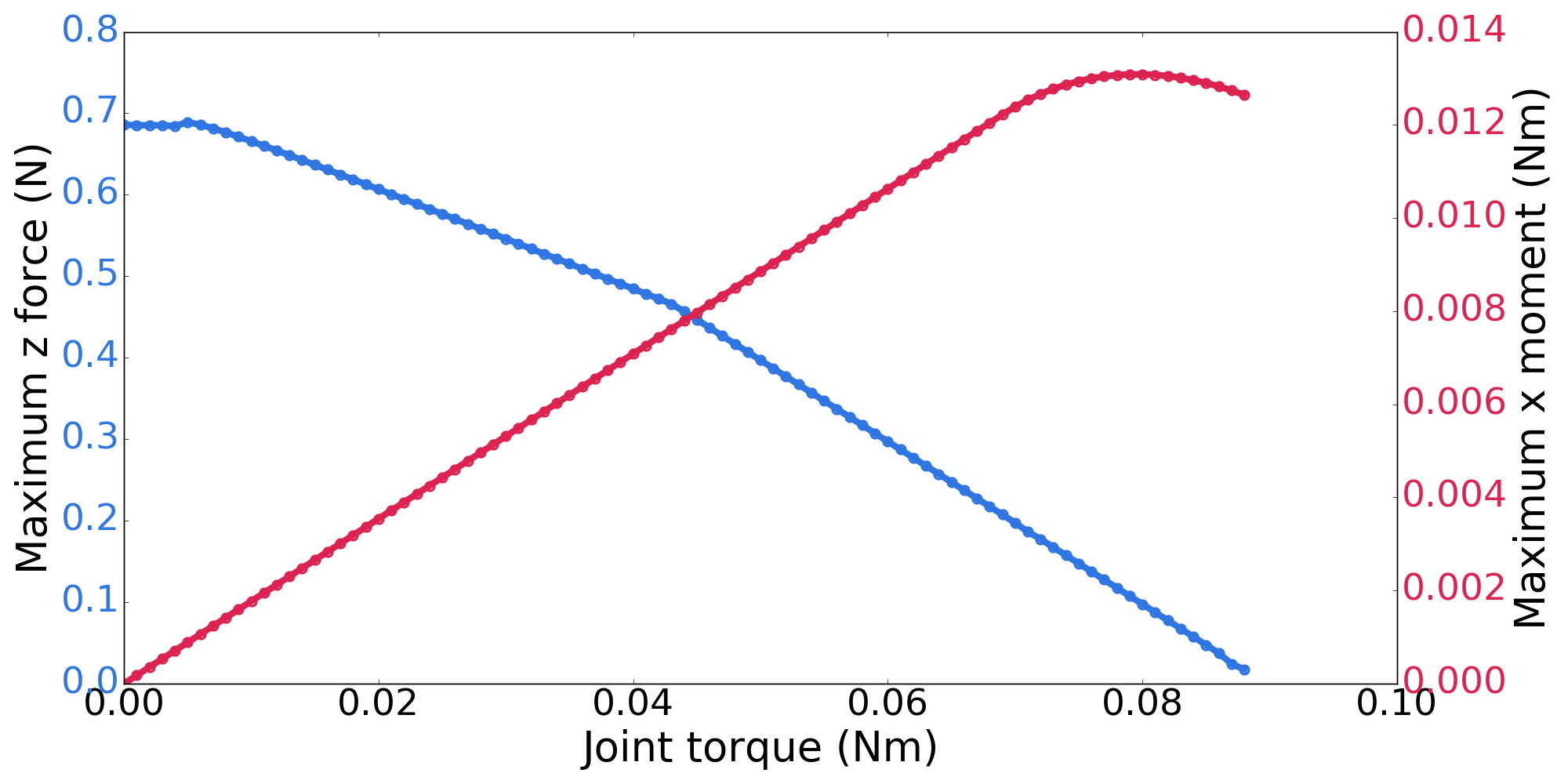}

\caption{Force in the $z$-direction (blue) and moment in the $x$-direction
(red) the grasp in Fig.~\ref{fig:cube} can resist for a range of preloads at
finger 3.}

\label{fig:preload_test}
\end{figure}

The passive stability of a grasp is not only determined by its geometry: the
actuator commands are equally important. Consider for example the grasp in
Fig.~\ref{fig:cube}. The three contacts the hand makes with the object all lie
approximately in the $xz$-plane. Contacts 1 and 2 lie approximately on the
$x$-axis and oppose each other. Let us assume we create a grasp by commanding
the proximal joints of fingers 1 and 2 to each apply 0.1Nm. Let us vary the
torque commanded at the proximal joint of finger 3 and observe the difference
in passive stability. Specifically, we will use our framework to investigate
the maximum disturbance on the object the grasp can resist in two directions.
Fig.~\ref{fig:preload_test} shows the resulting predictions from our
algorithm.

First we are interested in forces along the positive $z$-axis. As expected,
the resistance is largest if no motor torque is applied by finger 3. Any load
by this finger only adds to the disturbance and does not help in resisting it.
When the torque applied by finger 3 reaches 0.09 Nm, it has completely removed
any resistance to $z$-direction forces. This can be easily verified: The
coefficient of friction chosen for this example is 0.45 and the moment arms
from joint to contact are identical for all three fingers. As there cannot be
any out-of-plane forces, the normal forces at all contacts will be
proportional to the applied joint torque. Thus, applying 0.09Nm at finger 3
claims all possible contact friction at both contacts 1 and 2: no further
forces in that direction can be resisted.

Let us now consider passive resistance to torques applied to the
object around the $x$-axis. If we do not load finger 3 the object is
only held by contacts 1 and 2. As both these contacts lie on the
$x$-axis they cannot apply any torque to the object in that
direction. Thus, the grasp cannot resist any torques around the
$x$-axis unless we also load finger 3. The third finger provides the
contact necessary for resisting the torque on the object. The more we
load finger 3, the larger the force at contact 3 and the larger the
resistible torque. At some point however, as discussed above, the
forces at finger 3 begin to overwhelm fingers 1 and 2 and the object
slides out along the $z$ axis even without any external disturbances.

This example shows the importance of preload and passive stability:
for this grasp, loading finger 3 helps resistance against some
disturbances, but hurts against others. The right amount of preload
must thus be chosen based on the task. To the best of our knowledge,
no existing grasp analysis method can make such predictions.

Using our model, we can also find actuator commands that are optimal
with respect to any specific objective we choose. For example we may want to
minimize the maximum torque a single actuator must produce to resist a given
wrench $\bm{w}^m$. We now use Algorithm~\ref{alg:refinement} as follows:

\begin{eqnarray} 
\text{objective:}~ & \text{minimize}~ \max_{j}{\tau_j^c} \\
\text{additional constraints:}~ & \bm{w} = \bm{w}^m 
\end{eqnarray}

For the grasp in Fig.~\ref{fig:flasks}, we can compute the optimal actuator
commands for a force of $\bm{w}^m = [0,-1.21743,0,0,0,0]^T$, the largest force
in the negative $y$-direction that can be resisted when applying a preload of
0.1Nm at every proximal joint, according to our previous analysis. We find
that the optimal torques at these joints are actually $\bm{\tau}^c = [0,
0.042, 0.073]^T$. This shows that a large amount of the naive preload (0.1Nm
at every joint) is wasted in the sense that it does not increase disturbance
resistance in this particular direction. In fact it appears that loading
finger 1 was detrimental to grasp stability with respect to this disturbance.

In many practical applications it may be of interest to take into account
physical limits such as the maximum torque an actuator can apply or a maximum
permissible normal force in order to not break the grasped object. Such
constraints can be expressed as linear inequalities and are straightforward to
add to our model.

\subsection{Computational performance}

In Chapter~\ref{sec:refinement} we state that an accurate solution to the grasp
problem with discretized friction cones but without hierarchical refinement
requires a very large number of friction edges. We further argue that solving
such a problem becomes computationally intractable as a large number of
friction edges results in a large number of binary variables in the MIP. To
verify these hypotheses, we analyze the convergence of our algorithm with
varying levels of refinement of the friction approximation. We use the grasp
in Fig.~\ref{fig:flasks} as an example, with the task of finding the largest
force in the negative $x$-direction that the grasp can withstand.  We do so at
varying levels of refinement, and record the predicted force magnitude and
runtime.

We compare two approaches: The first method ("full resolution") directly uses
a friction cone approximation that is at the desired level of accuracy in its
entirety. The second method ("hierarchical refinement") always starts with a
coarse approximation and refines it as described in
Algorithm~\ref{alg:refinement}. Throughout all experiments, both methods (when
able to finish)  produced identical solutions, but the running times varied
greatly. All recorded data can be found in Table~\ref{tab:data}.

\begin{sidewaystable}[ph!]
\centering
\begin{tabular}{p{0.15\textwidth} | p{0.1\textwidth} p{0.1\textwidth} p{0.15\textwidth} | p{0.1\textwidth} p{0.1\textwidth} p{0.15\textwidth} }
\multirow{2}{0.15\textwidth}{Equivalent number of friction edges} & \multicolumn{3}{c|}{\textbf{Without robustness scheme}} & \multicolumn{3}{c}{\textbf{With robustness scheme}} \\
& Maximum wrench (N) & Time at full resolution (s) & Time with hierarchical refinement (s) & Maximum wrench (N) & Time at full resolution (s) & Time with hierarchical refinement (s) \\
\hline
4 & \textgreater 100 & 0.088 & 0.088 & \textgreater 100 & 0.11 & 0.11 \\
8 & 7.07435 & 0.22 & 0.43 & 7.07435 & 0.24 & 0.46 \\
16 & 6.34905 & 0.43 & 0.65 & 3.73377 & 0.45 & 1.02 \\
32 & 3.71036 & 0.72 & 1.92 & 3.70539 & 0.97 & 1.33 \\
64 & 3.70095 & 1.58 & 2.33 & 2.31479 & 1.83 & 3.09 \\
128 & 3.69775 & 5.80 & 3.50 & 1.21805 & 5.12 & 10.9 \\
256 & 3.69751 & \textgreater 600 & 4.87 & 1.21756 & 20.3 & 13.0 \\
512 & 3.69681 & - & 4.34 & 1.21743 & 102 & 17.8 \\
1024 & 3.07268 & - & 13.3 & 1.21740 & 394 & 22.9 \\
2048 & 3.07267 & - & 15.0 & 1.21739 & - & 32.2 
\end{tabular}

\caption{Performance of the analysis of the grasp in Fig.~\ref{fig:flasks}. We
compute the magnitude of the largest force that can be applied in the negative
$y$-direction without destabilizing the grasp. We do this for varying levels
of refinement expressed as the number of sectors the friction cone
approximation contains. We record compute times for both an approach where all
sectors are already of the desired size and our hierarchical refinement
approach.}

\label{tab:data}
\end{sidewaystable}

\begin{sidewaystable}[ph!]
\centering
\begin{tabular}{p{0.05\textwidth} p{0.1\textwidth} | c | c c | c c}
\multirow{2}{0.1\textwidth}{\textbf{Grasp}} & \multirow{2}{0.1\textwidth}{\textbf{Number of contacts}} & \textbf{Stability check} & \multicolumn{2}{c|}{\textbf{Maximum disturbance}} & \multicolumn{2}{c}{\textbf{Optimal torques}} \\
& & Time (s) & Mean time (s) & Median time (s) & Mean time (s) & Median time (s) \\
\hline
(a) & 3 & 0.246 & 0.797 $\pm$ 0.067 & 0.754 & 0.389 $\pm$ 0.063 & 0.343 \\
(b) & 3 & 0.234 & 0.785 $\pm$ 0.085 & 0.745 & 0.447 $\pm$ 0.074 & 0.408 \\
(c) & 3 & 0.294 & 0.893 $\pm$ 0.056 & 0.907 & 0.347 $\pm$ 0.017 & 0.345 \\
(d) & 3 & 0.259 & 0.664 $\pm$ 0.081 & 0.560 & 0.308 $\pm$ 0.036 & 0.281 \\
(e) & 4 & 1.07 & 10.0 $\pm$ 3.4~ & 6.39 & 3.26 $\pm$ 0.50 & 3.83 \\
(f) & 4 & 15.7 & 36.7 $\pm$ 6.5~ & 39.5 & 13.5 $\pm$ 4.8~ & 9.58 \\
(g) & 4 & 2.78 & 33.8 $\pm$ 9.3~ & 20.8 & 16.8 $\pm$ 5.3~ & 12.3 \\
(h) & 4 & 8.02 & 23.7 $\pm$ 4.2~ & 23.2 & 12.3 $\pm$ 3.3~ & 8.10 \\
(i) & 4 & 1.66 & 14.4 $\pm$ 2.7~ & 11.4 & 1.67 $\pm$ 0.12 & 1.60 \\
(j) & 5 & 6.06 & 150 $\pm$ 24~ & 105 & 22.8 $\pm$ 5.8~ & 19.0 \\
(k) & 6 & 51.0 & 398 $\pm$ 104 & 344 & 136 $\pm$ 54~ & 80.6 \\
(l) & 6 & 27.5 & 488 $\pm$ 99~ & 330 & 777 $\pm$ 288 & 626 \\
\end{tabular}

\caption{Runtime analysis of our method for the three tasks
demonstrated in this paper performed on a consumer desktop computer for the
grasps shown in Fig.~\ref{fig:comp_grasps}. Where multiple trials were
performed we report the mean $\pm$ standard error as well as the median
runtimes.}\label{tab:comp_data}

\end{sidewaystable}

We notice that, at high levels of refinement, full resolution becomes
intractable, whereas hierarchical refinement finds a solution efficiently. The
study of how the refinement level affects the returned solution is more
complex. The exact value of the solution generally reaches a point where
increasing the accuracy of the approximation (adding more friction edges)
stops making a significant difference. In some cases, as in the case of the
maximum wrench in the left side of Table~\ref{tab:data}, this happens for
accuracy levels that only hierarchical refinement can reach. In others, as in
the case of the maximum robust wrench (with 2.5\degree~normal uncertainty) in
the right side of Table~\ref{tab:data}, both methods are able to find good
approximations of the final value. At the more shallow levels, full resolution
will often outperform hierarchical refinement, but since we generally do not
know which of these cases any specific query might fall into, only
hierarchical resolution allows us to increase the accuracy without the risk of
compute time exploding. These results show both that high accuracy is actually
required in order to obtain meaningful results, and that solving such problems
without hierarchical refinement is computationally intractable.

Another interesting finding is that the predicted maximum resistible force is
exceedingly large when only four friction edges are used in the first step of
the refinement process. This is because at this stage the MDP is only enforced
such that the friction force and negative relative tangential contact motion
lie within a 90\degree~sector. This allows sufficient freedom to the solver to
use the rigidity of the robot hand when backdriven along with unphysical
object motions to create large contact forces - much like what was described
in Chapter~\ref{sec:3dfriction} where there are no constraints on the friction
direction at all. This illustrates again the need for a high accuracy solution
to the grasp stability problem including the MDP.

\begin{figure}[ph!]
\centering
\subfigure[]{\includegraphics[width=1.8in]{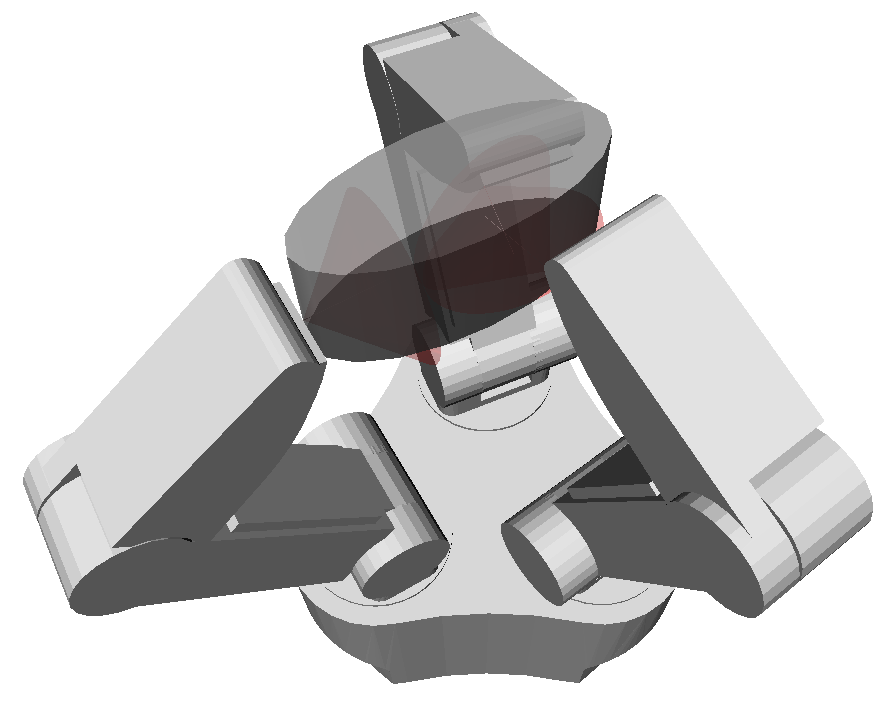}}
\subfigure[]{\includegraphics[width=1.8in]{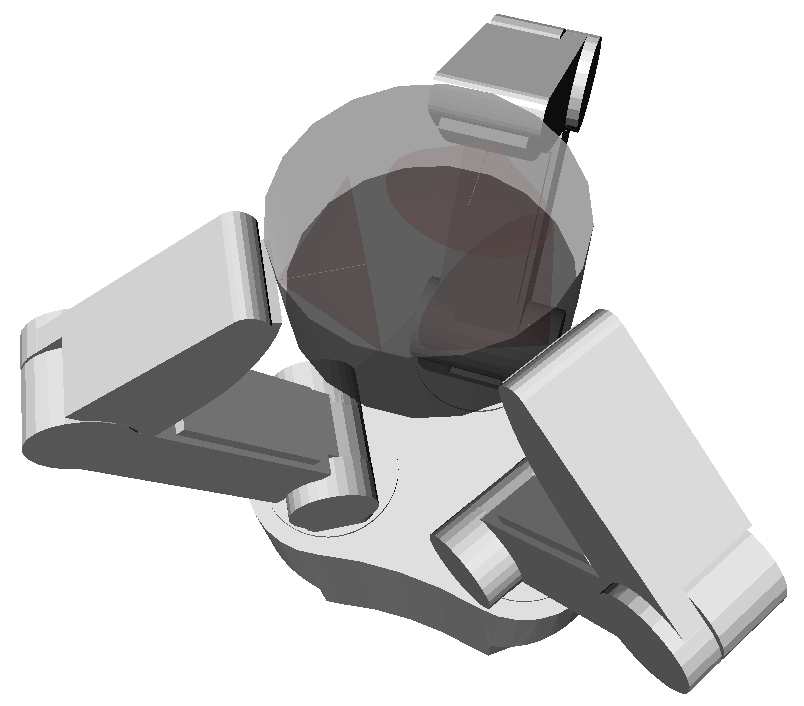}}
\subfigure[]{\includegraphics[width=1.8in]{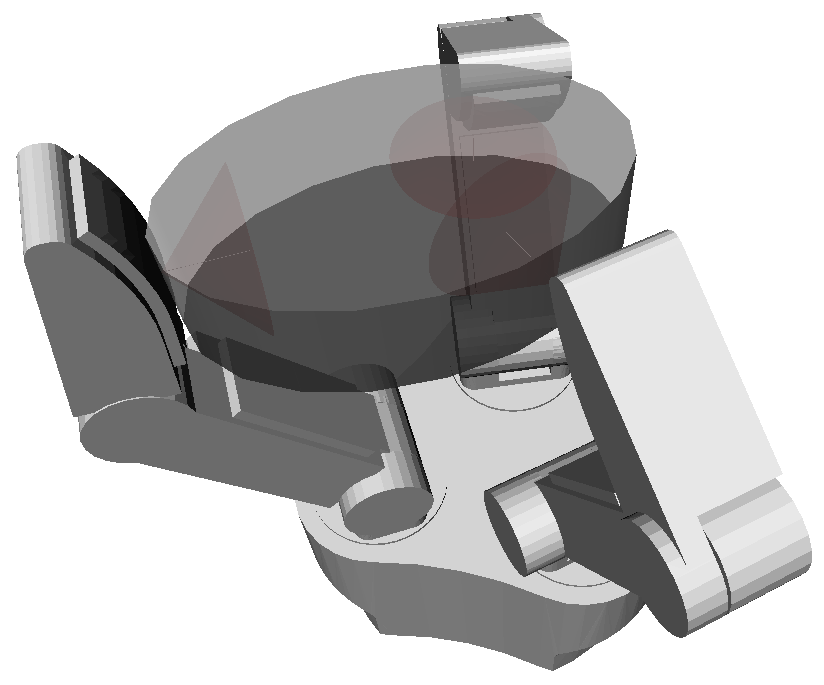}}
\subfigure[]{\includegraphics[width=1.8in]{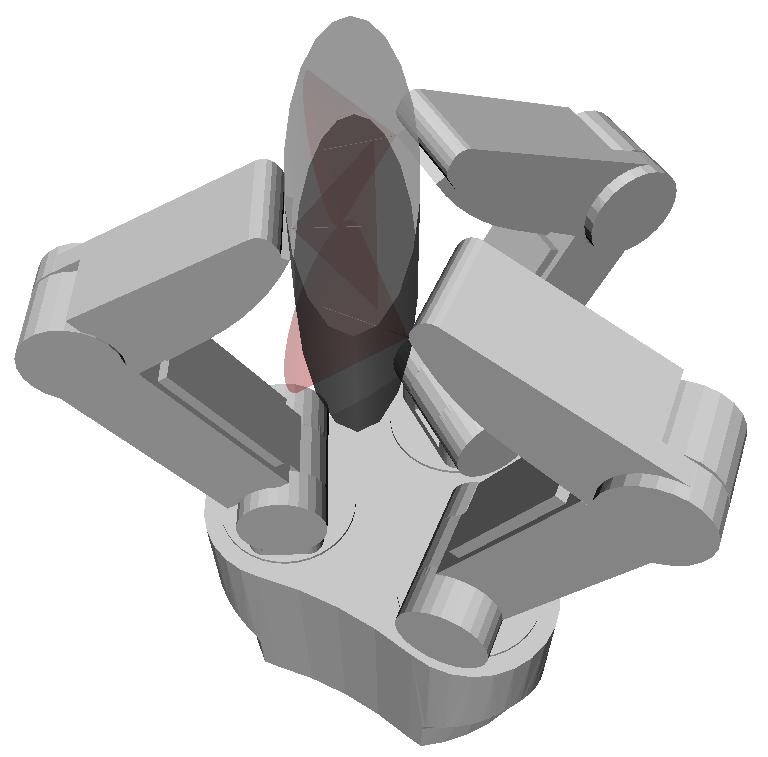}}
\subfigure[]{\includegraphics[width=1.8in]{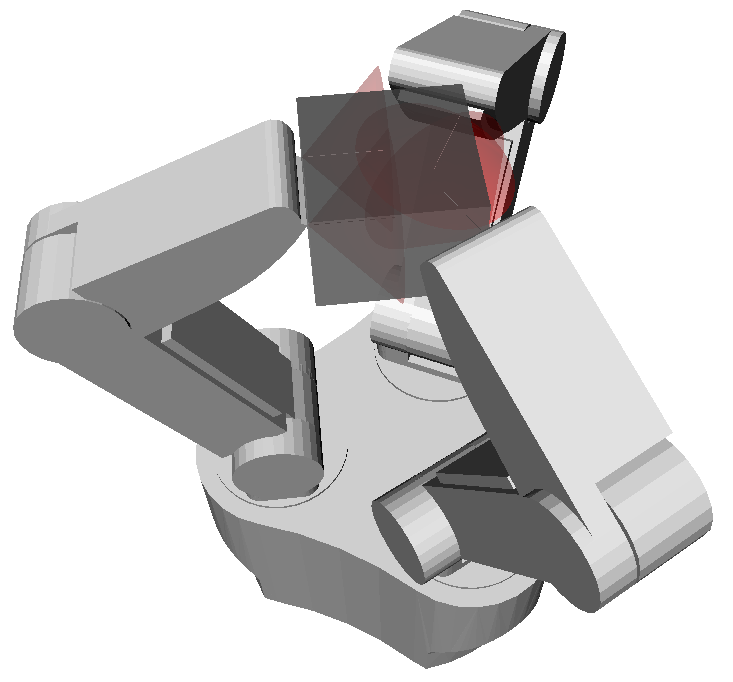}}
\subfigure[]{\includegraphics[width=1.8in]{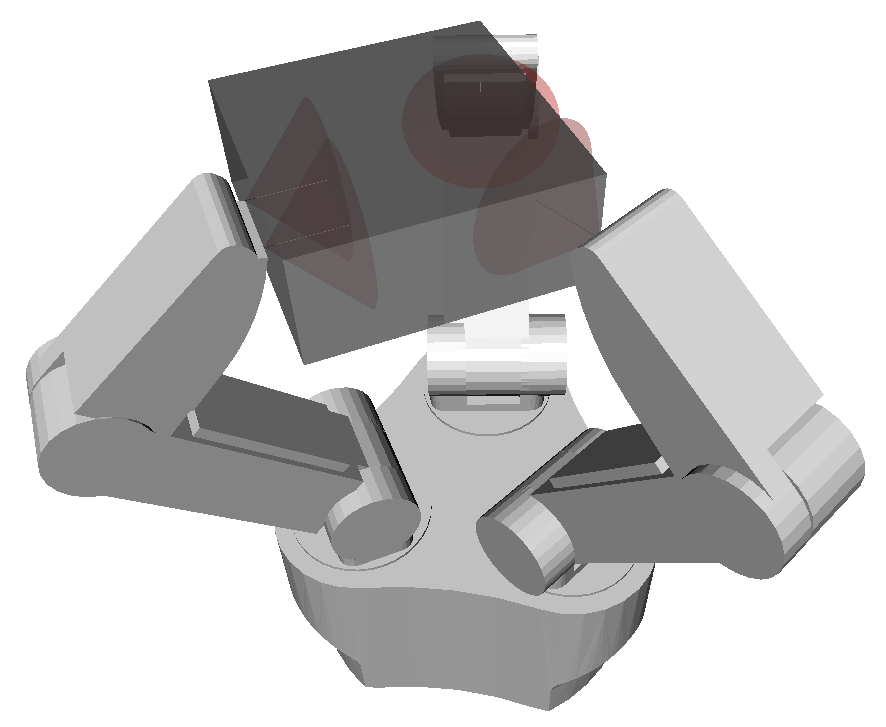}}
\subfigure[]{\includegraphics[width=1.8in]{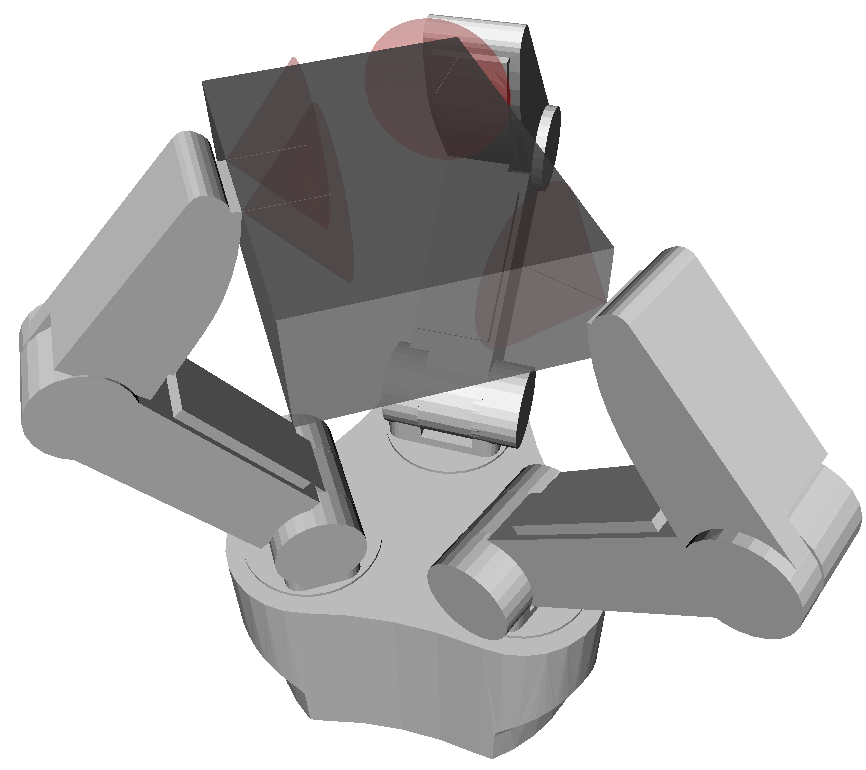}}
\subfigure[]{\includegraphics[width=1.8in]{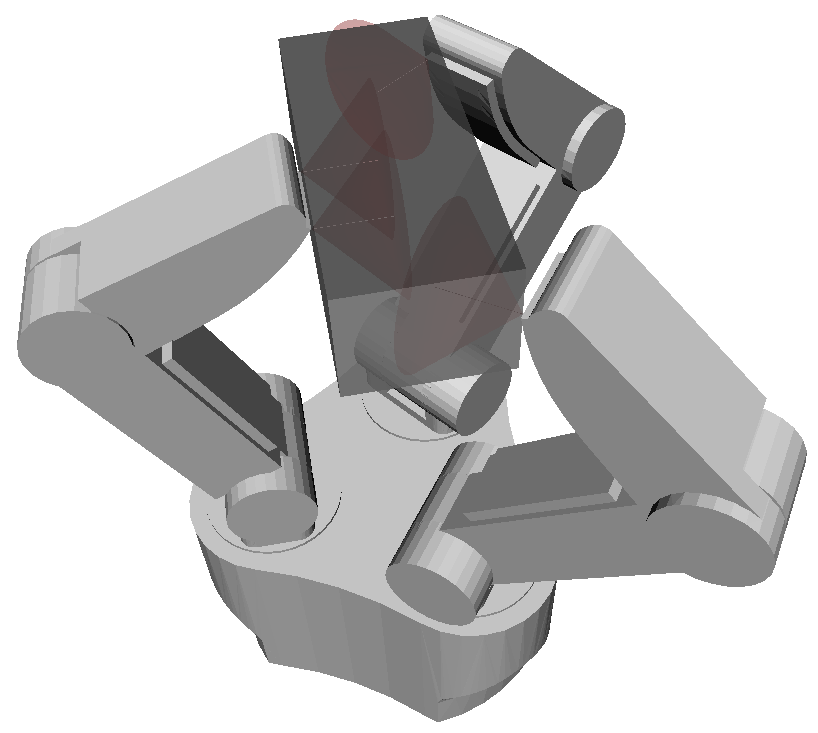}}
\subfigure[]{\includegraphics[width=1.8in]{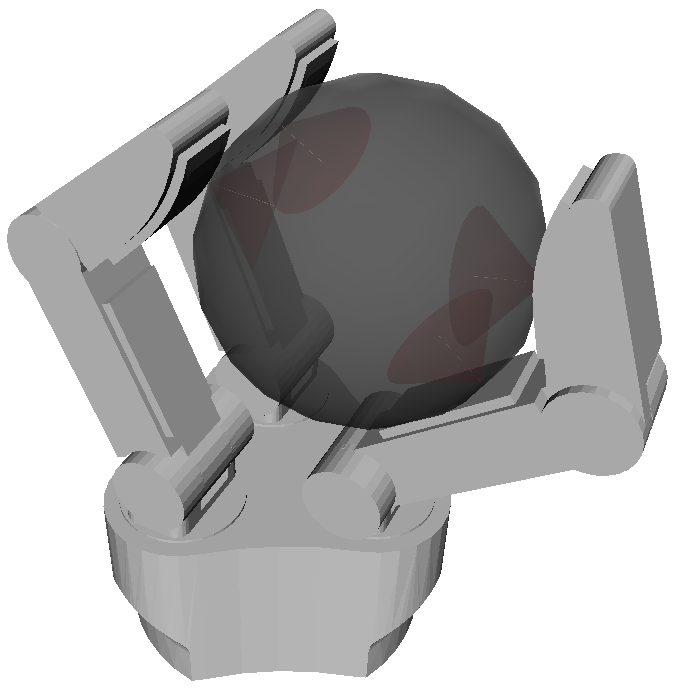}}
\subfigure[]{\includegraphics[width=1.8in]{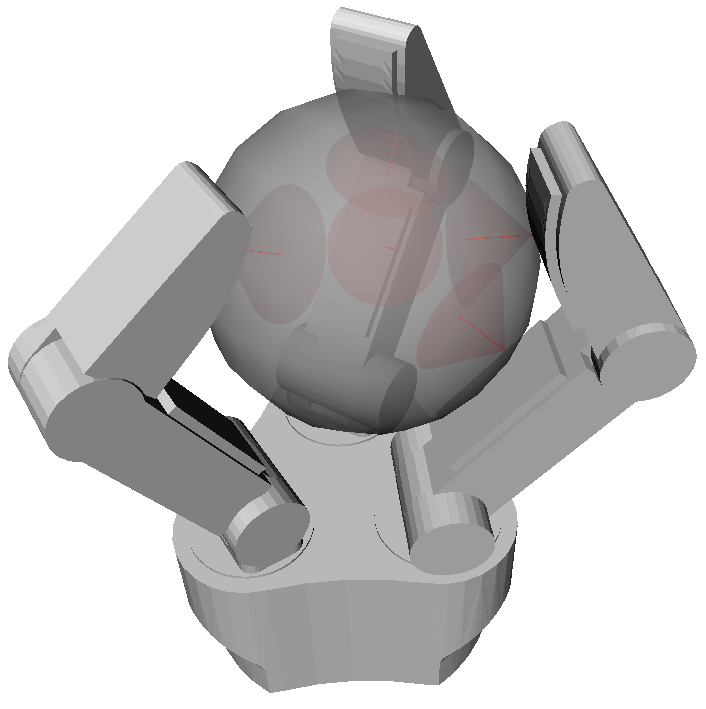}}
\subfigure[]{\includegraphics[width=1.8in]{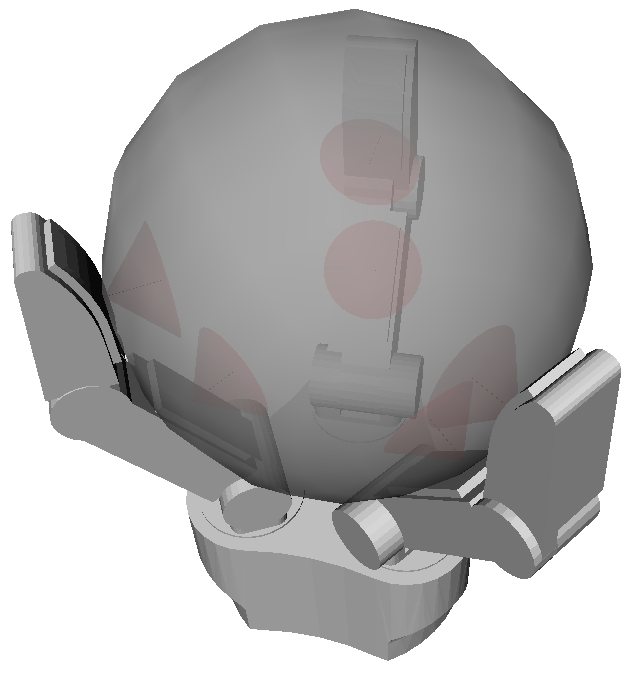}}
\subfigure[]{\includegraphics[width=1.8in]{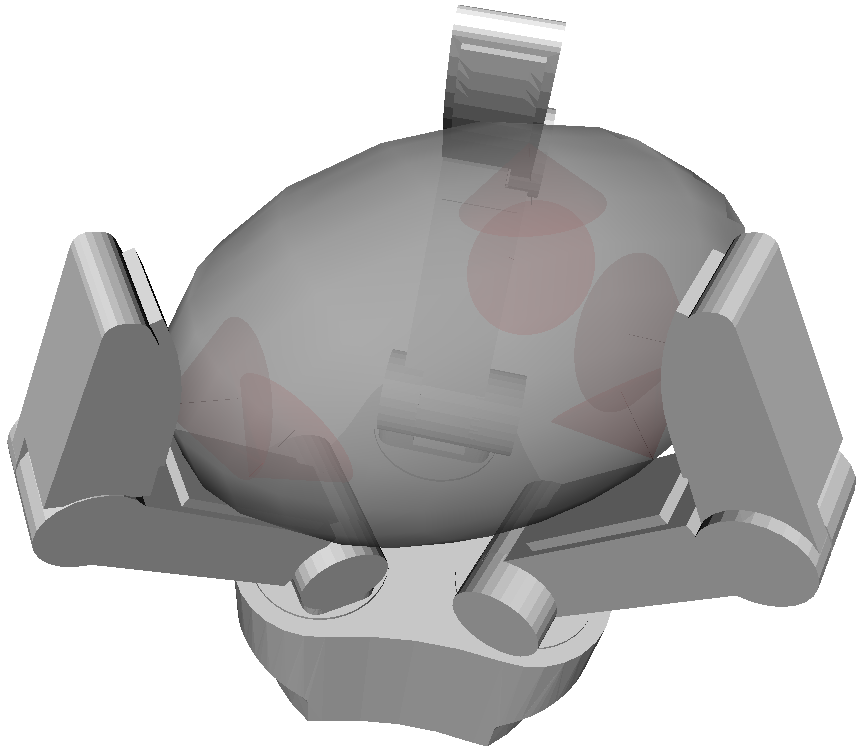}}
\caption{Algorithmically generated grasps to investigate the computational performance of our method.}\label{fig:comp_grasps}
\end{figure}

In order to investigate the practical applicability of our
method, we tested the runtime of our algorithm on a range of grasps that
could be encountered in a grasping task (see Fig.~\ref{fig:comp_grasps}.) We
generated these grasps on a range of differently shaped objects using a brute
force grasp planner~\cite{MEEKER19}. On each of these grasps we applied our
framework to perform the three tasks demonstrated above: Checking for stability,
finding the maximum resistible wrench in a given direction and computing the
optimal actuator commands. In the second task the direction along which to
find the maximum resistible disturbance can have an impact on the runtime of
the algorithm. We hence repeated the solution process with ten different
randomly generated directions to obtain more meaningful results. Similarly,
the time it takes to optimize the actuator commands depends on the external
wrench applied to the object and hence we also repeat these experiments with
ten randomly sampled external wrenches. The results are recorded in
Table~\ref{tab:comp_data}. In order to guarantee accurate results we continued
refinement until we reached a level equivalent to 2048 sectors in the friction
approximation. Empirically we have found this level of refinement guarantees
convergence for all grasps tested (although many converged sooner.)

We note that, as expected, the runtime of our algorithm grows with the
number of contacts. For grasps with three contacts, all queries
typically had sub-second runtime; up to and including five contacts
we noticed a runtime typically between 1 and 30 seconds. Also as
expected, stability checks (the most fundamental operation needed for
a grasp) are faster than computations like maximum resistible
disturbance or optimal joint torques. This suggests different
applications of our framework for different scenarios: pruning a
larger number of possible grasps using the faster stability analysis,
then computing the optimal torques only on the most promising
candidates. Even with six contacts, our method had a runtime on the
order of minutes, suitable for example for fixturing analysis in a
manufacturing line. Finally, we also note that, in the absence of our
hierarchical refinement, it is altogether intractable to approach this
level of accuracy for all except the simplest of grasps.

\section{Discussion}\label{sec:3d_furtherwork}

In this Chapter we described a grasp stability model that allows for efficient
and accurate solution methods under realistic constraints. Noting that an
exact formulation of Coulomb friction includes non-convex constraints (due to
the Maximum Dissipation Principle), we discussed how convex relaxations along
with successive hierarchical tightening can be used to solve non-convex
optimization problems. We specifically discussed McCormick envelopes and
pointed out their limitations.

Instead, we proposed a discretization method that allows the Coulomb friction
law to be reformulated as a piecewise convex Mixed Integer Program solvable
through branch and bound. However, such discretization methods traditionally
involve a trade-off: coarse discretizations provide only rough approximations
of the exact constraints, while high resolutions discretizations are
computationally intractable. To address this problem, we introduce a
hierarchical refinement method that progressively increases the resolution of
the discretization only in the relevant areas, guided by the solution found at
coarser levels. 

Our local refinement method remains efficient up to high discretization
resolution, and also similarly to traditional tightening methods for convex
relaxations provides strong guarantees: if a solution cannot be found at a
coarse approximation level, the underlying exact problem is guaranteed not to
have a solution either. Combined, these two features make our method efficient
for problems both with and without exact solutions. It is, to the best of our
knowledge, the first time that grasp stability models incorporating Coulomb
friction (along with the MDP) have been solved with such high discretization
resolution.

Our model and algorithm accepts many types of queries: for example, we can
analyze the space of wrenches applied to an object that a given grasp can
withstand, or compute optimal joint commands given a specific object wrench.
Thanks to the hierarchical refinement method, these can be solved efficiently
(on the order of seconds per query) even with very high resolution
approximations of the MDP.

Running our analysis method on a number of example grasps, we showed
that our method predicts effects both intuitive (pressing directly
against contacts is passively stable, but pulling the object away
requires preload torques) and more subtle (an object wedging itself in
a grasp in response to a disturbance for a given contact geometry.) In
contrast, grasp stability models that do not consider the MDP produce
unrealistic results, and fail to predict the dependence of disturbance
resistance on applied preloads.

A limitation of our method is that, while it performs well in
practice, its theoretical running time remains worst-case exponential
in the level of discretization for friction constraints. Furthermore,
for cases where a coarse discretization yields a sufficiently accurate
solution, hierarchical refinement might be outperformed by an
equivalent method with uniform resolution (although these cases are
generally unknowable in advance, without actually solving up to high
resolutions.)

From a practical perspective, we would like to explore additional applications
of our approach. In the context of grasp analysis and planning the runtime of our
algorithm is currently is too large for online grasp planning with more than a
few contact. Therefore, we see the main practical relevance of our method in enabling
practitioners to understand and utilize passive effects in grasping and in
providing labels for learning-based grasp planners. The model we introduced
can allow multiple types of queries, and in this paper we have only presented
some of the possible applications.

Our framework may also be applicable to problems encountered in the field of
robotic locomotion, for instance to determine the balance of a legged robot on
uneven terrain~\cite{BRETL08}. In fact, we believe that our model and
algorithm is applicable to many other fields concerned with the stability of
arrangements of rigid bodies with frictional contacts and other unilateral
constraints such as pushing in clutter and robotic construction.

From a theoretical perspective there are many interesting problems remaining.
Firstly, we took care that our convex relaxation satisfies the requirement
that a coarse relaxation contains the solution space of all successively
tighter relaxations. This characteristic is important for Branch and Bound
algorithms that prune parts of the solutions space based on the fact that the
minimum found in such a coarse relaxation provides a lower bound to the
minimum in any tighter relaxation. Instead of making use of this, we solve an
entire Branch and Bound problem at every refinement step. This means that
Branch and Bound is recomputing much of its search tree at every step.
Instead, we could save the tree generated at one refinement step and use it to
warmstart Branch and Bound at the next. This promises large improvements in
performance and the relaxation we proposed in this chapter satisfies all the
requirements for such an approach.

Furthermore, recent work by Huang et al.~\cite{Huang2020} built upon our own
work discussed in Chapter~\ref{sec:2d} and developed an algorithm that can
determine all possible combinations of contact slip states. The slip state of
a contact is defined as the sector of a discretized tangential plane
containing the relative tangential contact motion --- much the same as in our
work discussed above. They make use of the fact that not every combination of
contact slip states is possible under rigid body motion constraints. Their
algorithm could be a useful pre-processing tool that can prune large amounts
of the Branch and Bound search tree based on these rigid body
constraints.  

Finally, we have thus far turned to commercial solvers for the solution of the
Mixed Integer Programs that describe our grasp models. These solvers search
for an exact solution through computationally expensive methods such as the
Branch and Bound algorithm. Recently, however, there has been promising work
on the development of more efficient, albeit approximate solution
methods~\cite{WU18}. Reducing the computational demands of our algorithms
would greatly affect their practical applicability and hence we will
investigate these novel methods.

\chapter{Grasp Stability Analysis for Reinforcement Learning of in-hand~manipulation}\label{sec:shield}

\section{Introduction}\label{sec:shield_intro}

In-hand manipulation (i.e. reorienting a grasped object with respect to the
grasping hand) is a challenging task in robotics due to the problem's large
state-space and hard to model dynamics. While there are numerous analytical
approaches making use of planning and optimization algorithms (some of which
we mentioned in Chapter~\ref{sec:rel_ihm}) they rely on the ability to derive
analytic models of the kinematics and dynamics of the bodies involved as well
as the contact phenomena. This results in many complications as small errors
in the models can have dramatic effects. Accurately modeling friction forces
by itself is a large area of study (see Chapter~\ref{sec:related_dynamics}.)
Furthermore, partial observability of the state --- unknown geometries as well
as uncertainty in the contact positions and forces due to occlusion for
instance --- make it impossible to compute complete models of the grasp. 

In contrast, model-free reinforcement learning takes an end-to-end approach
treating the grasp and its dynamics as a black box: The grasped object, hand
kinematics and contact mechanics are assumed unknown. Reinforcement learning
reasons probabilistically: Through stochastic exploration a policy is trained
to learn which actions maximize rewards given the current partial observation
of the state. It can therefore be applied to problems with partial
observability such as most manipulation tasks. In consequence, however, such
methods forgo any formal guarantee of success.

While model-free RL has shown great promise in high-dimensional manipulation
tasks, its sample complexity remains problematic. Furthermore, policies
applied to in-hand manipulation on real robots thus far all require some level
of external support for the manipulated object. This is either done through
resting the object on a tabletop~\cite{7363524}, on the up-turned palm of the
hand~\cite{doi:10.1177/0278364919887447}\cite{openai2019solving}\cite{Rajeswaran-RSS-18},
or fixing some of its degrees of freedom through external
constraints~\cite{zhu2018dexterous}.

It is our intuition that the notion of maintaining a robust quasi-statically
stable grasp throughout a task greatly simplifies general in-hand
manipulation. While humans leverage dynamic phenomena to great effect
controlling them requires rich tactile feedback and great skill. Thus we will
focus on manipulation skills that maintain quasi-static stability as we
believe the are more easily attainable for robotic systems, albeit less
powerful than general manipulation leveraging dynamic effects.

We also believe that efforts to maintain quasi-static stability throughout a
manipulation task can improve the sample complexity of training RL policies
for dexterous manipulation\footnote{Bicchi~\cite{897777} defines dexterous
manipulation as "the capability of the hand to manipulate objects so as to
relocate them arbitrarily for the purposes of the task"}. Consider the process
of collecting data in order to train a policy: whenever the grasp becomes
unstable and the object is dropped the rest of the episode can no longer
contribute to learning the task. While resetting the environment is trivial in
simulation, on a real robot it requires a reset mechanism, which can be
complicated to design.

In Chapter~\ref{sec:bridge} we argued that robotic hands designed solely for
grasping tasks (such as picking an object from clutter) have evolved in such a
way as to trade off complexity for passive stability. Most researchers
applying reinforcement learning are currently leveraging vision and motion
capture instead of tactile
sensing~\cite{doi:10.1177/0278364919887447}\cite{openai2019solving}\cite{zhu2018dexterous}\cite{Rajeswaran-RSS-18}.
We believe the reason for this lies in the limited tactile sensing
capabilities of the current generation of robotic hands and expect tactile
sensing to begin playing a more prominent role as this technology evolves. In
fact, members of our group are actively developing and building tactile
sensors that will supply us with high fidelity contact position and force
information over the majority of the finger surface~\cite{9006916}.

As a new generation of highly sensorized multi-finger robotic hands is
beginning to emerge the use of rich sensor data from tactile and
proprioceptive sensors will become a viable alternative to purely vision based
approaches to dexterous in-hand manipulation. Note that, while the global
shape of the object may be unknown tactile sensors allow us to observe the
local shape of the object. This new generation of robotic hands can also
satisfy the assumption that is central to most of the grasp modeling
literature we discussed in Chapter~\ref{sec:related_closure}: Active control
over contact forces through actuator commands. Thus we can determine the
(local) stability of a grasp or actuator commands necessary for stability
using classical grasp stability theory.

However, it appears the theoretical works have had little impact in the
learning community so far. While we believe RL by itself to be able to train
policies that maintain stable grasps throughout a reorientation task (we show
this is the case in Chapter~\ref{sec:pure_RL}) it appears wasteful to spend
training time in order to learn an aspect of the task for which we already
have good analytical tools. Thus we argue there is merit to the idea of
applying such model-based approaches to the smaller problem of maintaining
quasi-static grasp stability (for which the necessary state variables are
locally fully observable) while using model-free methods for the larger
problem of global manipulation planning under partial observability.

In this chapter we present exploratory work. We will demonstrate that pure
model-free reinforcement learning can be successfully applied to in-hand
manipulation tasks in simulation. It was important to us to find a task on
which pure reinforcement learning can generate satisfactory policies to ensure
that RL is capable of solving the manipulation planning problem and to serve
as a baseline for later comparison.

We will then elaborate on the intuition behind introducing analytic grasp
modeling theory and propose a method that uses rich tactile feedback as well
as such analytical hand models to modify RL actions such that grasp stability
is maintained at all times. Work in this direction is ongoing, with no results
to report at the time of this writing, but numerous experiments currently
being undertaken. Finally, we will discuss some of the many open questions
that remain.

\section{In-hand manipulation with pure reinforcement learning}\label{sec:pure_RL}

\subsection{Experimental design}\label{sec:design}

\begin{figure}[t!]
\centering
\includegraphics[width=0.65\columnwidth]{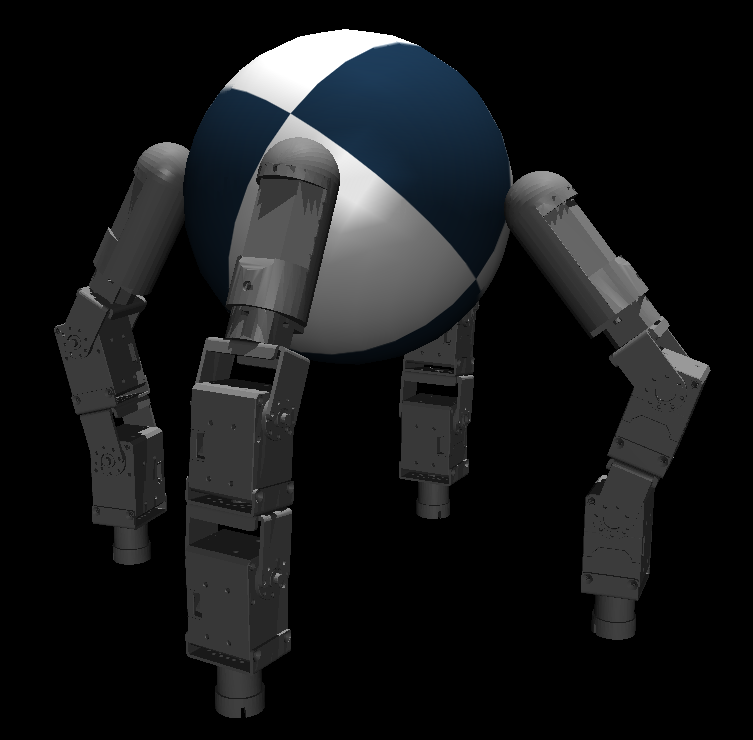}

\caption{Simulation of the four fingered hand and spherical object in
MuJoCo.}

\label{fig:ffh}
\end{figure}

Let us first describe the experimental setup we chose in order to investigate
robotic in-hand manipulation. We created a hand model in the MuJoCo physics
simulator~\cite{6386109}, which provides us both with a fast simulation of the
grasp dynamics as well as a method of determining contact positions and
forces. The hand we constructed has four identical fingers and 12 independent
fully actuated degrees of freedom (see Fig.~\ref{fig:ffh}.) As we are
currently also constructing a physical version of this hand we chose the
physical parameters such as the finger skin stiffness and actuator gains to
correspond to those of the physical hand design.

We chose to place a series elastic coupling between each joint and
corresponding actuator as series elastic couplings of known stiffness along
with encoders on both ends of the coupling provide a simple method of torque
measurement. This also provides us with multiple choices with regards to the
low-level control scheme of the hand. What we mean by this is choosing if the
setpoint provided to the actuator controller is in terms of joint torques or
joint positions for instance. While series elastic actuators with torque
sensing capabilities would allow for the implementation of torque control or
even more complex variants of impedance control there are convincing reasons
to use position control instead.

Consider a grasp in equilibrium using a torque controlled hand. Any change to
the actuator setpoint will disturb the equilibrium and lead to growing
accelerations and thus an unstable
grasp~\cite{doi:10.1177/027836499601500203}. In control theoretic terms such a
grasp is not in stable equilibrium as a small disturbance will cause the
system to accelerate away from equilibrium. In contrast, a position controlled
hand with compliant actuators allows for grasps that are in stable
equilibrium: A small change in setpoint likely results in the system moving to
a new equilibrium. We therefore chose position controlled actuators such that
we can achieve grasps that are in stable equilibrium.

For the sake of simplicity we chose a spherical object of mass 0.1kg for
these initial experiments. The local geometry of a sphere is uniform across
its surface and thus we hope to mitigate some effects of partial
observability. Using the Proximal Policy
Optimization~\cite{schulman2017proximal} algorithm we trained policies using
as a reward the instantaneous rotational velocity of the sphere around the
global vertical axis. We use a neural network with 4 hidden layers of 512
nodes each to encode the policy. 

The observation consists of the joint positions, actuator setpoints, contact
positions and contact normal forces. At every time step we sample an action
from the policy $\bm{a} \in \mathbb{R}^{12}$ from which we calculate the
actuator setpoint change $\Delta \bm{q} \in \mathbb{R}^{12}$. In order to
limit the velocity of hand joints we define vector $\bm{q}^{max}$, which
contains the maximum permissible setpoint change magnitude for each joint. A
value of 0.05 radians appeared to work well with a policy frequency of 20Hz.

\begin{equation}
\Delta q_j = max(-1, min(1, a_j)) \cdot q_j^{max},\quad \text{for all actuators j}
\end{equation}

\subsection{Results}\label{sec:RL_results}

\begin{figure}[t!]
\centering
\includegraphics[width=0.95\columnwidth]{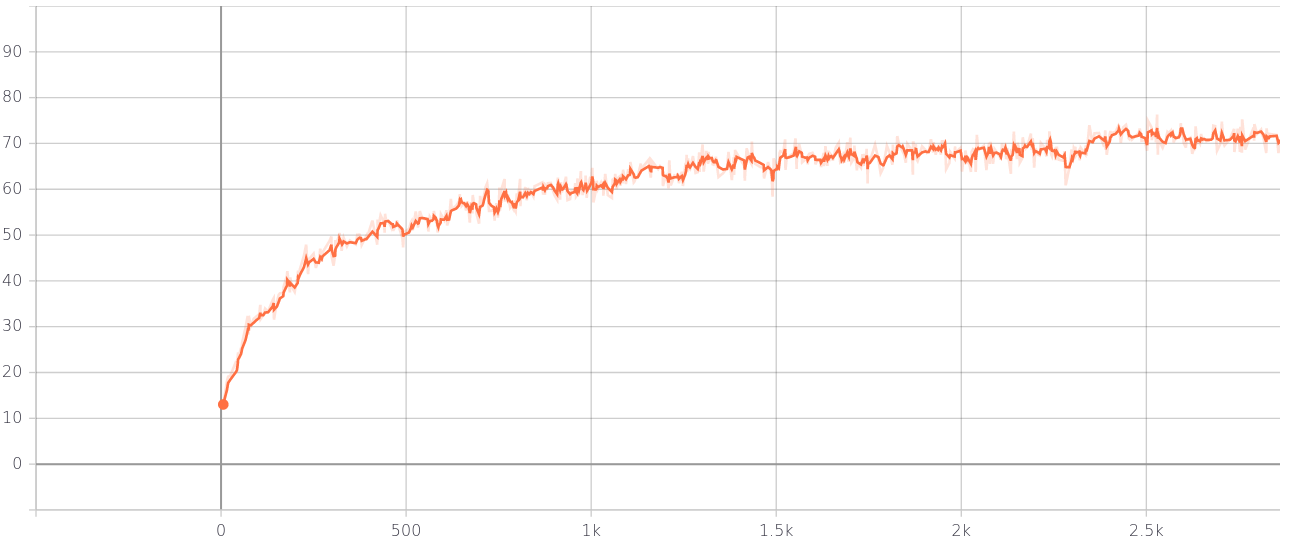}

\caption{Training curve for a successful in-hand manipulation task. Shown on
the horizontal axis is the number of policy updates. Each update consists of
96 episodes of 64 steps each.}

\label{fig:reward}
\end{figure}

\begin{figure}[t!]
\centering
\subfigure[]{\includegraphics[width=0.3\columnwidth]{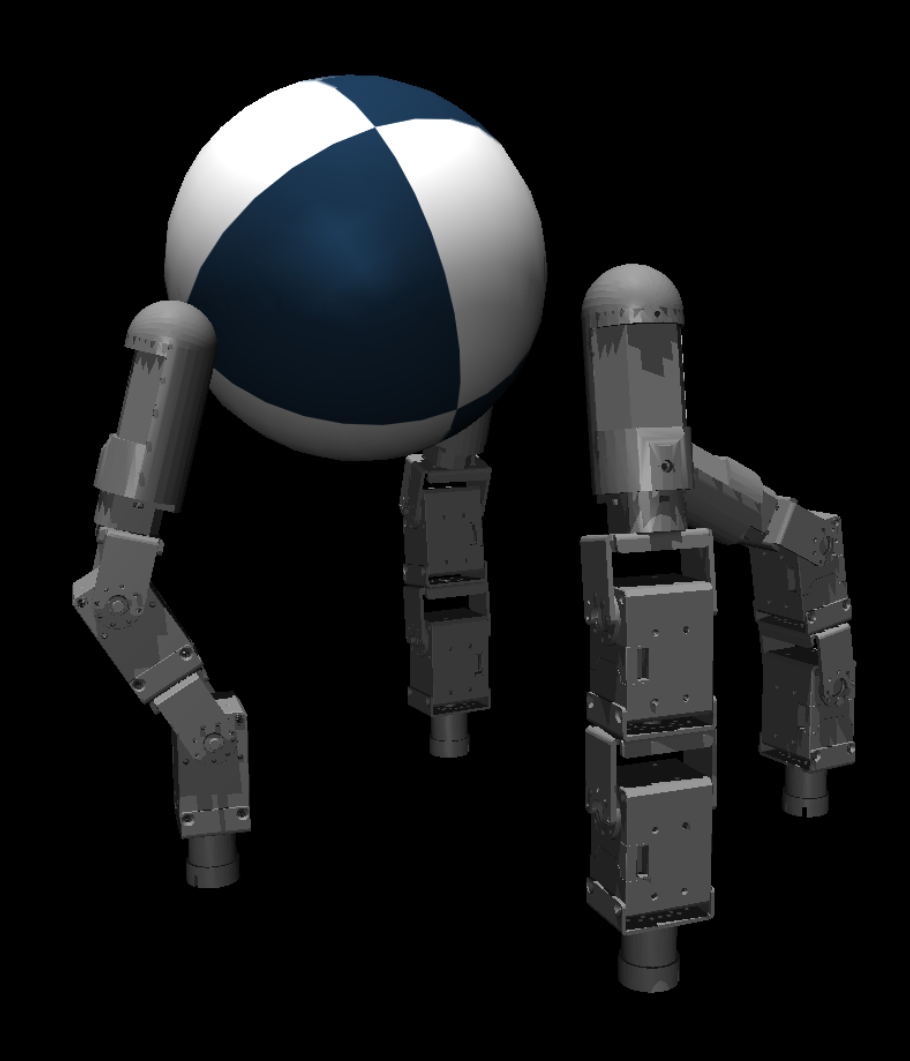}}\hfill%
\subfigure[]{\includegraphics[width=0.3\columnwidth]{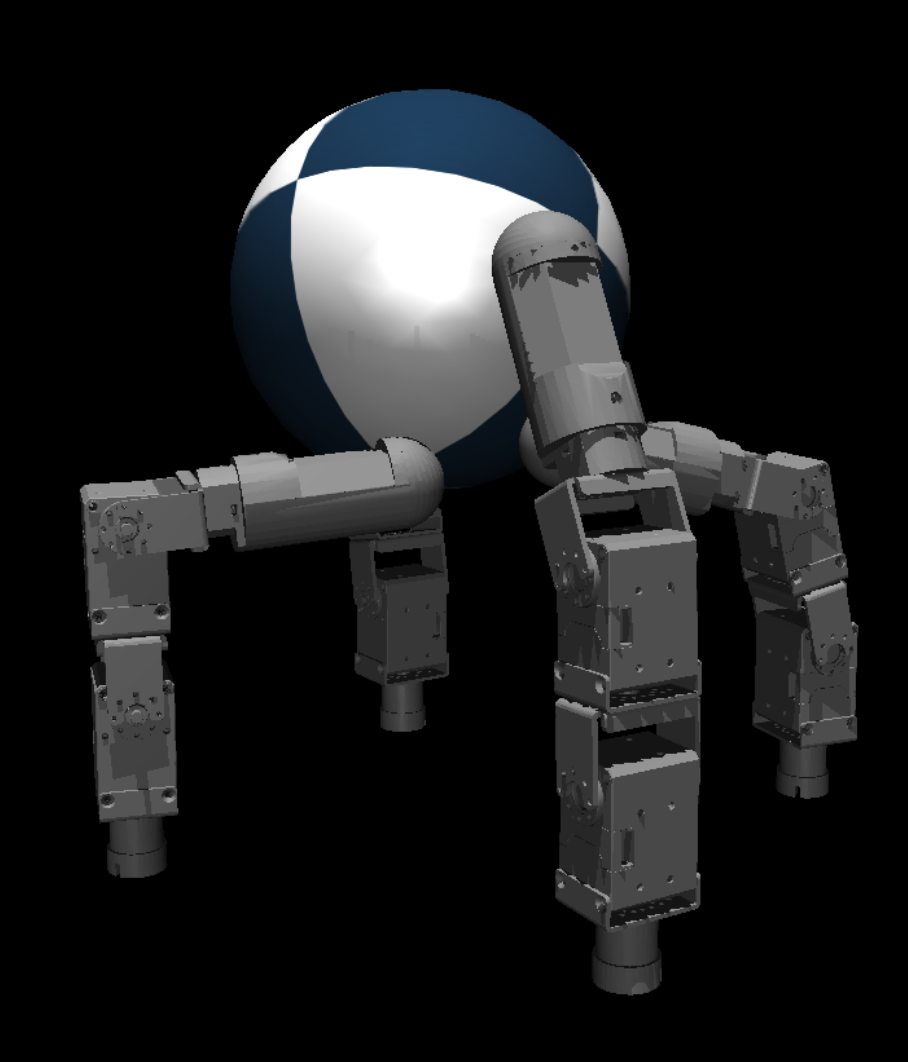}}\hfill%
\subfigure[]{\includegraphics[width=0.3\columnwidth]{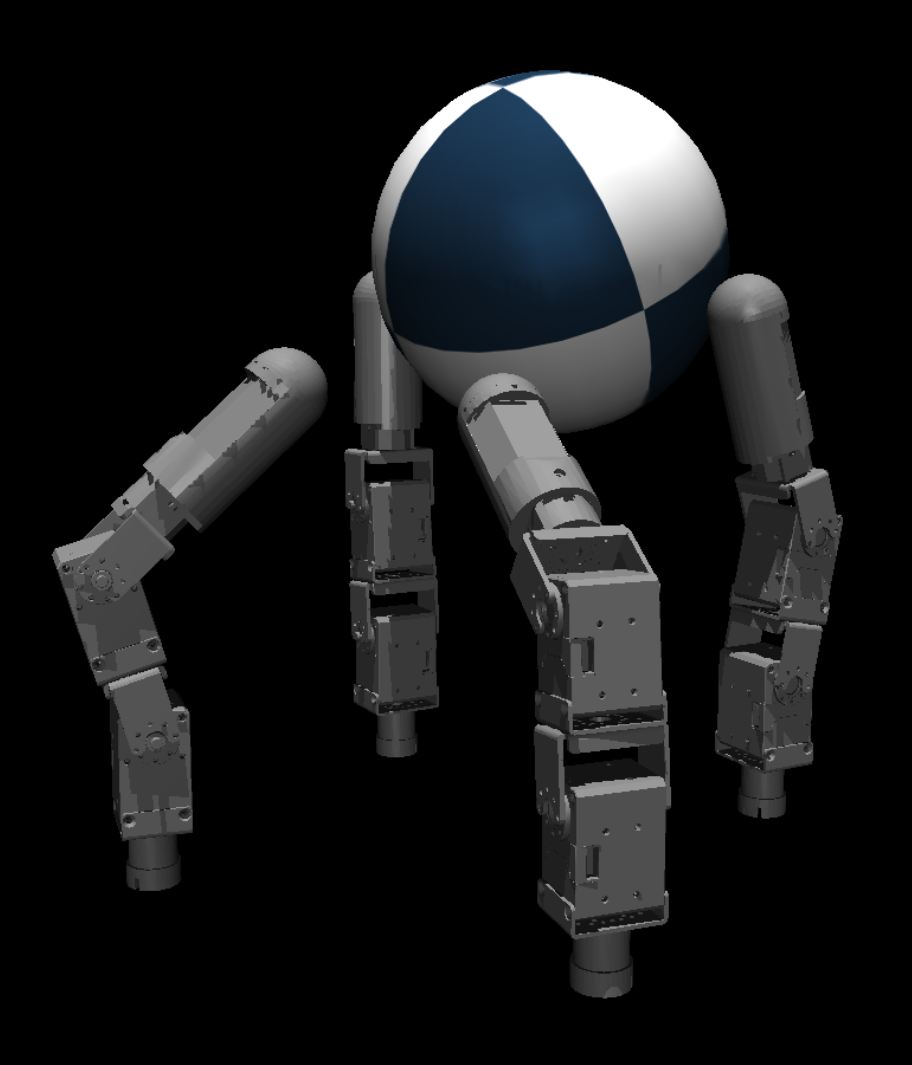}}\hfill%
\subfigure[]{\includegraphics[width=0.3\columnwidth]{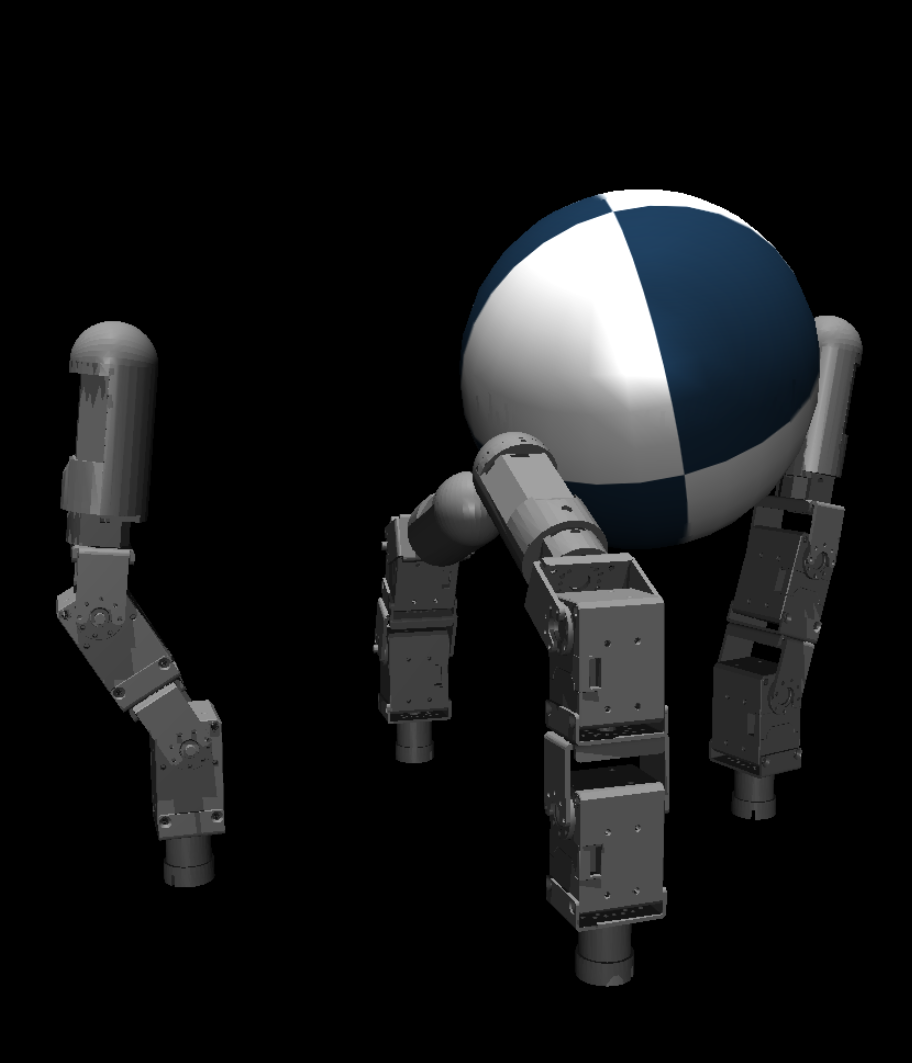}}\hfill%
\subfigure[]{\includegraphics[width=0.3\columnwidth]{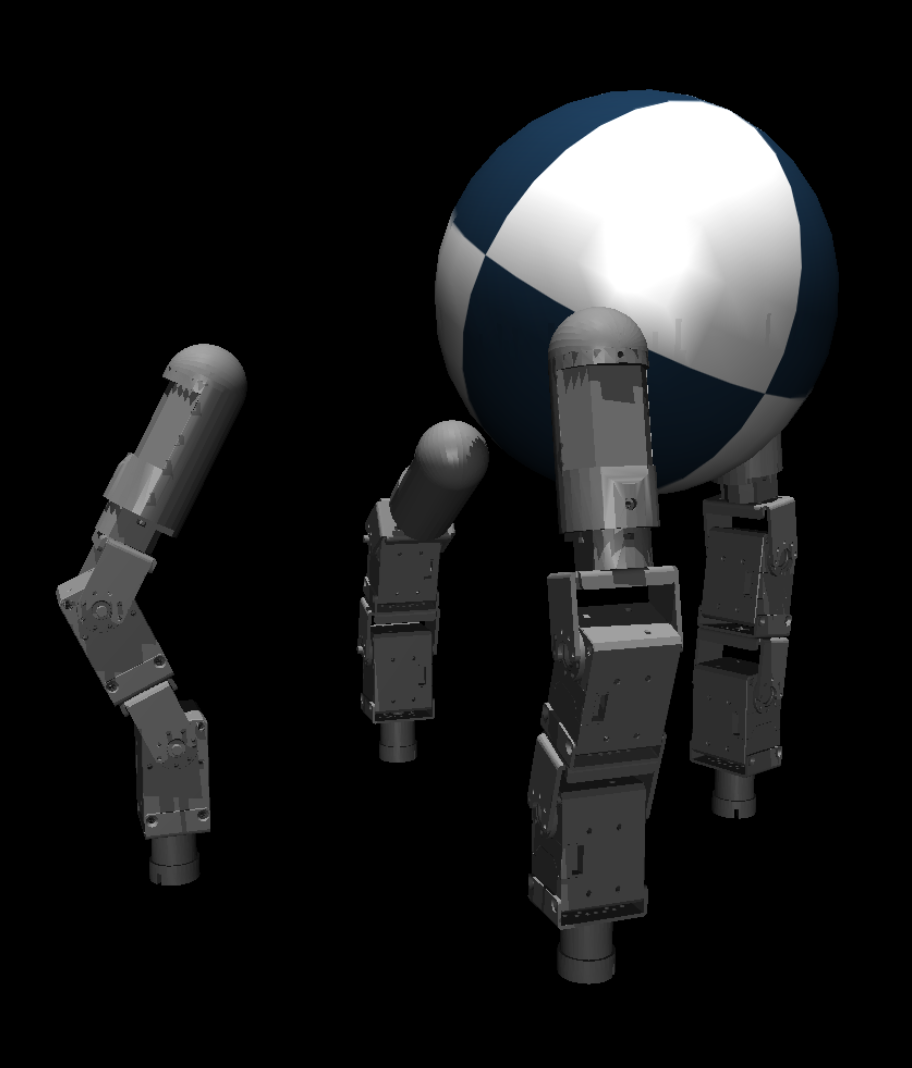}}\hfill%
\subfigure[]{\includegraphics[width=0.3\columnwidth]{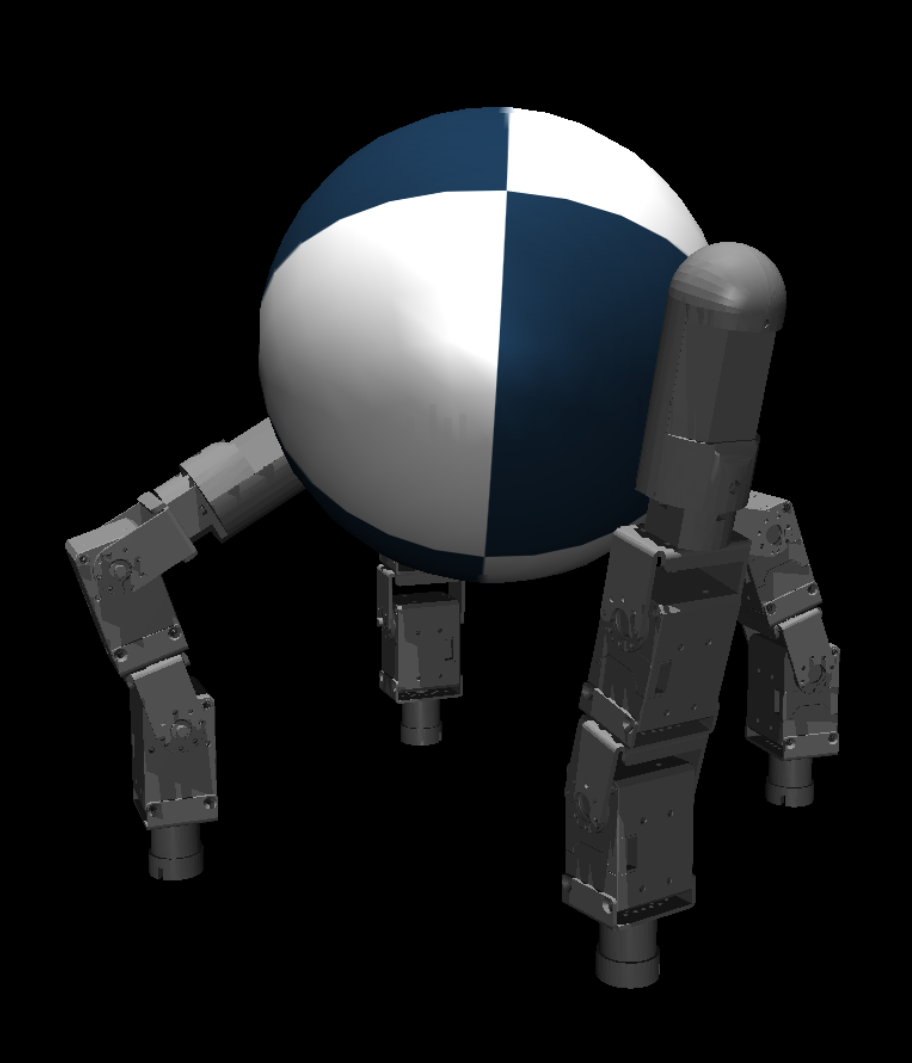}}\hfill%
\subfigure[]{\includegraphics[width=0.3\columnwidth]{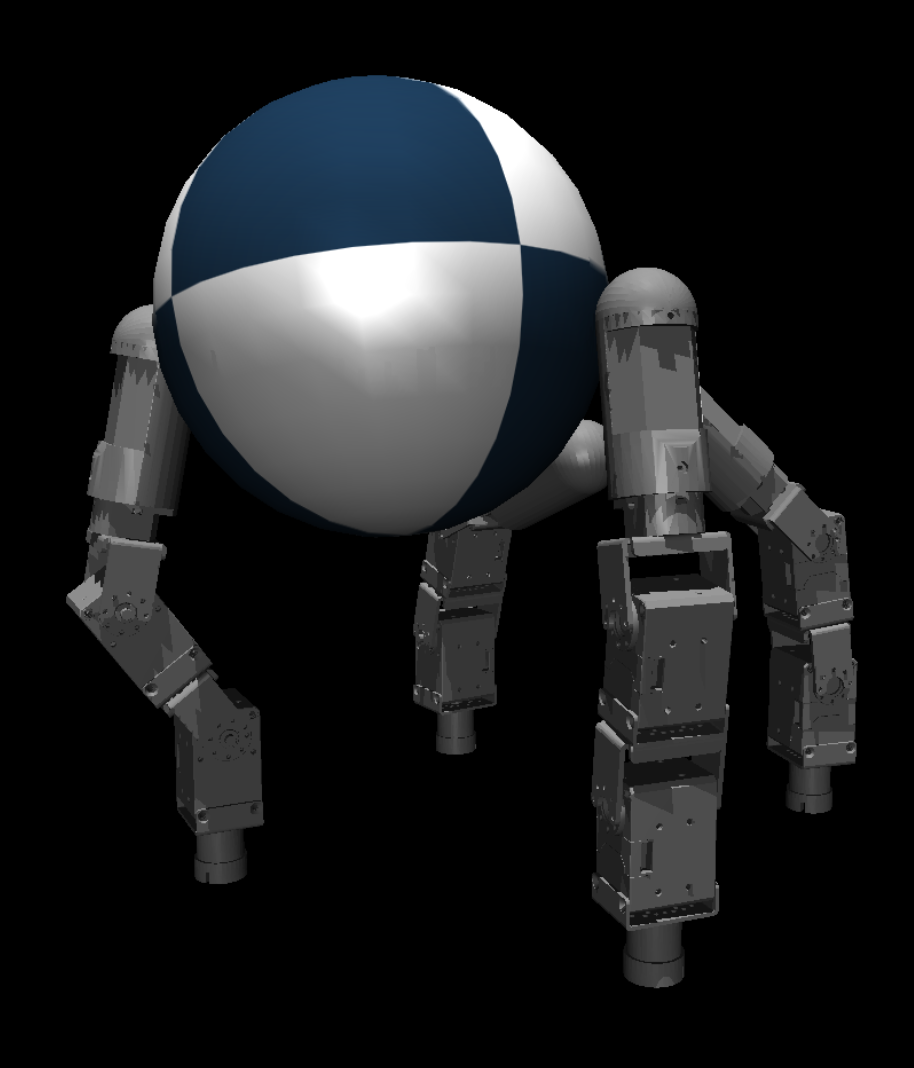}}\hfill%
\subfigure[]{\includegraphics[width=0.3\columnwidth]{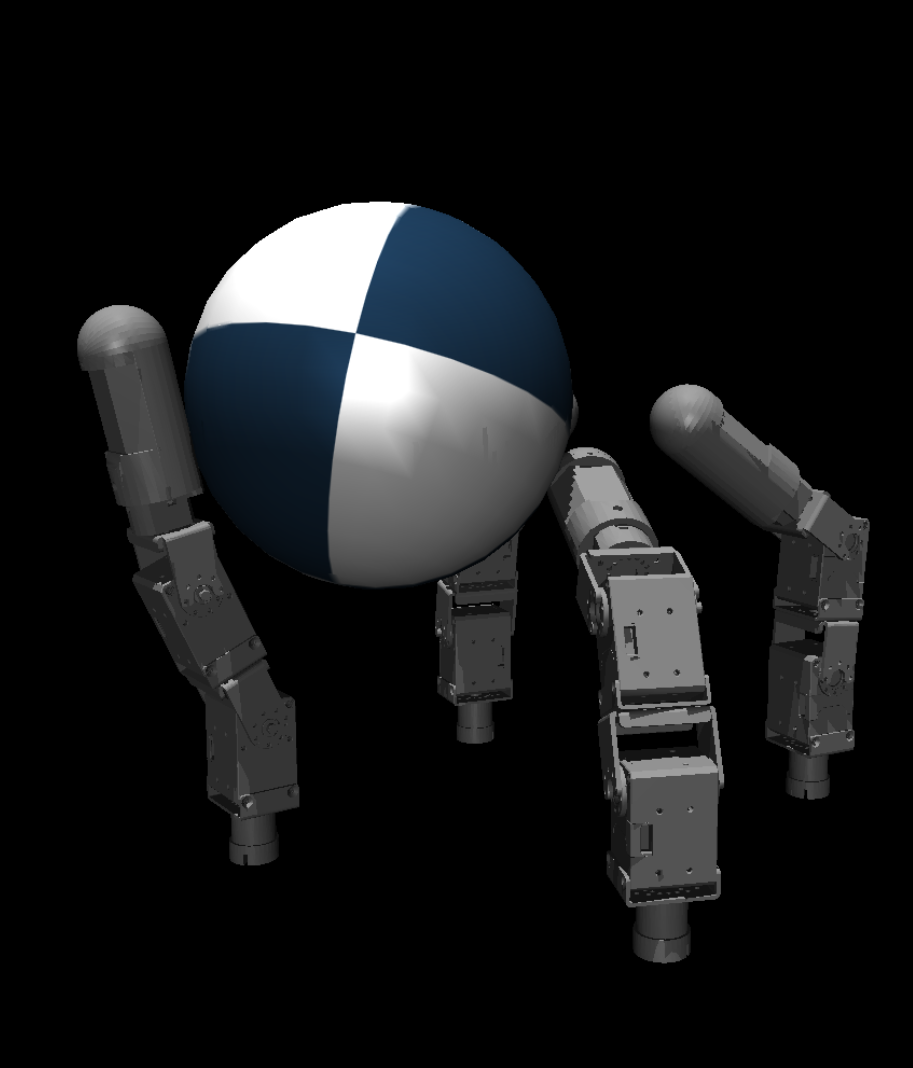}}\hfill%
\subfigure[]{\includegraphics[width=0.3\columnwidth]{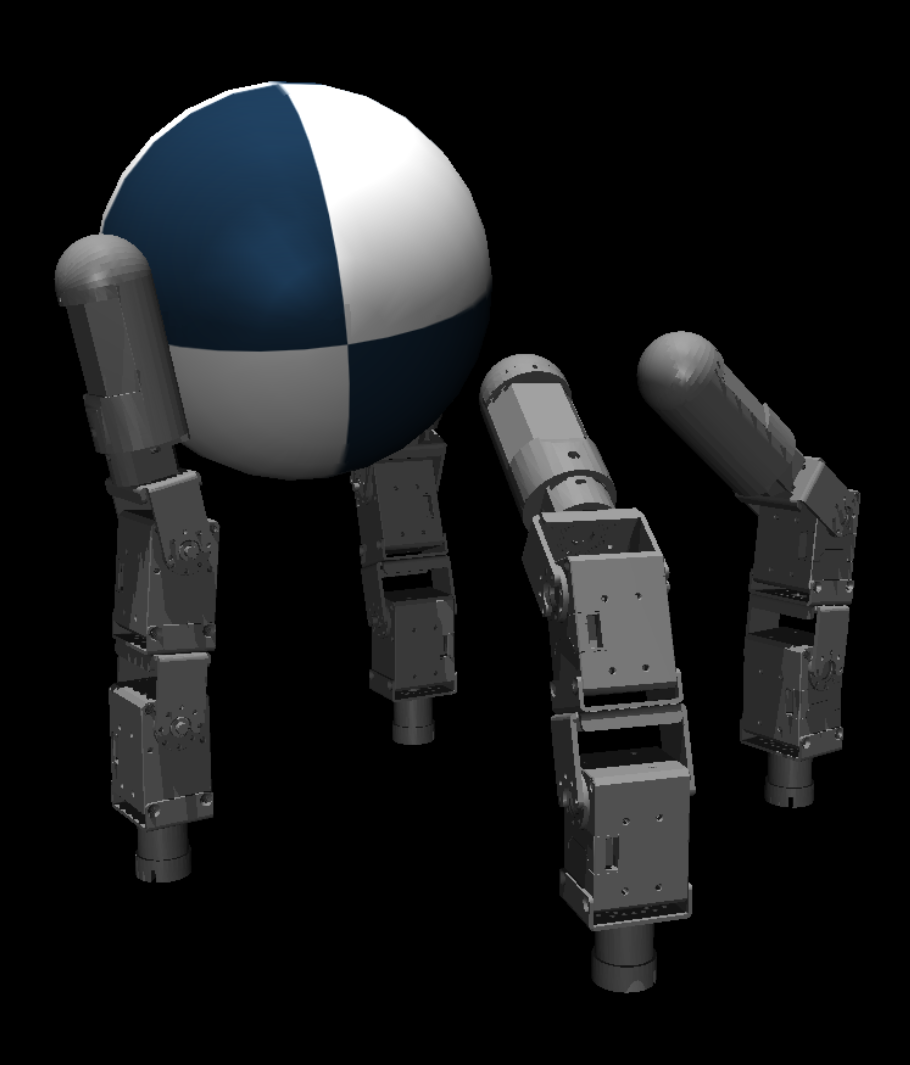}}\hfill%

\caption{One complete period of the manipulation gait learned with pure
reinforcement learning as seen from the side.}

\label{fig:manip1}
\end{figure}

\begin{figure}[t!]
\centering
\subfigure[]{\includegraphics[width=0.3\columnwidth]{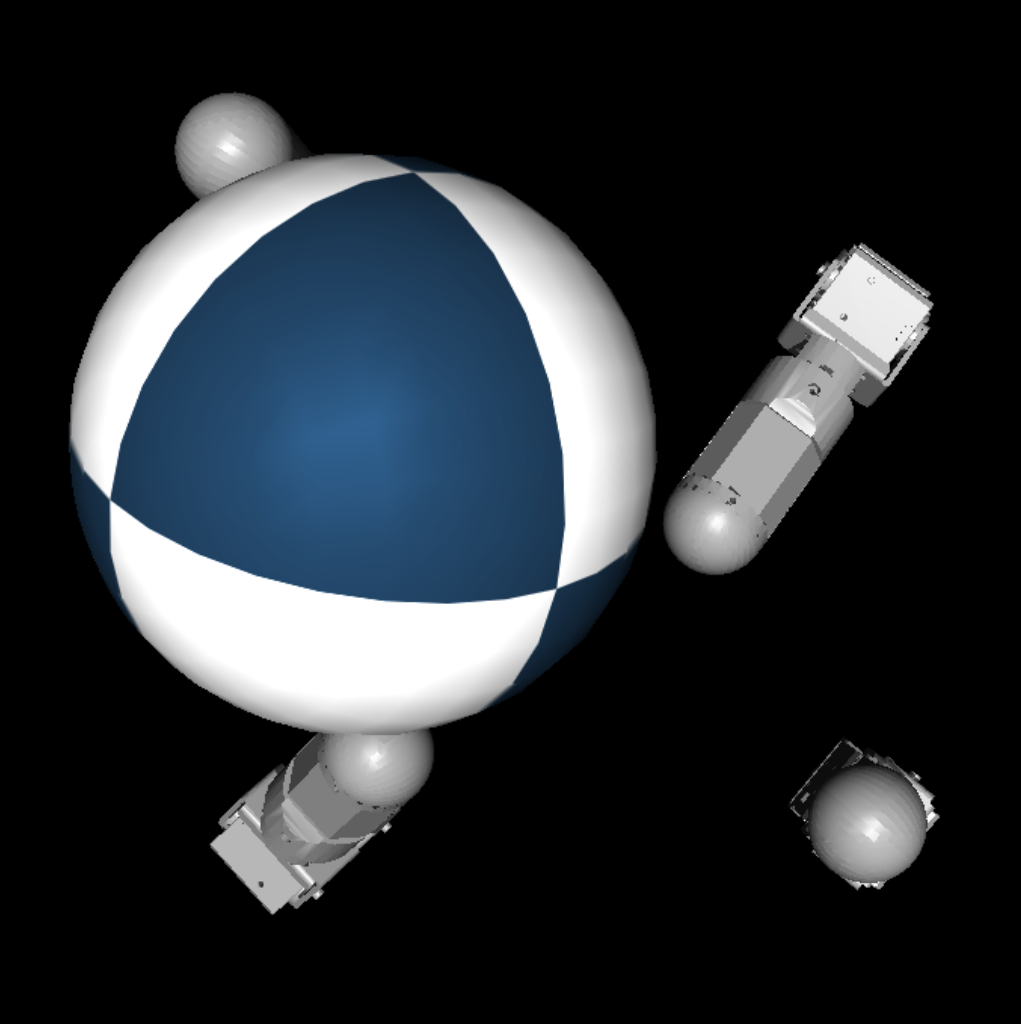}}\hfill%
\subfigure[]{\includegraphics[width=0.3\columnwidth]{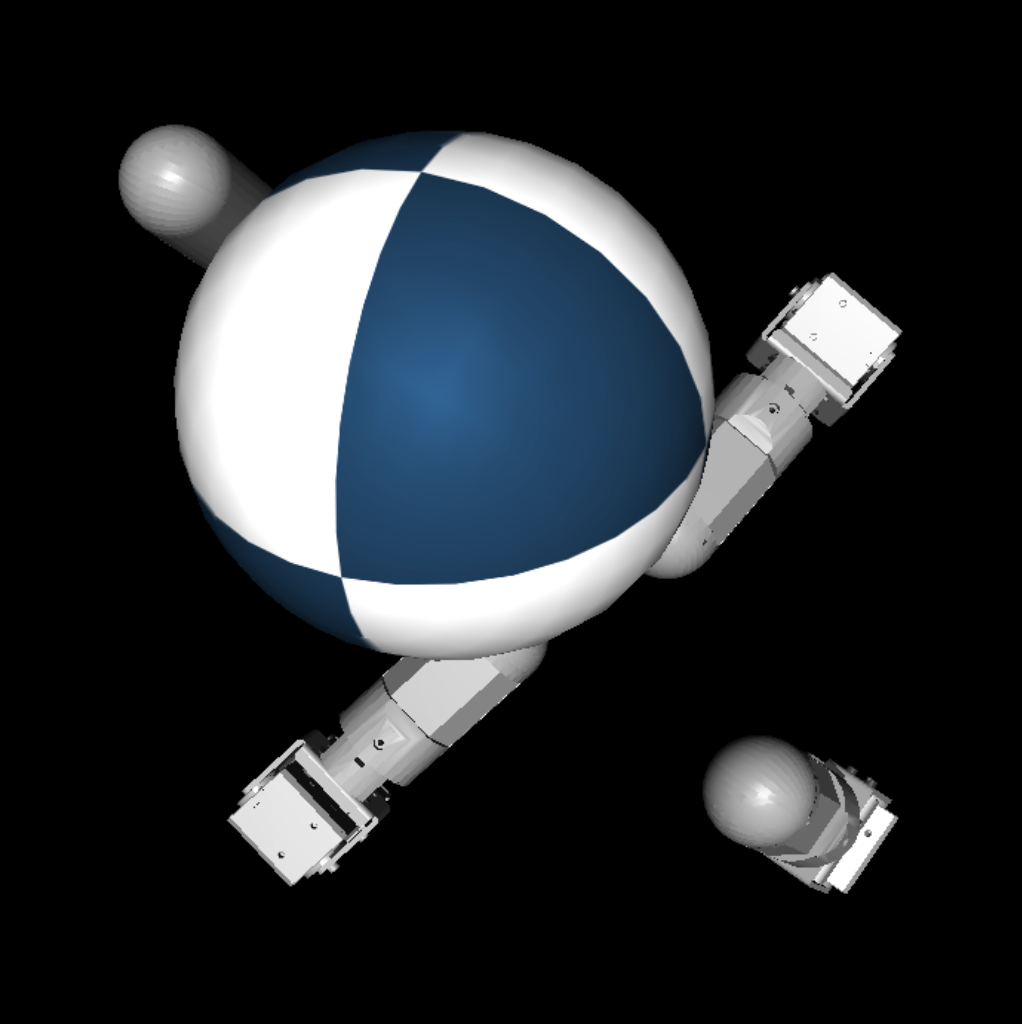}}\hfill%
\subfigure[]{\includegraphics[width=0.3\columnwidth]{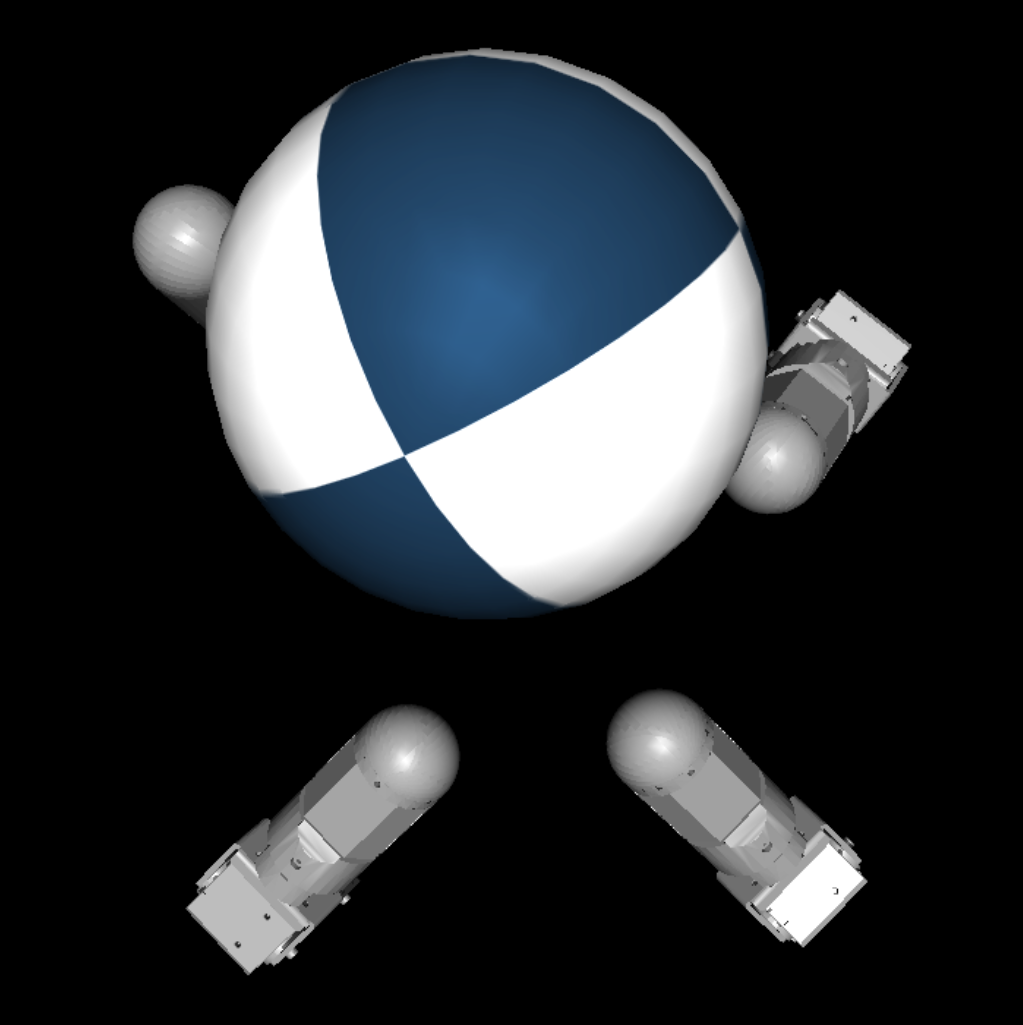}}\hfill%
\subfigure[]{\includegraphics[width=0.3\columnwidth]{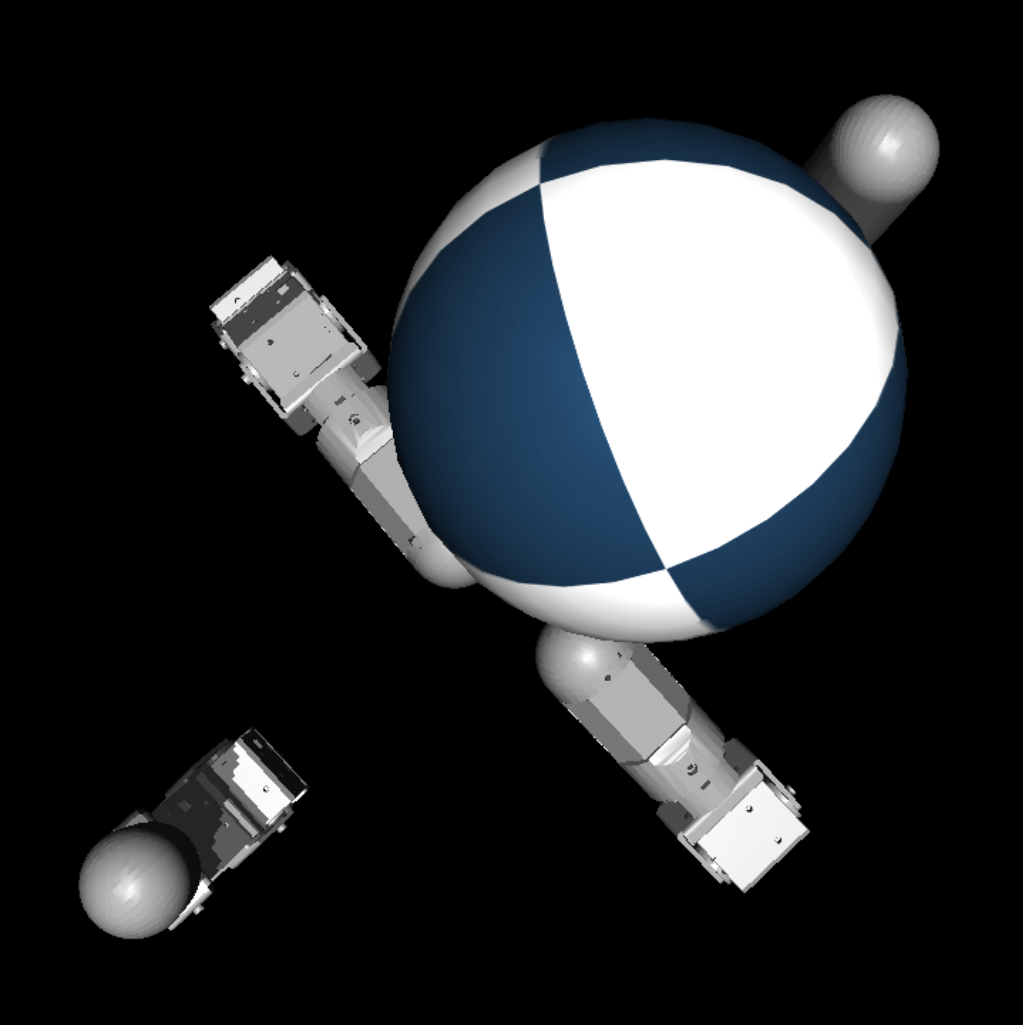}}\hfill%
\subfigure[]{\includegraphics[width=0.3\columnwidth]{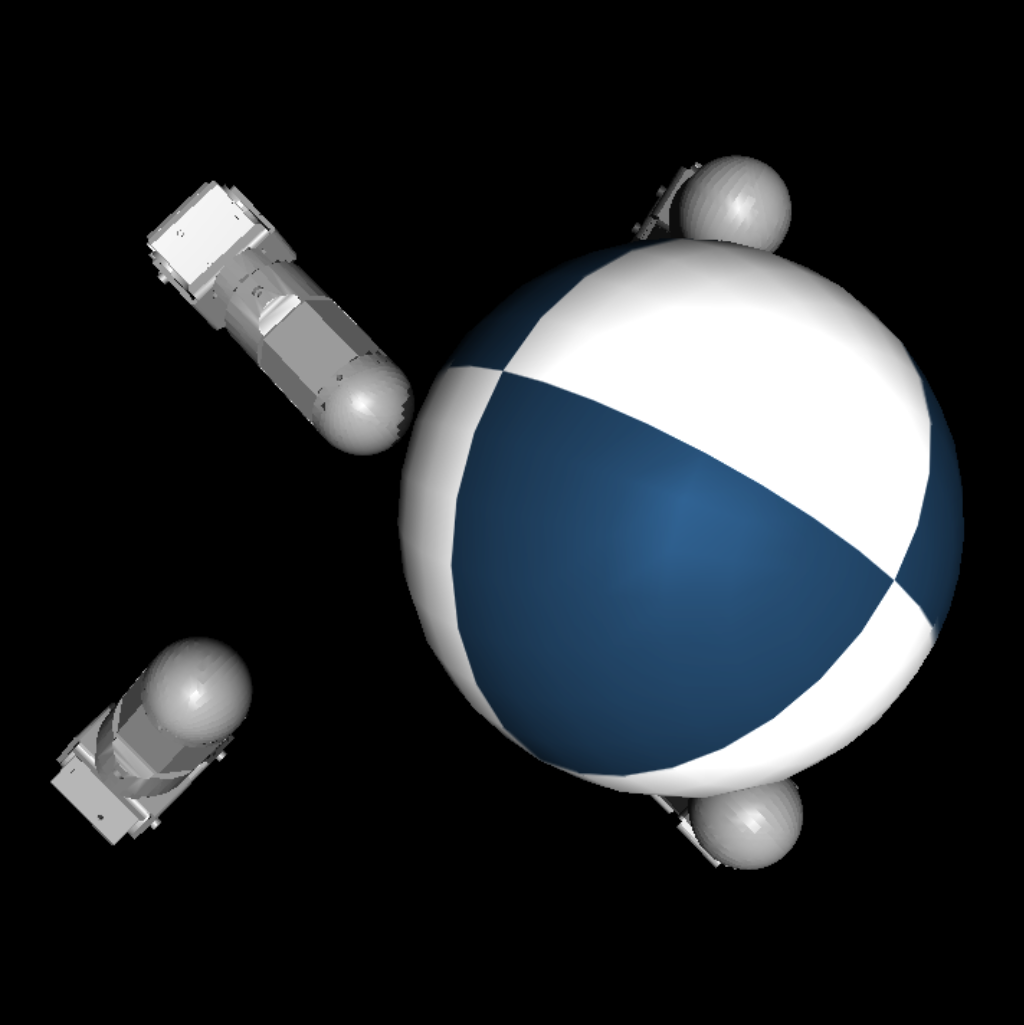}}\hfill%
\subfigure[]{\includegraphics[width=0.3\columnwidth]{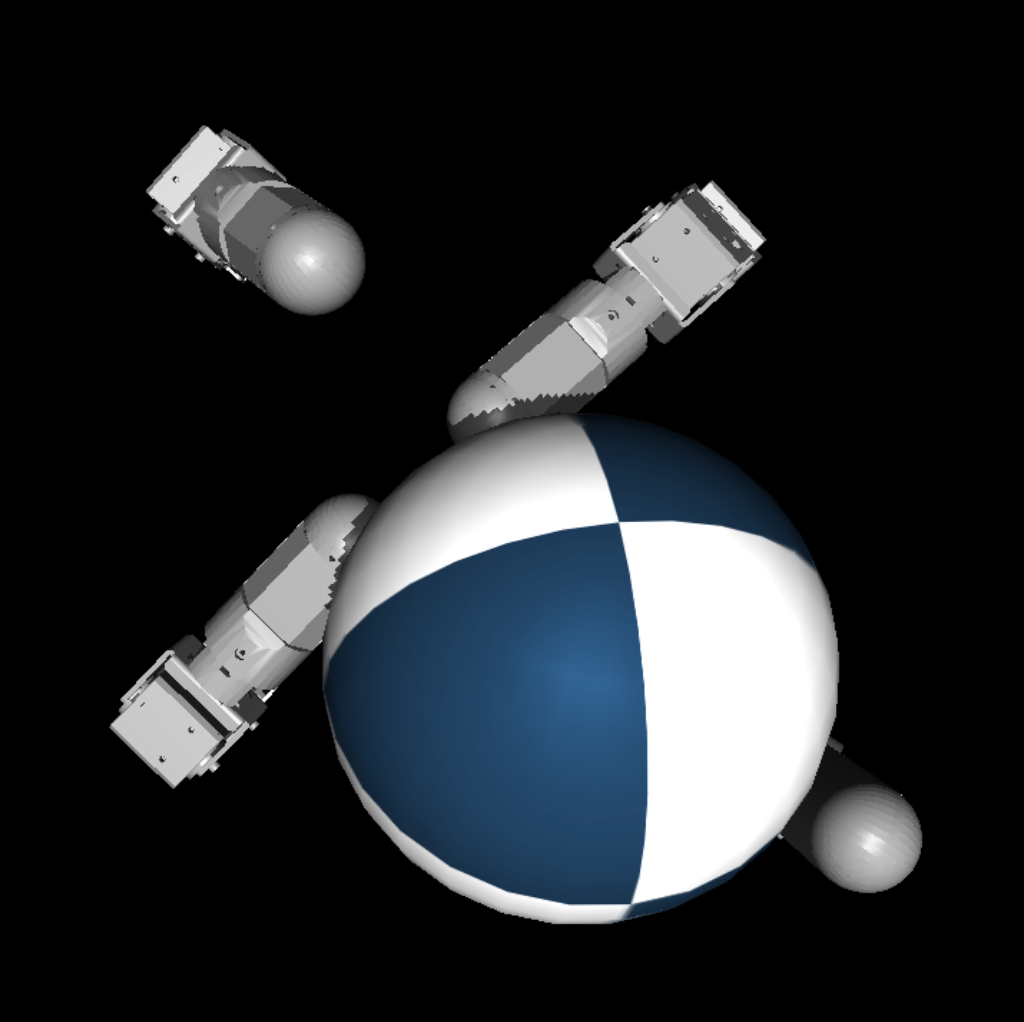}}\hfill%
\subfigure[]{\includegraphics[width=0.3\columnwidth]{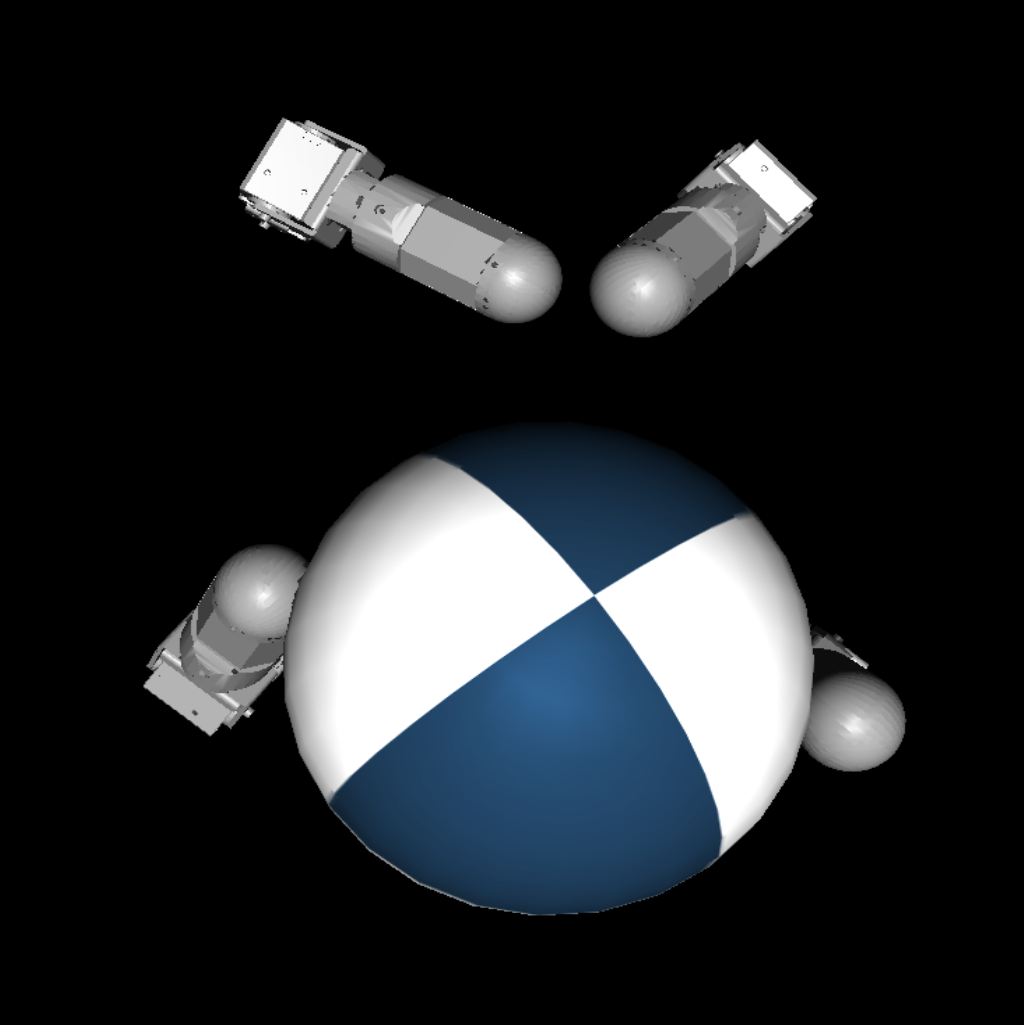}}\hfill%
\subfigure[]{\includegraphics[width=0.3\columnwidth]{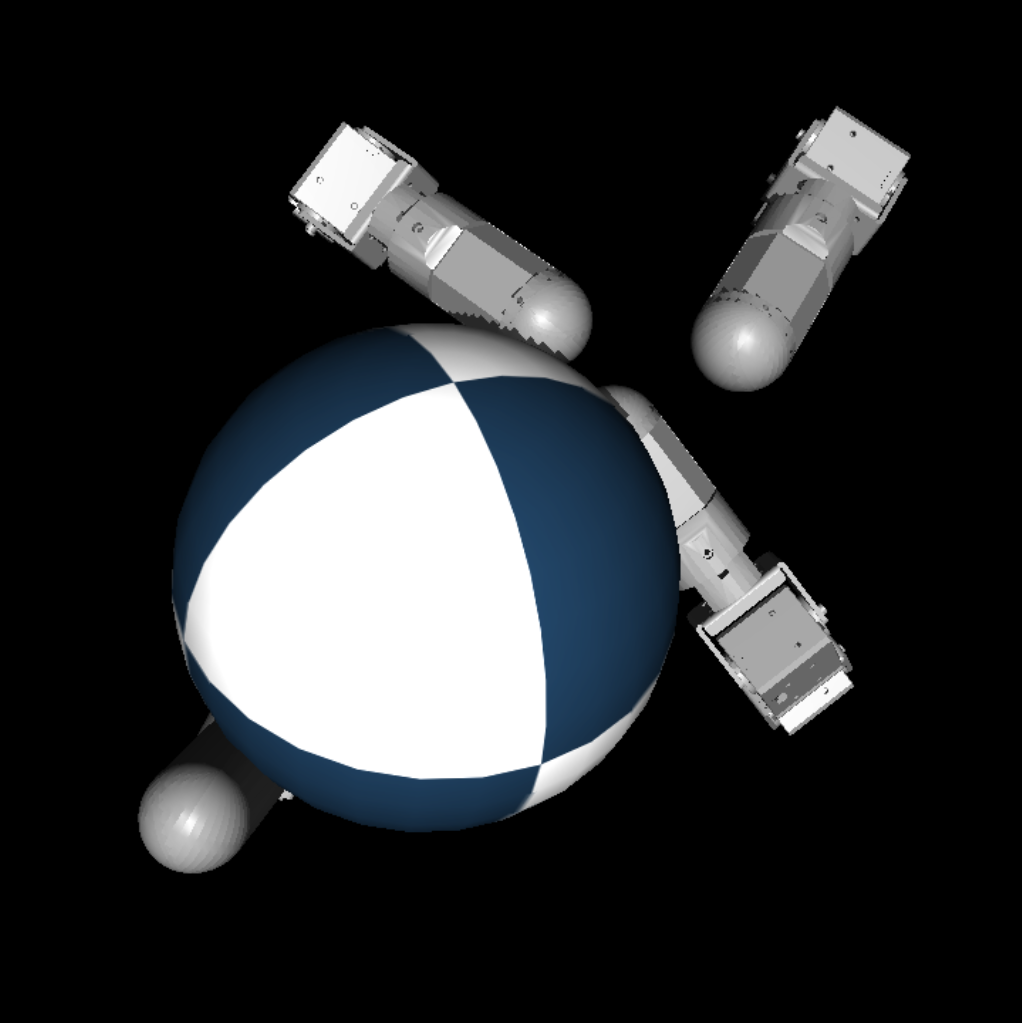}}\hfill%
\subfigure[]{\includegraphics[width=0.3\columnwidth]{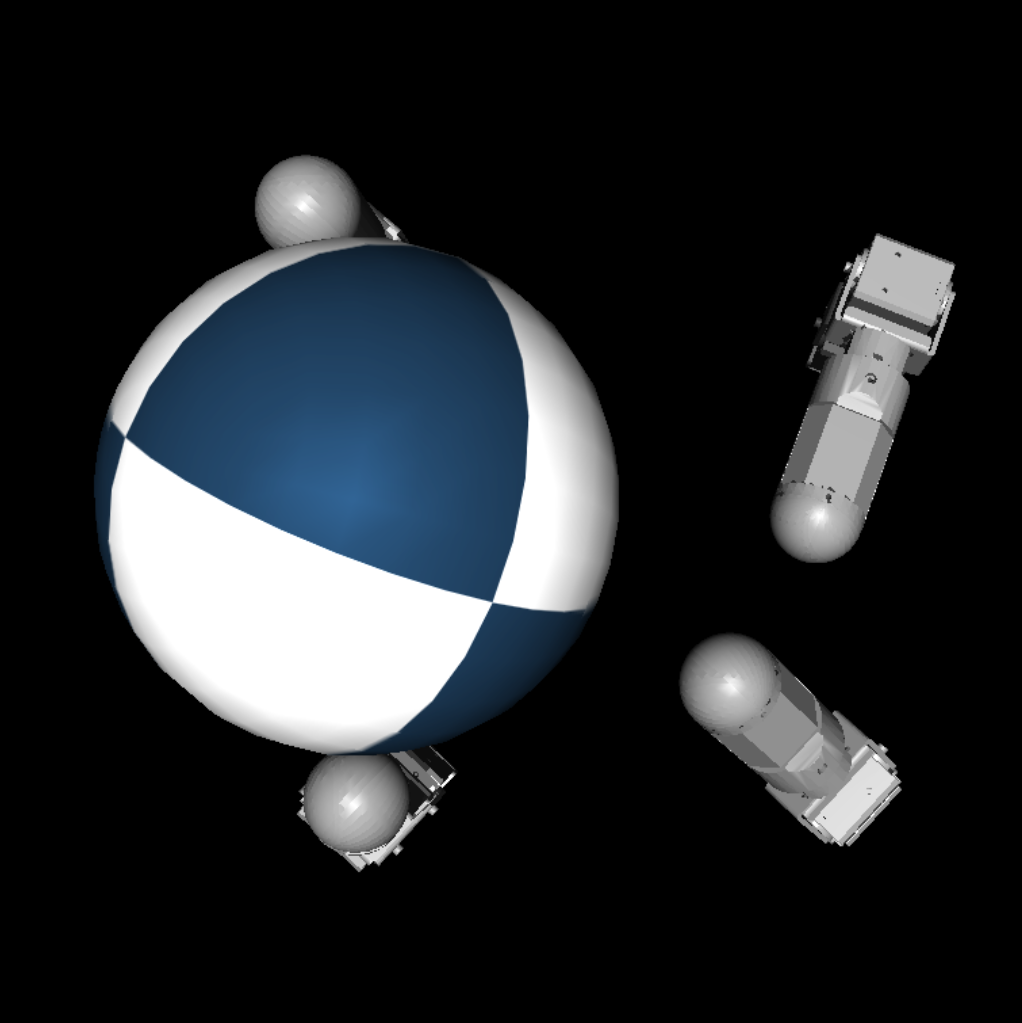}}\hfill%

\caption{One complete period of the manipulation gait learned with pure
reinforcement learning as seen from the top.}

\label{fig:manip2}
\end{figure}

We were able to successfully train policies to manipulate the sphere as
desired. The emergent behaviors are periodic and can therefore be likened to a
gait. This was expected as the reward function chosen encourages continued
motion of the grasped object. More specifically, we see the hand pinching the
sphere between two neighboring fingers and passing on to the next set of
neighboring fingers in order to rotate it (see
Figs.~\ref{fig:manip1} \& \ref{fig:manip2}.) We show the training curve of such
a successful experiment in Fig.~\ref{fig:reward}. It is clear that the gait
produced by the policy requires large movements of the sphere in the
horizontal plane in order to achieve rotation around the vertical axis.

We found that the randomization of initial states during training is of
profound importance. A good randomization that provides a distribution of
stable grasps that is as varied as possible allows for training with shorter
episode lengths, which in turn greatly improves sample complexity. This is not
a surprising finding. The intuition behind this is that without initial state
randomization the episode length would have to be at least as long as the
period of the emergent behavior we are trying to synthesize in order to be
able to learn a full period of the behavior. As we are trying to learn a
continuous rotation task we are expecting the policy to learn to gait the
fingers in such a periodic fashion. Initial state randomization means that the
policy can learn smaller sections of this gait in shorter episodes.

Furthermore, it appears that adding gravity in the simulation makes the task
much more difficult. Much more effort had to be spent tuning hyperparameters
in order to obtain behaviors that resemble periodic gaiting. Different objects
also appear to provide the task much difficulty. While learning policies to
manipulate a cylinder aligned with the vertical axis was relatively
successful, we could not train policies for more complex objects such as
cubes.

Being able to train adept policies for an in-hand manipulation task was an
important stepping stone for us. Perhaps unsurprisingly, the above results
show that reinforcement learning can be effectively leveraged to tackle the
difficulties due to fact that the in-hand manipulation task is only partially
observable. Fig.~\ref{fig:reward} however illustrates the sample complexity of
such an approach. While learning is fast at the beginning, it took over 2500
updates until learning appeared to plateau. These 2,500 updates correspond to
240,000 episodes, 15,360,000 steps or approximately 9 simulation days of
experience. This process took approximately 9 hours of wall time on a
commercial 11th generation server.

Note that the problem of rotating a sphere around a single axis is expected to
be much simpler than the task of moving general objects in arbitrary
directions. If it is possible to train policies to solve more general tasks
using reinforcement learning alone the number of samples required is expected
to be vastly greater than those we report above. Furthermore, while training
times on this scale are acceptable in simulation, they are prohibitively large
for training on real hardware. We believe that analytical grasp models can
help us improve sample complexity or even allow us to learn policies for tasks
for which pure reinforcement learning is insufficient. In the following we
will illustrate the intuition behind this using a simplified problem.

\section{Islands and Bridges: An equivalent problem?}\label{sec:islands}

Our central insight is that a defining characteristic of the state space of
manipulation problems is that only a small subset of it corresponds to stable
grasps. Defining the state space to consist of the object pose as well as the
joint angles of the hand then a majority of the space does not correspond to
grasps with valid contacts, much less a stable grasp. In fact, if the robotic
hand and object are rigid, then the contact conditions define a
lower-dimensional manifold in the full-dimensional state space~\cite{406939}.

A stochastic reinforcement learning algorithm operating in this space would
not be able to successfully traverse this manifold as stepping of it is
virtually guaranteed. In practice, of course, grasps are not perfectly rigid.
Series elastic actuators as well as soft contact interfaces are a popular
choice for exactly this reason: to provide some 'width' such that the
lower-dimensional manifold defined by the contact conditions turns into a
full-dimensional volume. A policy that changes the position setpoint of an
actuator connected to its joint through a series elastic coupling has a
greater chance of maintaining a valid contact state.

The fact remains, however, that the vast majority of the state space does not
correspond to valid grasps and that the regions of stability are seldom far
removed (in state space) from an unstable configuration. We like to think of
such a region of stability as an 'island'. In order to continuously reorient a
grasped object contacts must necessarily break to reestablish contact
somewhere else \cite{10.5555/2422356.2422377}. When a contact is broken a
grasp tends to become less stable --- that is until a new contact is
established somewhere else. We think of this process as crossing a 'bridge'
--- a region closer to instability that takes us to another, more stable
region. An island can thus also be thought of as corresponding to one grasp
geometry. When gaiting a finger we cross a 'bridge' to another grasp geometry.

\begin{figure}[t!]
\centering
\includegraphics[width=0.65\columnwidth]{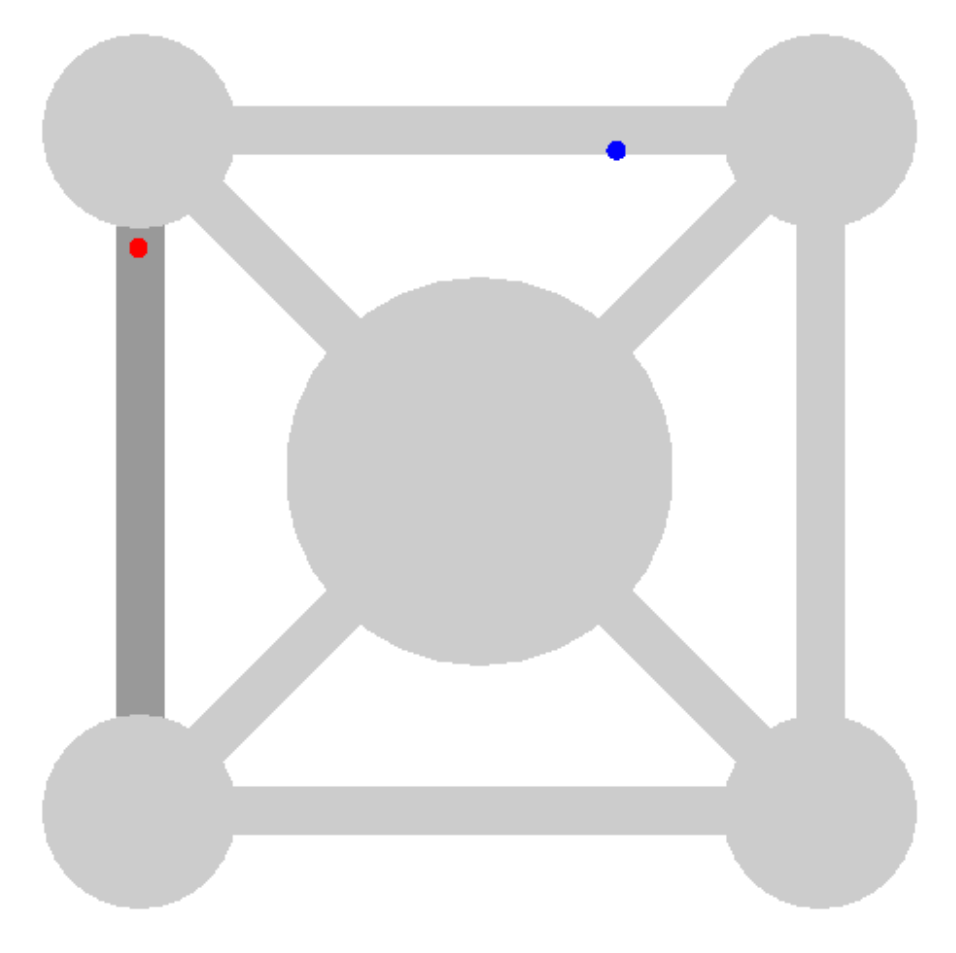}

\caption{Illustration of the 'Islands and Bridges' problem. The red and blue
dots corresponds to the current state and goal state respectively.}

\label{fig:islands}
\end{figure}

We therefore believe that there are parallels between the in-hand manipulation
problem and this problem which we call 'Islands and Bridges' (see
Fig.~\ref{fig:islands}.) For reasons of visualization we limit the problem to
two dimensions such that the current state is denoted as a cartesian position coordinate in this
environment made up of distinct islands and bridges connecting them. Any state
that lies on such an island or bridge corresponds to a stable grasp. A path
from one point in this environment to another hence corresponds to an in-hand
manipulation of the grasped object. Thus, the task of in-hand manipulation
from one grasp state to another corresponds to finding such a path.

Fig.~\ref{fig:islands} also illustrates the difficulty of solving such a task
with reinforcement learning: Any step off the thin bridges results in a
failure (i.e. dropping the object.) In order to perform in-hand manipulation
the policy must learn to navigate state space without ever entering these
unstable regions. Therefore, it is our intuition that if we can confine the
policy to the subset of the space which is useful, the task becomes easier to
learn. We use this simplified problem to investigate the efficacy of methods
that constrain the agent to the useful subset of the state space. We call a
method that modifies agent actions such that the given constraints are
satisfied a \textit{shield}. 

\begin{figure}[t!]
\centerline{
\includegraphics[width=.7\linewidth]{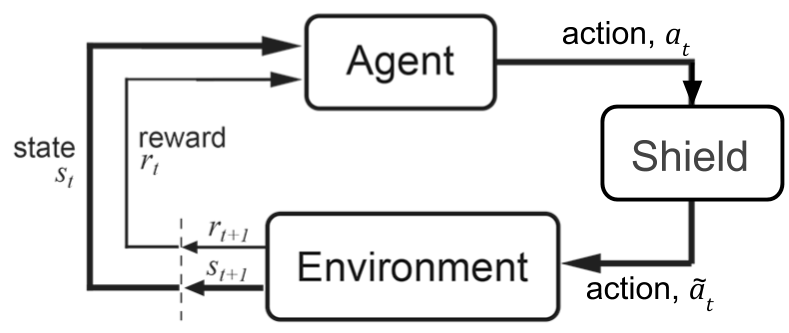}
}
\caption{Shield RL}
\label{fig:shieldRL}
\end{figure}

Let us formalize the notion of a shield. The method of shielding RL
\cite{ShieldRL} is shown in Fig.~\ref{fig:shieldRL}. All actions sampled
from the policy (deterministic or stochastic) are subject to the shield and,
if necessary, are modified by the shield to ensure 'safety'. Let us denote the
shield function as $\Psi$.

\begin{equation}
\tilde{a}^t = \Psi(a^t, s^t)
\label{eq:shield}
\end{equation}

where $a_t$ is the action sampled by the RL agent and $\tilde{a}_t$ is the
safe action given out by the shield function. It is infeasible to consider the
safety of an action over all future time steps, hence we perform a single-step
lookahead for determining the safety of an action. We declare an action a
safe action if the resulting state satisfies a set of \textit{shield
constraints} and thus define shield constraint function $\zeta$.

\begin{eqnarray}
\zeta(s^{t+1}) =
\begin{cases}
1, & \text{if constraints satisfied} \\
0, & \text{otherwise}
\end{cases}
\end{eqnarray}

Thus, we must be able to predict the state $s^{t+1}$ given $s^t$ and $a^t$. In
the case of our 'Islands and Bridges' problem this is trivial. Defining the
state as the cartesian position in the environment $s^t = [x,y]^T$ and the action as a
step in this cartesian space $a^t = [\delta x, \delta y]^T$ we can compute the
predicted state as follows:

\begin{equation}
s^{t+1} = s^t + a^t
\end{equation}

We can also define a distance function $M$ that is defined as the closest
distance of the current state to the edge of the safe region.

\begin{eqnarray}
\zeta(s^{t+1}) =
\begin{cases}
1, & \text{if } M(s^{t+1}) \geq 0 \\
0, & \text{otherwise}
\end{cases}
\end{eqnarray}

We can now define the shield function $\Psi$ for the islands-and-bridges
problem. A simple implementation would be to simply ignore any action if it
results in a state outside of the safe region.

\begin{eqnarray}
\tilde{a}^t = \Psi_{simple}(a^t, s^t) =
\begin{cases}
a^t, & \text{if } M(s^{t+1}) \geq 0 \\
[0,0]^T, & \text{otherwise}
\end{cases}
\end{eqnarray}

An issue with this formulation is that it does not work with deterministic
policies. Simply ignoring an action and remaining in the same state means that
the same action will be sampled again and rejected again for all future times.
Thus, a more sophisticated shield might be to find the action that results in
a state that satisfies $\zeta$ while maintaining the smallest deviation from the
originally sampled action.

\begin{equation}
\tilde{a}^t = \Psi_{smart}(a^t, s^t) = \argmin_{M(s^{t+1}) \geq 0}{\left\lVert \tilde{a}^t - a^t \right\rVert}
\end{equation}

We implemented both of these shields and applied them to the task of
navigating in the islands-and-bridges environment. Specifically, the task
consists of navigating from a randomly sampled initial position to a randomly
sampled goal position. We define two rewards $r$: A dense and a sparse reward
that only rewards the agent if it is within a distance $\epsilon$ from the
goal.

\begin{eqnarray}
r_{dense}^t =& -\left\lVert s^t - s^t_{goal} \right\rVert \\
r_{sparse}^t =&
\begin{cases}
1, & \text{if } \left\lVert s^t - s^t_{goal} \right\rVert \leq \epsilon \\
0, & \text{otherwise}
\end{cases}
\end{eqnarray}

\begin{figure}[t!]
\centering
\includegraphics[width=0.75\columnwidth]{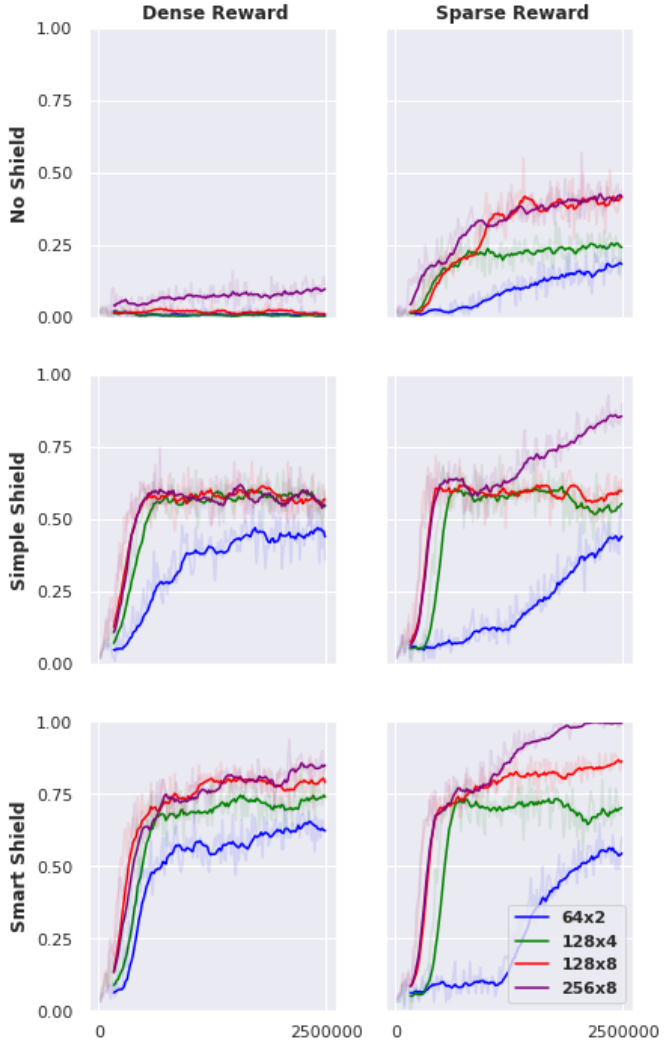}

\caption{Success rates of the PPO agent reaching the goal state in the islands
and bridges problem for different shield and reward functions. We also
experimented with the size of the policy and value networks (we used separate
networks of equal size.)}

\label{fig:curves}
\end{figure}

If the agent steps outside of the safe region (failure) or reaches the goal
region of radius $\epsilon$ (success) the episode is over and no more reward
may be collected. We used PPO to train a stochastic policy without a shield,
with the simple shield and with the smart shield as described above. After
training the shield remains in place. We show the success rates for all three
cases and both rewards throughout the training process in
Fig.~\ref{fig:curves}.

From these success rates it is evident that the shield provides large
advantages during training of the PPO agent. While a sparse reward results in
higher success rates for the unshielded case, it is still no match for the
success rates achieved using either the simple or smart shields. It is also
evident that the simple and smart shields perform similarly. However, only the
smart shield allows for policies to be trained that guarantee success. Perhaps
surprisingly it appears relatively large networks are also needed in order to
guarantee success even for such a small environment.

These findings suggest that a shield can be a great advantage in learning
tasks with RL that exhibit characteristics like the islands-and-bridges
environment. As we described above we believe that the task of in-hand
manipulation is such a task and therefore a shield can also be of great use
there.

\section{Constructing a shield for in-hand manipulation}

We now describe a method to construct a shield similar in spirit to those
described above for the in-hand manipulation problem described in
Chapter~\ref{sec:design}. We will make use of theoretical grasp
analysis in order to derive constraints for the shield to enforce as well as
the necessary means to predict future states. To this end we define the state
to consist of the pose of the object $\bm{u}$, the joint actuator setpoints
$\bm{q}$ as well as the contact forces $\bm{c}$.

\subsection{Grasp stability constraints}\label{sec:shield_constraints}

As we are concerned with maintaining a stable grasp at all times we chose a
set of grasp stability constraints as safety criteria to be enforced by the
shield. Specifically, we are concerned with maintaining sufficiently many
contacts for force-closure and preventing the contacts from slipping. Thus, we
want to ensure that the contact forces at a subset of contacts sufficient for
force closure remain within their respective friction cones.

For every such contact $i$ the following describe the constraints that ensure
the contact does not break or slip. Subscripts $n$ and $t$ refer to normal and
tangential components respectively.

\begin{align}
c_{i,n} &\geq 0 \label{eq:force_constr} \\
\left\lVert \bm{c}_{i,t} \right\rVert &\leq \mu t_{i,normal} \label{eq:fric_constr}
\end{align}

Furthermore, we may also want to prevent the object from moving into positions
from which it may be hard to make progress. Therefore we also confine the
object pose as part of the shield. We can enforce arbitrary convex constraints
on the object pose. For instance, we may confine the objects translational
degrees of freedom to a box:

\begin{align}
x_{min} \leq u_x \leq x_{max} \label{eq:box_start}\\
y_{min} \leq u_y \leq y_{max} \\
z_{min} \leq u_z \leq z_{max} \label{eq:box_end}
\end{align}

Rotational degrees of freedom are a somewhat more complex. We chose to
constrain the orientation of body axes in a global frame. Specifically, we
constrain the angle subtended between the projection of the body $z$-axis
$\hat{z}$ into the global $xz$ and $yz$ cardinal planes ($\hat{z}_x$ and
$\hat{z}_y$ respectively) and the global $z$-axis $\hat{Z}$ (all unit
vectors).

\begin{align}
r_{x,min} \leq sgn((\hat{z}_y \times \hat{Z}) \cdot \hat{X}) \cdot arccos(\hat{z}_y \cdot \hat{Z}) \leq r_{x,max} \\
r_{y,min} \leq sgn((\hat{z}_x \times \hat{Z}) \cdot \hat{Y}) \cdot arccos(\hat{z}_x \cdot \hat{Z}) \leq r_{y,max} \label{eq:rot_constr}
\end{align}

In short, the \textit{arccos} term gives us the angle between the two vectors
while the \textit{sgn} term tells us if the angle is in the negative or
positive sense with respect to the right hand rule. This angle is again
bounded to lie within a box.

Thus, given a state we can compute if it satisfies the above stability
constraints or not.

\subsection{Grasp model}\label{sec:shield_model}

Recall that the shield requires the ability to perform a 1-step lookahead.
Thus, in order for the above stability constraints to be useful we must be
able to predict future states from the current state-action pair. Ideally, we
would be able to solve an optimization problem such as that in the smart
shield described above. We could then find optimal actions such that stability
is guaranteed at the next step.

However, the agent can only control parameters such as joint positions or joint
torques, which in turn affect contact forces and object pose through complex
nonlinear relationships. Thus, exactly solving the optimization problem posed
by the shield is difficult as there is no closed form analytic expression for
the contact forces and object pose in terms of the agent action. We therefore
approximate this relationship through linearization, which also allows for
efficient computation of a global optimum.

We do this by making use of existing work by Cutkosky et
al.~\cite{CUTKOSKY_COMPLIANCE} analyzing the compliance characteristics of
robotic grasps as well as work by Bicchi~\cite{BICCHI94} who studied the
distribution of forces resulting from changes in position control setpoints.

\vspace{2mm} \noindent \textbf{Equilibrium:} The Grasp Map Matrix $\bm{G} \in
\mathbb{R}^{6 \times 3m}$ maps the contact forces at the $m$ contacts to the
corresponding wrench acting on the object and from (small) object motions to
the corresponding contact motions. It therefore expresses object force/torque
equilibrium and rigid body motion. The Hand Jacobian $\bm{J} \in
\mathbb{R}^{3n \times q}$ similarly maps from contact forces to joint torques
at the $l$ joints and from (small) joint motions to corresponding contact
motions. It expresses joint torque equilibrium and hand kinematics.

\begin{align}
\bm{w} &= - \bm{G} \bm{c} \label{eq:equilibrium2} \\
\bm{\tau} &= \bm{J}^T \bm{c} \\
\Delta \bm{d} &= \bm{J} \Delta \bm{q} - \bm{G}^T \Delta \bm{u}
\end{align}

Calculating contact forces that arise in response to an applied wrench
$\bm{w}$ is complicated by the fact that, in general, \textbf{G} will have
fewer rows than columns. Therefore it has a nullspace and no unique inverse.
This is known as static indeterminacy.

\vspace{2mm} \noindent \textbf{Grasp Stiffness:} We do, however, know that if
we apply a wrench \textbf{w} to a grasped object there must arise a unique set
of contact forces. Assuming perfectly rigid bodies the problem of computing
these contact forces is ill posed but we can resolve the static indeterminacy
by taking into account the \textit{compliance} of the grasp. By doing so we
are adding constitutive relations to resolve the static indeterminacy. 

\begin{figure}[t!]
\centering
\includegraphics[width=0.65\columnwidth]{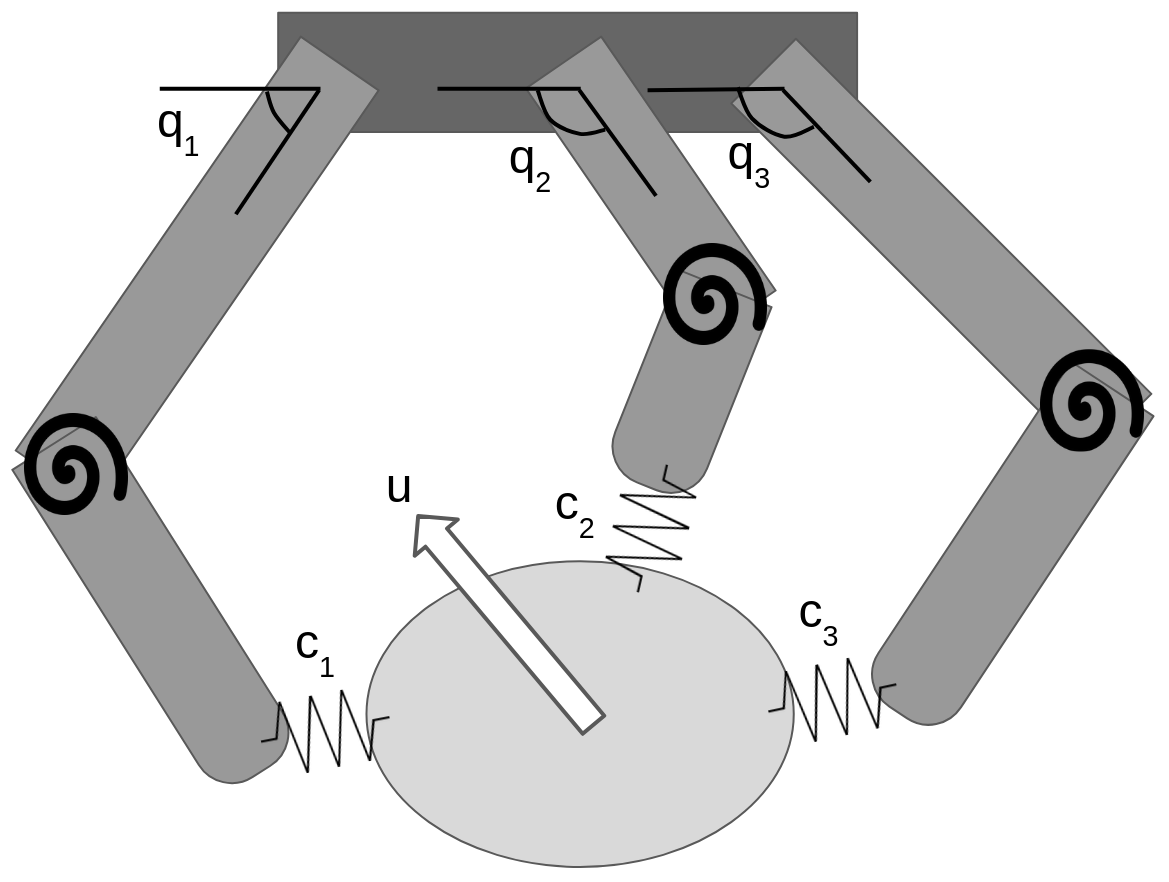}

\caption{Illustration of a grasp with its main sources of compliance. Linear
springs at the contacts denote compliance due to soft interfaces between the
fingers and the object. Torsional springs denote the compliance due to series
elastic actuation as well as the compliance of the low level controller of the
actuator.)}

\label{fig:compliance}
\end{figure}

Using the work of Cutkosky et al.~\cite{CUTKOSKY_COMPLIANCE} we can derive the
compliance in contact frame by constructing the mapping from a change in
contact forces $\Delta \bm{c}$ to contact motions $\Delta \bm{d}$. The main
sources of compliance are at the contacts (due to softness of the skin) and in
the joints (due to the proportional term of the servo controller as well as
any series elastic coupling. See Fig.~\ref{fig:compliance}.) Assuming linear
compliances we can construct diagonal matrices $\bm{C}_{contacts} \in
\mathbb{R}^{3m \times 3m}$ and $\bm{C}_{joints} \in \mathbb{R}^{l \times l}$
containing the compliances of contacts and joints respectively. 

Starting from an initial equilibrium position we can now apply a change to the
contact forces $\Delta \bm{c}$ and compute the resulting contact motion
$\bm{d}_{contacts}$ due to contact compliance.

\begin{equation}
\Delta \bm{d}_{contacts} = \bm{C}_{contacts} \Delta \bm{c}
\end{equation}
Computing the contact motion $\bm{d}_{joints}$ due to joint compliance is only
slightly more complex.
\begin{equation}
\Delta \bm{d}_{joints} = \bm{J} \bm{C}_{joints} \bm{J}^T \Delta \bm{c} \label{eq:motion}
\end{equation}
Due to the linearity of the compliance, overall contact motion is simply the
sum of the two components.
\begin{align}
\Delta \bm{d} &= \Delta \bm{d}_{contacts} + \Delta \bm{d}_{joints} \\
&= (\bm{C}_{contacts} + \bm{J} \bm{C}_{joints} \bm{J}^T) \Delta \bm{c}
\end{align}
From this we can determine the grasp stiffness --- mapping from contact motion
to the corresponding change in contact forces --- by inverting the overall
compliance matrix.
\begin{align}
\Delta \bm{c} &= (\bm{C}_{contacts} + \bm{J} \bm{C}_{joints} \bm{J}^T)^{-1} \Delta \bm{d} \\
&= \bm{K} \Delta \bm{d}
\end{align}
Using equation (\ref{eq:motion}) we can express the change in contact force as
a function of object motion and joint setpoint change.
\begin{equation}
\Delta \bm{c} = \bm{K} (\bm{J} \Delta \bm{q} - \bm{G}^T \Delta \bm{u}) \label{eq:K_matrix}
\end{equation}

\vspace{2mm} \noindent \textbf{Controllable Object Motions:} As
we only have direct control over the joint setpoint, Bicchi~\cite{BICCHI94}
was interested in predicting the motion of the object $\Delta \bm{u}$ that
will arise when we change the actuator setpoint by $\Delta \bm{q}$. Assuming
equilibrium and using (\ref{eq:equilibrium2}) \& (\ref{eq:K_matrix})

\begin{equation}
\bm{G} \Delta \bm{c} = \bm{G} \bm{K} (\bm{J} \Delta \bm{q} - \bm{G} \Delta \bm{u}) = 0
\end{equation}
which implies
\begin{equation}
\bm{G} \bm{K} \bm{J} \Delta \bm{q} = \bm{G} \bm{K} \bm{G} \Delta \bm{u}
\end{equation}
and hence
\begin{align}
\Delta \bm{u} &= (\bm{G} \bm{K} \bm{G})^{-1} \bm{G} \bm{K} \bm{J} \Delta \bm{q} \\
&= \bm{F} \Delta \bm{q}
\label{eq:F_matrix}
\end{align}

\vspace{2mm} \noindent \textbf{Controllable Internal Forces:} Armed with this
we can predict the contact forces that will arise when we change the actuator
setpoint by $\Delta \bm{q}$. Substituting (\ref{eq:F_matrix}) into
(\ref{eq:K_matrix}) we get

\begin{align}
\Delta \bm{c} &= \bm{K} (\bm{J} \Delta \bm{q} - \bm{G}^T (\bm{G} \bm{K} \bm{G})^{-1} \bm{G} \bm{K} \bm{J} \Delta \bm{q}) \\
&= (\bm{I} - \bm{G}^T (\bm{G} \bm{K} \bm{G})^{-1} \bm{G}) \bm{K} \bm{J} \Delta \bm{q} \\
&= \bm{E} \Delta \bm{q}
\label{eq:E_matrix}
\end{align}

\vspace{2mm} \noindent \textbf{State prediction:} All the relations described
above are linearizations that describe the behavior of a grasp with respect to
changes in actuator setpoints. Due to nonlinearities such as rolling contacts
etc. these linearization are only accurate in a small region around the
current state. They do, however allow us to make predictions of future states
given a current state-action pair given that the action is sufficiently small.
Recall that the components of the state we are interested in with respect to
satisfying grasp stability constraints are the contact forces as well as the
object pose. We now have the tools to express these parts of the state in
terms of the action defined as setpoint changes of the position controlled
actuators.

\begin{equation}
\begin{bmatrix}\bm{c} \\ \bm{u} \end{bmatrix}^{t+1} =
\begin{bmatrix}\bm{c} \\ \bm{u} \end{bmatrix}^{t} +
\begin{bmatrix}\bm{E} \\ \bm{F} \end{bmatrix}^{t} \Delta \bm{q}^t
\label{eq:linearization}
\end{equation}

\subsection{Complete shield formulation}

We can now put together the pieces and define the shield function $\Psi$ in
(\ref{eq:shield}) as a constrained optimization: What action satisfies all
shield constraints while minimizing the Cartesian distance from the original
action.

\begin{align}
minimize &: \norm{\bm{a}^t - \tilde{\bm{a}}^t} \\
such \ that &: (\ref{eq:force_constr})-(\ref{eq:rot_constr})
\label{eq:optimization}
\end{align}

Of course, in order to allow gaiting we must allow contacts to break
occasionally such that fingers can reestablish contact elsewhere. This is why
the constraints (\ref{eq:force_constr}) \& (\ref{eq:fric_constr}) may only be
applied to a subset of contacts --- we must sometimes exempt a contact from
the shield in order allow for contact relocation. We do this as follows:

\begin{enumerate}

\item Enumerate all combinations of contact states, where every contact is
either shielded or not.

\item Discard all combinations with fewer than 2 shielded contacts as at least
two contacts are required for stability.

\item For all remaining combinations check if the selection of shielded
contacts by themselves are sufficient for \textit{force closure} (are these
contacts sufficient for stability.) Discard all combinations that fail this
criterion. This can be done using simple force closure tests~\cite{FERRARI92}.

\item Perform the optimization (\ref{eq:optimization}) for every remaining
combination. For all unshielded contacts remove the stability constraints
(\ref{eq:force_constr}) \& (\ref{eq:fric_constr}) and add constraints
$\tilde{a}_j = a_j$ for all joints $j$ that make up the kinematic chain of an
unshielded contact.

\item Select the result with the lowest value optimal objective. This solution
gives us $\tilde{\bm{a}}$

\end{enumerate}

Thus, we have computed a safe action $\tilde{\bm{a}}^t$ that applies the
smallest correction to $\bm{a}^t$ such that stability constraints are
satisfied. This optimization formulation is an important distinction between
our work and the hierarchical control approach by Li et
al.~\cite{Li2020LearningHC}. Their approach relies on a finite number of low
level controllers from which the policy chooses, which constrains the possible
behaviors to those determined by those controllers. In our case the policy
retains all freedom to individually change actuator commands as long as the
resulting grasp is stable. It is our intuition that while this may result in a
more difficult task it ultimately allows for more dexterous behavior. 

We also want to note that the method of enumerating contact state combinations
in order to allow for finger gaiting bears similarities to the work by
Prattichizzo et al.~\cite{PRATTICHIZZO97}. In fact it is their proposed
\textit{Potential Grasp Stability}, which has inspired us to combine the
stability constraints described in Chapter~\ref{sec:shield_constraints} with
such an enumerative approach.

\section{Future work}

Recall that the experimental design described in Chapter~\ref{sec:design}
provides us with information of contact position, contact forces, joint
positions and joint torques. Thus, we can compute the grasp map matrix as well
as the grasp Jacobian and perform an optimization over contact forces as
described above. This is true for both the simulated hand as well as its
physical counterpart, once it becomes available to us. Natural next steps for
us to take include the application of the shield in simulation as well as
attempting to transfer the policies trained and described in
Chapter~\ref{sec:RL_results} to the real robotic hand we are constructing.

Assuming the islands-and-bridges concept is indeed an accurate representation
of the in-hand manipulation task and the shield proves to be an effective
tool, there are many further open questions that require investigation. So far
we have only considered tasks in which the policy was trained on a single
object and is tested on the same object. The concept of
\textit{generalization} is central to learning methods and in this case could
be applied to a policy that can successfully manipulate previously unseen
objects. Note that this could be another instance in which a shield as
described above becomes advantageous as the shield is entirely agnostic to the
object. A policy could learn general manipulation strategies while the
shield provides the intricate feedback control necessary for stability.

We believe that policies can be trained on a variety of objects and thus learn
a general structure of the underlying manipulation problem. Partial
observability becomes an even larger challenge during such tasks and thus it
may be advantageous to make use of LSTM architectures for policy and value
networks. This may allow the meta-learning of general in-hand manipulation
tasks as the experience collected is encoded in the internal state of the LSTM
and can be used to adapt to the specific object.

Considering that the navigation in previously unknown environments bears
similarities to maze navigation an interesting avenue of investigation might
be the application of RL maze solving
algorithms~\cite{7849365}\cite{mirowski2017learning}\cite{10.5555/3327144.3327168}.
Furthermore, finger gaiting bears many similarities with regrasping for object
reorientation. Manipulation for regrasping has been studied for decades (see
for example~\cite{Alami1989}\cite{774088}\cite{769979}\cite{Wan2015}) and many
different approaches such as regrasp maps~\cite{Alami1989} or
graphs~\cite{509176}\cite{Wan2015} have been proposed. These manipulation
representations appear to be related to the notion of 'islands and bridges'
described in Chapter~\ref{sec:islands} and we expect there to be many other
parallels with different aspects of the manipulation theory that may provide valuable
insights into the problem of in-hand manipulation.

\chapter{Conclusions}\label{sec:conclusion}

\section{Contributions}

In this dissertation we discuss the reasons for the relative obscurity of
theoretical grasp analysis within the applied manipulation
community. We argue that the most important reason is that none of the
existing theoretical approaches capture the passive effects that are the
fundamental characteristic of most hands commonly used in practice today.
Particularly the effects of nonbackdrivable actuators have not been previously
studied. This is despite the fact that this characteristic results in the
notion of \textit{passive stability}, which we argue has defined both the
design of multi-fingered robotic hands as well as the ways we use them in
practice.

We describe the limitations of the existing grasp theory, which prevent the
analysis of passive effects as well as alternative approaches from the rigid
body dynamics, contact modeling as well as physics simulations communities. We
argue that none of these existing approaches allow for the efficient
computation of grasp stability taking into account passive effects as they are
either too computationally complex or approximate in nature. Thus, we set out
to develop our own grasp models and algorithms in order to analyze the
stability of grasps with passive effects. 

Our first contribution is a grasp model for planar grasps that allows for the
computation of contact forces given the \textit{slip state} of the contacts.
The state of a contact is defined as either rolling, breaking or sliding in
one of two directions. We show that given a valid contact slip state
combination stability can be determined by a linear program.

We then show that the number of contact state combinations possible under
rigid body motion constraints is quadratic in the number of contacts and describe an
algorithm that allows for the enumeration of all such combinations in
polynomial time. Thus, we have developed the first algorithm that can
determine planar grasp stability taking into account passive effects in
polynomial time. Using an example grasp we show that this method produces
physically accurate results and demonstrate its computational efficiency.

Turning our attention to spacial grasps we continue our investigation by
describing a system of equations, which we call the \textit{exact problem}. It
comprises of the equations defining static equilibrium of a grasp with passive
effects due to nonbackdrivable joints and underactuation. As we show many of
these governing equations can be cast as constraints in a Mixed Integer
Program for which efficient solvers exist. Due to the Maximum Dissipation
Principle (MDP) the Coulomb friction law is shown to be non-convex and that it
cannot easily be reformulated using piecewise convex decompositions as was
done in the planar case.

As we show, a consequence of omitting the MDP from our treatment of static
grasp stability is that we introduce unphysical solutions (they violate the
law of conservation of energy) that make determination of stability
impossible. We thus propose a physically motivated iterative scheme that aims
to mitigate the effects of omitting the MDP. Constraining the motion of the
grasped object and minimizing at every step the resultant wrench acting on the
object approximates the effects of the MDP and thus leads to physically
plausible solutions.

This algorithm is the first to enable the stability analysis of spacial grasps
with passive characteristics such as nonbackdrivable joints and
underactuation. We demonstrate the application of our framework with various
grasps using fully actuated as well as underactuated robotic hands. Verifying
the predictions made by our algorithm with empirical data we illustrate its
practical applicability as a tool both for grasp selection as well as actuator
control.

Building on this work we investigate the use of piecewise convex relaxations
of the MDP in order to accurately enforce it and obtain solutions to the
\textit{exact problem}. We discuss the potential application of existing
convex relaxation techniques based on McCormick envelopes and describe an
important limitation in that they require knowledge of upper and lower bounds
on some variables. Our contribution here is a convex relaxation technique that
makes use of the structure of the friction constraint in order to alleviate
this limitation.

As our relaxation satisfies all the requirements for the application of
tightening methods we also contribute an algorithm that allows for the
successive hierarchical refinement of the convex relaxation in order to obtain
solutions converging to those of the \textit{exact problem}. An important
advantage of tightening methods such as ours is that it provides strong
guarantees: Specifically, we are guaranteed convergence to the solution of the
exact problem if one exists. Furthermore, if our algorithm does not find a
solution we are guaranteed that none exists. Combined, these two features
make our method efficient for problems both with and without exact solutions.
It is, to the best of our knowledge, the first time that grasp stability
models incorporating Coulomb friction (along with the maximum dissipation
principle) have been solved with such high discretization resolution.

We furthermore propose a method of accounting for uncertainties in grasp
geometry, specifically contact normals, in order to robustly guarantee
stability up to a given magnitude of uncertainty. As this framework allows for
multiple queries posed as optimization problems we demonstrate its application
not only in the determination of passive stability but also as a tool for
computing optimal actuator commands. We investigate the computational
performance on a range of algorithmically generated grasps and show that our
framework is computationally viable as a practical tool.

Implementations of the grasp models and algorithms presented in this
dissertation are publicly available as part of the open source GraspIt!
simulator~\cite{MILLER04}.

Having developed tools that fill a gap in the grasp theory with respect to
understanding and modeling the passive effects of robotic grasps we turn our
attention to recent developments in the robotic manipulation community ---
specifically the advent of deep reinforcement learning and its application for
in-hand manipulation. We argue that parts of the classical grasp modeling
theory is indeed applicable to the types of robotic hands considered by
practitioners in the learning community. As we believe there are synergistic
benefits to be discovered at the interface of classical grasp analysis and
reinforcement learning we contribute a method to constrain an RL agent such
that it maintains a stable grasp at all times.

We draw parallels to established terms such as 'Safe RL' and 'Shield RL' and
derive a shield using a linearization based on traditional grasp analysis.
While this work is highly exploratory we present preliminary results, which we
believe to be promising and calling for further research.

\section{Potential impact}

The work pertaining to the analysis of passively stable grasps presented in
this dissertation marks a clear theoretical advance. For the first time it is
possible to determine the passive behavior of grasps with commonly used
robotic hands such as the Barrett and Schunk hands. A potential practical
application is in grasp planners, which compute the grasp to be executed by a
robot. 

Typically, these grasp planners compute a list of candidate grasps and rank
them by a quality metric such as the Ferrari and Canny Grasp Wrench Space
metric~\cite{FERRARI92}. The grasp quality metrics commonly used by popular
grasp planners are concerned with contact locations only and do not reason
about contact forces. Thus, they cannot predict actual stability. There
already exist grasp quality metrics that take into account hand kinematics and
force generation capabilities, but as described in Chapter~\ref{sec:bridge}
they make assumptions that limit their applicability to most common robotic
hands.

Our frameworks could fill this gap and provide grasp planners with the means
to make accurate predictions of stability taking into account not only the
hand kinematics but also actuator commands. Thus, our methods provide a
further advantage in that they allow for the optimization of actuator
commands. Where existing grasp planners only generate grasp geometries, our
work can also compute the appropriate actuator commands for a given grasp and
task.

We stress that while we have investigated the problem of grasp stability from
our perspective as manipulation researchers the models and algorithms in this
dissertation are applicable to a much wider class of problems. The stability
of arrangements of rigid bodies in frictional contact, some fixed, some free,
some constrained by other unilateral or bilateral constraints is an important
aspect of many other problems from a variety of fields such as locomotion,
robotic construction and the simulation of rigid body structures. We hope that
our work may also be of use to researchers in these fields.

In the context of manipulation however, we note that interest in this kind of
traditional analytic grasp analysis has decreased over the last decade or so.
Modern grasp planning research tends to focus on the applications of machine
learning and the popularity of simple two-fingered grippers as well as suction
cups has diminished the need for complex quality metrics. 

So perhaps there is more progress to be expected in the field of data-driven
robotics. Recent results from the deep reinforcement learning community are
promising a new breakthrough in truly dexterous multi-fingered manipulation.
We hope that our perspective on these problems as analytical grasp stability
analysis researchers can provide valuable insights and that our current work
of integrating analytical grasp modeling with reinforcement learning will
contribute to this progress. 

\section{Challenges}

Throughout our work we have found that the accurate modeling of the Maximum
Dissipation Principle is perhaps the most complex and difficult aspect of any
passive stability formulation. It has thus provided us with many challenges in
developing the methods in this dissertation.

While we show a convenient piecewise convex decomposition of the MDP in two
dimensions, its treatment only remains of polynomial complexity under some
fairly limiting assumptions. We are limited to a single mobile body in the
grasp and can thus have to model the hand as a single rigid body; we cannot
model the kinematic effects of multi-fingered robotic hands and joint
actuation. Accurate treatment of the MDP is necessarily exponential in the
number of bodies involved and while our algorithm can be extended to such
multi-body problems it cannot retain its polynomial complexity.

For spacial contacts there is unfortunately no exact piecewise convex decomposition
of the MDP. Our iterative approach to approximating the MDP --- while
physically motivated --- unfortunately also loses many useful guarantees: It
is not guaranteed to converge and if it does is not guaranteed to converge to
a physically meaningful solution. As a result we report some outlier results,
which are difficult to diagnose, particularly for complex grasps. Furthermore,
experimental validation is difficult due to the difficulty of exactly
recreating grasps on a real hand and the uncertainty in many of the physical
parameters such as the torque provided by the actuators.

In order to alleviate some of these limitations we turned to convex
relaxations of the MDP in order to solve the exact problem. The price we pay
is an increased complexity in the grasp model and vastly larger optimization
problems. While our algorithm often beats its theoretical worst-case
exponential complexity the time taken to answer queries with more than 4
contacts is still significant. This is a major hurdle to the application of
our algorithm in practice.

A further limitation is an arguably narrow definition of grasp stability. An
initially unstable grasp may - through movement of the fingers and object
and hence changes in the grasp geometry - eventually settle in a different stable equilibrium
grasp. In such a case our framework can only determine that the initial grasp
is unstable and makes no prediction on stability of the final grasp. In order
to account for the motion of an initially unstable grasp we would have to
model the dynamics of the grasp. Using the currently available dynamics
engines to this end comes with its own difficulties.

Our secondary focus has been on the application of traditional grasp
analysis to reinforcement learning of in-hand manipulation tasks. While
results are preliminary and this work is still ongoing the amount of
engineering that has to be performed in order to obtain convincing policies
has been significant. This is expected to be even more pronounced once we
complete building the physical hand with which we plan on putting our insights
into practice. Furthermore, we foresee the discrepancies between the
simulation and reality to be a major challenge. 

\section{Future work}

We believe we have provided a solid foundation of theoretical work that, for
the first time, enables the analysis of passive effects in grasping. There is,
however, still room for improvement. An extension of the hyperplane approach
to contact state enumeration for spacial grasps is already being
investigated~\cite{Huang2020}. Another interesting and useful extension would
be to allow for multiple bodies such that hand kinematics can be accounted
for.

A more rigorous validation of our methods for stability determination in three
dimensions would be of value, albeit difficult. However, perhaps there are
other contact problems that exhibit similar characteristics that are more
amenable to empirical validation.

While we believe we described a method to solve the \textit{exact problem}
there remain many avenues of investigation in order to improve computational
performance. While we used Branch and Bound at every iteration of our
tightening approach we did not retain any information of the solution tree
between iterations. This means that many convex subproblems are being
recomputed at every tightening. Using a solver that allows for warm starting
the Branch and Bound process promises great performance gains. An alternative
approach might be to cast our relaxation in a formulation that can be solved
using existing spacial Branch and Bound algorithms, making the refinement part
of the branch and bound process itself.

Ultimately, in this dissertation we have highlighted what we believe to be
research topics holding influential results of great practical importance. We
believe that there is yet much to be learned about passive reactions in
grasping and how to effectively leverage these effects. If we want to move
beyond the use of suction grippers for bin picking tasks and achieve robust
manipulation with multi-fingered hands we have to develop appropriate tools to
model and analyze them. We hope that this field of study sees renewed interest
now that industries have evolved, which require large numbers of humans
fulfilling menial tasks such as bin picking.

Furthermore, we believe there lies much promise in bringing approaches from
the traditional grasping theory to the learning community. Our own endeavors
to that end will continue as we plan on continuing along the path outlined in
this dissertation. The increasing sensitivity of tactile sensors along with
their decrease in price will hopefully drive more practitioners to consider
the grasping theory that allows them to analyze the data. To this end we hope
the community will continue the development of a new generation of robotic
hands, which will provide practitioners with rich tactile and proprioceptive
information and become a viable alternative to purely vision-based systems.

If we want to see robots becoming commonplace in human spaces we need to learn
to endow robots with the skills that come to us so naturally. A task as simple
as picking an unknown object from a bin without vision is still an open
problem in robotics. Perhaps it is unsurprising that the methods that
currently appear the most promising and lead to the most human-like robotic
behaviors --- such as deep reinforcement learning --- are those inspired by
nature itself. And while roboticists are captivated by these new methods and
robots make up more and more ground on their human counterparts we hope that
researchers will not lose sight of the analytical models and algorithms.




\titleformat{\chapter}[display]
{\normalfont\bfseries\filcenter}{}{0pt}{\large\bfseries\filcenter{#1}}  
\titlespacing*{\chapter}
  {0pt}{0pt}{30pt}

\cleardoublepage
\phantomsection
\addcontentsline{toc}{chapter}{References}  
\begin{singlespace}  
	\setlength\bibitemsep{\baselineskip}  
		\normalem
	\printbibliography[title={References}]
\end{singlespace}


  



\end{document}